\documentclass[%
]{tumDiss}
\usepackage[utf8]{inputenc}
\usepackage{lipsum}
\usepackage{dsfont}

\university{University of Augsburg}
\faculty{Faculty of Applied Computer Science}

\degree{Dr. rer. nat.}

\title{
\texorpdfstring{Learning to Predict, Discover, and Reason in High-Cardinality Event Sequences\\}{Learning to Predict, Discover, and Reason in High-Cardinality Event Sequences\\}
}

\subtitle{Applications to Vehicle Diagnostics Automation}

\author{Hugo Math}
\erstpruef{Prof. Dr. rer. nat. Rainer Lienhart}
\zweitpruef{Prof. Dr. rer. nat. Elisabeth André}

\date{March 2026}

\datedefense{October 2026}

\usepackage{scrhack}
\usepackage{pdfpages}
\usepackage{tikz}
\usepackage{xcolor}
\usepackage{booktabs}
\definecolor{codegreen}{rgb}{0,0.6,0}
\definecolor{codegray}{rgb}{0.5,0.5,0.5}
\definecolor{codepurple}{rgb}{0.58,0,0.82}
\usetikzlibrary{matrix, positioning, calc, shapes.geometric, arrows.meta, fit, backgrounds, patterns, decorations.pathreplacing}

\usepackage{xargs}
\usepackage{graphicx}
\usepackage{pgfplots}
\usepackage{pgfplotstable}
\tikzset{>=stealth}
\usetikzlibrary{patterns}
\usetikzlibrary{pgfplots.statistics}
\usetikzlibrary{backgrounds,scopes}
\usetikzlibrary{backgrounds,scopes}
\usetikzlibrary{arrows}
\usetikzlibrary{positioning}
\usetikzlibrary{calc}
\usetikzlibrary{fit}
\usetikzlibrary{shapes.misc}
\usetikzlibrary{external}
\usepackage{amsthm}

\usepackage{booktabs}
\usepackage{xcolor}
\usepackage{colortbl}
\usepackage{array}
\usepackage{multirow}
\usepackage{makecell}

\newtheorem{theorem}{Theorem}
\newtheorem{lemma}{Lemma}

\newtheorem{definition}{Definition}
\newtheorem{assumption}{Assumption}
\newtheorem{remark}{Remark}
\newtheorem{proposition}{Proposition}

\usepackage{moreverb}

\usepackage{listings}

\usepackage{amsmath}
\usepackage{amssymb}
\usepackage{amsfonts}
\usepackage{upgreek}
\usepackage{dsfont}
\usepackage{accents}
\usepackage{mathtools, nccmath}
\usepackage{svg}
\usepackage{tabularx}
\usepackage{enumerate}

\definecolor{codebg}{RGB}{245,245,244}
\definecolor{codekw}{RGB}{0,0,180}
\definecolor{codestring}{RGB}{153,0,0}
\definecolor{codecomment}{RGB}{0,128,0}

\tikzset{
    event/.style={circle, draw=blue, fill=blue, inner sep=2pt},
    cnode/.style={circle, draw, minimum size=18pt},
    axis/.style={thick},
    causal/.style={->, thick}
}

\lstdefinestyle{oscar}{
  backgroundcolor=\color{codebg},
  basicstyle=\ttfamily\footnotesize,
  keywordstyle=\color{codekw}\bfseries,
  stringstyle=\color{codestring},
  commentstyle=\color{codecomment}\itshape,
  columns=fullflexible,
  frame=single,
  framerule=0pt,
  rulecolor=\color{black!20},
  breaklines=true,
  showstringspaces=false,
  tabsize=2,
}
\usepackage{siunitx}
\usepackage[all]{nowidow}

\usepackage{lipsum}

\usepackage{url}
\usepackage[
  hidelinks,
  bookmarksnumbered,
]{hyperref}

\usepackage{caption}

\lstdefinestyle{mystyle}{
    commentstyle=\color{codegreen},
    keywordstyle=\color{magenta},
    numberstyle=\tiny\color{codegray},
    stringstyle=\color{codepurple},
    basicstyle=\ttfamily\footnotesize,
    breakatwhitespace=false,
    breaklines=true,
    captionpos=b,
    keepspaces=true,
    numbers=left,
    numbersep=5pt,
    showspaces=false,
    showstringspaces=false,
    showtabs=false,
    tabsize=2
}

\usetikzlibrary{shapes,decorations,arrows,calc,arrows.meta,fit,positioning}
\tikzset{
    -Latex,auto,node distance =1 cm and 1 cm,semithick,
    state/.style ={ellipse, draw, minimum width = 0.7 cm},
    point/.style = {circle, draw, inner sep=0.04cm,fill,node contents={}},
    bidirected/.style={Latex-Latex,dashed},
    el/.style = {inner sep=2pt, align=left, sloped}
}

\usepackage[
  toc,
  acronym,
  style = long,
  nogroupskip,
  section
]{glossaries}

\makeglossaries

\usepackage[colorinlistoftodos,prependcaption,textsize=small]{todonotes}
\usepackage{makecell}
\usepackage{booktabs}
\usepackage{multirow}
\usepackage{subcaption}
\usepackage[normalem]{ulem}
\usepackage[export]{adjustbox}
\usepackage{tikz}
\usepackage{diagbox}
\usepackage{pdflscape}
\usepackage{afterpage}
\usepackage[lined,ruled,linesnumbered]{algorithm2e}
\SetKwComment{Comment}{/* }{ */}
\usepackage{array,multirow}
\usepackage{float}
\usepackage{placeins}
\usepackage{epsfig}
\usepackage{enumitem}
\usepackage{layout}
\usepackage{wrapfig}
\usepackage{nameref}
\usepackage{algorithmic}
\usepackage{changepage} %

\usepackage{tkz-euclide} %
\usetikzlibrary{positioning}
\allowdisplaybreaks %

\usepackage{pifont}%
\usepackage{rotating}
\usepackage{afterpage}

\usepackage{colortbl}
\usepackage{cleveref}

\usepackage[backend=bibtex, style=numeric, maxnames=100]{biblatex}
\DeclareNameFormat{highlightme}{%
	\nameparts{#1}%
	\def\formattedgiven{\namepartgiven}
	\def\formattedfamily{\namepartfamily}
	\ifthenelse{\equal{\namepartfamily}{Sch\"on}}{%
		\def\formattedgiven{\mkbibbold{\namepartgiven}}
		\def\formattedfamily{\mkbibbold{\namepartfamily}}
	}{}%
	\ifthenelse{\equal{\namepartfamily}{Sch\"on*}}{%
		\def\formattedgiven{\mkbibbold{\namepartgiven}}
		\def\formattedfamily{\mkbibbold{\namepartfamily}}
	}{}%
	\usebibmacro{name:given-family}%
	{\formattedfamily}
	{\formattedgiven}
	{\namepartprefix}
	{\namepartsuffix}%
	\usebibmacro{name:andothers}%
}
\DeclareFieldFormat{myaward}{\bibcplstring{award}\addcolon\space#1}

\DeclareCiteCommand{\mycite}
{\usebibmacro{prenote}}
{%
	\begingroup
	\begin{addmargin}[0pt]{0pt}
		\defcounter{maxnames}{99}
		\mkbibbold{\printfield[noformat]{title}}
		\cite{\thefield{entrykey}},
		\printnames[highlightme]{author},
		\printfield{booktitle},
		\printfield{year}.
		\printfield{award}
	\end{addmargin}
	\endgroup
}
{\multicitedelim}
{\usebibmacro{postnote}}

\usetikzlibrary{shapes,arrows,calc, fit, }

\definecolor{grey60} {RGB} {102, 102, 102} %

\graphicspath{{fig/}}

\definecolor{gray}{rgb}{0.4,0.4,0.4}
\definecolor{lightgray}{rgb}{0.89,0.89,0.89}
\definecolor{lightblue}{rgb}{0.75,0.90,0.957}
\definecolor{lightred}{rgb}{0.95,0.80,0.75}
\definecolor{lightyellow}{rgb}{0.95,0.9,0.8}
\definecolor{lightgreen}{rgb}{0.6,0.95,0.7}
\definecolor{richgreen}{rgb}{0.1,0.6,0.1}
\definecolor{grass}{rgb}{0.2,0.8,0.2}
\definecolor{empty}{rgb}{0.9,0.9,0.0}
\definecolor{stone}{rgb}{0.5,0.5,0.5}

\usepackage{xspace}

\newcommandx{\todored}[2][1=]{\todo[linecolor=red,backgroundcolor=red!25,bordercolor=red,#1]{#2}}
\newcommandx{\todoblue}[2][1=]{\todo[linecolor=blue,backgroundcolor=blue!25,bordercolor=blue,#1]{#2}}
\newcommandx{\todogreen}[2][1=]{\todo[linecolor=OliveGreen,backgroundcolor=OliveGreen!25,bordercolor=OliveGreen,#1]{#2}}
\newcommandx{\todoviolet}[2][1=]{\todo[linecolor=Plum,backgroundcolor=Plum!25,bordercolor=Plum,#1]{#2}}

\newcommand{\cmark}{\ding{51}}%

\newacronym[description={Standardized code generated by a vehicle's electronic control unit to indicate a precise event within the vehicle (malfunctions, software update, \(\cdots\))}]{dtc}{DTC}{Diagnostic Trouble Code}

\newacronym[description={A higher-level abstraction of DTC sequences defined by domain experts to characterize specific vehicle faults using Boolean rules, example shown in Fig.~\ref{fig:illustration}}]{ep}{EP}{Error Pattern}

\newacronym[description={Embedded system controlling one or more electrical systems or subsystems in a vehicle}]{ecu}{ECU}{Electronic Control Unit}

\newacronym[description={A probabilistic model for representing language on computers, e.g., a model that assigns probabilities to words of sentences, is called an LM}]{lm}{LM}{Language Model}

\newacronym[description={A special attention layer used in the Transformer architecture}]{mha}{MHA}{Multi-Head Attention}

\newacronym[description={A specialized version of an RNN cell that allows for generating language recurrently}]{lstm}{LSTM}{Long Short-Term Memory}

\newacronym[description={An artificial signal mostly consisting of sine and cosine waves that encode indices within a sequence}]{pe}{PE}{Positional Encoding}

\newacronym[description={An information-theoretic measure of dependency between random variables given a conditioning set}]{cmi}{CMI}{Conditional Mutual Information}

\newacronym[description={A measure of difference between two probability distributions}]{kl}{KL}{Kullback-Leibler}

\newacronym[description={A family of sampling-based inference methods for dynamic systems}]{smc}{SMC}{Sequential Monte Carlo}

\newacronym[description={A method for approximating probability distributions using weighted samples}]{sis}{SIS}{Sequential Importance Sampling}

\newacronym[description={A class of neural networks designed for sequence modeling}]{rnn}{RNN}{Recurrent Neural Network}

\newglossaryentry{transformer}
{
    name=Transformer,
    description={Neural network architecture based on self-attention mechanisms, widely used for sequence modeling}
}

\newglossaryentry{bicarformer}
{
    name=BiCarFormer,
    description={A bidirectional multimodal Transformer model for vehicle diagnostics}
}

\newglossaryentry{epredictor}
{
    name=EPredictor,
    description={An autoregressive Transformer decoder model for error pattern prediction in vehicles}
}

\newglossaryentry{oscar}
{
    name=OSCAR,
    description={\uline{O}ne-\uline{S}hot \uline{C}ausal \uline{A}uto\uline{R}egressive Discovery is a causal discovery method described in Chapter~\ref{c7:multi_label_one_shot_causal_discovery} that infers a directed acyclic graph representing events-to-outcome relationships from a single sequence of events leading to outcomes}
}

\newglossaryentry{trace}
{
    name=TRACE,
    description={\uline{T}emporal \uline{R}econstruction via \uline{A}utoregressive \uline{C}ausal \uline{E}stimation is a causal discovery method described in Chapter~\ref{c7:event_to_event} that infers causal graph describing event-to-event relationships from a single sequence of discrete events}
}

\newacronym[description={A field of machine learning that learns cause-and-effect relationships in data to improve robustness of machine learning models (intervention, counterfactual, causal graphs)}]{cml}{CML}{Causal Machine Learning}

\newglossaryentry{carep}
{
name=CAREP,
description={\uline{C}ausal \uline{A}utomated \uline{R}easoning for \uline{E}rror \uline{P}attern, a multi-agent system for automated error rule generation}
}
\newacronym[description={A Transformer model with billions of parameters, typically trained via self-supervised learning objectives on large-scale text corpora}]{llm}{LLM}{Large Language Model}
\newglossaryentry{cargo}
{
    name=CARGO,
    description={\uline{C}ausal \uline{A}ggregation via \uline{R}obust \uline{G}raph \uline{O}perations, a scalable multi-label causal discovery framework designed to construct global causal structures from a batch of high-cardinality event sequences}
}

\newacronym[description={A stochastic process used to model sequences of discrete events over continuous time}]{tpp}{TPP}{Temporal Point Process}

\newglossaryentry{hawkes}
{
    name=Hawkes Process,
    description={A self-exciting point process where the occurrence of an event increases the likelihood of future events in the short term}
}

\newacronym[description={The study of computational techniques for processing human language}]{nlp}{NLP}{Natural Language Processing}

\newacronym[description={Neural network architectures comprising multiple layers of parameterized transformations, enabling the hierarchical learning of representations from raw data}]{dnn}{DNN}{Deep Neural Network}

\newacronym[description={A technique that improves the accuracy of LLM answers by allowing the model to access and cite external data sources before generating a response}]{rag}{RAG}{Retrieval-Augmented Generation}

\newacronym[description={An autoregressive model that factorizes the joint distribution over a sequence into a product of conditionals, each parameterized by a neural network}]{nade}{NADE}{Neural Autoregressive Density Estimator}

\begin{document}
\frontmatter
\maketitle

\chapter{Abstract}
\label{c0:abstract}
\glsresetall
\Glspl{ecu} embedded within modern vehicles generate a large number of asynchronous events known as \glspl{dtc}. These discrete events form complex temporal sequences that reflect the evolving health of the vehicle’s subsystems. In the automotive industry, domain experts manually group these codes into higher-level error patterns (EPs) using Boolean rules to characterize system faults and ensure safety. However, as vehicle complexity grows, this manual process becomes increasingly costly, error-prone, and difficult to scale. Consequently, manufacturers urgently require automated solutions capable of anticipating malfunctions, reducing downtime, and lowering warranty and repair costs. Notably, the number of unique DTCs in a modern vehicle is on the same order of magnitude as the vocabulary of a natural language, often numbering in the tens of thousands. This observation motivates a paradigm shift: treating diagnostic sequences as a language that can be modeled, predicted, and ultimately explained. Traditional statistical approaches fail to capture the rich dependencies and do not scale to high-cardinality event data characterized by thousands of event types, large sample sizes, and long sequence lengths. Specifically, the high cardinality of categorical event spaces in industrial logs poses a significant challenge, necessitating new machine learning architectures tailored to such event-driven systems. This thesis addresses automated fault diagnostics by unifying event sequence modeling, causal discovery, and \glspl{llm} into a coherent framework for high-cardinality event streams. It is structured in three parts, reflecting a progressive transition from prediction to causal understanding and finally to reasoning for vehicle diagnostics. Part~I introduces Transformer-based architectures for predictive maintenance from DTC sequences.
Part~II develops a suite of three scalable causal discovery frameworks that leverage pretrained autoregressive models as neural density estimators. OSCAR and CARGO recover the direct causes of each outcome label at the sample- and population-level respectively, while TRACE addresses a complementary problem: recovering the event-to-event causal graph from a single sequence.
Finally, Part~III integrates these causal structures with large language models in a multi-agent system (\gls{carep}) that automates the synthesis of Boolean EP rules. Together, these contributions establish a coherent, scalable pipeline from raw event streams to human-readable diagnostic knowledge.

\chapter{Acknowledgments}
First and foremost, I would like to express my deepest gratitude to my supervisor, Prof. Rainer Lienhart, for his openness to ideas, constant investment throughout my PhD, and scientific rigor. His support and vision were essential in shaping both this thesis and my growth as a researcher. I am grateful for the opportunity to have conducted this research under his guidance.
I also extend my sincere thanks to Prof. Elisabeth Andre for her time and effort in reviewing this thesis.
I am also sincerely thankful to my PhD colleagues, Julian Lorenz, Daniel Kienzle,
Robin Schön and Mrunmai Phatak, for their help, support, and camaraderie. They made this journey not only productive but also enjoyable. The countless discussions, collaborations, and shared moments have been a source of motivation and inspiration.
I warmly thank my BMW supervisors, Stefan Oelsner for his deep expertise, creativity, and knowledge, and Stefan Mueller for his rigorous guidance, constructive feedback, and support in every phase of the Promotion program. I also thank Stefan Kirsch for his interest, support and the opportunity to connect academic research with real-world industrial challenges at BMW. I am thankful to Michael Decker and Ingo Stock for their support and time reviewing this thesis. Their insights and the collaboration between BMW and the University of Augsburg greatly enriched this work.
Additionally, I would like to acknowledge all my BMW colleagues, particularly Bernd Postmaier for his orientation and advice, Stefan Benesch for operational guidance, and Lisa Jobst for her positive spirit in the office.
Finally, I am deeply grateful to my family, especially my parents, Bertrand and Audrey Math, for their constant encouragement and unwavering support. They greatly helped me stay motivated and keep moving forward. I hope I can make them as proud as I am to have them as parents. I also want to thank my sister, Laura, for her presence, and my two grandfathers, Henri and Jean, whose memory continues to inspire me. And of course, my close friends, particularly Nathan Haudot, for his encouragement and companionship throughout this journey.
\label{c0:acknowlegments}

\tableofcontents

\renewcommand{\glossarysection}[2][]{} %

\chapter*{Glossary}
\printglossary[title={List of Abbreviations}]
\printglossary[type=\acronymtype, title={List of Symbols}] %
\chapter*{Notations}
We summarize the notation used throughout this thesis here. We followed the standard notation from \textit{Deep Learning} \cite{goodfellow2016deep} available at \url{https://github.com/goodfeli/dlbook_notation/}

\section*{Numbers and Arrays}
\begin{tabularx}{\textwidth}{lX}
$a$ & A scalar (integer or real) \\
$\boldsymbol{a}$ & A vector \\
$\boldsymbol{A}$ & A matrix \\
$\mathcal{A}$ & A tensor \\
\end{tabularx}

\section*{Sets and Graphs}
\begin{tabularx}{\textwidth}{lX}
$\mathbb{A}$ & A set \\
$\mathcal{X}$ & An alphabet \\
$\mathbb{R}$ & The set of real numbers \\
$\{0, 1\}$ & The set containing 0 and 1 \\
$\{0, 1, \dots, n\}$ & The set of all integers between $0$ and $n$ \\
$[a, b]$ & The real interval including $a$ and $b$ \\
$(a, b]$ & The real interval excluding $a$ but including $b$ \\
$\mathbb{A} \backslash \mathbb{B}$ & Set subtraction, i.e., the set containing the elements of $\mathbb{A}$ that are not in $\mathbb{B}$ \\
$\mathcal{G}$ & A graph \\
$\mathcal{D}$& A dataset \\
$\textit{Pa}_{\mathcal{G}}(X_i)$ & The parents of $X_i$ in $\mathcal{G}$ \\
$\mathcal{MB}(Y_j)$ & The Markov Boundary of the label $Y_j$; $\widehat{\mathcal{MB}}(Y_j)$ its estimate \\
\end{tabularx}

\section*{Probability and Information Theory}
\begin{tabularx}{\textwidth}{lX}
$X$ & A discrete random variable with alphabet \(\mathcal{X}\)\\
$X_t$ & A random variable referred to as an event occurring at position (time step) \(t\) in a sequence (Chapters~\ref{c7:multi_label_one_shot_causal_discovery} and~\ref{c7:event_to_event}); in the population-level Chapter~\ref{c7:multi_label_causal_discovery}, \(X_i\) instead denotes the event type \(i\) \\
$x_t$ & A realization of a random variable at a time step \(t\) \\
$Y$ & A discrete random variable characterizing an outcome or a label (e.g., an error pattern) drawn from \(\mathcal{Y}\)\\
$\mathbf{X}= \{X_1, \cdots, X_n\}$& A set of discrete random variables \\
$P(X)$ & A probability distribution over a discrete random variable \(X\)\\
$P_\theta(X)$ & A probability distribution parametrized by \(\theta\), e.g., a pretrained model\\
$\hat{P}(X)$ & A probability distribution over a discrete random variable \(X\) estimated empirically based on a data sample\\
$p(x)$ & The probability mass function for the random variable, \(X\) i.e., \(p(x) = P(X=x), x\in \mathcal{X}\). We denote \(p(x)\) instead of \(p_X(x)\) for convenience\\
$X \sim P$ & Random variable $X$ has distribution $P$ \\
$\mathbb{E}_{p(x)} [f(X)]$ or $\mathbb{E} f(X)$ & Expectation of $f(X)$ with respect to $P(X)$ \\
$\text{Var}(f(X))$ & Variance of $f(X)$ under $P(X)$ \\
$H(X)$ & Entropy of the random variable $X$ \\
$D_\text{KL}(P \Vert Q)$ & Kullback-Leibler divergence of $P$ and $Q$ \\
$\mathcal{N}(\boldsymbol{x}; \boldsymbol{\mu}, \boldsymbol{\Sigma})$ & Gaussian distribution over $\boldsymbol{x}$ with mean $\boldsymbol{\mu}$ and covariance $\boldsymbol{\Sigma}$ \\
\end{tabularx}

\section*{Functions}
\begin{tabularx}{\textwidth}{lX}
$f: \mathcal{A} \rightarrow \mathcal{B}$ & The function $f$ with domain $\mathcal{A}$ and range $\mathcal{B}$ \\
$\log{x}, \ln{x}$ & Natural logarithm of $x$, we follow the convention that \(\ln{x} = \log{x}\) \\
$\sigma(x)$ & Logistic sigmoid, $\frac{1}{1 + \exp(-x)}$ \\
\end{tabularx}

\mainmatter

\chapter{Introduction\label{ch:intro}}
\glsresetall

The increasing digitization of modern vehicles has transformed automotive systems into complex, interconnected networks of \glspl{ecu} continuously generating streams of Diagnostic Trouble Codes (DTCs). Each vehicle now asynchronously produces thousands of discrete events that document the behavior and health of its subsystems. These data consist of vast collections of sequences, each containing hundreds of events drawn from a large vocabulary of unique codes, effectively forming high-cardinality datasets. For decades, automotive manufacturers have relied on domain experts to manually analyze these codes and define higher-level Boolean rules called \glspl{ep} to characterize system faults and ensure safety. However, with the increasing complexity of vehicles and the proliferation of embedded software, this manual process has become prohibitively expensive, time-consuming, and error-prone.
This challenge illustrates a broader industrial problem: how can we automatically learn, understand, and reason about event-driven systems? %
The answer is not merely one of prediction, but of understanding why events occur and how they interact with each other, thereby transforming data-driven systems into explainable and autonomous diagnostic agents.

Over the past decade, the rise of \glspl{dnn} has profoundly reshaped the landscape of artificial intelligence. From image recognition to speech understanding, their ability to model complex, nonlinear relationships has led to significant progress across multiple fields. Among these advances, Transformer architectures have become the cornerstone of modern deep learning, spanning vision, audio, and sequence modeling. They underpin the development of \glspl{llm} that excel not only at compressing vast textual corpora but also at exhibiting emergent induction mechanisms. Beyond a certain scale of parameters, training data, and with appropriate fine-tuning, LLMs appear to exhibit remarkable reasoning capabilities. Additionally, language models embody a general principle: sequences, whether words, medical signals, or machine events, encode meaning through their temporal structure. This observation opens a new research frontier, extending the power of language modeling to machine-generated data in industrial settings. In such domains, event logs, sensor readings, medical test results, or diagnostic codes constitute a machine-generated “language”. By interpreting this language, deep learning models can uncover hidden patterns of operation and anticipate failures before they occur. The same architectures that have fundamentally changed natural language processing now hold the potential to transform how industries understand, monitor, and reason about the behavior of complex machines.

\section{From Industrial Challenge to Scientific Question}
The automotive industry presents one of the most demanding real-world contexts for developing such intelligent systems. Vehicles must be safe, reliable, and explainable, yet their internal states are observed indirectly through asynchronous, heterogeneous, and noisy event logs. Traditional statistical and rule-based approaches struggle to capture  the rich dependencies underlying these sequences; they remain applicable only when the event vocabulary is small and sequences are short. Additionally, diagnostic data exhibits all the hallmarks of complex event-driven domains: extremely high cardinality of event types, sparse temporal observations, and a high degree of redundancy. It is essential to develop techniques that not only anticipate such events but also explain their underlying causal mechanisms.
These characteristics make automotive diagnostics an ideal testbed for advancing machine learning at scale, particularly in event-sequence modeling, multimodal learning, causal discovery, and multi-agent systems. What begins as a challenge in fault detection thus becomes a general scientific endeavor: building models that not only predict, but also explain and reason over sequences of events.

\section{Research Areas, Challenges and Open Problems}
This thesis builds upon the convergence of three complementary research areas:
\paragraph{Sequence Modeling}
Modern language models have demonstrated remarkable ability to capture long-range dependencies and contextual relations in structured sequences. Beyond their success in natural language, they establish a general and practical paradigm. A model is first trained on a large-scale, self-supervised task such as next-token prediction, where the training signal comes from the data itself without manual labels. It is then adapted to downstream applications where labeled data are scarce. %

In contrast, machine-generated data in complex systems such as vehicles, aircraft, hospitals, or computer networks is inherently noisy, repetitive, and scattered across time. Vehicle diagnostic logs, for instance, comprise tens of thousands of distinct DTCs, each irregularly triggered by sensor anomalies, environmental fluctuations, or transient software behavior. Similar challenges arise in other domains: sequences of electrocardiogram spikes or clinical events in healthcare, network intrusions and security alerts in cybersecurity, flight logs in aeronautics, and transactional traces in finance—all of which consist of heterogeneous event streams. These data are often non-stationary, dominated by repetitive or redundant events, and corrupted by noise or incomplete information, which makes it difficult for classical methods to extract stable temporal dependencies or causal relations.

Modeling such event sequences with modern sequence models, such as Transformers~\cite{tf}, offers an alternative. Their self-attention mechanism enables the capture of long-range dependencies, modeling of multiscale temporal dynamics, and integration of heterogeneous modalities, such as sensor readings, timestamps, and textual metadata, through fusion mechanisms. Despite these advantages, the use of autoregressive and bidirectional Transformers for machine-generated event data, particularly in automotive diagnostics, remains underexplored. In addition, the applicability of Transformers to prognosis and long-horizon forecasting in event-driven systems remains poorly studied.

\paragraph{Causal Discovery in Event Sequences}
Predictive accuracy alone is insufficient in safety-critical domains. To truly anticipate failures, prevent accidents, or mitigate risks, one must understand \textit{why} events occur and how they influence one another. This is the central goal of causal discovery: recovering, from observational data, the underlying mechanisms that generate the observed phenomena. In practice, this involves learning a compact Directed Acyclic Graph (DAG), i.e., a graph of directed edges that contains no cycle, that captures direct causes, mediating effects, and potential confounders influencing an outcome such as a vehicle defect or a disease. Once established, this DAG can be used to perform causal inference, i.e., to answer counterfactual questions such as 'what if' or to measure the average causal effect of a variable on the outcome.

In event-driven domains, causal discovery becomes particularly intricate. Unlike static cross-sectional data or regularly sampled time series, event sequences are asynchronous, irregularly spaced, and highly context-sensitive. In these settings, the causal effect of an event is not fixed; it may vary significantly depending on its precise timing, its position within a sequence, and its co-occurrence with other markers. In general, two main perspectives emerge. The first, events-to-outcome causal discovery (or multi-label causal discovery), seeks to identify which events or combinations thereof cause a higher-level outcome, such as a disease from symptoms, a security breach from system alerts, or an EP from DTCs. The second, event-to-event causal discovery, focuses on uncovering causal influences between events themselves, forming the basis for understanding asynchronous system behavior and performing root-cause analysis.

Traditional causal discovery methods are ill-suited to this setting. They include constraint-based algorithms, which build the graph from conditional independence tests (PC, after its authors Peter Spirtes and Clark Glymour~\cite{constrainct_based_cd}), score-based optimization, which searches for the graph that maximizes a goodness-of-fit score (Greedy Equivalence Search, GES~\cite{ges}), and functional causal models, which assume a specific form of the causal mechanisms (Linear Non-Gaussian Acyclic Model, LiNGAM~\cite{lingam}). They typically assume a small number of variables. When applied to high-cardinality sequences with thousands of discrete event types, long temporal dependencies, and intense noise, these methods quickly become intractable due to their computational complexity and statistical assumptions.
Even recent approaches for time series and event sequences (THP~\cite{transformerhawkeprocess}, SHTP~\cite{shtp}, CAUSE~\cite{cause}, CASCADE~\cite{cueppers2024causal}) are applied to dozens or, at most, a hundred variables on short sequences which is not applicable in many scenarios. In practice, they fail to scale, restricting their applicability in industrial contexts where datasets are both massive and heterogeneous.
As a result, the field of causal discovery in large-scale event sequences remains largely underexplored. Practical adoption is hindered by three central challenges: (1) the combinatorial explosion of possible event interactions, (2) the lack of scalable, GPU-compatible algorithms, and (3) the absence of production-ready methods that reuse pre-existing architectures for inference, like Transformers or other autoregressive models. As a consequence, practitioners are seeking automated solutions to identify causal relationships in high-cardinality sequence datasets in a reasonable amount of time, leveraging existing pretrained sequence models. In the automotive domain, each EP is characterized by an associated \emph{EP rule}: a deterministic Boolean expression over \glspl{dtc} that encodes the causal relationships giving rise to it (see Section~\ref{sec:ep_def} for the precise distinction between an EP and its rule). By reinterpreting these rules as a causal discovery problem, it becomes possible to identify potential causes as DTCs for unknown EPs and hence create new rules based on the DTCs as causes of an EP.

\paragraph{Automated and Explainable Reasoning Systems}
\noindent Beyond discovering causal relations, a further goal in industrial AI is to enable autonomous reasoning over complex automation tasks. Such a system can not only infer causal structures but also explain, justify, and act upon them to automate manual processes. While causal discovery provides the skeleton of knowledge, it often remains abstract, numerical, or incomplete. Such causal graphs must be translated into interpretable reasoning chains that domain experts can trust and validate. This requires bridging two complementary worlds: data-driven inference from causal discovery algorithms and domain knowledge.
Recent advances in LLMs have opened new possibilities for reasoning, automation, and the integration of external knowledge to form intelligent systems. These models can process both structured and unstructured information, including textual descriptions, metadata, documentation, and relational rules, either directly in the prompt (in-context learning) or via \gls{rag}, making them ideal candidates for reasoning over causal evidence. Yet, using LLMs alone remains insufficient: without grounding in causal structure or domain constraints, they may hallucinate or reason inconsistently. Conversely, causal discovery methods alone provide semantic knowledge but lack the adaptability of LLMs.%

This gap motivates a hybrid paradigm that combines the structured and principled outputs of causal discovery algorithms with the natural-language reasoning capabilities of LLMs, often referred to as \gls{cml}. By organizing these components into multi-agent systems, each agent can specialize in one task. One inferring causal dependencies, another contextualizes them using metadata and textual descriptions. The result is a collaborative reasoning process where causal evidence is grounded, verified, and articulated in human-readable form. In the automotive industry, they could automatically generate interpretable Boolean rules that describe how diagnostic events combine to form EPs. These rules, which currently require extensive manual expert labor, could then be fully automated. More broadly, multi-agent systems powered by causal discovery could be extended to any domain where reducing \gls{llm}s' hallucinations and increasing interpretability are essential, from medical diagnostics and industrial health monitoring to cybersecurity incident analysis.

\section{Contributions}
The contributions of this thesis can be summarized as follows.

\subsection{Predictive Sequence Modeling of Vehicle Diagnostics}
\newsavebox{\carFigure}
\savebox{\carFigure}{%
  \includegraphics[width=0.7\columnwidth]{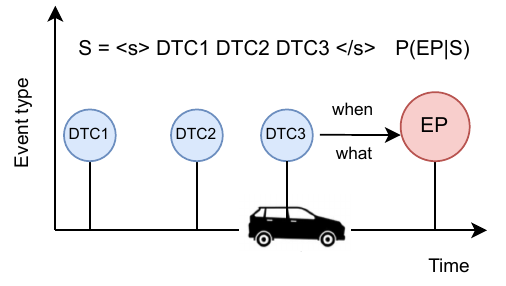}%
}

\noindent In the first contribution of this thesis, we explore how modern sequence models can anticipate system failures in vehicles before they occur \cite{math2024harnessingeventsensorydata}. %
\begin{figure}[!h]
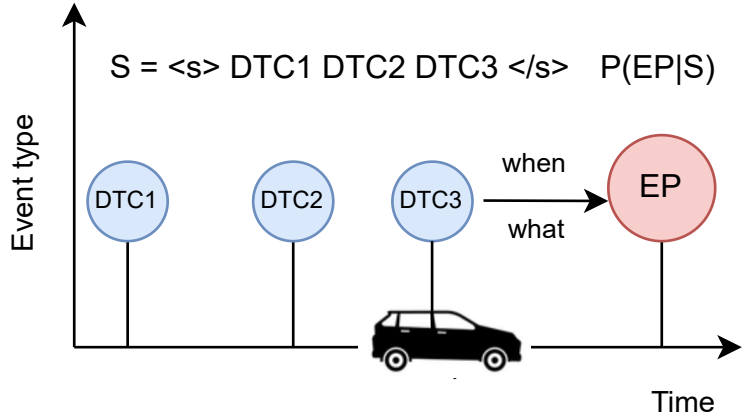

  \centering
  \usebox{\carFigure}
  \caption{\textbf{Error Pattern Prediction (when and what)}. Based on the past sequence $S$ of diagnostic trouble codes (DTCs), manufacturers want to prevent error patterns (EPs) from happening by predicting their likelihood and time of occurrence.}
  \label{fig:car_}
\end{figure}
\newline

\paragraph{Autoregressive Transformers for Vehicle Event Sequences}
We draw an analogy between natural language and vehicular diagnostic data: just as words that form structured sentences convey meaning, DTCs form spatio-temporal sequences encoding the vehicle’s state at each mileage and time.\footnote{We use the term \emph{spatio-temporal} loosely throughout this thesis: since a vehicle's mileage increases monotonically with usage in a manner analogous to a spatial coordinate, we treat the pair (time, mileage) associated with each DTC as its spatio-temporal context, rather than a literal geographic or physical spatial coordinate.} This analogy motivates the use of Transformer-based language models for automotive fault prediction. To this end, we introduce two autoregressive Transformer architectures \cite{tf}:

\begin{itemize}
    \item \textbf{CarFormer}, designed to model the generative process of DTC sequences by predicting the following DTC and its time of occurrence.%
    \item \textbf{EPredictor}, which extends this formulation to EPs by jointly predicting \emph{what} EP will occur and \emph{when}, given a trajectory of observed DTCs.
\end{itemize}
Together, these models learn to anticipate vehicle faults in an autoregressive manner, providing both event type and time of occurrence. Despite the complex data distribution of a stochastic process such as the DTCs and EPs appearance, their high event-type cardinality (\(>10^4\)), irregular sampling, and unbalanced label distributions, they demonstrate robust performance across multiple metrics. They outperform a standard autoregressive Transformer baseline trained without our EP-specific prediction head (see Section~\ref{c2:ep_prediction_based_on_live_data} for the exact baseline definition). In particular, they could anticipate a given EP appearance with 80\% F1-score within an interval of \(58.4 \pm 13.2\)h when predicting the time of occurrence.
\paragraph{Predictive Maintenance Metric for Autoregressive Models}
We introduce the Confident Predictive Maintenance Window (CPMW), a novel evaluation protocol that measures how early in an observed  sequence of events (DTCs) a model can issue a confident outcomes (EPs) prediction. Unlike accuracy-at-end-of-sequence metrics, CPMW rewards early, high-confidence predictions, directly reflecting the operational value of predictive maintenance.

\newsavebox{\carFigurePlus}
\savebox{\carFigurePlus}{%
  \includegraphics[width=0.74\columnwidth]{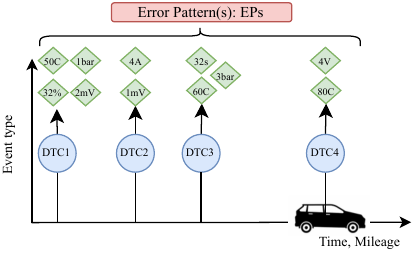}%
}
\begin{figure}[!t]
  \centering
  \usebox{\carFigurePlus}
  \caption{\textbf{Error Pattern Prediction using Multimodal Sequences}. Past diagnostic trouble codes only provide limited information about complex and overlapping error patterns. In addition, domain experts rely on environmental conditions (e.g., temperature, voltage, humidity, \dots) to identify more accurately error patterns.}
  \label{fig:dtc_env_car}
\end{figure}

\paragraph{Offline Error Pattern Prediction using Multimodal Learning}
While modern diagnostic systems primarily rely on sequences of vehicular DTCs, they overlook valuable contextual information such as raw sensory data (e.g., temperature, humidity, and pressure) known as environmental conditions. Experts often use such sensory information to separate overlapping EPs that might share the same DTCs in their Boolean rule definition. Fig.~\ref{fig:dtc_env_car} illustrates this setting: each DTC recorded over time and mileage comes with several environmental conditions measured at the same moment (e.g., \(50\)°C, \(1\)bar, \(4\)A), and the EPs must be predicted from both the DTCs and these conditions. To this end, we introduce \textbf{BiCarFormer}, a bidirectional multimodal Transformer that integrates environmental sensor data with DTC sequences. Through special fusion mechanisms, BiCarFormer learns interdependencies between discrete diagnostic events and environmental conditions, improving predictive performance compared to standalone bidirectional Transformers (+\(9\%\) precision on average) and enabling more interpretable predictions through attention scores interpretation.

\subsection{Multi-Label Causal Discovery in High-Cardinality Event Sequences}
Part~\ref{pa:p2} turns from prediction to explanation. We introduce two causal discovery frameworks for discrete sequences, targeting respectively \textit{sample}- and \textit{population-level} causal relationships, that is, discovered from one sequence or from a multitude.

\paragraph{Sample-Level}
To address this challenge, we introduce \gls{oscar} (\uline{O}ne-\uline{S}hot \uline{C}ausal  \\ \uline{A}uto\uline{R}egressive discovery). OSCAR identifies, for each outcome label, the minimal set of causal events (the so-called Markov Boundary\footnote{Formally defined in Definition~\ref{def:markov_boundary}, Chapter~\ref{c3:foundation_causal_discovery}.}) directly from a single observed sequence, as formalized in Table~\ref{tab:causal_framework}. We refer to this regime as sample-level causal discovery throughout this thesis to emphasize the key property: the causal graph is inferred from and specific to one individual sequence, without requiring a population of observations\footnote{This terminology supersedes the label 'one-shot' used in the associated conference publications~\cite{math2025oneshot, math2025towards}, where 'one-shot' was used informally to mean 'from a single sequence'.}.
OSCAR reuses the pretrained autoregressive Transformers from Part~I as \glspl{nade} to estimate the conditional mutual information (CMI) between events and labels given the past events. They quantify the extent to which each event contributes to a future outcome. This formulation enables efficient, GPU-parallelized computation of causal dependencies.
Theoretically grounded under standard assumptions, we show that OSCAR recovers the Markov Boundaries of each label within an associated Bayesian Network~\cite{koller2009probabilistic}. In practice, it also introduces causal indicators that distinguish excitatory and inhibitory effects between events, offering a more nuanced view than deterministic causal graphs~\cite{quantifyingcausalinfluence}.

\begin{figure}[!h]
    \centering
    \includegraphics[width=0.95\linewidth]{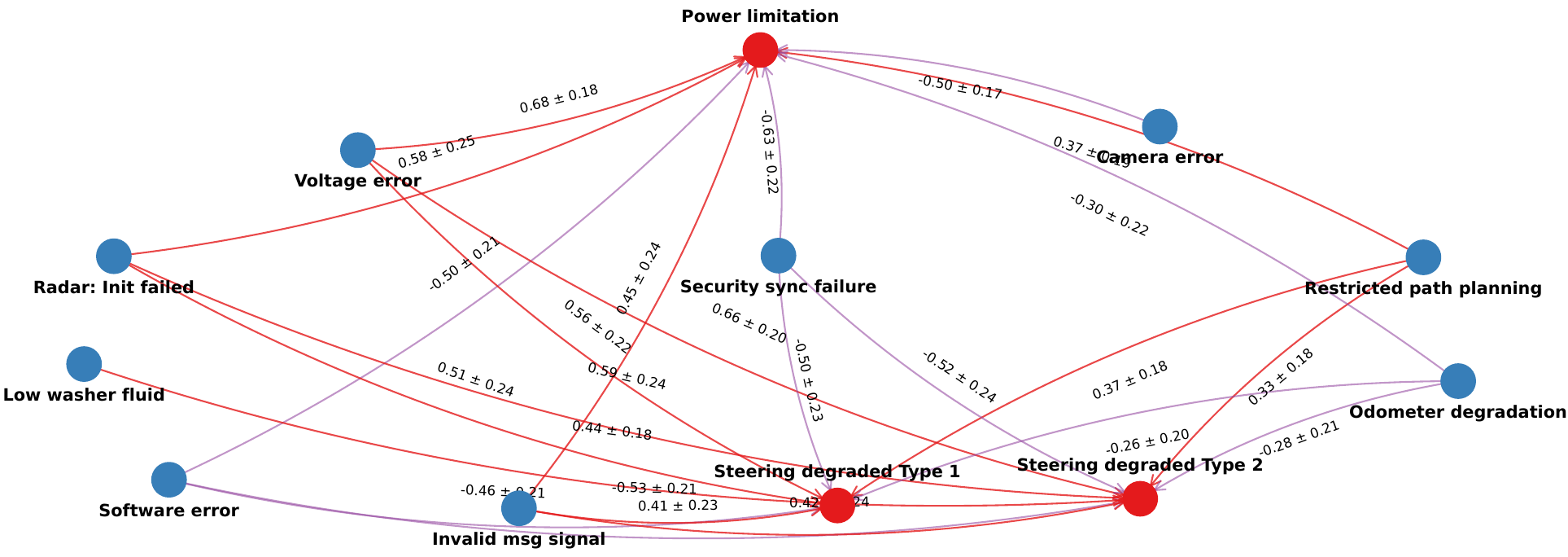}
\caption{\textbf{Anonymized Causal Graph.} Example of a sequence of events (\textcolor{blue}{DTCs}) that lead to a steering wheel degradation and a power limitation as outcome \textcolor{red}{labels}. The causal indicators are shown in \textcolor{violet}{violet} if inhibitory and \textcolor{orange}{orange} or \textcolor{red}{red} if excitatory, depending on the magnitude. The voltage error DTC is the biggest causal driver for the power limitation EP while the security sync failure seems to have the biggest inhibitory effect on all EPs.}
\label{fig:graph_steering_wheel_degradation}
\end{figure}

We validate OSCAR on a large-scale vehicular dataset comprising 29,100 DTCs and 474 Error Patterns (EPs), a regime where classical constraint-based causal discovery methods, which fail to scale beyond a few hundred variables, are inapplicable. OSCAR identifies meaningful causal relations between DTCs and EPs, revealing interpretable causal structures for phenomena such as steering wheel degradation or battery power limitation (Fig.~\ref{fig:graph_steering_wheel_degradation}).

\paragraph{Population-level}
\noindent While OSCAR performs causal discovery at the level of individual sequences (sample-level), real-world systems demand a broader perspective: understanding global causal relationships across the entire observational dataset (population-level).
To address this, we introduce \gls{cargo} (\uline{C}ausal \uline{A}ggregation via \uline{R}obust \uline{G}raph \uline{O}perations), a scalable framework that merges sample-level causal graphs into a global representation. CARGO treats causal discovery as a two-phase process. In the first phase, OSCAR extracts sample-level causal graphs. In the second phase, CARGO aggregates these causal graphs based on the frequencies of edge appearances across labels using a novel adaptive frequency-based fusion strategy. Motivated by long-tail distributions in labeled data, the edges are retained based on their frequency of appearance and the number of available samples per label.
This aggregation process enables the recovery of global Markov Boundaries for each label while filtering out spurious relations caused by noise, sparsity, or rare events. %
Empirical evaluation on 300,000 sequences with 29,100 event types shows that CARGO scales where classical algorithms fail, while also improving accuracy and recall over OSCAR alone; ablations confirm the robustness of the aggregation phase across multiple thresholding and scoring strategies.

\subsection{Scalable Sample-Level Event-to-Event Causal Discovery}
Multi-label causal discovery methods are limited to local structure learning, ignoring the dense information of inter-event dependencies. To fully diagnose complex phenomena like cascading vehicle failures, we must recover the complete causal graph. This presents a significantly more challenging scenario due to the combinatorial explosion of possible edge connections, as opposed to simple events-to-outcome relationships.
\begin{figure}[!t]
    \centering
    \includegraphics[width=0.7\linewidth, trim=0 0 0 28pt, clip]{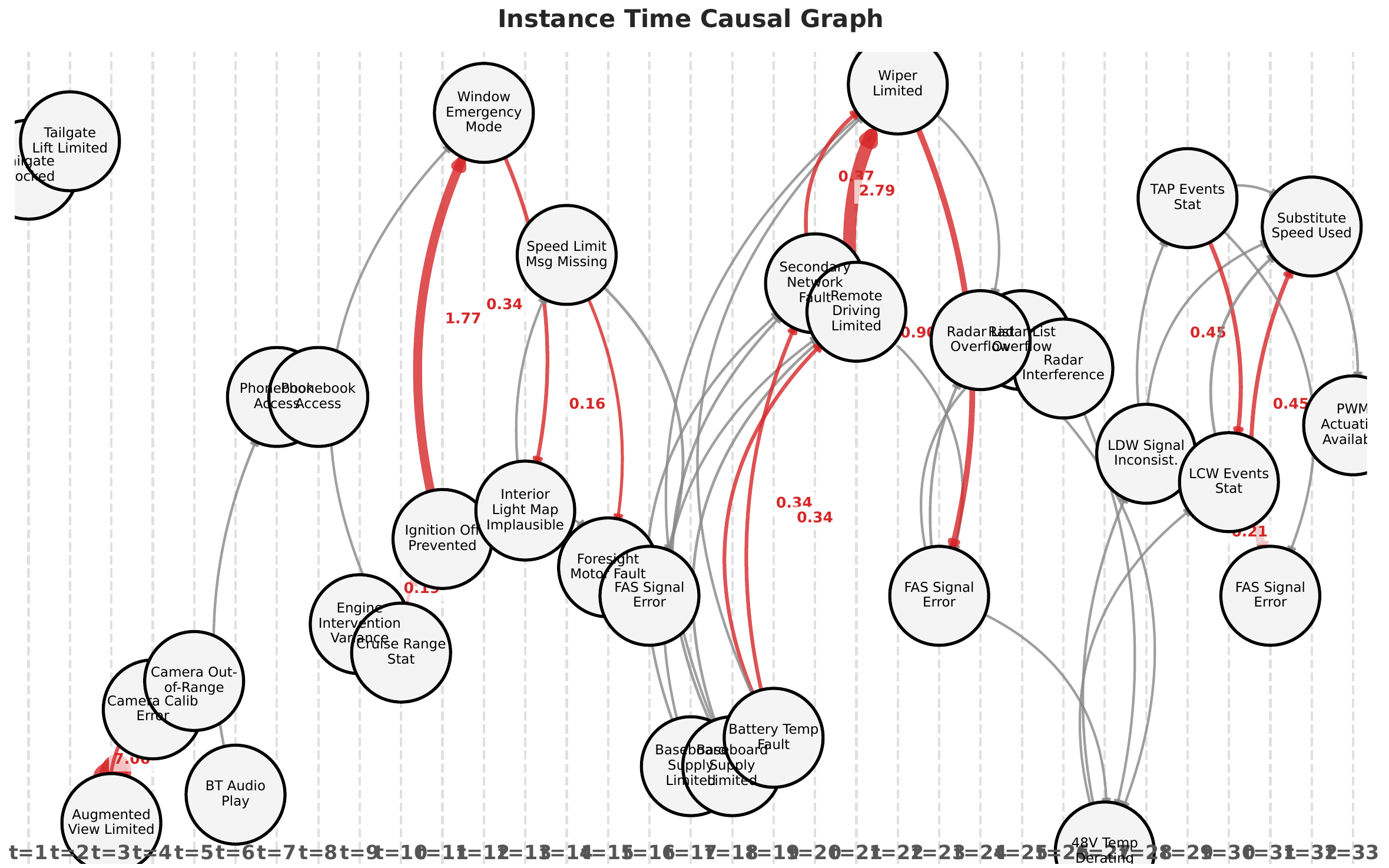}
  \caption{\textbf{Anonymized Sample Time Causal Graph for Diagnostic Defect Cascade.} Temporal evolution of a diagnostic defect cascade in a vehicle ($|\mathcal{X}| \approx 29,100$). TRACE effectively captures causal relationships, revealing distinct \textbf{error clusters} at different time steps (e.g., initial sensor failures at $t=3$ triggering mechanical faults at $t=12$, battery at issue \(t=17\)). This enables actionable root-cause analysis by isolating the specific onset of a failure mechanism and its strength using the conditional mutual information.}%
\end{figure}
We therefore introduce \gls{trace} (\uline{T}emporal \uline{R}econstruction via \uline{A}utoregressive \uline{C}ausal \uline{E}stimation), a framework that uses the density estimation capabilities of autoregressive models to recover the causal graph from a single sequence. By treating the event stream as a high-order Markov chain, TRACE performs scalable conditional independence testing via the model's learned distributions, inheriting OSCAR's linear scalability with the vocabulary size while remaining agnostic to the underlying autoregressive backbone (Transformers, Mamba, RNNs). We derive error bounds showing the recovered causal graph is correct up to a noise floor set by the model's convergence, and find empirically that causal structure becomes identifiable well before the autoregressive model fully converges during pretraining.

\definecolor{contrib}{RGB}{220, 235, 252}   %
\definecolor{future}{RGB}{245, 245, 245}    %
\definecolor{headerrow}{RGB}{40, 80, 140}   %
\definecolor{headertext}{RGB}{255,255,255}  %

\begin{table}[!t]
\centering
\renewcommand{\arraystretch}{2.2}
\setlength{\tabcolsep}{0pt}
\begin{tabular}{
    >{\centering\arraybackslash}m{3.4cm}  %
    | >{\centering\arraybackslash}m{5.4cm}
    | >{\centering\arraybackslash}m{5.4cm}
}

\rowcolor{headerrow}
\textcolor{headertext}{\textbf{Causal Scope}}
& \textcolor{headertext}{\makecell{\textbf{Events-to-Outcome} \\ \small (DTCs $\rightarrow$ Error Pattern)}}
& \textcolor{headertext}{\makecell{\textbf{Event-to-Event} \\ \small (DTCs $\rightarrow$ DTCs)}}
\\[2pt]
\hline

\cellcolor{headerrow}\textcolor{headertext}{\makecell{\textbf{Sample-Level} \\ \small (Single Sequence)}}
& \cellcolor{contrib}
    \makecell{
        \textbf{OSCAR} \\
        \small One-Shot Multi-Label \\
        \small Causal Discovery \\[3pt]
        \scriptsize Markov Boundary per label \\
        \scriptsize from a single sequence \\[3pt]
        \scriptsize \textit{Chapter~\ref{c7:multi_label_one_shot_causal_discovery}}
    }
& \cellcolor{contrib}
    \makecell{
        \textbf{TRACE} \\
        \small Scalable Amortized \\
        \small Causal Discovery \\[3pt]
        \scriptsize Sample time causal graph \\
        \scriptsize from a single sequence \\[3pt]
        \scriptsize \textit{Chapter~\ref{c7:event_to_event}}
    }
\\
\hline

\cellcolor{headerrow}\textcolor{headertext}{\makecell{\textbf{Population-Level} \\ \small (Full Dataset)}}
& \cellcolor{contrib}
    \makecell{
        \textbf{CARGO} \\
        \small Graph Aggregation for \\
        \small Causal Discovery \\[3pt]
        \scriptsize Global Markov Boundaries \\
        \scriptsize via adaptive graph fusion \\[3pt]
        \scriptsize \textit{Chapter~\ref{c7:multi_label_causal_discovery}}
    }
& \cellcolor{future}
    \makecell{
        \textbf{Open Problem} \\[3pt]
        \scriptsize Population-level summary \\
        \scriptsize causal graph over event types/outcomes \\[3pt]
        \scriptsize Requires aggregation of \\
        \scriptsize sample time and summary graphs \\
        \scriptsize with acyclicity constraints \\[6pt]
        \scriptsize \textit{Future work}
    }
\\
\end{tabular}

\caption{%
\textbf{The four regimes of causal discovery in discrete event sequences},
defined by the cross-product of \emph{causal scope}
(single-sequence inference vs.\ population-level aggregation)
and \emph{dependency type}
(events-to-outcome vs.\ event-to-event).
The three shaded contributions of this thesis --- \textbf{OSCAR},
\textbf{TRACE}, and \textbf{CARGO} --- share a common backbone:
pretrained autoregressive Transformers repurposed as neural
autoregressive density estimators (NADEs) for conditional
mutual information estimation.
The remaining cell, population-level event-to-event discovery,
constitutes an open problem and a direct avenue for future work.
}
\label{tab:causal_framework}
\end{table}

\subsection{Automated Causal Reasoning with Multi-Agent Systems for Error Pattern Rule Automation}
\noindent %
\noindent To fully automate the construction of EPs, we introduce \gls{carep} for \uline{C}ausal \uline{A}utomated \uline{R}easoning for \uline{E}rror \uline{P}atterns, a multi-agent system for the automatic synthesis of EP rules in high-cardinality automotive event sequences. It comprises three agents:
\begin{itemize}
    \item A \textbf{Causal Discovery Agent} that estimates candidate causes and causal indicators for each unknown EP using OSCAR and CARGO.
    \item A \textbf{Contextual Information Agent} that incorporates DTC descriptions, known EP rules, and metadata to simulate expert knowledge.
    \item An \textbf{Orchestrator Agent} that coordinates the two agents to synthesize five plausible rules for these EPs, alongside a natural language explanation.
\end{itemize}
Through its agent-based design, CAREP translates causal evidence and contextual information into symbolic rules, along with their explanations, so that engineers can use them directly for vehicle fault analysis. To evaluate CAREP, we masked a random subset of the known rules and tested the framework’s ability to identify them. Results show that CAREP consistently outperforms LLM-only baselines when reconstructing the rule, reaching \(83\%\) precision in recovering the correct elements in the rule, while offering interpretable explanations at scale.

The same backbone plays two distinct roles across the thesis. It is a domain-specific autoregressive predictor in Part~\ref{part1} and is repurposed without fine-tuning as a neural density estimator for causal discovery in Part~\ref{pa:p2}. In Part~\ref{pa:p3}, it is complemented by general-purpose Large Language Models that reason in natural language over the causal evidence produced by OSCAR and CARGO.

\subsection{Software}

In addition to the theoretical, methodological and architectural contributions presented in this dissertation, a significant effort has been dedicated to practical and scalable software. The core algorithmic frameworks have been formalized and open-sourced as a Python library named \texttt{seq2cause}~\cite{math2026seq2cause}\footnote{\url{https://pypi.org/project/seq2cause/}}.

Released under the permissive MIT License and publicly available via the Python Package Index (PyPI), \texttt{seq2cause} provides a universal, plug-and-play pipeline designed to transform standard autoregressive sequence models such as Transformers into robust causal discovery frameworks. This software contribution ensures the reproducibility of the experiments detailed in the subsequent chapters and equips the broader machine learning community with an open-source implementation.

\newpage
\section{List of Publications}

Every part of this thesis has been peer-reviewed and presented at an international venue. Only the AAAI paper appeared in archival conference proceedings (main track). The five other papers were accepted at \emph{workshops} held at NeurIPS, ICLR, and ICML. These workshops are non-archival: the papers are publicly available on OpenReview, but they are not part of the proceedings of the main conferences.

\subsection*{Conference Paper}
\begin{description}[style=nextline,leftmargin=0cm,labelsep=0em]

\item[\textbf{Harnessing Event Sensory Data for Error Pattern Prediction in Vehicles: A Language Model Approach.}] \cite{math2024harnessingeventsensorydata}
Hugo Math, Rainer Lienhart, Robin Schön, Proceedings of the AAAI Conference on Artificial Intelligence, vol. 39, no. 18, pp. 19423–19431, 2025, Philadelphia, USA, February 2025.
\end{description}

\subsection*{Workshop Papers}
\begin{description}[style=nextline,leftmargin=0cm,labelsep=0em]

\item[\textbf{One-Shot Multi-Label Causal Discovery in High-Dimensional Event Sequences.}] \cite{math2025oneshot}
Hugo Math, Robin Schön, Rainer Lienhart, Workshop \emph{CauScien: Uncovering Causality in Science} at NeurIPS 2025, San Diego, USA, December 2025.

\item[\textbf{Towards Practical Multi-label Causal Discovery in High-Dimensional Event Sequences via One-Shot Graph Aggregation.}] \cite{math2025towards}
Hugo Math and Rainer Lienhart, Workshop on \emph{Structured Probabilistic Inference \& Generative Modeling} at NeurIPS 2025, San Diego, USA, December 2025.

\item[\textbf{Context-Informed Sequence Classification: A Multimodal Approach to Vehicle Diagnostics.}]
\cite{math2026contextinformed}
Hugo Math, Rainer Lienhart, Workshop on \emph{Time Series in the Age of Large Models} at ICLR 2026, April 2026.

\item[\textbf{Neuro-Symbolic Rule Discovery: Empowering LLMs with Causality for Vehicle Diagnostics.}]
\cite{math2026multiagentcausalreasoningerror}
Hugo Math, Julian Lorenz, Rainer Lienhart, Workshop on \emph{Logical Reasoning of Large Language Models} at ICLR 2026, April 2026.

\item[\textbf{Your Autoregressive Model Already Reveals the Causal Graph.}]
\cite{math2026tracescalableamortizedcausal}
Hugo Math, Rainer Lienhart, Workshop on \emph{Structured Probabilistic Inference \& Generative Modeling} at ICML 2026, Seoul, South Korea, July 2026.
\end{description}

\newpage
\section{Thesis Outline}
This thesis is organized into three main parts. Part~\ref{part1} (Chapters \ref{c2:foundation_event_sequence_modeling}–\ref{c4:multimodal}) establishes the modeling foundations. Chapter~2 provides background on event sequence modeling and Transformers. Chapters 3 and 4 present our predictive architectures for DTC sequences, evaluated on large-scale BMW data. Part~\ref{pa:p2} (Chapters 5–8) develops three causal discovery frameworks. Chapter 5 provides mathematical foundations. Chapters 6, 7, and 8 present OSCAR, CARGO, and TRACE respectively, with theoretical guarantees and scalability experiments. Part III (Chapters 9–10) moves from causal discovery to automated reasoning and introduces CAREP, a multi-agent system.
The thesis concludes in Chapter~\ref{c7:conclusion_n_outlook} with a synthesis of the main findings and perspectives for future work. We summarize how predictive modeling, causal discovery, and automated reasoning form a coherent pipeline for scalable, interpretable diagnostics of vehicle error codes and patterns. Finally, we discuss open challenges and future directions.

\part{Predictive Sequence Modeling}
\label{part1}
\chapter{Foundations of Event Sequence Modeling}\label{c2:foundation_event_sequence_modeling}
\chaptermark{Foundations of Event Sequence Modeling}
\glsresetall
This chapter introduces the formal machinery used throughout the thesis: event sequence modeling, Transformer architectures, and evaluation metrics. We give a formal definition of a sequence of events and common ways of representing them. Then, we highlight challenges of traditional statistical methods and provide a brief introduction to Transformers \cite{tf, bert, gpt}. %
We introduce their primary components: positional embedding techniques, feed-forward layers, layer normalization, and the attention mechanism. At the end of this chapter, we investigate various metrics for evaluating language model predictions in classification and regression. Throughout this chapter we first introduce event, time, and sequence notation in its generic form. Table~\ref{tab:notation} in Section~\ref{sec:dataset} later specializes this notation to the automotive DTC dataset used from Chapter~\ref{c2:ep_prediction_based_on_live_data} onward.

\section{Hawkes Processes}
The Hawkes Process \cite{hawkeppp} has arguably been the most studied modeling technique for the occurrence of discrete events in a sequence, spanning financial and earthquake data~\cite{ogata1978statistical, hawke_process_and_app_2025}. Its key property, the so-called \textit{self-excitation}, is that the occurrence of one event may trigger a series of similar events. For instance, earthquakes trigger aftershocks, stock-market transactions may trigger a chain reaction, or a neuron fires and triggers brain activity elsewhere.

To do so, we model the number of event occurrences up to time \(t\) through a stochastic process \(N(t)\). In some extensions of the basic Hawkes process, a random variable called a \textit{mark}, which provides auxiliary information such as spatial coordinates or sensory measurements is attached to each event. In the automotive context, a vehicle generates a sequence of discrete events (DTCs), where each arrival is characterized by a timestamp $t$ and a specific mileage $m$ acting as the mark. Furthermore, we consider the multivariate setting where multiple Hawkes processes are \textit{mutually exciting}: an occurrence in one event stream increases the probability of future occurrences in other streams. For example, in a social network, a message may trigger further messages from different users.

This cross-excitation is the fundamental mechanism explored in this work. However, scaling these models remains a challenge: when faced with thousands of distinct event types across numerous long sequences, how can we efficiently learn a robust model and reason about the underlying interactions? In Part~\ref{part1}, we address the statistical and computational challenges of learning robust models from event sequences.

\subsection{Background}
 \begin{definition}[Counting Process]
     A stochastic process \(N(t)\), also noted \(N_t\), defined for \(t \geq 0\), is a \textit{counting process} starting at \(N(0)= 0\) if \(N(t)\) only takes values in \(\{0, 1, 2, \cdots\}\) and increases in jumps of size +1. The random jump times \(t_1, t_2, \cdots\) form a \textit{point process}, with \(0 < t_1<t_2 < \cdots\) and \(t_0 \triangleq 0\). The counting process can be defined in terms of the point process as:
     \[N(t) = \sum^{\infty}_{i=1}\mathds{I}\{t_i \leq t\} = \sum_{t_i \leq t} 1\]
 \end{definition}
\noindent The \textit{point process} representation enables us to formally define a \textit{marked event sequence} (Def.~\ref{def:marked_event_sequences}). Although event sequences can refer to multivariate time series as well~\cite{hasan2023a}, we abbreviate the term and refer solely to \textit{event sequences} as a discrete sequence of events, such as:

\begin{definition}[Event Sequence]\label{def:event_sequences}
Let each event be composed of a time of occurrence $t \in \mathbb{R}^+$ and an event type $x \in \mathcal{X}$ forming a pair $(x, t)$. $\mathcal{X}$ is a finite set of discrete event types.
A sequence of length \(L+1\) is constructed with multiple pairs of events, such as $S = \{(x_0, t_0), ... , (x_L, t_L)\}$ where $0<t_1< ... < t_L$ with \(t_0 \triangleq 0\).
\end{definition}

\begin{definition}[Marked Event Sequence]\label{def:marked_event_sequences}
A Marked Event Sequence is an event sequence (Def.~\ref{def:event_sequences}) where each event $x_i$ is augmented with a mark $m_i \in \mathcal{M}$. The event is then defined as a triple $(x_i, t_i, m_i)$.
\end{definition}
\noindent The inclusion of marks is mostly used in Part~\ref{part1} and Part~\ref{pa:p3} when dealing with the mileage and other sensory or metadata information per event \(x_i\).

\begin{definition}[Conditional Intensity Function]
The conditional intensity function of \(N(t)\) is defined, for \(t \geq 0\), by
\[\lambda^* \triangleq \lim_{\Delta \rightarrow 0}\frac{\mathbb{E}(N(t+\Delta) - N(t) | \mathcal{H}_t)}{\Delta}\]
if the limit exists. Here \(\mathcal{H}_t\) is the history of \(N(t)\) such that \(\mathcal{H}_t \triangleq \{(x_i, t_i) \in \mathcal{X} \times  \mathbb{R}^+|t_i<t\} \).
\end{definition}

\begin{remark}
In the context of vehicle logs or manufacturing trajectories, $S$ is a realization of a discrete-event stochastic process. While we acknowledge the Hawkes process framework common in the literature for modeling instantaneous rates as $\lambda^*(t)$, our primary interest lies in the conditional dependencies between the event types $x$ within the sequence.
\end{remark}

\subsection{Self-Exciting Property}
Hawkes~\cite{hawkeppp} defined the counting process (named after him) by specifying its conditional intensity process:
\begin{definition}[Hawkes Process]
    A Hawkes process is a counting process \(N(t)\) whose conditional intensity process for \(t\geq0\) is:
    \[\lambda^*(t) = \lambda + \sum_{t_i < t}\mu(t-t_i)\]
    where \(\lambda>0\) is a background arrival rate and \(\mu: \mathbb{R}^+ \rightarrow \mathbb{R}^+\) is the excitation function.
\end{definition}
\noindent The excitation function \(\mu\) controls how past event occurrence will affect the rate of future occurrences. A common and simple excitation function is \(\mu(t) = \alpha \exp(-\beta t)\) (exponentially decaying intensity), where \(\alpha>0\) sets the initial jump in intensity right after an event and \(\beta>0\) is the decay rate governing how quickly this excitation fades. In this case, \((N(t), \lambda^*(t))\) is a \textit{Markov process}~\cite{koller2009probabilistic} such that the current intensity \(\lambda(t)\) does not depend on the entire history of events.

\subsection{Mutually Exciting Hawkes Processes}

\begin{definition}[Mutually-Exciting Hawkes processes]
    Consider \(\mathbf{N}(t) = (N^1(t), \cdots, N^{|\mathcal{X}|}(t))\) as a collection of \(|\mathcal{X}|\) counting processes with \(N^k(t)\)'s occurrence times denoted as \(t^k_1, t^k_2\) etc. They are mutually-exciting Hawkes processes if \(N^k(t)\)'s conditional intensity follows:
    \[\lambda^*_k(t) = \lambda_k + \sum^{|\mathcal{X}|}_{j=1}\sum_{t^j_i < t} \mu_{j,k}(t-t^j_i) \; \text{for }\; k=1, \cdots, |\mathcal{X}|\]
    where \(\mu_{j,k}(s) \geq 0\) quantifies the excitation exerted by stream \(j\) (the triggering stream) on stream \(k\) (the triggered stream), and \(\lambda_k>0\) is the background arrival rate specific to stream \(k\).
\end{definition}
\begin{remark}
    Denote the integrals of the \(\mu_{j,k}\) excitation functions as \(\phi_{j,k} = \int^\infty_0 \mu_{j,k}(s)ds\) and the \emph{excitation matrix} as \(\Phi = (\phi_{j,k}) \in \mathbb{R}^{|\mathcal{X}|\times|\mathcal{X}|}\). The \(\Phi\) matrix is conventionally treated as an approximation of the causal adjacency matrix of a graph with \(|\mathcal{X}|\) nodes and is typically used to extract causal relationships~\cite{granger_causality_hawkes}. The \(|\mathcal{X}|^2\) parameter count renders this approach computationally intractable in our high-cardinality regime. Part~\ref{pa:p2} adopts an alternative that avoids this bottleneck.
\end{remark}

\subsection{Limitations}
\paragraph{Number of parameters.}
The \(N^j_t\) process has either an excitatory effect on \(N^k_t(\phi_{j,k}>0)\) or no effect on \(N^k_t(\phi_{j,k} = 0)\) for \(j = k \) or \(j \not= k\). Importantly, these processes have \(\mathcal{O}(|\mathcal{X}|^2)\) parameters to fit, which is often intractable in practice for many applications~\cite{hawke_process_and_app_2025}. For \(|\mathcal{X}|= 10^3\) the number of parameters is already \(1\) million. Moreover, the actual number of scalars being optimized in practice is more than \(|\mathcal{X}|^2\), since there are the \(|\mathcal{X}|\) stream-specific base intensities \(\lambda_k\), the parameters of the excitation functions \(\mu\) (decay rates) and interaction magnitude. The literature often assumes \(|\mathcal{X}|^2+|\mathcal{X}|\)~\cite{hawke_process_and_app_2025}.

\paragraph{Optimization.}
To understand why the Hawkes process is insufficient for our setting, we first show that the Maximum Likelihood Estimation (MLE) of its parameters is computationally intractable at scale. We therefore seek the parameters (base rate $\alpha$, decay $\beta$) that make the observed sequence of events most probable. The likelihood for any point process parametrized by \(\theta\) within a time horizon \(0 \leq t \leq T\) can be computed as the sum of the log-likelihood for each process. Denoting by \(N^k(T)\) the number of events observed in stream \(k\) by time \(T\) and by \(t^k_1, \cdots, t^k_{N^k(T)}\) their occurrence times (as introduced for the mutually-exciting Hawkes process above):
\begin{equation}\label{eq:mll_mutual_hawkes}
     \ln \mathcal{L} (\theta|T) = \sum^{|\mathcal{X} |}_{k=1}\sum^{N^k(T)}_{i=1}\ln{\lambda^{*}_k(t^k_i)} - \sum^{|\mathcal{X}|}_{k=1}\int^T_0 \lambda^{*}_k(t)dt
\end{equation}
The full derivation can be found in Appendix~\ref{appendix:derivation_ll_hawke_mutual}. Importantly, Eq.~\ref{eq:mll_mutual_hawkes} cannot be computed easily since, for each of the \(|\mathcal{X}|\) streams, evaluating \(\lambda^*_k(t^k_i)\) itself requires summing over all preceding events across all \(|\mathcal{X}|\) streams, leading to \(\mathcal{O}(|\mathcal{X}|^2\cdot N(T)^2)\) at worst and \(\mathcal{O}(|\mathcal{X}|^2\cdot N(T))\) if using exponential decay~\cite{hawke_process_and_app_2025}. This is a regime where classical MLE is computationally prohibitive and statistically prone to overfitting~\cite{hawke_sparse_mutual, sparsetemporalattention}.
In contrast, the standard self-exciting Hawkes process with exponential decay leads to \(\mathcal{O}(N(T))\) complexity~\cite{hawke_process_and_app_2025} but implies a greater parametric assumption.

\subsection{The High-Cardinality Discrete Event Stream Regime}

While the Mutually-Exciting Hawkes process provides a rigorous theoretical foundation for interacting events, it assumes that maintaining a continuous intensity function \(\lambda^*_k(t)\) for every dimension \(k \in \{1, \dots, |\mathcal{X}|\}\) is computationally feasible. However, in regimes of highly structured data with a large number of discrete events (patient trajectories~\cite{nlp_pred_mimic3}, event logs~\cite{draxler2025transformers}, vehicle diagnostics~\cite{math2024harnessingeventsensorydata}, NLP~\cite{shannon1951prediction}), representing each continuous intensity function or \textit{stream} separately is inefficient. For instance, in modern automotive diagnostics, the event space cardinality approaches \(|\mathcal{X}| \approx 10^4\), while the actual event occurrences are sparse due to structural relationship (event \(A\) always excites the occurrence of \(B\) despite the history). We assume such a high-cardinality\footnote{The titles of some of our publications use the term \emph{high-dimensional} to match the terminology common in the machine learning and causal discovery literature~\cite{assaad_survey_ijcai_cd_time_series, cd_temporaldata_review}. The precise term for the setting studied here, a large vocabulary of event types \(|\mathcal{X}|\), is \emph{high-cardinality}.} regime throughout the thesis.

This shift in perspective is crucial. It aligns the problem with \textit{Neural Sequence Modeling}~\cite{seq2seq_learning} and \textit{Language Modeling}~\cite{gpt}, where the focus is on predicting the next token \(x_{i+1}\) from a discrete sequence of words drawn from a high-cardinality vocabulary, rather than regressing \(|\mathcal{X}|\) parallel sequences. This justifies our use of autoregressive architectures (in Part~\ref{part1}) to approximate the conditional density \(P(x_{i+1} \mid \mathcal{H})\) and enables the scalable causal discovery techniques developed further in Part~\ref{pa:p2}.

\section{Language Models}
Although we employ the term event sequence modeling, we define it in the \gls{nlp} realm and adopt its terminology (such as tokens, vocabulary). Therefore, we give an overview of the important concepts in NLP as used in this work.
\subsection{Word Level}
\paragraph{Sentence and Words}Before introducing the modeling framework, we formally define the notion of a sentence as a special case of a unit-time event sequence:

\begin{definition}[Sentence as a Unit-Time Event Sequence]\label{def:sentence_sequence}
A sentence $S$ is a special case of an event sequence (Def.~\ref{def:event_sequences}) characterized by \textbf{unit-time arrivals}. Given a vocabulary (the event type space $\mathcal{X}$), a sentence \(S\) is a sequence of events defined as:
\begin{equation}
S = {(x_i, t_i)}_{i=0}^L \quad \text{where} \quad t_i = i, \quad \forall i \in \{0, \dots, L\}
\end{equation}
where events are so-called 'words'.
\end{definition}

\paragraph{Tokenization} To translate words into machine-interpretable data, a step called \emph{tokenization} converts a sequence of words into \emph{tokens}. These tokens can be seen as word fragments. Consequently, a word is often split into smaller units (\emph{sub-word tokenization}) to optimize the compression of a natural language~\cite{bert} into a fixed-size \emph{vocabulary}. In event sequence modeling, it is common to treat events as whole tokens, since we do not know their underlying semantic structure. %

\paragraph{Language Preprocessing}
To obtain all distinct tokens, there is often a preprocessing stage where we iterate over the entire observational dataset \(\mathcal{D} = \{S_0, \cdots, S_m\}\) of sequences and sort the different tokens by their frequency, i.e., we obtain a vocabulary with the most common words. The vocabulary is referred to as \(\mathcal{X}\). The words that do not appear in the vocabulary are replaced by the unknown token, often defined as: \([UNK]\). A start-of-sequence token \(<S>\) or \([CLS]\) in BERT \cite{bert} is used as \(x_0\), and an end-of-sequence token \(</S>\) or \([EOS]\) for \(x_L\) the last word.

\subsection{Word Embeddings}
To get a dense representation of our sequence, we embed each token into a $ d$-dimensional space using a learnable embedding matrix $\mathbf{W}^e \in \mathbb{R}^{V \times d}$ where $V=|\mathcal{X}|$ is the number of distinct tokens (vocabulary). To perform this projection in parallel, we create a sequence of one-hot encoded vectors from the tokens $\{\boldsymbol{x}_i\}^L_{i=0}$ as $\mathbf{Y} \in \mathbb{R}^{L \times V}$. The word embedding \(\boldsymbol{E}\) is obtained via a linear projection:

\begin{equation}\label{eq:word_embedding}
\mathbf{E} = \mathbf{Y} \mathbf{W}^e \in \mathbb{R}^{L \times d}
\end{equation}
where \(d \ll V\). Typically, \(W^e\) is learned from scratch, especially adapted for event sequence modeling, where there is no pretrained word embedding.
\subsection{Transformers}
Throughout the thesis, we will primarily employ Transformer-based architectures. This architecture underpins the vast majority of modern deep learning, whether in vision \cite{vilbert}, time series \cite{ansari2024chronoslearninglanguagetime}, or language \cite{touvron2023llamaopenefficientfoundation}. It relies on several key concepts such as the attention mechanism, feed-forward layers, layer normalization \cite{ba2016layernormalization}, position embeddings, and self-supervised learning. We provide a brief overview of these elements.

\subsubsection{Attention}

\paragraph{Scaled Dot-Product Attention}
The majority of research utilizing Transformer models for sequence data employs the architecture introduced by \cite{tf}.
We define three linear projection matrices $\mathbf{Q} =\mathbf{U} \mathbf{W}^Q \in \mathbb{R}^{L \times d}$, $\mathbf{K} =\mathbf{U} \mathbf{W}^K \in \mathbb{R}^{L \times d}$, and $\mathbf{V} =\mathbf{U} \mathbf{W}^V \in \mathbb{R}^{L \times d}$ which are called respectively \textit{query}, \textit{key}, and \textit{value}.
$\mathbf{W}^Q$, $\mathbf{W}^K$, $\mathbf{W}^V$ are trainable weights.
Scaled dot-product attention computes a similarity between queries and keys to weight the importance of each token's value, allowing the model to focus on relevant parts of the sequence when making predictions. It is computed as:
\begin{equation}
\label{eq:vanilldotprodattention}
 \mathbf{A} = \text{Attention}(\mathbf{Q}, \mathbf{K}, \mathbf{V}) = \text{softmax}\Big(\frac{\mathbf{Q}\mathbf{K}^T}{\sqrt{d_k}}\Big)\mathbf{V}
\end{equation}
where $\mathbf{A} \in \mathbb{R}^{L \times d}$. The softmax is applied row-wise. %
\noindent In the original paper, \cite{tf} argue that scaling the \(\mathbf{Q} \mathbf{K}\) dot product enables a better stability of the softmax function since the gradients become extremely small. Therefore the \emph{scaling} factor \(d_k\), the dimensionality of the key vectors \(\mathbf{K}\) (equal to \(d\) in the single-head case), is introduced in Eq.~\ref{eq:vanilldotprodattention}. For \emph{causal} or \emph{autoregressive} Transformers, a lower triangular matrix \(\tilde{\mathbf{A}} \in \{0, -\infty\}^{L \times L}\) is applied additively to the pre-softmax logits of Eq.~\ref{eq:vanilldotprodattention}, setting future positions to \(-\infty\) to prevent tokens from attending to future tokens \cite{gpt}.

\paragraph{Multi-Head Attention}
In the same paper, \cite{tf} extended the single scaled dot-product into a \textit{multi-head attention} (MHA). They divided the attention scores computation into multiple heads to perform multiple attentions in parallel. In this case, the feature dimension \(d\) is divided into \(h\) parallel attention heads of size \(\frac{d}{h} \in \mathbb{N}\). It is usually done in practice by reshaping the query, key matrices into respectively \(Q \in \mathbb{R}^{L \times \frac{d} {h}\times h}\) and \(K \in \mathbb{R}^{L \times \frac{d}{h}\times h}\) before feeding them to the attention scores calculation:

\begin{equation}\label{eq:mha}
    \text{MHA}(\mathbf{Q}, \mathbf{K}, \mathbf{V}) = [\text{head}_1, \cdots, \text{head}_h] \cdot \mathbf{W}^{\text{O}}
\end{equation}
with:
\begin{equation}\label{eq:head_mha}
    \text{head}_i = \text{Attention}(\mathbf{Q}_i, \mathbf{K}_i,  \mathbf{V}_i)
\end{equation}
where \(\mathbf{Q}_i \in \mathbb{R}^{L \times \frac{d}{h}}, \mathbf{K}_i \in \mathbb{R}^{L \times \frac{d}{h}}, \mathbf{V}_i \in \mathbb{R}^{L \times \frac{d}{h}}\) and \(\mathbf{W}^{\text{O}} \in \mathbb{R}^{d \times d}\) is a learnable projection matrix, where \(d = h \cdot d_V\) since the concatenation \([\text{head}_1, \cdots, \text{head}_h]\) yields a matrix in \(\mathbb{R}^{L \times d}\). The authors show that each head learns different token-to-token relationships from a single sequence.

\subsubsection{Feed-forward Layers}
The multi-head-attention scores are processed using a combination of layer normalization, position-wise feed-forward neural network (FFN), and residual connections to ensure a proper gradient flow and to learn the attention scores patterns between tokens:
\begin{equation}\label{eq:ffn_hidden_state_tf}
\begin{split}
    \mathbf{U}'= \text{LayerNorm}(\mathbf{A}+\mathbf{U}) \\
    H = \text{LayerNorm}(\mathbf{U'}+\text{FFN}(\mathbf{U}'))
\end{split}
\end{equation}
where \(H \in \mathbb{R}^{L \times d}\) is the resulting hidden state. The so-called position-wise FFN layers are typically two layers that project the \(d\)-dimensional input to an intermediate size, often \(4 \times d\). These FFNs are applied independently for each position:
\begin{equation}
    \text{FFN}(\mathbf{U}) = f(\mathbf{U}\mathbf{W}_1+\boldsymbol{b}_1)\mathbf{W}_2 + \mathbf{b}_2
\end{equation}
where \(f\) an activation function (ReLU \cite{relu} in the original paper), \(\mathbf{W}_1, \mathbf{W}_2\) learnable weight matrices, and \(\boldsymbol{b}_1, \mathbf{b}_2\) their corresponding bias vectors. Then, residual connections \cite{he2016residual} are applied in Eq.~\ref{eq:ffn_hidden_state_tf} to improve gradient flow and training stability.
This layer is stacked \(k\) times, leading to a Transformer with \(k\) layers.
\subsubsection{Normalization}
The normalization layer helps stabilize the dynamics of the deep neural network's hidden states \cite{ba2016layernormalization} by rescaling each output neuron before the activation function. This mitigates the problem of \emph{internal covariance shift} \cite{bs_norm_covariate_shift}, which arises from the fact that the weights of one layer are highly dependent on the outputs of the neurons in the previous layer.
Modern Transformers such as LLaMA \cite{touvron2023llamaopenefficientfoundation} use root-mean-square (RMS) normalization~\cite{rmsnorm} where each activation is divided by the RMS of the layer's inputs, omitting the mean-centering step of classical layer normalization.%

\subsubsection{Positional Encoding}
\paragraph{Deterministic}
Due to the nature of the attention scores, the architecture does not have a sense of ordering, as the input is just an array of vectors. Therefore, \cite{tf} added a so-called \emph{positional encoding} to the word embedding vector for each sequential step. This vector comprises a fixed periodic function, such as:
\begin{equation}\label{eq:pos_embedding_vanilla}
\mathbf{PE}_{i,j} := \begin{cases}
    \sin(i \times \omega_0^{j/d}) & \text{if } j\mod 2 = 0 \\
    \cos(i \times \omega_0^{(j-1)/d}) & \text{if } j\mod 2 = 1\\
\end{cases}
\end{equation}
where \(i\) is the index of the \(i-\)th token, \(j \in \{0, \ldots, d-1\}\) indexes the embedding dimension, and \(\omega_0\) the initial frequency (usually \(10^{-4}\)).
Then, it is added to the word embedding to form the input embedding:
\begin{equation}\label{eq:pe}
    \mathbf{U} = \mathbf{PE} + \mathbf{E}
\end{equation}
\paragraph{Learnable}
In most PyTorch \cite{pytorch} and Hugging Face \cite{wolf2020huggingfacestransformersstateoftheartnatural} implementations of Transformers (e.g., BERT \cite{bert}, GPT \cite{gpt}), a learnable positional embedding is obtained via a learnable lookup table  \(\mathbf{W}^{\text{PE}} \in \mathbb{R}^{(L+1) \times d}\), one row for each of the \(L+1\) sequential steps \(\{0, 1, 2, \cdots, L\}\). It is added to the word embeddings in Eq.~\ref{eq:pe}

\subsubsection{Autoregressive Transformers}
Autoregressive or causal Transformers have a causal attention mask applied in Eq.~\ref{eq:vanilldotprodattention}.  They perform the next token prediction via the chain-rule factorization of the joint distribution \(P(X_0, \cdots, X_L)\) \cite{NIPS1999_e6384711}, such as the probability of observing a sequence of event types \(s = (x_0, \cdots, x_L)\) is:
\begin{equation}\label{eq:autoreg_factorization}
    P(s) = \prod_{i=0}^{L} p(x_i \mid x_0, x_1, \ldots, x_{i-1})
\end{equation}
where \(p(x)\) is the probability mass function, i.e., \(P(X=x)\). These models are trained using next-token prediction \cite{gpt}, by minimizing the negative log-likelihood of observing sequences.

\section{Metrics}
\noindent We measure the prediction of language models using common classification metrics that are interpretable and anchored in the machine learning domain. We also define our own metric later in Chapter~\ref{c2:ep_prediction_based_on_live_data} for predictive maintenance.

\subsection{Next-Event Prediction}
Next-event prediction is often evaluated using the Accuracy of the inferred next event \cite{selfatthawke, transformerhawkeprocess, sparsetemporalattention}. However, in the presence of class imbalance, Accuracy alone can be misleading \cite{reviewmultilabellearning}. Therefore, we include Precision and Recall to provide a general view of model performance.

\paragraph{Accuracy} is defined as the proportion of correct predictions among the total number of cases examined:
\begin{equation}\label{metrics:acc}
    ACC = \frac{TP + TN}{TP + TN + FP + FN}
\end{equation}
\noindent where \(TP\), \(TN\), \(FP\), and \(FN\) denote, respectively, the number of true positives, true negatives, false positives, and false negatives.

\paragraph{Precision} (or positive predictive value) quantifies the reliability of the positive predictions, measuring the proportion of inferred events that are correctly identified:
\begin{equation}\label{metrics:precision}
  P = \frac{TP}{TP + FP}
\end{equation}

\paragraph{Recall} (or sensitivity) measures the completeness of the discovery, capturing the proportion of actual positive events that were recovered by the model:
\begin{equation}\label{metrics:recall}
 R = \frac{TP}{TP + FN}
\end{equation}

\subsection{Next-Time Prediction}
\paragraph{Mean Absolute Error}
MAE is the average of the absolute difference between the actual \(t_i\) and predicted values \(\hat{t_i}\):
\begin{equation}\label{eq:metrics_mae}
    \text{MAE} = \frac{1}{m}\sum^m_{i=0} |t_i - \hat{t}_i|
\end{equation}

For instance, if one finds an MAE of \(5h\), it indicates that the model has, on average, an error of 5 hours when predicting the time of occurrence. Eq.~\ref{eq:metrics_mae} weighs all errors equally, making it less sensitive to outliers.

\paragraph{Root Mean Square Error} or RMSE, adds a square function inside the sum, averages across all points, and takes the square root:
\begin{equation}\label{eq:metrics_rmse}
    \text{RMSE} = \sqrt{\frac{1}{m}\sum^m_{i=0} (t_i - \hat{t}_i)^2}
\end{equation}

\noindent Unlike MAE, RMSE is more sensitive to outliers as it disproportionately penalizes larger errors. Nevertheless, RMSE remains less interpretable as the scaling of values changes.

\subsubsection{Confident Predictive Maintenance Window}
\label{metrics:cpmwauc}

Let $\zeta : \{0, \ldots, L\} \to \mathbb{R}$ be a performance metric evaluated as a function
of the number of observed events (e.g., precision or mean absolute error). Given a
confidence threshold $\theta \in \mathbb{R}$, we define the \emph{onset of confident
prediction} $x_\theta$ as the earliest observation index at which $\zeta$ exceeds (or
falls below, for error metrics) $\theta$ and remains so on average:

\begin{equation}
    x_\theta = \min \bigl\{ i \in \{0, \ldots, L\} \mid \zeta(i) \bowtie \theta \bigr\},
    \label{eq:cpmw_onset}
\end{equation}

\noindent where $\bowtie\, \in \{{\geq},{\leq}\}$ depending on whether $\zeta$ is a
quality score (e.g., F1, $\bowtie\, = {\geq}$) or an error metric (e.g., MAE,
$\bowtie\, = {\leq}$). Let $\mu_{\mathrm{seq}}$ denote the average sequence length
across the evaluation set.

\begin{definition}[Confident Predictive Maintenance Window]
\label{def:cpmw}
The \emph{Confident Predictive Maintenance Window} (CPMW) is the interval
$[x_\theta,\, \mu_{\mathrm{seq}}]$ within which the model produces confident predictions,
as determined by threshold $\theta$.
\end{definition}

To quantify the overall quality of a model within the CPMW, we define the
\emph{CPMW Area Under the Curve} ($\mathrm{CPMW}_\zeta$) as the normalised integral of
$\zeta$ over this interval:

\begin{equation}
    \mathrm{CPMW}_\zeta =
        \frac{1}{\mu_{\mathrm{seq}} - x_\theta}
        \int_{x_\theta}^{\mu_{\mathrm{seq}}} \zeta(x)\, \mathrm{d}x,
    \label{eq:cpmwauc}
\end{equation}

\noindent where $\zeta$ is assumed to be continuous and integrable on
$[x_\theta, \mu_{\mathrm{seq}}]$. In practice, $\zeta$ is estimated at discrete
observation steps and Eq.~\eqref{eq:cpmwauc} is approximated by the trapezoidal rule.
Importantly, \emph{A higher $\mathrm{CPMW}_{\mathrm{F1}}$ indicates reliable early classification};
a lower $\mathrm{CPMW}_{\mathrm{MAE}}$ indicates accurate early time-of-occurrence
estimation.

\section{Dataset}\label{sec:dataset}
This thesis uses proprietary BMW vehicle datasets. Due to confidentiality constraints, we will briefly describe the dataset in terms of its nature, dimensions, and scale, but no specific numerical values, vehicle identifiers, or customer information can be disclosed. Appendix~\ref{appendix:bmw_confidentiality} reproduces a confirmation statement from our BMW Group supervisor specifying the exact conditions under which this data was accessed, the anonymization and synthetic-data procedures applied to all publicly released artifacts, and the rationale for this limited disclosure.
\uline{All mentioned GitHub repositories contain synthetic or anonymized data only; no BMW data or mappings are included}. Table~\ref{tab:notation} below specializes, for this automotive setting, the generic event (\(x\)), event-type alphabet (\(\mathcal{X}\)), and time (\(t\)) notation introduced earlier in this chapter (Def.~\ref{def:event_sequences}).

\begin{table}[ht]
    \centering
    \begin{tabular}{|c|p{6cm}|}
        \hline
        \textbf{Notation} & \textbf{Description} \\
        \hline
        $x$ & Discrete event type, in our case a DTC. \\
        \hline
        $d$ & Absolute mileage of the vehicle in km.\\
        \hline
        $m$ & Mileage of the vehicle in km since the first DTC ($x_0$) occurred in a sequence $S$ such as $m_i = d_i - d_0$\\
        \hline
        $ts$ & Unix timestamp attached to each DTC.\\
        \hline
        $t$ & Number of hours passed since the first DTC occurred in a sequence. More specifically, \(t_i\) is defined as \(t_i = ts_i - ts_0\). \\
        \hline
        $S$ & Sequence of triplets (event type, time, mileage) defined as \(S = \{(x_i, t_i, m_i)\}_{i=0}^{L}\) of length $L$ with index starting from 0. \\
        \hline
        $S_l$ & Multi-labeled sequence of triplets (event type, time, mileage) leading to an outcome defined as \(S_l = \{(x_i, t_i, m_i), \boldsymbol{y}\}_{i=0}^{L}\) of length $L$ with index starting from 0 and \(\boldsymbol{y} \in \{0, 1\}^{|\mathcal{Y}|}\). \\
        \hline
        $S^{(i)}$ & The \(i\)-th sequence in a dataset \(\mathcal{D}\).\\
        \hline
        $S_l^{(i)}$ & The \(i\)-th multi-labeled sequence in a dataset \(\mathcal{D}\).\\
        \hline
        $s$ & Sequence of length \(L\) containing only the event types, e.g., $(x_0, x_1, \cdots, x_L)$. \\
        \hline
        $i\in \{0,..., L\}$ & index of element in (event) sequence. \\
        \hline
    \end{tabular}
    \caption{List of dataset symbols and their respective meanings for the datasets used in this thesis.}
    \label{tab:notation}
\end{table}

\subsection{Vehicular Fault Sequences}
A vehicular fault sequence is commonly referred to as an event sequence, where each event includes an error code and a timestamp. If specified, the vehicle's mileage at the time the error code was triggered can be included as \(m_i\), forming a marked event sequence (Def.~\ref{def:marked_event_sequences}). To obtain a full marked event sequence $S = \{(x_i, t_i, m_i)\}^L_{i=0}$, we obtain the last known timestamp $ts_L$ and mileage $d_L$ and select all DTCs that are no further than: (1) a given period in the past ($ts_L-ts_i \leq 30$ days) and (2) a given distance in the past ($d_L-d_i \leq 300$km). In addition to that, the dataset is joined with \glspl{ep} per sequence. In most of our experiments we have labeled sequences, where the labels are the EPs. Let \(\mathcal{D} = \{S_l^{(1)}, \cdots, S_l^{(m)}\}\) be a dataset of multi-labeled sequences \(S_l^{(k)} = (\{(x_0, t_0), \cdots, (x_L, t_L)\}, \boldsymbol{y})\) where \(\boldsymbol{y} \in \{0, 1\}^{|\mathcal{Y}|}\) is a multi-one hot vector (multiple EPs can happen per sequence).

For each chapter, we will recall the used dataset and its characteristics, such as the number of distinct DTCs \(|\mathcal{X}|\) and EPs \(|\mathcal{Y}|\), the number of sequences, and their average lengths. An overview of the notations can be found in Table~\ref{tab:notation}.

\subsubsection{Diagnostic Trouble Code: DTCs}\label{sec:dtc_composition}
We construct a \textit{\gls{dtc}} indicating the precise error from 3 pieces of information arriving at the same timestamp and mileage: (1) the ID number of the ECU, (2) an error code (\textit{Base-DTC}), and (3) a \textit{Fault-Byte}. A single DTC token comprises these three elements:
\begin{equation}\label{eq:dtc_def}
   DTC = ECU||\textit{Base-DTC}||\textit{Fault-Byte}
\end{equation}
where \(||\) denotes a concatenation.
\subsubsection{Error Patterns}\label{sec:ep_def}
Error patterns (\gls{ep}s) are a higher-level abstraction of DTC sequences to characterize specific vehicle faults. They are defined as Boolean combinations over DTCs. When denoting \textit{error pattern}, we refer to it as their intrinsic name (e.g., a precise battery defect, a PCB fault), while when denoting \textit{error pattern rules}, we refer directly to the Boolean combinations that define this error pattern. In practice, this distinction tracks the task at hand: whenever an EP is \emph{predicted} (Chapters~\ref{c2:ep_prediction_based_on_live_data}--\ref{c4:multimodal}) or used as the \emph{outcome} of a causal discovery procedure (Chapters~\ref{c7:multi_label_one_shot_causal_discovery}--\ref{c7:multi_label_causal_discovery}), it is treated as a labeled target \(y \in \{0,1\}\), i.e., only its name/identity matters. Whenever an EP is \emph{discovered} or \emph{synthesized} (Chapter~\ref{c8:carep}), it is treated as its underlying Boolean rule over DTCs, i.e., the object being reconstructed. A short footnote recalls this distinction at the start of each chapter where the usage changes. An example of an EP rule for \((y_1)\) based on some diagnosis trouble codes \((x_i)\) is shown in Eq.~\ref{eq:ep_def} such as:
\begin{equation}\label{eq:ep_def}
  y_1 = x_1 \;  \& \; x_5 \;\&\; x_8 \; \& \; (x_{12} \; | \; x_3) \; \& \; !x_{10} \;\& \; !x_{20}
\end{equation}
A tangible EP example for a battery failure could be:
\begin{equation}
   (\text{voltage drop} \; \& \; \text{temp high})\; | \;(\text{charging error} \;\&\; ! \text{battery replaced})
\end{equation}

\subsubsection{Experimental Datasets used Across Chapters}
We used the following dataset settings across chapters%
\footnote{We will recall them in their respective experiment sections.}:
\begin{itemize}
    \item \textbf{(Chapter 3)}: 1.7 million sequences with
          $|\mathcal{X}| = 8{,}710$ DTCs and $|\mathcal{Y}| = 254$ EPs
          from one BMW model range.
    \item \textbf{(Chapter 4)}: 5 million sequences with
          $|\mathcal{X}| = 22{,}137$ DTCs and $|\mathcal{Y}| = 360$ EPs
          from different BMW model ranges.
    \item \textbf{(Chapters 6--9)}: $300{,}000$ sequences with
          $|\mathcal{X}| = 29{,}100$ DTCs and $|\mathcal{Y}| = 474$ EPs
          from different BMW model ranges.
\end{itemize}
\chapter{Autoregressive Architectures for Error Pattern Prediction}\label{c2:ep_prediction_based_on_live_data}
\glsresetall

Anticipating both \emph{what} error pattern will occur and \emph{when} it will appear is an important step toward predictive maintenance and improved vehicle safety.\footnote{Throughout this chapter, an EP is treated as a \emph{labeled target} to be predicted, i.e., only its name/identity matters, not its underlying Boolean rule over DTCs; see Section~\ref{sec:ep_def} for the formal distinction.} In this chapter, we present an approach that draws an analogy between natural language processing and modeling multivariate vehicle event streams. Similar to words in a sentence, DTCs unfold as sequences over time, enriched with contextual information such as mileage and time of occurrence. To capture these dependencies, we introduce two autoregressive Transformer-based architectures: \emph{CarFormer}, an autoregressive Transformer trained with a novel self-supervised strategy, and \emph{EPredictor}, a decoder that jointly predicts the type and timing of future EPs.
This chapter is based on the following publication:
\newline
\newline
\noindent\textbf{Harnessing Event Sensory Data for Error Pattern Prediction in Vehicles: A Language Model Approach.} \cite{math2024harnessingeventsensorydata} Hugo Math, Rainer Lienhart, Robin Schön.
\textit{Proceedings of the AAAI Conference on Artificial Intelligence}, vol. 39, no. 18, pp. 19423–19431, 2025, Philadelphia, USA, February 2025.
\newline

\section{Introduction}
A central challenge in vehicle diagnostics lies in predicting both \emph{what} kind of error will manifest and \emph{when} it is likely to occur. Unlike individual DTCs, which may be noisy or repetitive (e.g., due to recurring electrical glitches or software updates), EPs characterize entire fault sequences, such as engine or battery failures. Fig.~\ref{fig:car_} illustrates this setting, where the task is to infer, given a past sequence of DTCs, the probability distribution over future EPs and their expected time of occurrence.

\noindent Recent research has begun to address predictive maintenance using DTC sequences, for instance, with recurrent neural networks (RNNs)~\cite{faultpredmultivariatevehicule} or Transformer-based architectures~\cite{Hafeez2024DTCTranGruIT} that aim to predict the next DTC. However, focusing solely on the next DTC is insufficient for two reasons. First, minor or noisy DTCs can obscure critical patterns, making it difficult to distinguish between trivial and critical faults. Second, as the event vocabulary grows often into the tens of thousands, the accuracy of next-token prediction deteriorates rapidly, much like in natural language modeling with large vocabularies and long sequences \cite{bachmann2024pitfallsnexttokenpred}.

Historically, as seen in the foundation, Hawkes Processes and their neural variants have advanced the state of the art in event modeling for next event and time prediction tasks \cite{hawkeppp, rrnembedding, nppreview}. Transformer-based models like BERT \cite{bert} and GPT-3 \cite{gpt3} have gained overwhelming popularity due to their attention-based architecture, flexibility, parallelization, and state-of-the-art performance in sequence modeling. Consequently, models adapted to discrete-time sequences using Transformers have emerged naturally \cite{selfatthawke, transformerhawkeprocess, shou2024selfsupervisedcontrastivepretrainingmultivariate}, achieving state-of-the-art performance in next-event prediction benchmarks. Inspired by the analogy between natural language processing and vehicle event streams, we adopt this perspective for DTCs and EPs: DTCs play the role of words, while EPs correspond to a sentence label (e.g., sequence classification for sentiment analysis). Formally, we represent a DTC sequence as a sentence of the form
\[
"<s>\ DTC_1\ DTC_2\ \ldots\ DTC_n\ </s>"
\]
We make several modifications to the vanilla Transformer from \cite{tf}, incorporating continuous-time and mileage positional embeddings as additional context. Using two distinct training phases, we introduce two models. CarFormer is a pretrained model acting as an encoder. EPredictor is a Transformer-based decoder that generates a probability distribution over a set of error patterns for each event step \(i\) to determine \textit{what} EPs will most likely happen and estimates a time for \textit{when} it will occur.

\section{Background}
To model asynchronous discrete events in a sequence, we typically reuse the vanilla Transformer \cite{tf} and create two embeddings.

\paragraph{Time Embedding} Time embedding replaces the traditional positional encoding, which grants the Transformer model positional information of each token within the sequence. Instead of using the token position \(i\) in Eq.~\ref{eq:pos_embedding_vanilla}, the continuous event time occurrence \(t_i\) is used. This provides an asynchronous representation of the event positions in a sequence.
\paragraph{Event-Type Embedding}\label{eventypeemb} To get a dense representation of our sequence, we embed each event into a $ d$-dimensional space using an embedding matrix $\mathbf{L}^{V \times d}$ where $V=|\mathcal{X}|$ is the number of distinct events. As it is done for word embeddings in Eq.~\ref{eq:word_embedding}, we create a sequence of one-hot encoded vectors from the event types $\{x_i\}^L_{i=0}$ as $\mathbf{Y} \in \mathbb{R}^{L \times V}$. The event-type embedding $\mathbf{E}$ is defined as $\mathbf{E} = \mathbf{Y} \mathbf{L} \in \mathbb{R}^{L \times d}$ and the input embedding $\mathbf{U}$ as $\mathbf{U} =\mathbf{E} + \mathbf{P} \in \mathbb{R}^{L \times d}$.

\section{Dataset Description}
\paragraph{Overview}
In this chapter, we use an anonymized vehicular DTC sequence dataset of $1.7 \times 10^6$ sequences with, on average, 150 DTCs per sequence. Each sequence belongs to a unique vehicle and the dataset is focused on a specific BMW model range.
An overview of the DTC elements\footnote{DTC composition defined in \ref{sec:dtc_composition} and notation in~\ref{tab:notation}} can be found in Table \ref{tab:data_details_dtc}.
\begin{table}[h!]
\centering
 \begin{tabular}{c c c}
 \hline
 Data & \# of values & Description \\ [0.5ex]
 \hline
 DTC & 8710 & Diagnostic Trouble Code  \\
 ECU & 61 & Electronic Control Unit  \\
 Base-DTC & 7726 & Error Code \\
 Fault-Byte & 2 & Binary Value  \\
 \hline
 \end{tabular}
 \caption{Number of Distinct values of the DTC elements}
 \label{tab:data_details_dtc}
\end{table}
\begin{figure}[!b]
    \centering
    \includegraphics[width=0.9\columnwidth]{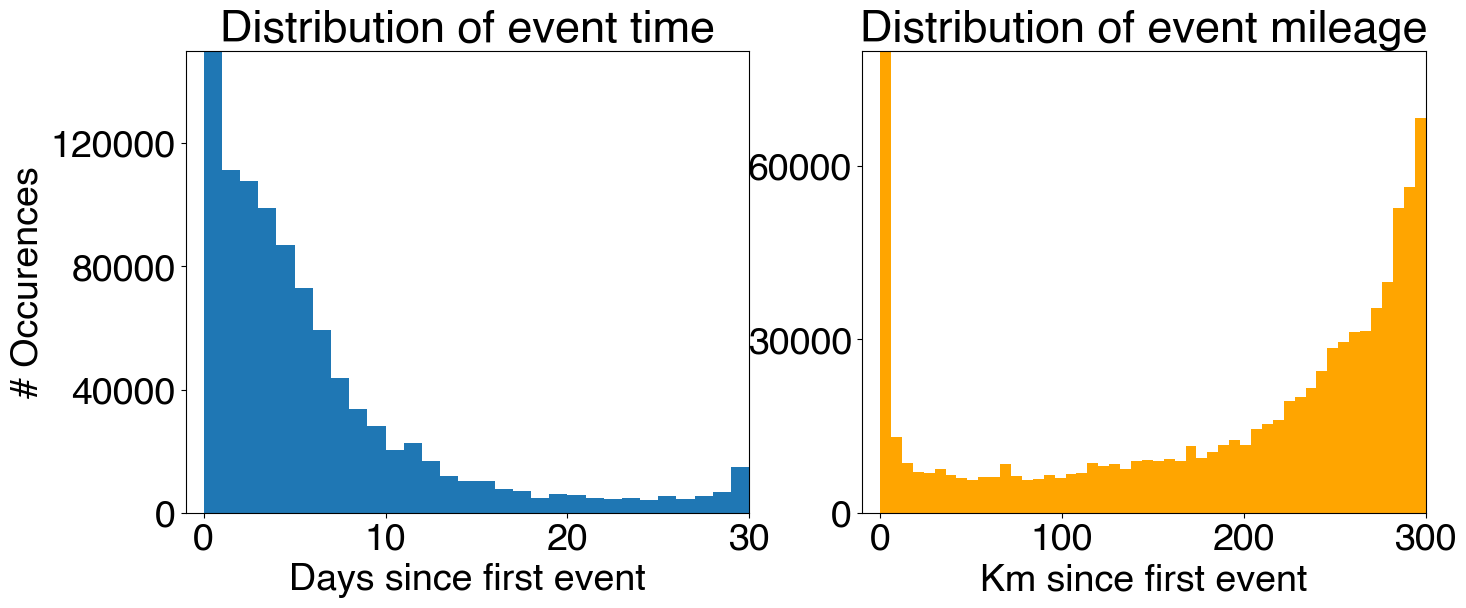}
    \caption{\textbf{Time and Mileage Distribution}. Distribution of $t_i$ and $m_i$ (relative time and mileage) in our dataset.} %
    \label{fig:aai_time_mileage}
\end{figure}
\subsection{Time and Mileage}
We plot the distribution of $t_i$ and $m_i$ in Fig.~\ref{fig:aai_time_mileage}.
We observe peaks at zero on both distributions due to truncation and missing values.

To feed our model with the time $t$ feature, we need a scaling method to mitigate the left-tail skew. Using $\log (t + 1)$ is a natural choice. At the same time, we want to approximately map $t$ into the range of $[-1, 1]$. Therefore, we apply the following non-linear function $f_t: \mathbb{R}^+ \to \mathbb{R}$ to $t$:
\begin{equation}\label{eq:ttransform}
    t' = f_t(t) = \log (t + 1) - 1\: \forall t \in \mathbb{R}^+
\end{equation}
In \glspl{tpp} the time is usually represented as inter-event time or its logarithm\cite{nppreview, rrnembedding}, making our approach anchored in the literature.

\section{Methodology}
We introduce the following two models: \emph{CarFormer} and \emph{EPredictor} for autoregressive DTC and EP prediction.

\subsection{CarFormer}
The CarFormer model can be seen in Fig.~\ref{fig:carformer}. The following defines the main architectural ideas.

\subsubsection{Embeddings}\label{c2:section:embeddings}
CarFormer uses four different embeddings to capture the spatial and temporal dependencies of irregular event appearance. These embeddings differ from the positional embedding $\mathbf{P}$ used in TPPs  \cite{transformerhawkeprocess}, \cite{multilabelpredictionfault}, and next-DTC prediction studies \cite{Hafeez2024DTCTranGruIT}, \cite{faultpredmultivariatevehicule}. We embed both time $t$ and mileage $m$ and use a rotation matrix to induce absolute and relative event positions such as:
\begin{itemize}
  \item \textbf{Event-type embedding} $\mathbf{E} \in \mathbb{R}^{L \times d}$ is obtained like described in section \ref{eventypeemb}.
  \item \textbf{Absolute time embedding} $\mathbf{T} \in \mathbb{R}^{L \times d}$ is constructed on-the-fly at each forward pass by a linear transformation $t_{i,j} = t'_i w_j + b_j$ where $w_j, b_j$ are learnable parameters and $t'_i$ is the scaled time at event step $i$.
  \item \textbf{Mileage embedding} $\mathbf{M} \in \mathbb{R}^{L \times d}$ is obtained via a learnable lookup table $\mathbf{W}^{m_{\text{max}} \times d}$ where $m_\text{max}=300$ km. Each row $\mathbf{w}_m \in \mathbb{R}^{d}$ corresponds to the learnable embedding vector for the discrete mileage $m$. The continuous mileage $m_i \in \mathbb{R}^{+}$ is cast to an integer value $m=\lfloor m_i \rfloor$. We attempted to leverage the continuous mileage by appropriately linearly interpolating the embeddings $w_m$ and $w_{m+1}$, but the time complexity increased without observing any benefits.

  \item \textbf{Rotary Position Embedding (RoPE)} $\mathbf{R}^d_{\Theta}$ Due to the permutation invariance of the Transformer model and the scattered time $t$, we still need to integrate positional event information. To do so, $\mathbf{Q, K}$ are rotated using the orthogonal matrix $\mathbf{R}^d_{\Theta}$ from \cite{Roformer} (with frequency parameters $\Theta$ defined below) as a function of the absolute event position $i$ in the sequence $S$. This method has two advantages: (1) it is not learnable (less likely to overfit), and (2) it integrates natively the relative position instead of altering $\mathbf{A}$ with a learnable bias like in \cite{relposatt}.
\end{itemize}

\noindent Some papers suggest using only a time embedding plus an event embedding \cite{transformerhawkeprocess, selfatthawke}. In contrast, and as in \cite{shou2024selfsupervisedcontrastivepretrainingmultivariate}, we argue that encoding solely the position with a time embedding leads to inefficient sequential learning for our Transformer model and to an early plateau in our pretraining.
We make the distinction between our event-type embedding $\mathbf{E}$ and the other information per event (time and mileage), which we call context embedding $ \mathbf{CE} = \mathbf{T + M}$.

\paragraph{Continuous Time Mileage Aware Attention }\label{customattention}
We modify the vanilla Transformer from \cite{tf} by (1) adding the context embedding to the projected event-type embedding at every layer \cite{touvron2023llamaopenefficientfoundation}, and then (2) a Rotary Position Embedding (RoPE) \cite{Roformer} is applied to both query \(\mathbf{Q}\) and key \(\mathbf{K}\):

\[
\mathbf{Q} = \mathbf{R}_{\Theta}^d (\mathbf{W}^Q \mathbf{E} + \mathbf{CE}),
\]
\[
\mathbf{K} = \mathbf{R}_{\Theta}^d (\mathbf{W}^K \mathbf{E} + \mathbf{CE})
\]

where\footnote{Note the reciprocal relationship between the standard PE frequency \(\omega_0\) and the RoPE base \(\theta_0\)} \(\Theta = \{\theta_i = \theta_0^{-2(i-1)/d}, i \in [1, 2, \ldots, d/2]\}, \theta_0 = 10^4\).
More specifically, the inner product between query \(\mathbf{q}_m\) and key \(\mathbf{k}_n\) takes the event-type embedding \(\mathbf{e}_m, \mathbf{e}_n\) where \(m-n\) is their relative position with context \(\mathbf{CE}_m\) and \(\mathbf{CE}_n\):
\begin{align}\label{eq:qkproduct}
     \mathbf{q}^T_m \mathbf{k}_n &= (\mathbf{R}^d_{\Theta, m}(\mathbf{W}_q \mathbf{e}_m + \mathbf{CE}_m))^T
    \mathbf{R}^d_{\Theta, n} (\mathbf{W}_k \mathbf{e}_n + \mathbf{CE}_n) \nonumber \\
    &= \mathbf{e}^T_m \mathbf{W}_q \mathbf{R}^d_{\Theta, n-m} \mathbf{W}_k \mathbf{e}_n +
    \mathbf{e}^T_m \mathbf{W}_q \mathbf{R}^d_{\Theta, n-m} \mathbf{CE}_n + \nonumber \\
    &\quad \mathbf{CE}^T_m \mathbf{R}^d_{\Theta, n-m} \mathbf{W}_k \mathbf{e}_n +
    \mathbf{CE}^T_m \mathbf{R}^d_{\Theta, n-m} \mathbf{CE}_n \nonumber \\
    &= \text{(1): query-to-key} + \text{(2): query-to-ce} \nonumber \\
    &\quad \text{(3): ce-to-key} + \text{(4): ce-to-ce}\nonumber \\
\end{align}
where $\mathbf{R}^d_{\Theta, n-m} = (\mathbf{R}^d_{\Theta, m})^T\mathbf{R}^d_{\Theta, n}$ is a sparse orthogonal matrix. The terms provide (1) standard attention capturing similarity between token embeddings, (2) query attending to the spatiotemporal information of the other token (vice versa for (3)) and (4) the interaction of the spatiotemporal information only of both token only. Consequently, they provide a richer query and key representations when computing the attention scores. The final the attention scores is computed as:
\begin{equation}
\mathbf{A} = \text{softmax}\left( \frac{(\mathbf{R}_{\Theta}^d (\mathbf{W}^Q \mathbf{E}+ \mathbf{CE})) (\mathbf{R}_{\Theta}^d (\mathbf{W}^K \mathbf{E} + \mathbf{CE}))^T}{\sqrt{3d}}\right)
\end{equation}
\noindent Adding \(\mathbf{CE}\) after the projection to query and key can be seen as a refinement of
\(\mathbf{Q}\), \(\mathbf{K}\) by \(\mathbf{T}\), \(\mathbf{M}\), providing additional context to the attention scores. We also add a scaling factor to compensate for the additional terms\footnote{Verified empirically in the ablation experiments (Table~\ref{tab:epredictor_overall}) via the \textit{rotcross-key-value-scaled-ce-2} model.} in Eq.~\ref{eq:qkproduct}.

\begin{figure}[t]
    \centering
\input{inc/part1/arch_styles}
\resizebox{0.72\columnwidth}{!}{\begin{tikzpicture}[x=1cm, y=1cm, -]
  \draw[archarr] (0.2,0) -- (8.6,0) node[right, archlbl] {Time};
  \foreach \x/\s in {2.4/rectangle, 3.6/diamond, 4.8/circle, 6.0/rectangle, 7.6/circle}
    \node[draw, \s, minimum size=2.2mm, inner sep=0pt, fill=white] at (\x,0) {};
  \node[archlbl] at (6.8,0.15) {$\cdots$};
  \foreach \x in {2.4, 3.6, 4.8, 6.0, 7.6} \draw[archarr] (\x,0.15) -- (\x,0.65);
  \node[archemb, minimum width=5.8cm] (ete) at (5.0,0.9) {Event-type Embedding $\mathbf{E}$};
  \node[archemb] (ce) at (0.8,0.9) {Context\\Embedding $\mathbf{CE}$};
  \draw[archarr] (0.8,0) -- (ce.south);

  \node[archnorm, minimum width=3.4cm] (n1) at (5.0,2.1) {RMS Norm};
  \draw[archarr] (ete.north) -- (n1.south);
  \node[archplus] (pk) at (5.0,2.95) {};
  \node[archplus] (pq) at (4.1,3.3) {};
  \node[archrope] (rq) at (4.1,3.95) {$\sim$};
  \node[archrope] (rk) at (5.0,3.95) {$\sim$};
  \node[archatt, minimum width=3.4cm] (att) at (5.0,4.85) {Masked Multi-Head Attention};
  \draw[thick] (4.1,2.35) -- (pq.south);
  \draw[thick] (5.0,2.35) -- (pk.south);
  \draw[thick] (pq.north) -- (rq.south);
  \draw[thick] (pk.north) -- (rk.south);
  \draw[archarr] (rq.north) -- (4.1,4.6);
  \draw[archarr] (rk.north) -- (5.0,4.6);
  \draw[archarr] (5.9,2.35) -- (5.9,4.6);
  \node[archlbl, anchor=east] at (4.05,2.62) {Query};
  \node[archlbl, anchor=west] at (5.05,2.62) {Key};
  \node[archlbl, anchor=west] at (5.95,3.5) {Value};
  \node[archlbl, anchor=east] at (3.85,3.95) {RoPE $\mathbf{R}^d_\Theta$};
  \draw[thick] (ce.north) -- (0.8,3.3);
  \draw[archarr] (0.8,3.3) -- (pq.west);
  \draw[archarr, preaction={draw=white, line width=3.5pt, -}] (0.85,2.95) -- (pk.west);
  \draw[thick] (0.8,2.95) -- (0.9,2.95);
  \fill (0.8,2.95) circle (0.045);
  \node[archlbl, anchor=south west, text=red!60!black] at (0.85,3.3) {$+\,\mathbf{CE}$};
  \node[archplus] (r1) at (5.0,5.65) {};
  \draw[archarr] (att.north) -- (r1.south);
  \fill (5.0,1.6) circle (0.045);
  \draw[thick] (5.0,1.6) -- (7.3,1.6) -- (7.3,5.65);
  \draw[archarr] (7.3,5.65) -- (r1.east);
  \node[archnorm, minimum width=3.4cm] (n2) at (5.0,6.55) {RMS Norm};
  \node[archffn, minimum width=3.4cm] (ffn) at (5.0,7.35) {Feed Forward};
  \node[archplus] (r2) at (5.0,8.15) {};
  \draw[archarr] (r1.north) -- (n2.south);
  \draw[archarr] (n2.north) -- (ffn.south);
  \draw[archarr] (ffn.north) -- (r2.south);
  \fill (5.0,6.05) circle (0.045);
  \draw[thick] (5.0,6.05) -- (2.9,6.05) -- (2.9,8.15);
  \draw[archarr] (2.9,8.15) -- (r2.west);
  \draw[dotted, very thick] (0.1,1.4) rectangle (8.4,8.5);
  \node[font=\small] at (1.2,7.2) {$N\times$};

  \node[archnorm, minimum width=3.4cm] (n3) at (5.0,9.1) {RMS Norm};
  \draw[archarr] (r2.north) -- (n3.south);
  \node[archlbl] (hs) at (5.0,9.85) {$H(S)$};
  \draw[archarr] (n3.north) -- (hs.south);
  \node[archhead] (h1) at (2.2,11.05) {Next Event Head};
  \node[archhead] (h2) at (5.0,11.05) {Next Time Head};
  \node[archhead] (h3) at (7.8,11.05) {Random Event Head};
  \draw[thick] (hs.north) -- (5.0,10.4);
  \fill (5.0,10.4) circle (0.045);
  \draw[thick] (2.2,10.4) -- (7.8,10.4);
  \draw[archarr] (2.2,10.4) -- (h1.south);
  \draw[archarr] (5.0,10.4) -- (h2.south);
  \draw[archarr] (7.8,10.4) -- (h3.south);
\end{tikzpicture}}
    \caption{\textbf{CarFormer Architecture}. A decoder-only Transformer with three predictive heads learns the dynamics of the DTC sequences. In every layer, the context embedding \(\mathbf{CE} = \mathbf{T} + \mathbf{M}\) (time and mileage) is added to the projected query and key, and the RoPE rotation \(\mathbf{R}^d_\Theta\) is applied afterwards, i.e., \(\mathbf{Q} = \mathbf{R}^d_\Theta(\mathbf{W}^Q\mathbf{E} + \mathbf{CE})\) and \(\mathbf{K} = \mathbf{R}^d_\Theta(\mathbf{W}^K\mathbf{E} + \mathbf{CE})\) (Section~\ref{customattention}). The values are not modified.}
    \label{fig:carformer}
\end{figure}

\subsubsection{Multi-Task Learning}

\paragraph{Next-Event Prediction.}
We use a standard language modeling objective which aims to minimize the cross-entropy loss between our output distribution $\hat{x}_i$ generated by our model's \textit{Next Event Head} and the next event $x_{i+1}$. \cite{shou2024selfsupervisedcontrastivepretrainingmultivariate} used a BERT \cite{bert} model trained on a masked event modeling task, which is commonly used for bidirectional models, in which each token attends to both preceding and following tokens. However, they simultaneously applied a causal mask, resulting in a loss of the bidirectional property. We argue that, by doing so, we lose substantial sample efficiency; thus, we will retain a standard next-token prediction.
$\hat{x}_i$ is the probability distribution produced by the \textit{Next Event Head}, which integrates an RMS normalization \cite{rmsnorm} and one linear layer.
The cross-entropy loss between $\hat{x}_i$ and $x_{i+1}$ (= a one-hot vector in $\{0,1\}^V$) is obtained by:
\begin{equation}
    \label{eq:nexteventtype}
    \mathcal{L}_c := -\sum_{i=0}^L \sum_{j=1}^V x_{i+1, j}  \log(\hat{x}_{i,j}) (1-\delta_{i,r})
\end{equation}
where $\delta_{i,r}$ is the Kronecker delta, iff event type \(x_i\) belongs to the set of random event types \(R\) and \(0\) otherwise.

\paragraph{Next Event Time Prediction.} In addition, we compute the Huber loss \cite{regloss} between the estimated inter-event time $\Delta \hat{t'}_i$ for the event $x_i$ and the ground truth $\Delta t'_i = f_t(t_{i+1}) - f_t(t_i)$, with $\beta=1, \epsilon_i = \Delta \hat{t}'_i - \Delta t'_i$. The Huber loss is quadratic for small errors and linear for large ones; it is therefore robust to outliers and prevents exploding gradients.
\begin{equation}
\mathcal{L}_t := \sum^L_{i=0} (1-\delta_{i,r})\begin{cases}
0.5 \epsilon^2_i  & \text{if } |\epsilon_i| < \beta, \\
 (|\epsilon_i| - 0.5) & \text{otherwise},
\end{cases}
\end{equation}
$t'$ is obtained using a log, hence we are essentially computing a kind of Mean Squared Logarithmic Error (MSLE) but with a $\beta$, useful to stabilize training and help convergence.

\paragraph{Random Event Prediction.}\label{randomevent} Finally, a binary classifier that predicts whether an event was true or randomly generated is added. This choice is motivated by several papers \cite{shou2024selfsupervisedcontrastivepretrainingmultivariate, guofakeevent} stating that a model should learn when an event does not happen to reinforce the negative evidence of no observable events within each inter-event.
At each step \( i \), a random event is injected with probability \( p \). If a random event is successfully injected, the process continues until a failure occurs (i.e., the event is not injected). This allows for multiple random events to be injected in a row, allowing for more complexity. The $\mathcal{L}_r$ loss is defined as the binary cross-entropy loss between the probability distribution $\hat{y}^r_i$ generated by our \textit{Random Event Head} and the ground truth $y^r_i$ at event step $i$:
\begin{equation}
\mathcal{L}_r := -\sum^{|R|}_{i=0} y^r_i \log(\hat{y}^r_i) + (1-y^r_i) \log (1-\hat{y}^r_i)
\end{equation}
\paragraph{Total Loss.}
The total loss is defined as follows:
\begin{equation}
 \mathcal{L} =  \frac{1}{L- |R|}(\mathcal{L}_c + k_1 \mathcal{L}_t) + k_2 \frac{1}{|R|} \mathcal{L}_r
\end{equation}
where $k_1, k_2$ are trade-off between the different loss terms, $L$ the sequence length, $R$ the set of random events injected in $S$.

\subsection{EPredictor}\label{epredictor}
We now describe the architecture of EPredictor in detail (Fig.~\ref{fig:epredictor}). We motivate this problem formulation and leverage the autoregressive property of CarFormer to introduce EPredictor.

\begin{figure}[t]
    \centering
\input{inc/part1/arch_styles}
\resizebox{0.62\columnwidth}{!}{\begin{tikzpicture}[x=1cm, y=1cm, -]
  \tikzset{archgap/.style={preaction={draw=white, line width=3.5pt, -}}}
  \node[archemb] (emb) at (5.0,-0.6) {Output Event-type Embedding (CarFormer)};
  \node[archnorm, minimum width=3.2cm] (n1) at (5.0,0.4) {RMS Norm};
  \draw[archarr] (emb.north) -- (n1.south);
  \node[archrope] (r1q) at (4.0,1.4) {$\sim$};
  \node[archrope] (r1k) at (5.0,1.4) {$\sim$};
  \node[archatt, minimum width=3.2cm] (att1) at (5.0,2.25) {Masked Multi-Head Attention};
  \draw[thick] (4.0,0.65) -- (r1q.south);  \draw[archarr] (r1q.north) -- (4.0,2.0);
  \draw[thick] (5.0,0.65) -- (r1k.south);  \draw[archarr] (r1k.north) -- (5.0,2.0);
  \draw[archarr] (6.0,0.65) -- (6.0,2.0);
  \node[archlbl, anchor=east] at (3.95,0.95) {Q};
  \node[archlbl, anchor=east] at (4.95,0.95) {K};
  \node[archlbl, anchor=east] at (5.95,0.95) {V};
  \node[archlbl, anchor=east] at (3.75,1.4) {RoPE};
  \node[archnorm, minimum width=3.2cm] (a1) at (5.0,3.05) {Add \& RMS Norm};
  \node[archffn, minimum width=3.2cm] (f1) at (5.0,3.85) {Feed Forward};
  \draw[archarr] (att1.north) -- (a1.south);
  \draw[archarr] (a1.north) -- (f1.south);
  \fill (5.0,-0.1) circle (0.045);
  \draw[thick] (5.0,-0.1) -- (7.3,-0.1) -- (7.3,3.05);
  \draw[archarr] (7.3,3.05) -- (a1.east);

  \node[archemb] (hs) at (1.15,4.7) {Output $H(S)$\\CarFormer};
  \node[archemb] (ce) at (1.15,5.6) {Output Context\\Embedding $\mathbf{CE}$};
  \node[archplus] (pk) at (4.0,5.35) {};
  \node[archplus] (pq) at (5.0,5.75) {};
  \node[archrope] (r2k) at (4.0,6.45) {$\sim$};
  \node[archrope] (r2q) at (5.0,6.45) {$\sim$};
  \node[archatt, minimum width=3.2cm] (att2) at (5.0,7.3) {Masked Multi-Head Attention};
  \draw[thick] (f1.north) -- (pq.south);
  \fill (5.0,4.35) circle (0.045);
  \node[archlbl, anchor=east] at (4.95,5.2) {Q};
  \draw[thick, archgap] (hs.east) -- (6.0,4.7);
  \fill (4.0,4.7) circle (0.045);
  \draw[thick] (4.0,4.7) -- (pk.south);
  \draw[archarr] (6.0,4.7) -- (6.0,7.05);
  \node[archlbl, anchor=east] at (3.95,5.0) {K};
  \node[archlbl, anchor=west] at (6.05,5.9) {V};
  \draw[archarr] (ce.east |- pk) -- (pk.west);
  \draw[archarr, archgap] (ce.east |- pq) -- (pq.west);
  \draw[thick] (pk.north) -- (r2k.south);  \draw[archarr] (r2k.north) -- (4.0,7.05);
  \draw[thick] (pq.north) -- (r2q.south);  \draw[archarr] (r2q.north) -- (5.0,7.05);
  \node[archlbl, anchor=east] at (3.75,6.45) {RoPE};
  \node[archnorm, minimum width=3.2cm] (a2) at (5.0,8.1) {Add \& RMS Norm};
  \node[archffn, minimum width=3.2cm] (f2) at (5.0,8.9) {Feed Forward};
  \draw[archarr] (att2.north) -- (a2.south);
  \draw[archarr] (a2.north) -- (f2.south);
  \draw[thick] (5.0,4.35) -- (7.3,4.35) -- (7.3,8.1);
  \draw[archarr] (7.3,8.1) -- (a2.east);

  \node[archhead] (h1) at (3.3,10.0) {Next EP Head};
  \node[archhead] (h2) at (6.7,10.0) {Next EP Time Head};
  \draw[thick] (f2.north) -- (5.0,9.45);
  \fill (5.0,9.45) circle (0.045);
  \draw[thick] (3.3,9.45) -- (6.7,9.45);
  \draw[archarr] (3.3,9.45) -- (h1.south);
  \draw[archarr] (6.7,9.45) -- (h2.south);
  \draw[thick] (h1.north) |- (5.0,10.6);
  \draw[thick] (h2.north) |- (5.0,10.6);
  \fill (5.0,10.6) circle (0.045);
  \draw[archarr] (5.0,10.6) -- (5.0,11.1);

  \draw[archarr] (0.3,11.25) -- (8.7,11.25) node[right, archlbl] {Time};
  \foreach \x in {0.8,1.5,2.2,2.9,3.6,4.3} \node[draw, circle, minimum size=2mm, inner sep=0pt, fill=white] at (\x,11.25) {};
  \node[draw, circle, minimum size=2.6mm, inner sep=0pt, fill=red!40] at (5.0,11.25) {};
  \node[draw, circle, minimum size=2mm, inner sep=0pt, fill=white] at (5.7,11.25) {};
  \node[archlbl] at (6.6,11.25) {$\cdots$};
  \node[archnorm, minimum width=3.4cm, minimum height=0.35cm, font=\scriptsize] at (2.2,11.7) {Minimum context $c$};
  \node[archlbl, anchor=west] at (5.1,11.7) {$P(\mathrm{EP}\mid S')$, $\hat{t}_{\mathrm{EP}}$};
  \draw[archarr, red!70!black] (0.5,12.15) -- (5.0,12.15) node[right, archlbl] {$S'$};
  \draw[archarr, red!70!black, densely dashed] (0.5,12.5) -- (5.7,12.5) node[right, archlbl] {$S''$ (next trajectory)};
\end{tikzpicture}}
    \caption{\textbf{EPredictor Architecture}. The first layer processes the normalized event-type embedding \(\mathbf{E}\) output by CarFormer. The second layer takes its query from the first layer, and its key and value from the hidden state \(H(S)\) extracted from CarFormer. As in CarFormer, the context embedding \(\mathbf{CE}\) carrying the spatio-temporal information is added to the projected keys and queries of the second layer \emph{before} the RoPE rotation. For a trajectory \(S'\) longer than the minimum context \(c\), the two heads jointly output the EP probabilities \(P(\mathrm{EP}\mid S')\) and the predicted time of occurrence \(\hat{t}_{\mathrm{EP}}\); the same prediction is then repeated on the next trajectory \(S''\), and so on.}
    \label{fig:epredictor}
\end{figure}

\subsubsection{Limitations of Next-Event Prediction}
Predicting only the next DTC in a\ sequence of DTC faults has inherent limitations and remains a difficult task. For instance, in~\cite{Hafeez2024DTCTranGruIT},
they used DTCs, whose ECU, Base-DTC, and Fault-Byte data have a cardinality of 83, 419, and 64, respectively, and only report a $81\%$ top-5 accuracy for next DTC prediction. This is because DTCs are not always correlated or causally linked.
Instead, we are also using repair and warranty data
to predict more important events, such as EPs (error patterns).
Repair and warranty data differ from DTCs since they are defined after observing all DTCs and characterizing a whole sequence $S$ and not an individual event $x_i$.

\subsubsection{Multi-Label Event Prediction}
We define that occurrence of error pattern $y$ at index $i=L$ in a sequence $S$. Note that multiple EPs can occur at the same time. We can now define a supervised multi-label classification learning problem of predicting EPs. Multi-label classification has attracted growing attention in the context of event prediction. \cite{multilabelpredictionfault} uses an LSTM for fault detection. More recently, \cite{Shou2023ConcurrentMP} treats concurrent event predictions as a multi-label classification problem and models such data using a Transformer architecture.

To define the multi-label event prediction task with $N$ labels ($\equiv$EPs), we reuse each $S = \{(x_i, t_i ,m_i)\}^L_{i=0}$ and attach a binary vector $\boldsymbol{y} \in \{0,1\}^N$ to indicate the EPs occurring at time $t_L$. Critically, \(\boldsymbol{y}\) is invariant per sequence $S$, meaning for all events within $S$ the ground truth $\boldsymbol{y}$ will be the same.
By applying a causal mask to both CarFormer and EPredictor, we enable predictive maintenance, since tokens can only attend to previous tokens, as shown in Figure \ref{fig:epredictor}.

With EPredictor, we leverage the \textit{seq2seq} (sequence-to-sequence) nature of Transformers: an input sequence is encoded into hidden states from which an output sequence is produced. Here, CarFormer outputs a sequence $H(S)$ of tokens encoded in a high-dimensional space $d$, positioning tokens with similar characteristics nearby. Then, this hidden representation $H(S)$ is fed into EPredictor, which acts as an autoregressive multi-label classifier for EPs. We approach EP prediction as a machine translation task. By utilizing the contextualized hidden states $H(S)$ from CarFormer as key and value (i.e., through \textit{cross-attention}), this effectively transitions our model from a "seq2seq" to a "dtc2ep" framework.

\subsubsection{Imbalanced Labels}

In our case, and in most real-world problems, EPs are highly imbalanced across the dataset, which makes the event prediction task considerably more challenging, especially when the least frequent classes are the most important to detect \cite{multilabelpredictionfault}. In natural language processing, traditional upsampling methods involve perturbing $S$ by shuffling and replacing tokens. However, in event data, we cannot afford to lose spatial and temporal information. Therefore, we inject random events $(x_i, m_i, t_i)$ with the same probability of $p=0.05$ as in Section \ref{randomevent}. We upsample the different EP classes to a minimum of $\theta_1 = 6000$ and downsample the most popular ones to a maximum of $\theta_2 = 12000$. We also drop classes below 100 appearances across the dataset.

\subsubsection{Losses}
We define a minimum context $c=30$  which acts as a “minimum trajectory” of DTCs to retain.
The \textit{Next EP Head} outputs a vector of probabilities  $\hat{y}_i = \text{sigmoid(MLP}_c( H_i))$ for each trajectory $\{H_1, ..., H_i\}$, $i \in {c,...,L}$, where $\text{MLP}_c$ is a multi-layer perceptron and $H_i \in \mathbb{R}^d$ is the generated hidden representation from EPredictor at step $i$. For the regression task we forecast the time until the EP(s) occurrence \( \Delta \hat{t'}_i \) = $\text{MLP}_t$(\( H_i\)) where the ground truth is \( \Delta t'_i = f_t(t_L, 30) - f_t(t_i, 30)\). Formally, we define our binary cross-entropy loss over the $N$ possible EPs for one step $i$ as follows:
\begin{equation}
\mathcal{L}^{ep}_i := -\frac{1}{N}\sum^{N}_{j=1} y_{j} \log(\hat{y}_{i,j}) + (1-y_{j}) \log (1-\hat{y}_{i,j})
\end{equation}
The total loss across $S$ with $L$ events and context $c$ is:
\begin{equation}
\mathcal{L}^{ep} := \frac{1}{L-c+1}\sum^{L}_{i=c} \mathcal{L}^{ep}_i
\end{equation}

We use the Huber loss \cite{regloss} and define $\epsilon_i = \Delta \hat{t}'_i - \Delta t'_i$, with $\beta=1$ thus:
\begin{equation}
\mathcal{L}^t := \frac{1}{L-c} \sum^{L}_{i=c} \begin{cases}
0.5 \epsilon^2_i  & \text{if } |\epsilon_i| < \beta, \\
|\epsilon_i| - 0.5 & \text{otherwise},
\end{cases}
\end{equation}
Our final loss to minimize is then: $\mathcal{L} = \mathcal{L}^{ep} + \gamma \mathcal{L}^ t$

\section{Experiments}\label{experiments}
We implemented CarFormer and EPredictor models using PyTorch, the code is publicly available\footnote{https://github.com/Mathugo/AAAI2025-CarFormer-EPredictor}.

\subsection{CarFormer Pre-training}\label{appendix:pretraining}
We trained the CarFormer model for 70,000 steps with a learning rate of \(5 \times 10^{-4}\), scheduled using a cosine warm restart with 10,000 warm-up steps, and a weight decay of 0.1. The loss coefficients were set to \(\alpha = 1\) and \(\beta = 1\). The model architecture included 12 attention heads, 6 layers, and a feature size of 600, utilizing the GELU (Gaussian Error Linear Unit) activation function in all feed-forward layers. We employed the AdamW optimizer with a batch size of 192 and a sequence length of 258, resulting in approximately 34 million parameters. It is important to note that on all experiments we fixed the same number of parameters when introducing new embeddings for all evaluated model to provide a fair comparison, same for EPredictor. The training data was split into $85\%$ training and $15\%$ testing without up-sampling, and random events were injected with a probability of \(p = 0.05\) per sample. On average, each training session lasted about 20 hours on an Nvidia A10G GPU.

\subsubsection{Ablation I: CarFormer Embeddings}
Multiple CarFormer models with different embedding choices were evaluated on next-token prediction accuracy (ACC) and on the regression task using mean absolute percentage error (MAPE) and root mean square error (RMSE) to determine the best-performing CarFormer model. The different choices were  \textit{rot} (RoPE), \textit{time} (absolute time embedding added to the input \(\mathbf{U}\)), \textit{mileage} (also added to \(\mathbf{U}\)), \textit{m2c}, and \textit{c2m} are additional dot-products (\cite{Deberta}).
In our case, we are more interested in the next-event prediction task, thus we are willing to sacrifice some MAPE \% over the ACC (\%).
\begin{table}[ht]
  \centering
  \begin{tabular}{lccc}
    \toprule
    \textbf{Model}&\textbf{ACC(\%)}&\textbf{MAPE(\%)}&\textbf{RMSE}\\
    \midrule
    \textbf{rot-ce} & \textbf{22.64} & 3.2 & 0.04770 \\
    time & 21.48 & \textbf{2.9} & \textbf{0.04762} \\
    time-mileage & 21.38 & 3.0 & 0.04785 \\
    time-c2m-m2c & 21.58 & 3.5 & 0.04794 \\
    time-m2c & 21.52 & 3.6 & 0.04823 \\
    GPT & 19.89 & - & - \\
    \bottomrule
  \end{tabular}
  \caption{Overall prediction performance of CarFormer with different embeddings. Best results are in bold.}
  \label{tab:pretraining_exp}
\end{table}

\noindent Using only \textit{time} gave the best MAPE (\(2.9\)) but not the best ACC (\(21.48\)), suggesting that other features might improve the model predictions. For the mileage integration, our intuition was that doing an early summation of the two embeddings ($\mathbf{T}, \mathbf{M}$) seemed to denature the input $\mathbf{U}$ (ACC of \textit{time-mileage} $<$ ACC of \textit{time}), but the mileage of the vehicle could help differentiate between different DTCs. We tried to modify the attention dot products, which seemed to help the ACC a bit (\textit{time-c2m-m2c}, \textit{time-m2c}), but increased the MAPE drastically. So we incorporated $\mathbf{CE}$ directly in $\mathbf{Q, K}$. Then, by applying a RoPE to the transformed input, the ACC increased while preserving the RMSE, leading to the best performing model in terms of ACC, namely: $\textit{rot-ce}$.

\begin{table*}[ht]
  \centering
  \footnotesize %
  \setlength{\tabcolsep}{4pt} %
  \begin{tabular}{lcccccc}
    \toprule
    \textbf{Model} & \textbf{F1} & \textbf{MAPE} & \textbf{MAE} & $\textbf{CPMW}_{F1} \uparrow$ & $\textbf{CPMW}_{MAE} \downarrow$ \\
    \midrule
    rotcross-query-key-ce-1-2 & 82.69 & 32.45 & 0.0268 & 52.80 & 0.888 \\
    rotcross-query-key-ce-2 & 82.69 & \textbf{31.18} & 0.0254 & 56.61 & 0.882 \\
    rotcross-query-ce-2 & 82.69 & \textbf{31.18} & 0.0254 & 53.00 & 0.884 \\
    rotcross-key-value-ce-2 & \textbf{84.38} & 33.47 & 0.0263 & 65.06 & 0.904 \\
    \textbf{rotcross-key-value-scaled-ce-2} & \textbf{84.38} & 31.44 & \textbf{0.0252} & \textbf{67.63} & \textbf{0.874} \\
    rotnocross-ce-1-2 & 80.74 & 37.61 & 0.0275 & 42.95 & 0.927 \\
    cross-speed & 83.53 & 34.73 & 0.0260 & 49.28 & 0.877 \\
    cross-mixffn & 83.41 & 33.37 & 0.0270 & 54.39 & 0.891 \\
    time-cross-query & 83.34 & 35.89 & 0.0275 & 45.71 & 0.896 \\
    \bottomrule
  \end{tabular}
  \caption{EPredictor evaluation results with different model architectures on the test set (no up- nor down-sampling). The MAPE and F1 are shown in (\%)}
  \label{tab:epredictor_overall}
\end{table*}

\subsection{EPredictor Experiments}
\begin{figure}[ht]
    \centering
    \includegraphics[width=0.8\columnwidth]{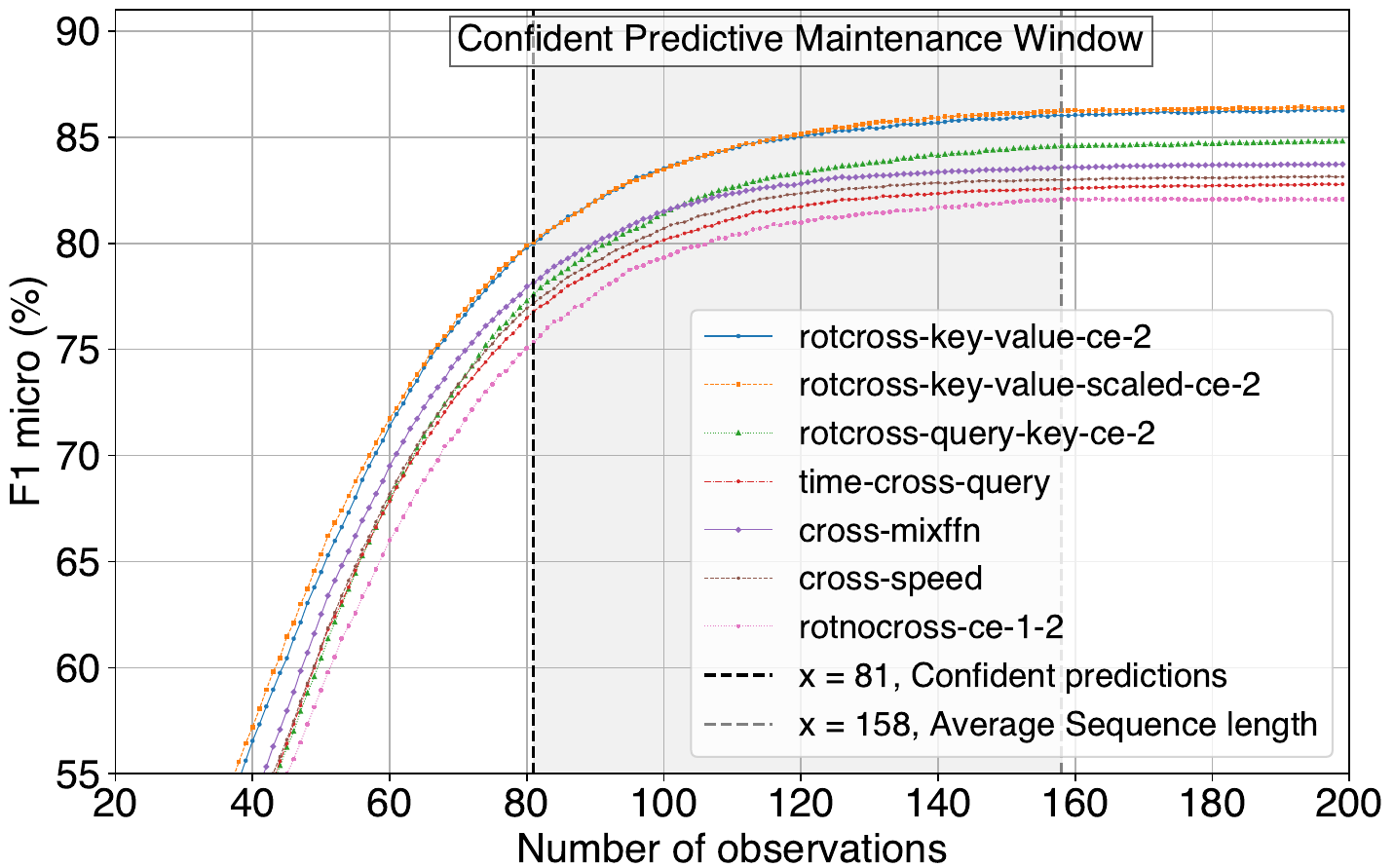}
    \caption{\textbf{Classification Performance Comparison Between Architectural Changes}. F1 Score comparison with multiple EPredictor architectures as a function of the number of observations. This answers \textit{what} EP is most likely to occur.}
    \label{fig:exp_epredictor_arch}
\end{figure}

We evaluate multi-labeled classification using the micro-F1 score~\cite{reviewmultilabellearning}. To better understand and enhance our model's predictive maintenance capabilities, we introduce the concept of \textit{Confident Predictive Maintenance Window (CPMW, Def.~\ref{def:cpmw})}, which represents the interval within which our model can make reliable predictive maintenance predictions.
We quantify this with the \textit{CPMW Area Under Curve} (\(\textit{CPMW}_\zeta\)). The F1 score, MAE, and MAPE have been calculated on average for all observations in Table \ref{tab:epredictor_overall}, and additionally for each trajectory $H^i_t$ in Fig.~\ref{fig:exp_epredictor_arch} to understand how each model performs with different numbers of observations. To monitor the predictive maintenance capability of each model, the $\text{CPMW}_{\text{F1}}$ and $\text{CPMW}_{\text{MAE}}$ were computed.
\subsection{Ablation II: EPredictor Architecture \& CPMW}
\paragraph{Embedding Architecture and Cross-attention}
We explored several architectural changes and their impact on the CPMW. We applied a RoPE (\textit{rot}) and a cross-attention with \textit{query} or \textit{key} or \textit{value} to the second multi-head attention block (\textit{cross}), and added the context embedding $\mathbf{CE}$ (\textit{ce}) to layer \textit{1} and/or \textit{2}. We also applied a scaling factor of $\sqrt{3d}$ to $\mathbf{Q, K}$ as shown in Eq.~\ref{eq:vanilldotprodattention} (\textit{scale}), injected a relative matrix $\mathbf{S}_{rel}$ into the attention scores (\textit{speed}),
and applied a mixed feed-forward network (\textit{mixffn}) \cite{NEURIPS2021_64f1f27b} to the mileage embedding. The \textit{time} model refers to $\mathbf{T}$, which is also added to $\mathbf{E}$. Finally, we trained a model with and without the \textit{Random Event Head}.

\paragraph{Results}
Our experiments revealed several key insights:
By applying a \textit{cross} attention, we can see improvement in all metrics (\textit{rotnocross-ce-1-2}), which is consistent with the machine translation analogy \textit{"dtc2errorpattern"}. The best cross-attention results were obtained when $H(S)$ was used as the \textit{key-value}. Adding the mileage via an MLP layer (\textit{cross-mixffn}) seemed to help the MAPE (-2.5\%), the $\text{CPMW}_{\text{F1}}(+9)$ and also the $\text{CPMW}_\text{MAE}(-0.005)$ compared to \textit{time-cross-query} model, suggesting that mileage is beneficial for both tasks. This is consistent since EPs are also dependent on the traveled distances between DTCs and the different stationary behavior of the vehicle. Furthermore, models incorporating a RoPE (\textit{rot}) performed significantly better in both regression and classification tasks, as in the pre-training, highlighting the performance of RoPE in machine translation tasks \cite{Roformer}. Adding $\mathbf{CE}$ to the last layer (\textit{ce-2}) yielded the best results, as opposed to adding it to both layers (\textit{ce-1-2}). Surprisingly, doing feature engineering on $\mathbf{T}, \mathbf{M}$ with a \textit{speed} matrix $\mathbf{S}_\text{rel}$ did not help the metrics, which could indicate some missing modalities (e.g., mileage) during training, thus $\mathbf{CE}$ is more adapted for real-world scenarios. We monitored the need for our \textit{Random Event Head} and noticed $+1.2\%$ in the F1 Score and $+10.1$ in the $\text{CPMW}_{f1}$.

\begin{figure}[ht]
    \centering
    \includegraphics[width=0.8\columnwidth]{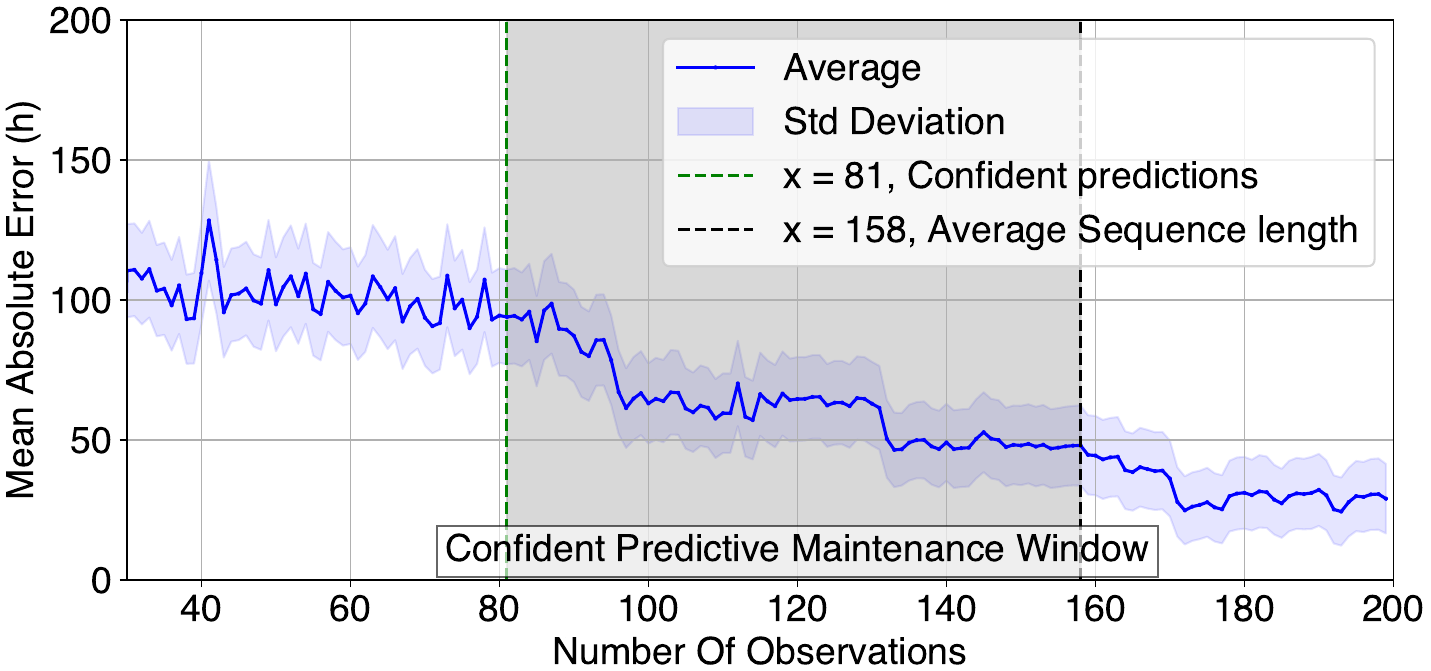}%
    \caption{\textbf{Regression Error As a Function of the Number of Observations}. Evolution of the MAE as a function of the number of observations for the best performing model, this effectively tests for \textit{when} an EP is most likely to occur.}
    \label{fig:mae_h}
\end{figure}

\noindent When taking the best performing model (\textit{rotcross-key-value-scaled-ce-2}), the model entered the CPMW after 81 observations, i.e., \textbf{half of the sequence}. Within this window, the model obtained an error of $\mathbf{\approx 65 \pm 14}$\textbf{h} when estimating the time of EP occurrence (Figure \ref{fig:mae_h}), highlighting the model's predictive capability within the CPMW. Fig.~\ref{fig:mae_h} shows the MAE (line) and its standard deviation (shaded band) as a function of the number of observed events: the grey area marks the CPMW, from \(81\) observed events to the average sequence length of \(158\) events. The error stays around \(100\)h before this window, decreases to about \(50\) to \(65\)h inside it, and to about \(30\)h for longer sequences. Otherwise, the average absolute error across all observations was approximately $58.4 \pm 13.2$h.
By experimenting with these modifications, we aimed to identify the optimal architecture for predictive maintenance. The findings reveal that cross-attention, context embedding in the second layer, and scaled attention significantly improve performance within the CPMW.

\section{Summary}
This chapter presented a novel approach to predictive maintenance in vehicles by modeling diagnostic event sequences as a language.
By introducing two Transformer-based architectures, \emph{CarFormer} and \emph{EPredictor}, we demonstrated that it is possible to jointly forecast both \emph{when} and \emph{what} error patterns are likely to occur, even in the presence of highly unbalanced, irregular, and high-cardinality diagnostic data.
Our results show that autoregressive modeling of DTC sequences provides a powerful means to capture the temporal and contextual dependencies underlying vehicle faults.
The proposed framework effectively reconciles traditional event sequence modeling and large-scale fault prediction, moving toward a unified, data-driven foundation for predictive maintenance.  Beyond its experimental performance, the proposed system has clear practical implications.
In principle, once deployed, \emph{EPredictor} can operate in real-world scenarios by continuously analyzing DTC streams and estimating the probability and timing of forthcoming error patterns.
When a sufficiently confident prediction is made, the vehicle can proactively notify the user of a potential critical fault and recommend a visit to a service center.
Such a system has the potential to enhance vehicle safety, reduce maintenance costs, and ultimately support the broader vision of intelligent and self-diagnostic automotive systems.  The pretrained CarFormer and EPredictor backbones are directly applicable for a broad range of downstream tasks, including sequence clustering, anomaly detection, and causal discovery, as demonstrated in the subsequent parts of this thesis. %

\section{Outlook}
Several limitations remain. First, predictive performance is sensitive to the quality and convergence of CarFormer's pretraining; a poorly trained backbone degrades EPredictor accordingly. Second, the CPMW metric shows that reliable predictions require observing approximately half the sequence, which may be too late for certain use cases. Third, the highly imbalanced distribution of error patterns means that rare but critical EPs remain harder to detect, despite the upsampling strategy employed. These constraints motivate the work presented in the following chapter, which incorporates additional modalities to improve the classification of overlapping and rare error patterns.

\chapter{Multimodal Sequence Modeling for Error Pattern Classification}\label{c4:multimodal}
\glsresetall
\noindent
Chapter~\ref{c2:ep_prediction_based_on_live_data} demonstrated that DTC sequences alone provide a strong predictive signal, but highlighted a key failure mode: error patterns sharing identical Boolean DTC rules cannot be disambiguated from the event stream alone. Chapter~\ref{c4:multimodal} addresses this limitation directly, by introducing environmental sensor data as a complementary modality. Naturally, contextual information such as raw sensory data (e.g., temperature, humidity, and pressure) is beneficial for engineers to classify sequences of error codes into vehicle failures. Yet, it introduces unique challenges due to its complexity and the noisy nature of real-world data. This chapter is based on the following publication:
\newline

\noindent\textbf{Context-Informed Sequence Classification: A Multimodal Approach to Vehicle Diagnostics.}
\cite{math2026contextinformed}
Hugo Math, Rainer Lienhart, Workshop on \emph{Time Series in the Age of Large Models} at ICLR 2026, April 2026.
\newline

\noindent This chapter introduces BiCarFormer: a multimodal approach to multi-label sequence classification of error codes into EPs that integrates DTC sequences and environmental conditions. BiCarFormer is a bidirectional Transformer model tailored for vehicle event sequences, employing embedding fusions and a co-attention mechanism, in which the DTC and environmental sequences attend to each other, to capture the relationships between diagnostic codes and environmental data. %

\section{Introduction}

Diagnostic experts routinely find it necessary to incorporate contextual information as complex EPs seem to overlap in their rule definition (Eq.~\ref{eq:ep_def}).
Prior work relies only on DTCs to infer the next DTCs (using Transformers \cite{tf, Hafeez2024DTCTranGruIT} and RNNs \cite{faultpredmultivariatevehicule}).
While environmental data can enhance the classification or clustering of EPs, it poses integration challenges due to its volume, variability, and high cardinality. Modern time series foundation models (TSFMs) such as Chronos~\cite{ansari2024chronoslearninglanguagetime} operate on one modality and on a continuous event sequence. Vehicle data is intrinsically multimodal, consisting of continuous signals (voltage, engine speed in revolutions per minute) interleaved with discrete event codes (DTCs), a modality mix that purely continuous TSFMs still struggle to integrate. Our goal is to uncover the appearance of DTCs and environmental conditions that correlate with certain EPs.
We demonstrate that our approach significantly improves the multi-label sequence classification of EPs, outperforming traditional models that rely solely on DTCs \cite{Hafeez2024DTCTranGruIT, faultpredmultivariatevehicule} and classical Transformers such as BERT \cite{bert}. We further explain classification predictions by interpreting cross-attention scores. We show that BiCarFormer learns to detect fluctuations of quantized continuous values, despite encoding different units and thousands of different environmental conditions.

\section{Related Work}
\subsection{Failure Detection in Vehicles}
\noindent Historically, machine learning models have been employed to detect machine failures. It includes techniques based on correlation \cite{7998309}, Bayesian networks \cite{LANGSETH200792}, decision tree models \cite{logbasedpredictivemaitnenancetree}, and more recently neural state-space models \cite{HE2023109598} to predict machine failures or remaining useful life (RUL). These methods rely on historical data to identify patterns and predict potential failures. Bayesian networks can model the probabilistic dependencies among various components of the vehicle. Some papers explore the use of Deep Neural Networks (DNNs). For example, \cite{Zhou2023} uses a combination of convolutional, fully connected, and Transformer modules within one architecture for the classification of machine event logs.

In Chapter~\ref{c2:ep_prediction_based_on_live_data}, we suggested using the warranty data of vehicles to create a supervised learning problem where we attach EPs to each DTC sequence. We further model these multivariate vehicle event sequences and focus on classifying error patterns to benchmark the proposed methods.

\subsection{Multimodal Fusion \& Learning}
\noindent Introducing other modalities in a Transformer model has been extensively studied~\cite{transformermultimodallearningreview, transformerbottleneck}.
One common and simple way to fuse two token embeddings $\boldsymbol{X}_A \in \mathbb{R}^{L_a \times d}$ and $\boldsymbol{X}_B \in \mathbb{R}^{L_b \times d}$ from modality A and B (assumed to be sequences) is via \emph{early summation}. Specifically, \emph{token-wise weighted summing} of multiple embeddings at the input level is defined as:
\begin{equation}
\boldsymbol{U} = \alpha \boldsymbol{X}_a \oplus \beta \boldsymbol{X}_b
\end{equation}
In MedBERT \cite{MedBERT}, the authors summed three types of embeddings (diagnostic codes, the order, and the position of each visit) to form input $\boldsymbol{U}$. This method has several advantages: it is simple to integrate into a Transformer and does not significantly alter the computation. However, the different tokens need to be aligned or projected to a latent space if $L_a \neq L_b$, and it is not clear how one should weigh the different embeddings when summing using $\alpha, \beta$. For example, BERT uses an early summation of its token embedding $\mathbf{E}$ and its position $\mathbf{PE}$ such that $\boldsymbol{U} = \boldsymbol{E} + \boldsymbol{PE}$.

In the diagnostic domain, DTC-TranGRU~\cite{Hafeez2024DTCTranGruIT} uses three separate DTC components embeddings that they concatenate along $d$ to form a global DTC embedding $\boldsymbol{D} \in \mathrm{R}^{L \times d}$ (we refer to this method as \emph{early concatenation}).

\begin{equation}
    \boldsymbol{U} = \textit{concat}(\boldsymbol{X}_a, \boldsymbol{X}_b) \in \mathbb{R}^{L \times (d_b+d_a)}
\end{equation}
\noindent This enables finer-grained integration of additional features by selecting their embedding sizes, $ d_a$ and $ d_b$. It also preserves the distinct characteristics of each feature by maintaining separate dimensions, thus potentially learning more nuanced representations and interactions in the deeper attention layers. However, it usually requires more computation since we concatenate along $d$, resulting in more parameters~\cite{transformermultimodallearningreview} for the point-wise feed-forward layers and the projection matrices. It also does not work well with low cardinality features such as token type ids from BERT~\cite{bert} because it will only span across a very small size in the hidden dimension $d$, thus early summation is preferable for this case.

By combining embeddings at an early stage, models can learn a unified representation that captures the interactions between different features. This can be particularly useful if the features are highly correlated. These methods are used in a wide variety of domains, in particular in medicine \cite{MedBERT, multidimpatientacuityestimation} with electronic health records (EHR).
When dealing with complex modalities such as audio or images that are not necessarily aligned, the attention mechanism of Transformers \cite{tf} is usually used for fusion. Late fusion in Transformers usually involves encoding the different modalities through independent Transformer encoders to extract high-level representations $\boldsymbol{Z}_a, \boldsymbol{Z}_b$:
\begin{align*}
     \boldsymbol{Z_a} &= \textit{Tf}_a(X_a) \in \mathbb{R}^{L_a\times d}
     \\
    \boldsymbol{Z}_b &= \textit{Tf}_b(X_b) \in \mathbb{R}^{L_b \times d}
\end{align*}
Then, these representations can be concatenated and encoded via another third Transformer (\emph{Hierarchical Attention}) to output a fused representation using multi-head attention:
\begin{equation*}
\boldsymbol{Z} = \textit{Tf}(\textit{concat}(\boldsymbol{Z}_a, \boldsymbol{Z}_b)) \in \mathbb{R}^{L_a \times (d_a + d_b)}
\end{equation*}

\noindent One can also fuse $\boldsymbol{Z}_a, \boldsymbol{Z}_b$ with similarity products like in CLIP \cite{clip}.
The CLIP model introduces a new self-supervised multimodal learning task where the model learns which caption goes with which image and demonstrates state-of-the-art performance for 30 different existing computer vision datasets. The problem with this method is its limited cross-modal interaction, where we first encode separately the modalities to fuse them later. Perhaps the model would benefit from learning multimodal dependencies earlier in the architecture.

\noindent At the other end, middle fusion (or \emph{mid fusion}) methods involve \emph{cross-attention} and more generally \emph{co-attention} mechanisms. In ViLBERT \cite{vilbert} and LXMBERT \cite{lxmbert}, the attention models enable computation of the attention scores $\boldsymbol{A}$ as a function of the image and text input. For example, using $\boldsymbol{X}_a$ as query $\boldsymbol{Q}_a$ and $\boldsymbol{X}_b$ as $\boldsymbol{K}_b, \boldsymbol{V}_b$, we can compute the resulting $\boldsymbol{C}_{a \longrightarrow b} \in \mathbb{R}^{L_a \times d}$ and vice versa (one input attends to another). In \emph{co-attention} both inputs attend to each other by computing two attentions simultaneously ($\boldsymbol{C}_{a \longrightarrow b}, \boldsymbol{C}_{b \longrightarrow a}$), which enables multimodal learning in both ways. One caveat to these methods is that by not projecting $\boldsymbol{X}_a, \boldsymbol{X}_b$ into a fixed attention scores matrix but computing a new $\boldsymbol{A} \in \mathbb{R}^{L_a \times L_b}$, if $L_b \gg L_a$, we drastically increase the computational overhead due to the quadratic time complexity $\mathcal{O}(L^2)$ of the \emph{vanilla} attention. We also get two hidden representations for each modality: $\boldsymbol{H}_{a}, \boldsymbol{H}_{b}$.

The product $\boldsymbol{Q}\boldsymbol{K}^T$ carries most of the computation, thus one may consider selecting specific query and key multiplications, resulting in a sparse attention variant like LongFormer \cite{beltagy2020longformer}, or BigBird \cite{bigbird}. These models use a sliding, global, and random attention combination, reducing $\mathcal{O}(L^2)$ to a linear complexity of $\mathcal{O}(w \times L)$ for LongFormer. Another method is to project the keys and values into a lower-dimensional space, like in Linformer \cite{wang2020linformer}, or use a multimodal bottleneck Transformer \cite{transformerbottleneck} to compute the cross-attention partially in a restricted latent space.
Feature fusion in event sequence models has received comparatively little attention in the literature. The majority of papers aim at fusing the time $t_i$ information per event type $x_i$ by performing an \emph{early summation} \cite{Zhou2023, selfatthawke, transformerhawkeprocess} to integrate the time component. Some omit the positional embedding in favor of the time embedding \cite{transformerhawkeprocess}, some sum both \cite{shou2024selfsupervisedcontrastivepretrainingmultivariate, sparsetemporalattention}, and others do not take time into account \cite{Hafeez2024DTCTranGruIT}.

\section{Dataset Extension with Environmental Conditions}
We extend the Chapter~\ref{c2:ep_prediction_based_on_live_data} dataset with 5 million sequences, containing an average of $L \approx 150 \pm 90$ DTCs per sequence from different BMW model ranges. Each sequence $S_{raw}$ belongs to a unique vehicle. In $S_{raw}$, each DTC occurrence is attached additionally with a small sequence $S_e^{(i)} = \{(d_j, v_j, c_j)\}_{j=0}^{L^{(i)}_e}$ of environmental conditions, respectively with elements description, value, and unit.

These elements construct a single event $(x_i, t_i, m_i, S^{(i)}_e)$. To obtain a complete marked event sequence $S_{raw} = \{(x_i, t_i, m_i, S^{(i)}_e)\}_{i=0}^L$ (Def.~\ref{def:marked_event_sequences}), we apply the same 30-day, 300-km observation window as in Chapter~\ref{c2:foundation_event_sequence_modeling}.
We then split $S_{raw}$ into two distinct sequences: $S = \{(x_i, t_i, m_i)\}_{i=0}^L$ for the DTCs, and $S_e = \{(d_i, v_i, c_i)\}^{L_e}_{i=0}$ for only the environmental conditions with length $L_e \approx 2275 \pm 2310$, making it highly unaligned with \(S\) and orders of magnitude longer. An overview of the elements is provided in Table~\ref{notation} as well as the number of distinct values for each feature in Table~\ref{distinct_value_des}.
\begin{table}[!b]
\begin{center}
    \begin{tabular}{ll}
        \hline
        \textbf{Notations} & \textbf{Description} \\
        \hline
        $S_e$ & Sequence of environmental conditions triplets defined as \\
             & $S_e = \{(d_i, v_i, c_i)\}_{i=0}^{L_e}$ of length $L_e \gg L$ \\
        $d$ & Environmental condition description, e.g., temperature \\
            & increase, vehicle speed, pressure increase \\
        $v$ & Environmental condition value can be int, float, string, null \\
        $c$ & Environmental condition unit, e.g., A, V, bar, °C, sec\\
        \hline
    \end{tabular}
    \end{center}
    \caption{New symbols introduced in this chapter. For base DTC notation, see Table~\ref{tab:notation}}
\label{notation}
\end{table}

\subsection{Integration} \label{data_env}
\noindent Incorporating sensory information efficiently, such as temperature or pressure value alongside discrete codes, remains an open problem \cite{Zhou2023, Hafeez2024DTCTranGruIT}. Intuitively, it might seem trivial to rely on temperature for diagnosing a specific defect in a vehicle engine or a voltage measurement in a battery to identify a cell failure. Experts often analyze this data to make decisions about EPs in vehicles. However, the integration of environmental conditions comes with three significant challenges:
\begin{enumerate}
    \item \emph{Cardinality:} there are many distinct condition descriptions ($>10^3$), and values can be multi-types: strings, Booleans, integers, floats, or NaN.
    \item \emph{Variability:} differs across each DTC: e.g., we cannot be certain that we will observe 'temperature' for DTC1 and 'pressure' for DTC2.
    \item \emph{Volume:} there are multiple environmental conditions per DTC, which can be redundant, noisy, and duplicated across the sequence $S_e$.
\end{enumerate}

\begin{table}[!b]
\center
\begin{tabular}{|l|r|p{6cm}|}
\hline
\textbf{Data} & \textbf{\# of values} & \textbf{Description} \\ \hline
DTC & 22,137 & Diagnostic Trouble Code \\
ECU & 132 & Electronic Control Unit \\
Base-DTC & 17,044 & Error Code \\
Fault-Byte & 2 & Binary Value \\
E. Condition Description & 2,559 & (See Table 1: $d$) \\
E. Condition Value & 3,288 & (See Table 1: $v$) \\
E. Condition Unit & 18 & Most popular units: $u$\\ \hline
\end{tabular}
\caption{Number of distinct values after filtering for each feature.}
\label{distinct_value_des}
\end{table}

The \emph{Cardinality} problem increases model complexity and often necessitates extensive data engineering. The \emph{Variability} makes integration into a machine learning model challenging. Lastly, the \emph{Volume} requires substantial data infrastructure to manage environmental conditions, since it is often several times the size of DTC data and significantly increases the sequence length, a known challenge in Transformers \cite{sparsetemporalattention}. We partially address these challenges by:
\begin{itemize}
    \item removing \textit{null} values, units, or descriptions.
    \item removing duplicates per DTC: the same $(d, v, c)$ triplet occurring multiple times within the same $S_e^{(i)}$ is removed, retaining only one instance. This significantly reduced redundancy and noise in the data, though we still observed $L_e \approx 2275 \pm 2310$ triplets per $S_e$. One caveat of this approach is the \textit{nonalignment} of environmental conditions and DTC when multiple DTCs occur at the same $t_i$. In such cases, we drop subsequent environmental conditions after the first occurrence.
    \item selecting the top-$18$ most popular units.
\end{itemize}

\noindent To ingest the resulting $S_e$ into a Transformer, we map our continuous values $v \in \mathbb{R}^+$ into discrete \textit{tokens} \cite{transformermultimodallearningreview} with a limited vocabulary. Therefore, we search for the optimal bins for each unit $c$. A variation of the Greenwald-Khanna algorithm \cite{spaceefficientonlinecomputationofquantile} %
up to a maximum of $4000$ tokens per unit $u$ to limit a potential $18 \times \theta = 72000$ values.
In practice, they overlap and do not reach the maximum (Table \ref{distinct_value_des}).

\section{BiCarFormer}

\subsection{DTC Embeddings}
\label{sec:dtc_model_section}
\noindent We use a Bidirectional Transformer model \cite{bert} that we train with mask language modeling. Each DTC element is embedded in a specific feature space, namely: \emph{ECU} ($\boldsymbol{D}_{ecu} \in \mathbb{R}^{L \times d_{ecu}}$), \emph{Base-DTC} ($\boldsymbol{D}_{base} \in \mathbb{R}^{L \times d_{base}}$), and the \emph{Fault-Byte} ($\boldsymbol{D}_f \in \mathbb{R}^{L \times d}$) using separate lookup tables. The first two are concatenated along the feature dimension, like in \cite{Hafeez2024DTCTranGruIT}, and $\boldsymbol{D}_f$ is added to the result like a token type id in BERT \cite{bert}. This reduces the vocabulary size and trainable parameters compared to \cite{math2024harnessingeventsensorydata} and ensures that the embeddings with high cardinality are preserved independently, thus capturing more relationships between the \textit{ECU}, the \textit{Base-DTC}, and the \textit{Fault-byte}. We obtain the input DTC embedding $\boldsymbol{D} \in \mathbb{R}^{L \times d}$:
\[
\boldsymbol{D} = \text{concat}(\boldsymbol{D}_{base}, \boldsymbol{D}_{ecu}) + \boldsymbol{D}_f
\]
During the pretraining phase, we mask only the Base-DTC token, since it provides the primary information.
\subsection{Positional Embeddings}
\begin{figure}[!h]
      \centering
      \includegraphics[width=3.9in]{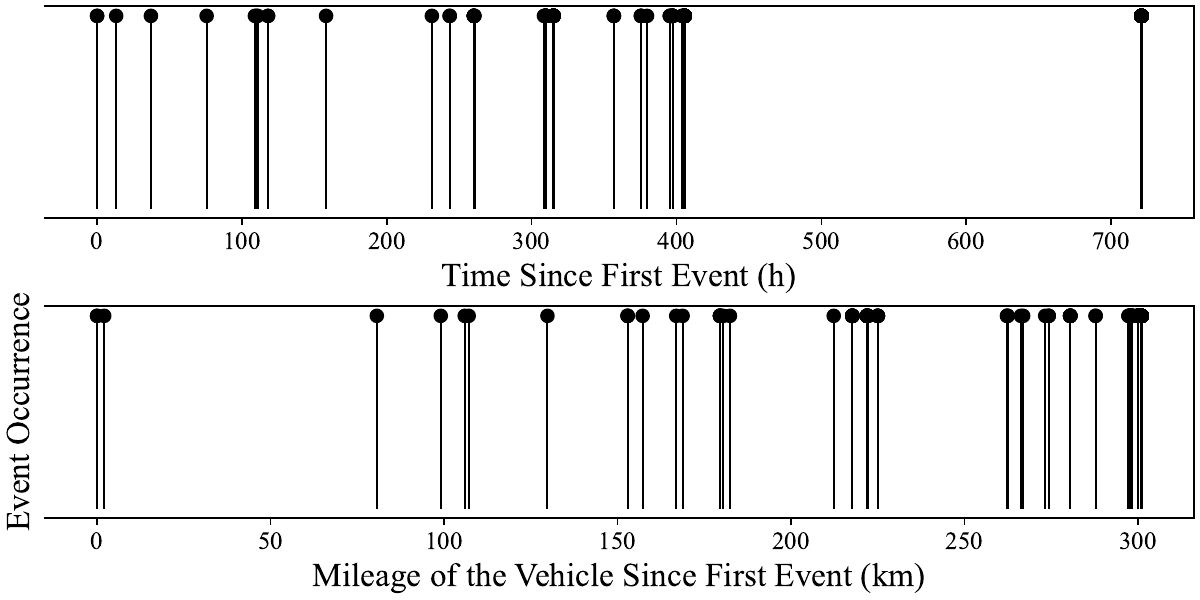}
      \caption{\textbf{Spatio-temporal Point Process Representation of a Vehicle.} Spatial (mileage) and Temporal (time) point process representation of events from a vehicle. Bold vertical lines indicate multiple events occurring at the same time $t_i$ or mileage $m_i$.}
      \label{fig:tpp}
\end{figure}

\noindent Understanding the positioning of failure events in both the temporal and spatial (mileage) dimensions is beneficial for predictive maintenance. Failure events in our dataset exhibit spatial and temporal patterns (Figure \ref{fig:tpp}), which could indicate stationary behavior of the vehicle or recurring temporal failures \cite{math2024harnessingeventsensorydata}. Prior work, such as \cite{Zhou2023}, employs adaptive binning to discretize time in long event sequences. However, given our moderate sequence length (258 DTCs with 30-day and 300-km intervals), binning is unnecessary. Additionally, lookup table embeddings, often used in discrete event modeling, would be computationally expensive, introduce unnecessary parameters, and fail to preserve meaningful distance relationships in $\mathbb{R}^+$.

\noindent We adopt a continuous time embedding \cite{selfatthawke} which is the deterministic positional encoding introduced in Chapter~\ref{c2:foundation_event_sequence_modeling}~(Eq.~\ref{eq:pos_embedding_vanilla}) where we replace the token position with the continuous time \(t_i\). %
We do it similarly to create a mileage embedding $\boldsymbol{M} \in \mathbb{R}^{L \times d/2}$.
Both embeddings, $\boldsymbol{T}, \boldsymbol{M}$, are concatenated along the feature dimension to form a positional representation. This preserves their independent contributions and prevents interference when fused with $\boldsymbol{D}$. Our final fused input $\boldsymbol{U} \in \mathbb{R}^{L \times d}$ is given by:
\begin{equation}
\boldsymbol{U} = \text{concat}(\boldsymbol{D}_{base}, \boldsymbol{D}_{ecu}) + \boldsymbol{D}_f +  \text{concat}(\boldsymbol{T, M})
\end{equation}

\subsection{Environmental Embeddings}
\noindent To better capture relationships between the different environmental conditions elements $(d, v, u)$, we create three distinct learnable embeddings: (1) the description of the environmental conditions $\boldsymbol{D}_e \in \mathbb{R}^{L_e \times d_d}$ (2) the discretized value $\boldsymbol{V}_e \in \mathbb{R}^{L_e \times d_v}$ (3) its unit $\boldsymbol{U}_e \in \mathbb{R}^{L_e \times d}$.
We use a mix of early summation and concatenation to fuse these embeddings at the input level and obtain the total environmental conditions embedding $\boldsymbol{E} \in \mathbb{R}^{L_e \times d}$:
\[
\boldsymbol{E} = \text{concat}(\boldsymbol{V}_e, \boldsymbol{D}_e) + \boldsymbol{U}_e
\]
To fuse the environmental conditions, we cannot simply align them with their respective DTC to perform an early token-wise summation or concatenation with $\boldsymbol{D}$. Given that we utilize real-world data (noisy, redundant) and perform rigorous data filtering, there are too many missing environmental conditions for each DTC.
 Consequently, we chose to create a separate sequence $S_e$ that is much longer than $S$, where we will concatenate all environmental conditions and fuse them into one embedding $\boldsymbol{E}$. This straightforward method is flexible, and $\boldsymbol{E}$ can be employed in a middle fusion manner inside the attention mechanism \cite{transformermultimodallearningreview}.

\begin{figure*}[!b]
    \centering
    \includegraphics[width=5.6in]{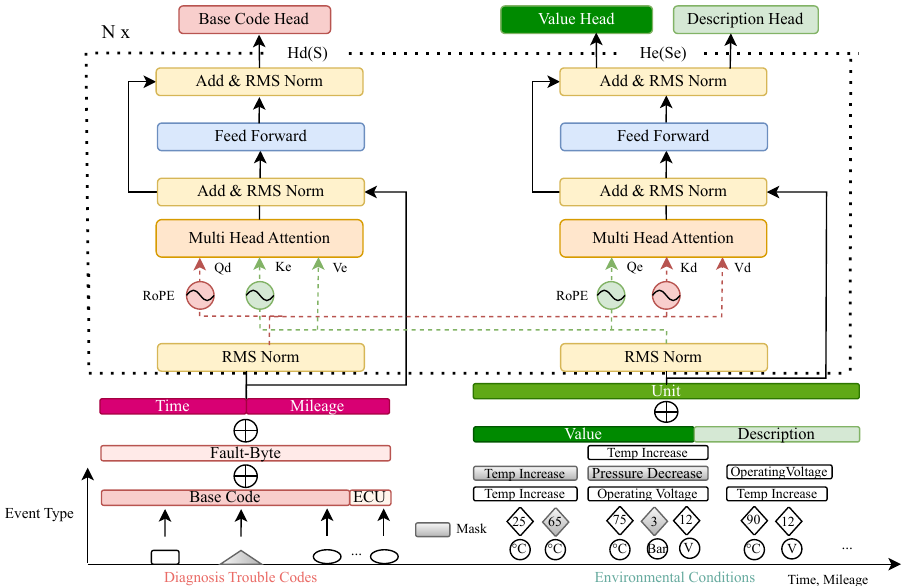}
    \caption{\textbf{BiCarFormer Architecture}. Both parallel Transformers are computing cross-attention scores conditioned on each modality $\boldsymbol{Q, K, V}$. Two final representations are generated for each modality: DTC ($\boldsymbol{H}_d)$ and environmental conditions ($\boldsymbol{H}_e$). Multiple hierarchical embeddings are defined at the input level to account for DTC-specific features (ECUs, Fault-Bytes, Base-Code).}
    \label{bicarformer}
\end{figure*}

\subsection{Co-attention for Vehicle Event Sequences}
\noindent The overall architecture of BiCarFormer is shown in Figure \ref{bicarformer}.
Our architecture is directly inspired by the co-attention mechanism of ViLBERT \cite{vilbert}, which enables multimodal learning by computing attention scores conditioned on each modality.
Two multi-head attention layers are processing $S$ and $S_e$, resulting in two attention scores computation conditioned on: $\boldsymbol{D}$ (DTCs) and $ \boldsymbol{E}$ (environmental conditions) fused embeddings.
We apply a RoPE (Rotary Position Encoding) \cite{Roformer} like in the previous chapter, on two sets of queries and keys (one for each cross-attention) to induce the absolute and relative position of tokens (Figure \ref{bicarformer}).

\noindent More specifically, the fused input embedding vectors $\boldsymbol{u}_m \in \mathbb{R}^d$ and $\boldsymbol{e}_n \in \mathbb{R}^d$ from token positions $m, n$ are projected through weights $\boldsymbol{W}_q, \boldsymbol{W}_k, \boldsymbol{W}_v$.
The query and key are given by applying two different RoPE~\cite{Roformer}:
\begin{align}
\boldsymbol{q}_m &= e^{im\theta_u} \boldsymbol{W}_q \boldsymbol{u}_m \\
\boldsymbol{k}_n &= e^{in\theta_e} \boldsymbol{W}_k \boldsymbol{e}_n
\end{align}
where $\theta_u = \text{diag}(\theta_1, \dots, \theta_{d/2})$ with $\theta_{i} = \theta_{0,u}^{-2i/d}$, $\theta_{0, u} = 5000$. Same for $\theta_e$ but with $\theta_{0, e} = 80000$ due to $L_e \gg L$. We use an alignment function \cite{survey_att_mechanisms} $f: \mathbb{R} \rightarrow [0, 1]$ to produce the attention weights between two tokens:
\begin{align}
    a(\boldsymbol{u}_m, \boldsymbol{e}_n, m,n) &= f(\boldsymbol{q}_m^T \boldsymbol{k}_n) = a_{dtc \rightarrow env}\notag\\
    &= f\left(\boldsymbol{u}_m^T \boldsymbol{W}_q^T e^{i(n\theta_e - m\theta_u)}\boldsymbol{W}_k \boldsymbol{e}_n\right) \label{eq:rope_q_k}
\end{align}
We also compute Eq.~\ref{eq:rope_q_k} for $a_{env \rightarrow dtc} = a(\boldsymbol{e}_m, \boldsymbol{u}_n, n, m)$ with separate weights for $\boldsymbol{q}_m, \boldsymbol{k}_n$. Finally, two cross-attended context vectors are produced:
\begin{align}
    \boldsymbol{c}_{dtc \rightarrow env}(m) &= \sum^{L_e}_{n=0} a_{dtc \rightarrow env}(m,n) \boldsymbol{v}_{e,n} \label{eq:coatt_dtc_env} \in \mathbb{R}^d\\
    \boldsymbol{c}_{env \rightarrow dtc}(n) &= \sum^{L}_{m=0} a_{env \rightarrow dtc}(n,m) \boldsymbol{v}_{u, m} \label{eq:coatt_env_dtc} \in \mathbb{R}^d
\end{align}
where values $\boldsymbol{v}_u, \boldsymbol{v}_e$ are obtained from $\boldsymbol{V}_e = \boldsymbol{W}^e_{v} \boldsymbol{E}, \boldsymbol{V}_u = \boldsymbol{W}^u_v \boldsymbol{D}$.
\noindent We tried two different alignment functions $f$ \cite{survey_att_mechanisms} for Eq.~\ref{eq:rope_q_k}. The widely used \textit{softmax} and \textit{1.5-entmax} \cite{peters-etal-2019-sparse}.
The standard \emph{softmax} outputs dense attention scores, which might take too many environmental conditions into account. Since our input data is noisy and redundant, we would like to extract salient information; a sparse alignment function seems like an intuitive choice.
However, in practice, we did not find a benefit using 1.5-entmax in the pretraining and classification where we observed a 30\% slowdown in iterations per second. Consequently, we retained the \textit{softmax}.
As a consequence, two hidden states $\boldsymbol{H}_{d} \in \mathbb{R}^{L\times d}$ and $\boldsymbol{H}_{e} \in \mathbb{R}^{L_e\times d}$ (Figure \ref{bicarformer}) for each modality are computed after fully connected layers, residual connections and root-mean-squared normalizations \cite{rmsnorm, touvron2023llamaopenefficientfoundation}.

\subsection{Multimodal Learning}
\noindent To reinforce the relationship between DTC events and environmental conditions, we not only mask the DTC but also the environmental conditions $(d, v)$ and let the unit unmask to reduce training complexity. This enables multimodal learning by reconstructing $S_e$ using $S$ and vice versa. As a result, BiCarFormer learns to benefit from this extra modality, which can be confirmed by inspecting the cross entropy loss of the \textit{Base-DTC} classification: $\mathcal{L}_{\text{dtc}}$ where using both modalities helps to reduce the pretraining loss (Appendix.~\ref{fig:pretraining_comp}). However, due to multitasking, the losses are less stable and may require gradient clipping and a smaller learning rate to stabilize training. BiCarFormer is trained with three cross-entropy losses balanced by static coefficients: $ \alpha=0.5, \beta=0.3, \gamma=0.2$:
\begin{equation}
    \mathcal{L} = \alpha \mathcal{L}_{\text{dtc}} +  \beta \mathcal{L}_{\text{value}} +  \gamma \mathcal{L}_{\text{description}}
\end{equation}
\section{Experiments}
\subsection{Settings}
\noindent We evaluated our model against established Transformer architectures, including BERT \cite{bert} (without the Next Sentence Prediction task) and DTC-TranGRU \cite{Hafeez2024DTCTranGruIT}. Since DTC-TranGRU is autoregressive \cite{Hafeez2024DTCTranGruIT}, we trained it using next-token prediction and three separate heads, each classifying one of the three DTC components. To assess each model's performance, we focused on multi-label classification of error patterns in the downstream task.
For BERT and BiCarFormer we masked 15\% of the tokens during the pretraining. The different models comprised approximately 25 million parameters, with the same hidden size $d=600$, ensuring a fair comparison.
For our downstream task of multi-label sequence classification, we used either the [CLS] token to perform classification or in the case of the BiCarFormer, the $[CLS]_{dtc}$ and $[CLS]_{env}$ after simply concatenating them along the feature dimension $d$ and fusing with a small $MLP$. We used the [EOS] token for DTC-TranGRU rather than the [CLS].
Each classifier has 3,461,160 trainable parameters and consists of a small MLP with layer normalization, residual connections, and a sigmoid activation. During the downstream classification task, all backbone weights were kept frozen. The dataset was partitioned into training, validation, and testing sets, adhering to a ratio of 75\%, 15\%, and 15\%, respectively, with about 360 labeled error patterns.
For the pretraining, we used a learning rate of $10^{-4}$ with a cosine scheduler and warm-up of 2000 steps for all models. We used the AdamW optimizer \cite{loshchilov2018decoupled} with $\beta_1$ set to 0.9 and $\beta_2$ to 0.999. We applied a weight decay of 0.1 and a percentile gradient clipping of 5. Finally, we trained on an NVIDIA A10G GPU using 16-bit floating-point precision (FP16) with a batch size of 32 for the pretraining and 192 for the classifiers.

\subsection{Metrics} The evaluations were conducted across 360 different error patterns. To measure classification performance, we employed multiple metrics, namely AUROC (Area Under the Receiver Operating Characteristic), Precision, Recall, and F1 Score \cite{classificationmetrics}. The last three were computed using a confidence threshold of 0.8. The metrics were aggregated using different averaging: micro, macro, and sample.
Micro computes metrics broadly and favors frequent classes. It provides an overall picture of model performance, but it does not reflect the performance \textit{per-class} or \textit{per-instance}. In contrast, macro averaging treats all classes equally by computing metrics independently for each class and averages the results. In multi-label classification, the focus is often on \textit{per-instance} performance rather than \textit{per-class} performance. This means we prioritize the overall quality of multi-label predictions for each instance, rather than optimizing for the accuracy of individual labels in isolation. For this, the sample averaging computes metrics \textit{per-instance} and then averages across all samples.

\noindent

\begin{table}[!b]
\centering
\begin{tabular}{|c|c|c|c|}
\hline
\multicolumn{4}{|c|}{\textbf{Micro and Macro Averaging}} \\
\hline
Model & AUROC (Micro) & F1 Score (Micro) & F1 Score (Macro) \\
\hline
\textbf{BiCarFormer} & \textbf{0.809} & \textbf{0.77} & \textbf{0.71} \\
\hline
DTC-TranGRU \cite{Hafeez2024DTCTranGruIT} & 0.602 & 0.36 & 0.28 \\
\hline
BERT \cite{bert} & 0.768 & 0.71 & 0.63 \\
\hline
\multicolumn{4}{|c|}{\textbf{Sample Averaging}} \\
\hline
Model & Precision (Sample) & Recall (Sample) & F1 Score (Sample) \\
\hline
\textbf{BiCarFormer} & \textbf{0.68} & \textbf{0.62} & \textbf{0.64} \\
\hline
DTC-TranGRU \cite{Hafeez2024DTCTranGruIT} & 0.23 & 0.19 & 0.20 \\
\hline
BERT \cite{bert} & 0.59 & 0.53 & 0.55 \\
\hline
\end{tabular}
\caption{Downstream evaluation of multi-label error pattern classification. Each model has the same number of parameters (25M). Results are grouped by averaging type (micro/macro vs. sample).}
\label{tab:bicarformer_table_combined}
\end{table}

\subsection{Multi-label Classification Performance Analysis}
\noindent Table~\ref{tab:bicarformer_table_combined} presents a comparative analysis of BiCarFormer against established sequence-to-sequence models for multi-label error pattern classification. Multi-label classification is particularly challenging due to the varying
complexity of error patterns, where a single instance may be associated with multiple diagnostic trouble codes (DTCs) and environmental conditions. DTC-TranGRU \cite{Hafeez2024DTCTranGruIT} underperforms, particularly in \textit{per-class} and \textit{per-instance} evaluations, likely due to the limitations of masked attention mechanisms and model architecture. These constraints make it difficult for the model to capture long-range dependencies in the input sequence, leading to a lower AUROC (0.602) and poor F1 scores (Micro: 0.36, Macro: 0.28, Sample: 0.2). While BERT \cite{bert} is already achieving relatively high AUROC and F1 Micro,
BiCarFormer significantly enhances the AUROC (Micro) by +4\%, and the F1 Score (Micro) by +6\%.

\begin{figure}[!b]
      \centering
      \includegraphics[width=4.2in]{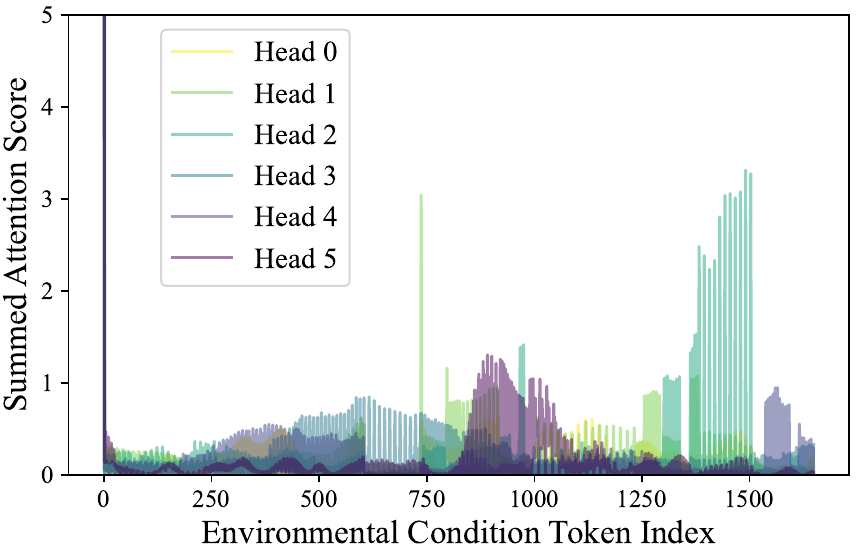}
      \caption{\textbf{Attention Distribution between Environmental Condition Tokens.} Amount of attention received by each environmental condition token from the DTCs. The y-axis was truncated to improve clarity as well as the number of heads printed. We take $\boldsymbol{A}_{dtc \rightarrow env}$ of the last layer.}
      \label{fig:att_distrib}
\end{figure}

\noindent The improvement of BiCarFormer is especially visible for \textit{per-class} averaging. When dealing with many error patterns, some are more difficult to distinguish due to their natural complexity and overlaps.
BiCarFormer, however, better differentiates rare classes, suggesting an improved capacity to generalize across a diverse range of error patterns.
This is confirmed in this experiment, where we see the gaps in classification performance when averaging across classes, with a +8\% F1 Macro compared to BERT. This means that classes with small instances, usually hard to classify using only the DTCs, can be better differentiated using environmental conditions.
Finally, \textit{per-instance} precision and recall are the biggest improvement with +9\% in F1 Score, Precision, and Recall compared to BERT. These results confirm that BiCarFormer effectively leverages this additional information and highlights the predictive improvement compared to standard Transformers.

\begin{figure*}[!b]
    \makebox[\textwidth][c]{\includegraphics[width=1.4\textwidth]{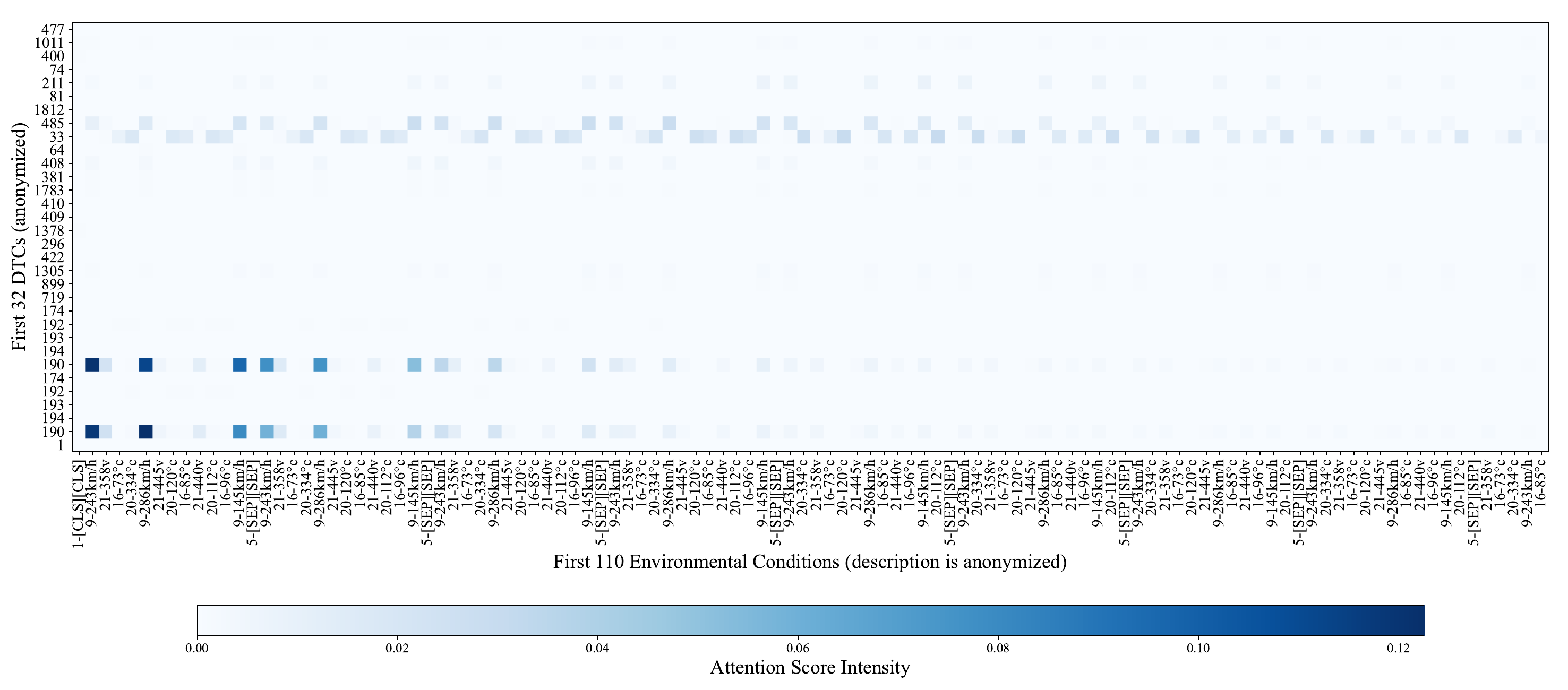}}

    \caption{\textbf{Cross-Attention Scores for $\boldsymbol{A}_{dtc \rightarrow env}$.} The DTCs are shown on the y-axis (anonymized), and the environmental conditions with their 3 elements $(d, v, u)$ concatenated are shown on the x-axis (the description $d$ is anonymized). The intensity of each cell reflects the attention weight, where darker shades indicate higher attention values.}
    \label{att_scores}
\end{figure*}

\subsection{Cross-Attention Scores Interpretations}
\noindent We would like to understand how the co-attention mechanism enables multimodal learning and enhances downstream tasks. We took a random test sample with a specific battery issue and ran it on BiCarFormer to analyze the different cross-attention score patterns.
Figure \ref{att_scores} presents a heat-map visualization of $\boldsymbol{A}_{dtc \rightarrow env}$. This helps us to understand general attention patterns from DTCs to environmental conditions.
Cross-attention uncovers natural relationships from certain DTCs linked to specific environmental conditions. Consider DTC 190, we can see that this token focuses more on a local series of environmental conditions with unit \textit{km/h} rather than other units. Whereas DTC 33 and 485 attend more to \textit{°C}.
This result is attributable to specific environmental conditions that characterize certain DTCs, i.e., the temperature might define DTC1 while DTC2 might indicate a rise in the voltage.
Due to the duplicated environmental conditions unit across $S_e$, one DTC might attend to the next environmental conditions 'voltage' also later in the sequence, creating these line patterns on Figure \ref{att_scores} with DTC 190 \& 33 \& 485.
As a consequence, we notice the local temporal patterns between DTCs and their surrounding environmental conditions (Figure \ref{att_scores}). Therefore, the initial DTCs should prioritize the initial environmental conditions in $S_e$, with an offset resulting from the difference in sequence length between $L$ and $L_e$. For instance, DTC 190 at positions 1 and 6 should prioritize the initial environmental conditions.

We would like to assess whether the DTCs attend to a few key environmental conditions. From the same test sample, Figure \ref{fig:att_distrib} plots the amount of attention that DTCs give to each of the environmental conditions. We observed different clusters per head. Each head learns to focus on specific environmental conditions, sometimes particular indices like \emph{head 1} ($i \approx 749$ and $800$), and \emph{head 2} ($i \approx 960)$. The [CLS] token is traditionally heavily attended. Interestingly, the last part of the sequence seems to be more attended by specific heads: $i \in [1300, 1500]$, signaling that we can capture long-range dependencies between DTCs and tokens of environmental conditions even with a big difference in sequence length (i.e., with $L_e \gg L$). Moreover, late environmental conditions might carry a lot of defect signals, which makes sense as the sequence approaches the time of the critical failure event occurring at time $t_i = t_L$. Thus, attention scores increase correspondingly.
\begin{figure}[!b]
      \centering
      \includegraphics[width=4.8in]{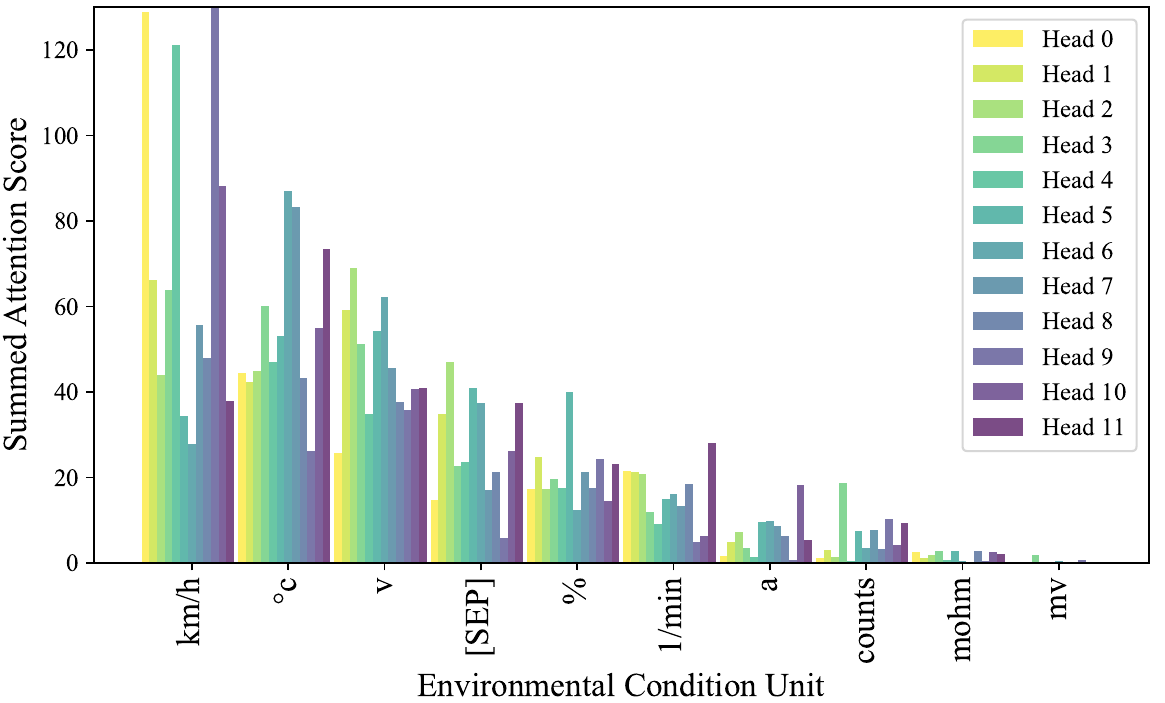}
      \caption{\textbf{Attention Distribution between Environmental Conditional Units.} Amount of attention received by each environmental unit from the DTCs in a \textit{battery aging} error pattern from a battery electric vehicle.  We take $\boldsymbol{A}_{dtc \rightarrow env}$ of the last layer and print all heads.}
      \label{fig:att_distrib_value}
\end{figure}
\subsection{Error Pattern in Battery Electric Vehicles}

\begin{figure}[h]
      \centering
      \includegraphics[width=5.2in]{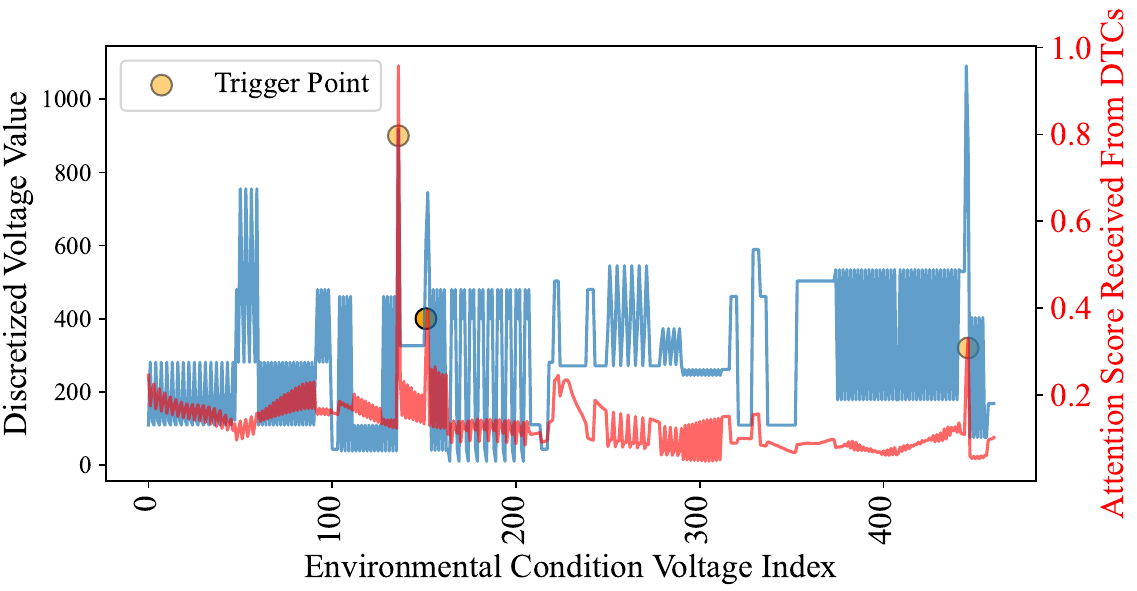}
      \caption{\textbf{Discretized Voltage and Attention Scores Evolution between Environmental Token Position}. Discretized voltage variation as a function of environmental condition tokens and the associated amount of attention received from DTCs. We only take triplets $(d,v, u)$ from $S_e$ with $u = $ 'v' and extract $\boldsymbol{A}_{dtc \rightarrow env}$ from the last layer and \emph{head 2}. The sample is taken from a \emph{battery aging} error pattern. BiCarFormer implicitly learns the fluctuation of discretized continuous values to make a prediction.}
      \label{fig:att_distrib_unit}
\end{figure}

\noindent With the increasing popularity of \textit{BEV} (battery-electric vehicle), there is a pressing need to diagnose failures in these vehicle types.
We take the example of a concrete error pattern, such as a \textit{battery aging}, which is a common phenomenon in electric vehicles. And draw the cumulative attention of e. condition units for the same test sample as before.
Since attention heads focus on a certain part of $S_e$ (Figure \ref{fig:att_distrib}), some should focus on special units (Figure \ref{fig:att_distrib_unit}) such as km/h (\emph{head 0, 4, 9}) while others focus on °C (\emph{head 6, 7}) and voltage (\emph{head 1, 2}).
Now, more interestingly, we take a head that tends to specialize in one unit like \emph{head 2} for voltage and draw the evolution of the discretized voltage value as a function of their index in $S_e$ in Figure \ref{fig:att_distrib_value}.
 We essentially try to answer if we have a fluctuation of voltage values and a correlated change in the attention pattern.
We also draw the associated attention scores received from DTCs and obtain multiple \emph{trigger points} where the voltage value strongly correlates with attention scores. We note that this is not the same depending on the heads; some exhibit exponentially decreasing attention scores shapes, while others exhibit noisy signals.
As a consequence, BiCarFormer is able to extract specific fluctuations of continuous voltage values from a tokenized input. Engineers may extract trigger points and analyze how these specific variations affect the vehicle, providing a more explainable result.

\section{Applications to Downstream Tasks}
\noindent Having a separate hidden state for the environmental conditions also enables unsupervised learning in settings where EPs are unavailable. The hidden state $H_e$ could provide a more accurate dimensionality reduction and observe better decision boundaries between unlabeled EPs using clustering techniques. Also, if provided with domain knowledge, we could directly filter $S_e$ beforehand to take specific environmental conditions. Our model and overall paradigm enable retention of this architectural choice and reduce the computation by injecting domain knowledge directly into the preprocessing steps.
Another possible useful downstream application is explainability. We would like to see the contributions of environmental conditions and DTCs to the EPs classification. Since error patterns are hard-coded defined rules, we could try to derive new rules for unknown error patterns based on feature attribution methods \cite{shap}, cross-attention scores, perturbation-based methods or multi-label causal discovery with Transformers.
BiCarFormer model classifies EPs a posteriori (offline setting); thus, the inference optimization might not be needed since we do not rely on edge computing capabilities within the vehicles. Nevertheless, due to the quadratic time complexity of cross-attention scores, there is an extensive need to optimize Eq.~\ref{eq:coatt_dtc_env}, and Eq.~\ref{eq:coatt_env_dtc}. Alternatives like sparse attention \cite{sparsetemporalattention, bigbird, beltagy2020longformer}, token merging \cite{bolya2023token} or pooling should be adapted for co-attention in event sequences.
\section{Summary}
\noindent We introduced BiCarFormer, a bidirectional multimodal Transformer designed to classify EPs by jointly encoding discrete DTC sequences and continuous environmental conditions. Through a co-attention mechanism and hierarchical embedding fusion, BiCarFormer learns cross-modal dependencies between failure events and their surrounding sensory context. On a real-world automotive dataset comprising 22,137 DTCs and 360 error patterns, BiCarFormer achieves an AUROC of 0.809, a micro F1 score of 0.77, and a sample-level F1 score of 0.64, representing improvements of +4\%, +6\%, and +9\% respectively over a standard BERT baseline trained on DTC sequences alone. The gains are most pronounced in per-class and per-instance evaluations, confirming that environmental conditions provide discriminative signal for rare and overlapping error patterns that share identical DTC rule definitions.
Interpretability analysis of the cross-attention scores reveals that individual attention heads specialize in specific environmental units, such as voltage or temperature, and implicitly detect fluctuations in discretized continuous values that correlate with fault onset. This provides an additional layer of transparency beyond what DTC-only models offer.
However, two practical limitations merit acknowledgment. First, the quadratic complexity of the co-attention mechanism with respect to the environmental condition sequence length (\(L_e \gg L\)) constitutes a computational bottleneck for long sequences, motivating future work on sparse or pooled cross-attention variants. Second, BiCarFormer classifies EPs offline (a posteriori), in contrast to EPredictor's online, causal-mask-based operation; it is therefore better suited to retrospective diagnostic analysis than real-time on-board deployment.

\section{Outlook}
Part~\ref{part1} has demonstrated that large-scale Transformers can serve as effective predictive and diagnostic models for high-cardinality vehicle event streams. The introduced models demonstrated that NLP techniques such as word embeddings, positional encoding, and transformer models can be adapted to event sequences to capture the hierarchical dependencies among diagnostic trouble codes (DTCs). The connection is not merely sequential: the autoregressive models of Part~I serve as neural density estimators in Part~II, a repurposing that distinguishes this thesis from prior work. By interpreting attention scores, automotive engineers can better understand BiCarFormer predictions. However, we must establish cause-and-effect relationships between DTCs and EPs rigorously, as relying on attention scores alone lacks reliability \cite{att_not_explain_2019, filippova-2020-elephant}. While predictive models enable early fault detection, they remain fundamentally \emph{correlational}: a high F1 score tells us what will fail and when, but not why. Without understanding the underlying causal mechanisms, any prediction remains fragile, susceptible to distributional shift, and incapable of supporting counterfactual reasoning or root-cause analysis. Establishing such causal relationships requires moving beyond correlation-based models toward methods that explicitly encode the data-generating process. This demands a formal framework for causal discovery, grounded in probability and information theory, that is capable of operating on the scale of tens of thousands of event types over multiple samples. Part II introduces precisely this foundation.

\part{Causal Discovery in High-Cardinality Event Sequences via Neural Density Estimation}
\label{pa:p2}

\chapter{Foundations of Information Theory and Causal Discovery}\label{c3:foundation_causal_discovery}
\chaptermark{Foundations of Information Theory and Causal Discovery}
\glsresetall

Having established predictive models in Part~I, this chapter turns to understanding. Causal discovery requires measuring statistical dependencies between random variables in graphical models, a task for which information theory provides the right tools. We therefore introduce the related background, including the common definitions and modeling techniques required for Part~\ref{pa:p2}.

Let \(\mathbf{U}\) denote the set of all (discrete) random variables. We define the event set \(\mathbf{X} = \{X_1, \ldots, X_n\} \subset \mathbf{U}\), and the label set \(\mathbf{Y} = \{Y_1, \ldots, Y_l\} \subset \mathbf{U}\); in these generic, non-sequential definitions, the subscript \(i\) of \(X_i\) simply enumerates the variables.

\noindent\textbf{Indexing convention for sequences.} Throughout Part~\ref{pa:p2}, a single convention is used for sequences: \(X_t\) denotes the event observed at position (time step) \(t \in \{0, \ldots, L\}\) of a sequence and \(x_t\) its realization, so that the trajectory preceding step \(t\) is \(X_{<t} = \{X_0, \ldots, X_{t-1}\}\). Labels keep the subscript \(j\), which indexes the label \emph{identity} \(Y_j\), \(j \in \{1, \ldots, |\mathcal{Y}|\}\), not a position; when the step at which a label is evaluated matters, it is written as a superscript, \(Y^{(t)}_j\). Chapter~\ref{c7:multi_label_one_shot_causal_discovery} (OSCAR) and Chapter~\ref{c7:event_to_event} (TRACE) both follow this convention. Chapter~\ref{c7:multi_label_causal_discovery} (CARGO) aggregates graphs over many sequences and therefore reasons about event \emph{types} rather than positions; there, \(X_i\) denotes the event type \(i \in \{1, \ldots, |\mathcal{X}|\}\), as restated in that chapter.

\section{Probability and Information Theory}
In this section, we present the primary mathematical tools from probability and information theory that we will use in Part~\ref{pa:p2}.

\subsection{Marginal, Joint, Conditional, and Relative Entropy}
\subsubsection{Definitions}

\begin{definition}[Entropy]\label{def:entropy}
The entropy \cite{shannon1951prediction} of a discrete random variable \(X\) with a probability mass function \(p(x) = P(X=x)\) is defined by
\begin{equation}\label{eq:entropy}
    H(X) = -\sum_{x \in \mathcal{X}} p(x)\log p(x)
\end{equation}
\end{definition}
\noindent It can be thought of as the amount of surprise of the random variable \(X\) and thus quantifies its uncertainty. We measure it in \textit{nats} as we use the cross-entropy loss of PyTorch~\cite{pytorch} in our experiments.

\begin{definition}[Conditional Entropy]\label{def:cond_entropy}
We can define the conditional entropy as the entropy of a random variable given another random variable
\begin{equation}\label{eq:cond_entropy}
    H(Y|X) = - \sum_{x \in \mathcal{X},y \in \mathcal{Y}} p(x, y)\log{\frac{p(x,y)}{p(x)}}
\end{equation}
\end{definition}

\noindent This can also be written as a function of the posterior \(p(y|x)\) using the chain rule of probability:

\begin{equation}
    H(Y|X) = - \sum_{x \in \mathcal{X},y \in \mathcal{Y}} p(x)p(y|x)\log{p(y|x)} = - \mathbb{E}_{p(x,y)} \log{p(Y|X)} \label{eq:cond_entropy_post}
\end{equation}
It can be thought of as the \uline{remaining uncertainty} in \(Y\) when we already know \(X\).

\begin{definition}[Relative Entropy]\label{def:dkl}
The relative entropy is defined as the distributional difference or \textit{Kullback-Leibler-Divergence} \((D_{\text{KL}})\) \cite{cover1999elements} between two probability distribution \(P\) and \(Q\):
\begin{equation}\label{eq:relative_entropy}
    D_{\text{KL}}(P||Q) = \sum_{x \in \mathcal{X}} p(x) \log{\frac{p(x)}{q(x)}}
\end{equation}
\end{definition}
\noindent It is an asymmetric “distance” measure between two probability distributions.

\begin{definition}[Joint Entropy]\label{def:joint_entropy}
    The joint entropy \(H(X, Y)\) of a pair of discrete random variables \((X, Y)\) with joint distribution \(P(X,Y)\) is defined as:
    \begin{equation}\label{eq:joint_entropy}
        H(X, Y) = -\sum_{x \in \mathcal{X}} \sum_{y\in \mathcal{Y}}p(x,y)\log{p(x,y)}
    \end{equation}
\end{definition}
\noindent which is also the expected value of \(\log{p(x, y)}\):
\[H(X,Y) = -\mathbb{E}_{p(x,y)}\log{p(X, Y)}\]

\begin{definition}[Information Gain]\label{def:information_gain}
    The information gain \(I_G\) of a random variable \(Y\) obtained from an observation of the realization \(X=x\) is the divergence between the posterior and the prior distribution of \(Y\):
    \begin{equation}\label{eq:information_gain_dkl}
        I_G(Y, x) = D_{\text{KL}}(P({Y|X=x})||P(Y))
    \end{equation}
\end{definition}
\noindent As a KL divergence, \(I_G(Y, x) \geq 0\), with equality if and only if \(P(Y|X=x) = P(Y)\). Its expectation over \(X\) is the difference of entropies, i.e., the mutual information (Def.~\ref{def:dkl_mi}):
\begin{equation}\label{eq:information_gain}
    \mathbb{E}_{p(x)}[I_G(Y, x)] = H(Y) - H(Y|X).
\end{equation}
The pointwise difference \(H(Y) - H(Y|X=x)\) has the same expectation but, unlike \(I_G(Y, x)\), it can be negative for a single realization \(x\); we therefore always use the divergence form.
\noindent Information gain is used to construct decision trees \cite{quinlan:induction}. It will be our primary tool for performing causal discovery from posterior distributions.

\subsubsection{Properties}
The following describes the key properties regarding the different types of entropy.
\begin{lemma}[Non-Negativity of the Entropy]\label{lemma:entropy_geq_0} \cite{cover1999elements}
The entropy of a random variable is always non-negative
    \[H(X) \geq 0\]
\end{lemma}

\begin{lemma}[Chain Rule]\label{lemma:entropy_chain_rule} \cite{cover1999elements}
We can decompose the joint entropy into a sum of marginal and conditional entropy
    \[H(X, Y) = H(X) + H(Y|X)\]
\end{lemma}

\begin{lemma}[Conditioning Reduces Entropy]\label{lemma:entropy_conditionning} \cite{cover1999elements}
   Conditioning the entropy of a random variable \(X\) on multiple random variables reduces its entropy
    \[H(X) \geq H(X|Y) \geq H(X|Y,Z) \geq \cdots \geq 0\]
\end{lemma}

\begin{lemma}[Bounded]\label{lemma:entropy_bounded} \cite{cover1999elements}
   The entropy of a random variable \(X\) is bounded by the log of its support\(|\mathcal{X}|\)
    \[0 \leq H(X) \leq \log{|\mathcal{X}|}\]
\noindent The lower bound holds because \(-\log p(x) \geq 0\); the upper bound follows from Jensen's inequality, since \(H(X)\) is concave in \(p\) and maximized by the uniform distribution.
\end{lemma}

\subsection{Mutual Information and its variants}
\subsubsection{Definitions}

\begin{definition}[Mutual Information]\label{def:dkl_mi}
The reduction in uncertainty due to another random variable is called the mutual information (MI) \cite{cover1999elements}. For two random variables \(X, Y\) with joint probability mass function \(p(x,y)\), it is given as the relative entropy between the joint distribution \(P(X,Y)\) and the product distribution \(P(X)P(Y)\)
    \begin{equation}\label{eq:dkl_mi}
    I(Y,X) = D_{\text{KL}}(P(X,Y)||P(X)P(Y)) =  \sum_{x \in \mathcal{X},y \in \mathcal{Y}} p(x,y)\log{\frac{p(x,y)}{p(x)p(y)}}
\end{equation}
It can also be defined as the difference between \(H(Y)\) and \(H(Y|X)\)
\begin{equation}\label{eq:mi}
    I(Y,X) = H(Y) - H(Y|X)
\end{equation}
Consequently, the mutual information \(I(Y, X)\) is the reduction in uncertainty of \(X\) due to the knowledge of \(Y\). We write \(I(Y, X)\) for mutual information, the semicolon notation \(I(X;Y)\) is equivalent. We now define the conditional mutual information (CMI), which is the reduction in uncertainty of \(X\) due to the knowledge of \(Y\) given \( Z\).
\end{definition}

\begin{definition}[Conditional Mutual Information]\label{def:cmi}
The conditional mutual information of random variables \(X\) and \(Y\) given \(Z\) is defined by
\begin{align}
        I(Y, X|Z) &= H(Y|Z) - H(Y|X,Z)\label{eq:cmi_with_entropy} \\
        &= \mathbb{E}_{p(x,y,z)}\log{\frac{p(X,Y|Z)}{p(X|Z)p(Y|Z)}}
\end{align}
\end{definition}
\noindent \(Z\) can also be a set of random variables \(\mathbf{Z}\); in the general definitions of this chapter, \(\mathbf{Z}\) denotes any conditioning set. In sequences, the conditioning set is the trajectory, \(\mathbf{Z} = X_{<t} = \{X_0, \ldots, X_{t-1}\}\), and Chapters~\ref{c7:multi_label_one_shot_causal_discovery} and~\ref{c7:event_to_event} write \(X_{<t}\) directly. Now that we have defined the conditional mutual information, we can define conditional independence between random variables.

\begin{definition}[Conditional Independence]\label{def:conditional_independence} Variables \(X\) and \(Y\) are said to be conditionally independent given a variable set \(\boldsymbol{Z}\), if \(P(X, Y|\boldsymbol{Z}) = P(X|\boldsymbol{Z})P(Y|\boldsymbol{Z})\), denoted as \(X \bot \space \space Y| \boldsymbol{Z}\). Inversely, \(X \not\perp \space \space Y| \boldsymbol{Z}\) denotes the conditional dependence.
Using the conditional mutual information \cite{cover1999elements} to measure the independence relationship, this implies that \[\text{I}(Y, X|\boldsymbol{Z}) = 0 \Leftrightarrow X \perp Y | \boldsymbol{Z}\]
\end{definition}

\subsubsection{Properties}
\begin{lemma}[Symmetry of Mutual Information]\label{lemma:mi_symmetry}
In its standard definition, the MI is symmetric:
\[I(Y,X) = I(X,Y) = H(Y) - H(Y|X) = H(X) - H(X|Y)\]
\end{lemma}

\begin{lemma}[Non-negativity of mutual information]\label{lemma:mi_nonnegative}
For any two random variable \(X,Y\),
\begin{align}
    I(Y,X) &\geq 0 \label{eq:mi_nonnegative}
\end{align}
It is also the case for the conditional mutual information:
\begin{align}
    I(X,Y|Z) &\geq 0\label{eq:cmi_nonnegative}
\end{align}
\end{lemma}

\section{Graphical Models}
Graphical models \cite{koller2009probabilistic} provide a simple way to represent probabilistic dependencies across random variables. For instance, a physician must infer the most likely diagnoses, given the patient's symptoms, test results and genetics. A diagnostic expert might want to evaluate the causes of defects in a vehicle based on the series of DTCs that preceded it.

\subsection{Bayesian Networks}
\subsubsection{Motivation}
Our goal is to represent the joint distribution P over the set of random variables \(\mathbf{U}\). To reduce the specification of every possible random variable assignment, the independence properties in the distribution can be used to represent distributions via directed acyclic graphs \cite{koller2009probabilistic}, i.e., graphs that have oriented edges and no cycles.

 Bayesian Networks \cite{pearl_1998_bn} have served as a modeling technique for a variety of decision problems and benefit from the extensive literature on causal discovery associated with such directed acyclic graphs (DAGs).
In particular, we build upon Pearl's model of causality, which assumes that the underlying causal structure is in the form of a directed acyclic graph (DAG) \cite{pearl_2009}. Such a graph describes the causal relationships between the random variables.

\subsubsection{Definitions}
\begin{definition}[Bayesian Network]\label{def:bn} \cite{pearl_1998_bn}
Let \(P\) denote the joint distribution over a variable set \(\mathbf{U}\) of a directed acyclic graph (DAG) \(\mathcal{G}\). The triplet \(<\mathbf{U}, \mathcal{G}, P>\) constitutes a Bayesian Network (BN) if it satisfies the Markov condition: every random variable is independent of its non-descendant variables given its parents in \(\mathcal{G}\). Each node \(X_i \in \mathbf{U}\) represents a random variable. The directed edge \((X_i \rightarrow X_j)\) encodes a probabilistic dependence. The joint probability distribution can be factorized as
\begin{equation}\label{eq:BN_factorization}
P(X_0, \cdots, X_n) = \prod^n_{i=0} P(X_i|\text{Pa}_{\mathcal{G}}(X_i))
\end{equation}
with \(\text{Pa}_{\mathcal{G}}(X_i)\) the parents of node \(X_i\) in \(\mathcal{G}\).

\noindent We now define \emph{I-maps}, which will help us identify graphs that represent the same set of conditional independence.
\begin{definition}[I-maps]\label{def:imaps}\cite{koller2009probabilistic}
    Let \(P\) be a distribution over \(\mathbf{U}\), we define \(\mathcal{I}(P)\) to be the set of independence assertions of the form \(X \perp Y | Z\) that hold in \(P\)
\end{definition}
\noindent We can rewrite the statement that \(P\) satisfies the local independencies associated with the graph \(\mathcal{G}\) as \(\mathcal{I}(\mathcal{G}) \subseteq \mathcal{I}(P)\). The graph \(\mathcal{G}\) encodes a set of conditional independencies \(\mathcal{I}(\mathcal{G})\), where each element corresponds to a conditional independence relation \(X \perp Y | \boldsymbol{Z}\), meaning that \(X\) and \(Y\) are conditionally independent given the variable \(Z\) (Def.~\ref{def:conditional_independence}).

\begin{definition}[Faithfulness]\label{def:bn_faithfulness}\cite{Spirtes2001CausationPA} Given a BN \(<\mathbf{U}, \mathcal{G}, P>, \mathcal{G}\) is faithful to \(P\) if and only if every conditional independence present in \(P\) is entailed by \(\mathcal{G}\) i.e., \(\mathcal{I}(P) \subseteq \mathcal{I}(\mathcal{G})\) and the Markov condition holds. \(P\) is faithful if and only if there exist a DAG \(\mathcal{G}\) such that \(\mathcal{G}\) is faithful to \(P\).
\end{definition}
\noindent That is, the conditional independencies in \(P\) correspond exactly to those implied by the DAG.
\end{definition}

\begin{definition}[Markov Equivalence Class]\label{def:mec}
Two distinct graphs \(\mathcal{G}, \mathcal{G'}\) are said to belong to the same Markov Equivalence Class (MEC) if they have the same set of conditional independencies, i.e., \(I(\mathcal{G}) = I(\mathcal{G'})\).
\end{definition}
\noindent Causal Discovery algorithm often recovers not the exact graph \(\mathcal{G}\) but rather an MEC \cite{ges, constrainct_based_cd, cd_temporaldata_review}, since orienting the edges is perilous and involves, in general, domain knowledge or additional assumptions.

\begin{definition}[Markov Boundary]\label{def:markov_boundary} \cite{optimal_feature_set_cd}
In a faithful BN \(<\mathbf{U}, \mathcal{G}, P>\), for a set of variables \(\boldsymbol{Z} \subset \mathbf{U}\) and label \(Y \in \mathbf{U}\), if every variable \(X \in \mathbf{U} \setminus (\boldsymbol{Z} \cup \{Y\})\) is independent of \(Y\) conditioned on \(\boldsymbol{Z}\), and any proper subset of \(\boldsymbol{Z}\) does not satisfy the condition, then \(\boldsymbol{Z}\) is the Markov Boundary of \(Y\): \(\mathcal{MB}(Y)\).
\end{definition}
\noindent Many local structure learning (LSL) algorithms, which learn only the neighborhood of selected target variables instead of the full graph, seek a reduced Bayesian Network in the form of the \textit{Markov Boundary} of certain random variables. This serves primarily for feature selection \cite{feature_selection_review}.

Equipped with these information-theoretic tools and the graphical-model vocabulary introduced above, we now turn from representation to inference. The remainder of this chapter formalizes causal discovery as a concrete estimation problem, states the assumptions under which it becomes tractable, and defines the evaluation metrics used throughout Part~\ref{pa:p2}.

\section{Causal Discovery: Tasks, Assumptions, and Evaluation}\label{sec:tasks}

\subsection{Causal Discovery}\label{sec:cd_task}
Causal discovery aims to recover the underlying causal structure among a set of random variables \(\mathbf{X}\) from observational data and or interventions \cite{pearl_2009}.
Formally, the goal is to identify a directed acyclic graph (DAG)
\(\mathcal{G} = (\mathbf{X}, E)\) whose edges encode the direct causal relations such that
the joint distribution \(P(\mathbf{X})\) satisfies the Markov condition with respect to
\(\mathcal{G}\) (Definition~\ref{def:bn}).
Depending on the assumptions and available information, the tasks consist of recovering either:

\begin{itemize}
    \item The \textbf{full DAG} \(\mathcal{G}\),
    \item A \textbf{Markov Equivalence Class (MEC)} (Def.~\ref{def:mec}),
    \item Or the \textbf{Markov Boundary} of selected target variables (Def.~\ref{def:markov_boundary}),
\end{itemize}

\noindent
Each representation offers a different level of identifiability.
In high-cardinality settings such as error codes in vehicles, recovering the complete DAG is often infeasible;
Many methods, therefore, focus on local structures (Markov Boundaries) or on equivalence classes of DAGs.

\subsubsection{Assumptions}
To enable identifiability, we must make several assumptions that we  causal discovery algorithms typically rely on classical assumptions such as:

\begin{assumption}[Causal Sufficiency]
\label{assumption:causal_sufficiency}
All relevant variables are observed, and there are no hidden confounders afecting the events and labels.
\end{assumption}
Furthermore, to be able to detect any conditional independencies in \(\mathcal{G}\) via \(P\) (i.e., through a statistical test) the faithfulness assumption is adopted~(Def.~\ref{def:bn_faithfulness}). For sequences, we assume temporal precedence (or priority~\cite{Hume1978Treatise}) which simply edge orientation by making causality asymmetric in time. That is:
\begin{assumption}[Temporal Precedence]\label{assumption:temporal_precedence}
Given a perfectly recorded sequence of events \((x_0, x_1, \ldots)\), ordered by non-decreasing time of occurrence, with the labels \(\boldsymbol{y}_L\) observed at its final position \(L\), an event \(x_t\) is allowed to influence any subsequent event \(x_{t'}\) (or label) only if it occurs at an earlier step, \(t < t'\). Formally, for the graph \(\mathcal{G} = (\mathbf{U}, \boldsymbol{E})\): \((x_t, x_{t'}) \in \boldsymbol{E} \implies t < t'\).
\end{assumption}
\noindent Violations of these assumptions (hidden confounders, cycles, non-stationarity) are common in practice due to the inaccurate recording of events, hidden confounders, and so forth.

\subsubsection{Classes of Causal Discovery Algorithms}

\paragraph{Constraint-Based Methods.}
These algorithms determine graph structure by performing a series of conditional independence tests (Def.~\ref{def:conditional_independence}). They are mostly employed in tabular data where a lot of data samples are presents with a few number of variables.
Examples include PC \cite{constrainct_based_cd}, and local structure learning methods such as IAMB \cite{iamb}, MBB-BY-MBB \cite{mbb-by-mbb}, or HITON \cite{hinton_markov_blanket_alg}.

\paragraph{Granger Methods.}
In temporal data, Granger \cite{granger_causality} causality is commonly employed to assess pairwise dependencies \cite{granger_causality_hawkes, shtp}. It is based on the assumption that \emph{causes precede effects and should improve the predictability of the effect}. Recently, \cite{han2025root} proposed a Granger-inspired causal discovery framework in multivariate time series using an encoder-decoder architecture. Their method achieves state-of-the-art root-cause identification via autoregressive modeling of structured time series. \cite{shtp} explores Granger causality under low-resolution temporal data using Hawkes processes \cite{hawkeppp} and shows gains in F1 across time granularities. However, their setup also assumes relatively small event vocabularies. CASCADE \cite{cueppers2024causal} recovers DAGs from temporal event data under a Poisson process assumption but is limited to smaller event spaces (\(\sim200\) types). All these approaches remain limited and are applied in poor experimental settings in which the number of variables (or events) does not reflect real-world and industrial applications. Additionally, Granger causality refers more to probabilistic causation \cite{Eells_1991} than Pearl causality~\cite{pearl_1998_bn}. It is often considered a relatively weak notion of causality that excludes instantaneous and inhibitory effects and is usually unable to discover true causal dependencies. In Granger causality, baking a cake is causal to a birthday.

\paragraph{Score-Based Methods.}
These methods search over graph structures to maximize a scoring function (e.g.,\ the Bayesian Information Criterion, BIC, which trades off likelihood against the number of parameters, or the Bayesian Dirichlet equivalent uniform score, BDeu).
The Greedy Equivalence Search (GES) \cite{ges} is the most popular method and relies on decomposable metrics (the score is a sum of per-node terms) that are score-equivalent (graphs encoding the same conditional independencies receive the same score), formally stated in Definitions~\ref{def:decomp}–\ref{def:local_consistency} (Appendix~\ref{appendix:part2}).

\paragraph{Optimization-Based and Neural Structural Learning Methods.}
A recent line of work formulates causal discovery as a continuous optimization problem, enabling gradient-based learning of DAG structures. The NOTEARS framework \cite{dag_no_tears} introduced a smooth acyclicity constraint that relaxes the discrete DAG search into a differentiable objective, making it possible to learn graph structures via standard gradient descent. Then, GOLEM \cite{golem} proposed a more scalable likelihood-based formulation, while DAG-GNN \cite{neural_dag} leveraged variational autoencoders and graph neural networks.

\paragraph{Temporal and Point-Process-Based Models.}
In sequential domains, Hawkes Processes and Temporal Point Processes model temporal causation via excitation kernels \cite{hawke_process_and_app_2025}. In Part~\ref{part1} we mentioned that the excitation matrix \(\Phi \in \mathbb{R}^{|\mathcal{X}|\times |\mathcal{X}|}\) was used as a proxy for causality. This excitation matrix is usually binarized to produce the adjacency matrix of the corresponding graph.
These approaches impose parametric assumptions and are interesting to discover causal relationships between different event streams \cite{cause}, rather than a single high-cardinality one like in vehicle fault sequences.

\subsection{Multi-Label Causal Discovery}
In multi-label settings, the objective is to recover the direct causes of each label \(Y_j\) among a large set of events \(\boldsymbol{X}\).
Instead of estimating the complete graph, one seeks the Markov Boundary:
\(
\mathcal{MB}(Y_j) \subseteq \mathbf{X}.
\)

\subsection{Evaluation Metrics for Multi-Label Causal Discovery}
In the multi-label setting (events-to-outcome), we aim to identify the set of direct temporal causes $\mathcal{P}_j \subseteq \mathcal{X}$ for each outcome label $y_j$. Following the Markov assumption, this set corresponds to the Markov Boundary $\mathcal{MB}(Y_j)$ of the label. We evaluate the alignment between the ground-truth boundary $\mathcal{MB}(Y_j)$ and the inferred boundary $\widehat{\mathcal{MB}}(Y_j)$.
\subsubsection{Set-Based Metrics}
To measure the local recovery of the causal neighborhood for each label, we utilize the following metrics:\begin{itemize}\item \textbf{Precision} ($P$) quantifies the reliability of the inferred causes, measuring the proportion of discovered event types that are true causal parents:
$$P = \frac{|\widehat{\mathcal{MB}}(Y_j) \cap \mathcal{MB}(Y_j)|}{|\widehat{\mathcal{MB}}(Y_j)|}
$$
\item \textbf{Recall} ($R$) (or Sensitivity) measures the completeness of the discovery, capturing the proportion of the true causal mechanism recovered by the model:
\[
R = \frac{|\widehat{\mathcal{MB}}(Y_j) \cap \mathcal{MB}(Y_j)|}{|\mathcal{MB}(Y_j)|}
\]

\item \textbf{F1-Score} ($F_1$) provides the harmonic mean of precision and recall. It is our primary metric for causal discovery, as it penalizes both the hallucination of spurious causes and the omission of weak causal signals:
\[
F_1 = 2 \cdot \frac{P \cdot R}{P + R}
\]
\end{itemize}

\subsection{Evaluation Metrics for Event-to-Event Causal Discovery}
For global structure recovery (Chapter~\ref{c7:event_to_event}), we reconstruct the full event-type adjacency matrix $\hat{A} \in \{0, 1\}^{|\mathcal{X}| \times |\mathcal{X}|}$. Unlike set-based evaluation, graph-based evaluation must account for the structural topology of the system.

\subsection{Structural Hamming Distance (SHD)}
The Structural Hamming Distance is the standard metric for comparing two causal graphs $\mathcal{G}$ and $\hat{\mathcal{G}}$. It counts the minimum number of edge operations (additions, deletions, or orientation reversals) required to transform the estimated graph into the ground truth. In our directed temporal setting, where orientation is fixed by time, SHD reduces to the $L_1$ distance between the flattened adjacency matrices:%

$$\text{SHD}(A, \hat{A}) = \sum_{i=1}^{|\mathcal{X}|} \sum_{j=1}^{|\mathcal{X}|} \left| A_{ij} - \hat{A}_{ij} \right|$$\noindent

\chapter{Sample-Level Multi-Label Causal Discovery}\label{c7:multi_label_one_shot_causal_discovery}
\chaptermark{Sample-Level Multi-Label Causal Discovery}
\glsresetall
Building upon the predictive foundations established in the previous chapter, where autoregressive Transformer models were used to anticipate \emph{when} and \emph{what} error patterns occur, we now turn to a complementary and equally essential question: \emph{why} these failures arise. We observe that autoregressive models trained on next-event prediction already encode the conditional distributions required by information theory metrics such as information gain (Def.~\ref{def:information_gain}), potentially usable for conditional independence testing in a graphical model. We therefore reuse the pretrained CarFormer and EPredictor weights, without fine-tuning, as \glspl{nade}. Consequently, in this chapter we address the challenge of causal discovery in large-scale event sequences, in which outcomes, such as error patterns,\footnote{As in Part~\ref{part1}, an EP here denotes the labeled outcome being causally explained, i.e., only its name/identity matters; it is not yet expressed as its Boolean rule. Chapter~\ref{c8:carep} recovers that rule.} emerge from preceding diagnostic events in a single sample during inference (\textit{sample-level}). It is based on the following publication:
\begin{description}[style=nextline,leftmargin=0cm,labelsep=0em]
\item[\textbf{One-Shot Multi-Label Causal Discovery in High-Dimensional Event Sequences.}] \cite{math2025oneshot}
Hugo Math, Robin Schön, Rainer Lienhart, Workshop \emph{CauScien: Uncovering Causality in Science} at NeurIPS 2025, San Diego, USA, December 2025.
\end{description}

\section{Introduction}

Understanding causal relations among diagnostic events is a crucial next step toward interpretable and automated reasoning in large-scale event-driven systems. Causal discovery in sequential data has broad relevance across domains such as healthcare \cite{MedBERT, bihealth}, cybersecurity \cite{MANOCCHIO2024122564}, flight operations \cite{flight_service_cd}, and vehicle diagnostics \cite{pdm_dtc_feature_extraction}.
Although recent progress in sequence modeling, particularly through Transformers \cite{tf, gpt, touvron2023llamaopenefficientfoundation} has improved our ability to capture temporal dependencies, most methods remain focused on prediction rather than explanation.
Traditional causal discovery methods, such as constraint-based or Granger-causal approaches~\cite{granger_causality}, face severe computational bottlenecks when applied to industrial-scale event data \cite{feature_selection_review, causality_based_feature_selection_2019, hasan2023a, cd_temporaldata_review}.
Their reliance on repeated conditional independence tests grows combinatorially with the number of event types, rendering them impractical for modern diagnostic systems.

Interpreting attention weights as a measure of causality is a promising direction \cite{tcdf, tf_causalinterpretation_neurips_2023} but lacks theoretical guarantees and often admits strong parametric assumptions as in \cite{tf_causalinterpretation_neurips_2023}. Some authors strongly criticize the reliance on attention scores, arguing that they provide no explanatory value \cite{att_not_explain_2019}.
On the other hand, repurposing Transformers as neural density estimators \cite{im2024usingdeepautoregressivemodels, moghimifar-etal-2020-learning} offers a promising direction, enabling the modeling of sequences using graphical models \cite{koller2009probabilistic} and the use of the learned joint probability distribution.

Nevertheless, all of these methods often aim to recover a single global causal graph, that is, one fixed graph defined over the entire vocabulary of event types (e.g., all \(29{,}100\) DTCs in our dataset) and assumed to hold identically across every sequence or individual. This graph is analogous to the type-level summary graph \(\mathcal{G}_s\) of Chapter~\ref{c7:event_to_event}, except that \(\mathcal{G}_s\) is recovered for one sequence at a time rather than imposed on the whole population. Such a graph does exist as a well-defined mathematical object, but committing to it as the \emph{only} representation is not only difficult to interpret for a lot of different events but also misaligned with how practitioners reason about causality in practice. In operational settings, reasoning about causality typically occurs at the \emph{sample-level}: given a single temporal sequence of observed events, one seeks to identify the specific causes leading to an outcome such as a system failure, defect, or disease with one sample during inference.
For instance, in vehicle diagnostics, engineers may ask: \emph{“Which diagnostic events most likely caused this particular error pattern?”}
This local perspective motivates a more efficient, explainable form of sample-level causal discovery.

To address this underexplored problem this chapter introduces OSCAR, to the best of our knowledge, the first sample-level causal discovery method that extends the predictive modeling framework of the previous chapter toward causal reasoning. OSCAR performs a single-pass, fully parallelized causal extraction on GPUs and recovers an interpretable causal graph with quantified causal indicators and uncertainty measures. This enables practical multi-label causal discovery at a scale previously unattainable, supporting thousands of event types and hundreds of outcome labels simultaneously.
This chapter, therefore, marks a conceptual shift: from purely predictive modeling to causal understanding, laying the foundation for the automated reasoning systems developed in the following chapters.

\section{Related Work}

\subsection{Neural Autoregressive Density Estimation}\label{c7:section:nades}
\gls{nade}s were initially introduced for density estimation via chain-rule factorization of the joint distribution using neural networks in the original paper of Bengio~\cite{NIPS1999_e6384711}, later extended through recurrent architectures \cite{cho-etal-2014-learning, lstm} and Transformers \cite{tf}. These models are trained using next-token prediction by minimizing the negative log-likelihood of observing the sequence \(s = (x_0, \ldots, x_L)\). Recall that the joint probability \(P(s)\) can be expressed using the chain rule as in Eq.~\ref{eq:autoreg_factorization}. Recent work has explored autoregressive models as tools for causal inference. For example, \cite{density_estimator_2O21_journal_ci} leverages density estimators to simulate interventions and compute average treatment effects. \cite{im2024usingdeepautoregressivemodels} shows that autoregressive language models can approximate Bayesian networks (Fig.~\ref{fig:markov_boundary_identification}), treating the model itself as a statistical engine for causal inference. These findings motivate our use of pretrained Transformers to estimate conditional mutual information (Def.~\ref{def:cmi}) between events and labels.

Specifically, we repurposed these models as \gls{nade}s for both the events and labels, allowing us to quickly estimate the conditional probabilities of the next event \(x_t\) and labels \(\boldsymbol{y}\) given past events \((x_0, \cdots, x_{t-1})\).

\subsection{Transformers as Causal Learners}\label{c7:section:transformers}
Transformer-based models have gained growing attention in causal discovery literature. \cite{tf_learns_gradient_structure} showed that when trained on sequences generated from in-context Markov chains, they can implicitly learn latent causal graphs, where attention weights align with the adjacency matrix of the actual causal structure.
For sequential data, \cite{tf_causalinterpretation_neurips_2023} analyzes self-attention under the assumption that data is generated by a linear-Gaussian structural causal model (SCM) \cite{Spirtes2001CausationPA}. They relate the covariance of endogenous variables to attention scores and apply conditional independence (CI) tests to the final layer’s outputs to recover a partial ancestral graph. Our work builds on this idea by leveraging Transformers to grant full access to the trajectory \(H_t\) using the self-attention mechanism. %

\subsection{Multi-label Causal Discovery}\label{c7:section:multi_label_cd}
Multi-label causal discovery seeks to identify the Markov Boundary or \(\mathcal{MB}\) (Def.~\ref{def:markov_boundary}) of each label, i.e., its minimal set of parents, children, and spouses—such that the label is conditionally independent of all other variables given its \(\mathcal{MB}\) \cite{optimal_feature_set_cd}. This boundary constitutes an optimal feature set for tasks such as explainable modeling and feature selection under the faithfulness assumption.

However, rather than learning the full joint causal graph, which is known to be NP-hard \cite{bn_np_hard}—\emph{we focus on recovering local causal structure} (LCS) \cite{causality_based_feature_selection_2019}: discovering minimal sub-graphs from events to labels within a single sequence. This formulation makes the problem tractable for high-cardinality event vocabularies and better suited for real-world production scenarios. Unlike event-to-event causal learning, multi-label causal discovery remains unexplored in event sequences \cite{cd_temporaldata_review, hasan2023a}, yet its potential applications are broad across various domains.

\section{Methodology}\label{sec:oneshotmarkov}
Fig.~\ref{fig:oscar} depicts OSCAR, our causal discovery method for event sequences. Concretely, the pipeline unfolds in five steps, matching the panels of Fig.~\ref{fig:oscar} from left to right. (1) The pretrained backbone \(\text{Tf}_x\) (CarFormer) infers the next-event distribution over \(\mathcal{X}\) for the observed trajectory (Eq.~\eqref{eq:prob_x}). (2) \(N\) alternative trajectories are drawn via combined top-\(k\)/nucleus sampling (each next event is drawn only among the \(k\) most probable event types whose cumulative probability stays below \(p\)), forming a batch of \(N\) plausible trajectories \(x^{(l)}_{<t}\) (Appendix~\ref{abl:sampling}). (3) The \(N\) trajectories are fed to the label model \(\text{Tf}_y\) (EPredictor), which reuses the hidden states of \(\text{Tf}_x\) as-is, without any additional fine-tuning. (4) \(\text{Tf}_y\) infers, for every sampled trajectory, the conditional probability of each label \(Y_j\) with and without the candidate event \(X_t\) (Eq.~\eqref{eq:prob_y}). (5) The resulting \(N\) estimates are averaged into the Monte Carlo estimator \(\hat{I}_N\) (Eq.~\eqref{eq:cmi_approx}) and the causal indicator \(\mathcal{C}\) (Eq.~\eqref{eq:causal_indic}), and compared against a per-label dynamic threshold \(\tau_j\) to decide whether \(X_t\) belongs to the Markov Boundary \(\mathcal{MB}(Y_j)\). In Fig.~\ref{fig:oscar}, blue boxes are the two pretrained Transformers and green boxes are the operations vectorized across the \(N\) sampled trajectories. The lemmas and theorems below are directly followed by their proofs.

\begin{figure}[!h]
    \centering
\begin{adjustbox}{width=1.2\textwidth, center}
    \begin{tikzpicture}[x=1cm, y=1cm, -, font=\footnotesize]
  \tikzset{
    mdl/.style={draw=blue!55!black, fill=blue!10, rounded corners=4pt, align=center, thick},
    cmp/.style={draw=green!45!black, fill=green!10, rounded corners=4pt, align=center, thick},
    arr/.style={-{Stealth[length=1.8mm]}, thick},
    stepn/.style={circle, draw, fill=white, inner sep=0.8pt, font=\scriptsize\bfseries},
    gnode/.style={circle, draw, thick, minimum size=7mm, inner sep=0pt, fill=white},
  }
  \def\ys{3.1, 2.3, 0.9, 0.1}
  \node[mdl, minimum width=1.9cm, minimum height=3.9cm] (tfx) at (2.4,1.6) {$\mathrm{Tf}_x$\\(CarFormer)};
  \node[stepn] at ($(tfx.north west)+(0.25,-0.25)$) {1};
  \foreach \yy/\ll in {3.1/x_0, 2.3/x_1, 0.9/x_{t-1}, 0.1/x_t}
    \draw[arr] let \p1=(tfx.west) in (0.65,\yy) node[left] {$\ll$} -- (\x1,\yy);
  \node at (0.35,1.6) {$\vdots$};
  \node[cmp, minimum width=2.3cm, minimum height=3.9cm] (smp) at (5.3,1.6) {Sample $N$\\trajectories\\$x^{(l)}_{<t}$, $l{=}1..N$\\(top-$k$/nucleus)};
  \node[stepn] at ($(smp.north west)+(0.25,-0.25)$) {2};
  \foreach \yy in \ys
    \draw[arr] let \p1=(tfx.east), \p2=(smp.west) in (\x1,\yy) -- (\x2,\yy);
  \node[mdl, minimum width=1.9cm, minimum height=3.9cm] (tfy) at (8.3,1.6) {$\mathrm{Tf}_y$\\(EPredictor)\\[3pt]{\scriptsize reuses the}\\{\scriptsize hidden states}\\{\scriptsize of $\mathrm{Tf}_x$}};
  \node[stepn] at ($(tfy.north west)+(0.25,-0.25)$) {3};
  \foreach \yy in \ys
    \draw[arr] let \p1=(smp.east), \p2=(tfy.west) in (\x1,\yy) -- (\x2,\yy);
  \node[cmp, minimum width=2.7cm, minimum height=3.9cm] (cmi) at (13.5,1.6) {Average over\\the $N$ trajectories:\\[2pt]$\hat{I}_N(Y_j, X_t\mid X_{<t})$\\and $\mathcal{C}(Y_j, X_t)$};
  \foreach \yy in \ys
    \draw[arr] let \p1=(tfy.east), \p2=(cmi.west) in (\x1,\yy) -- (\x2,\yy);
  \node[above, font=\scriptsize] at ($(tfy.east)!0.5!(cmi.west)+(0,-0.7)$) {$p(Y_j\mid x^{(l)}_{<t})$};
  \node[above, font=\scriptsize] at ($(tfy.east)!0.5!(cmi.west)+(0,-1.5)$) {$p(Y_j\mid x_t, x^{(l)}_{<t})$};
  \node[stepn] at ($(tfy.east)!0.5!(cmi.west)+(0,1.75)$) {4};
  \node[stepn] at ($(cmi.north west)+(0.25,-0.25)$) {5};
  \draw[-{Stealth[length=2.6mm, width=3.2mm]}, line width=2.2pt, gray!60] let \p1=(cmi.east) in (\x1+0.15cm,1.6) -- (\x1+1.45cm,1.6)
    node[midway, above=4pt, text=black] {$\hat{I}_N > \tau_j$};
  \begin{scope}[shift={(17.0,0)}]
    \fill[orange!20, draw=orange!70!black, dotted, thick] (0.25,2.35) ellipse (0.75 and 0.55);
    \fill[gray!15, draw=gray!70!black, dotted, thick] (0.25,0.35) ellipse (0.8 and 1.15);
    \node[gnode] (xc) at (0.25,3.55) {$X_c$};
    \node[gnode] (x2) at (0.25,2.35) {$X_{t-2}$};
    \node[gnode] (x1) at (0.25,0.95) {$X_{t-1}$};
    \node[gnode] (x0) at (0.25,-0.25) {$X_t$};
    \node[gnode] (y1) at (2.9,2.35) {$Y_1$};
    \node[gnode] (y2) at (2.9,0.6) {$Y_2$};
    \draw[densely dotted, thick] (xc) -- (x2);
    \draw[arr] (x2) -- (y1) node[midway, above, font=\scriptsize] {$\mathcal{C}\pm\sigma$};
    \draw[arr] (x1) -- (y2) node[midway, above, font=\scriptsize] {$\mathcal{C}\pm\sigma$};
    \draw[arr] (x0) -- (y2) node[midway, below right=-1pt, font=\scriptsize] {$\mathcal{C}\pm\sigma$};
    \node[font=\scriptsize, text=orange!50!black] at (-0.85,2.85) {$\widehat{\mathcal{MB}}(Y_1)$};
    \node[font=\scriptsize, text=gray!50!black] at (-0.9,-0.75) {$\widehat{\mathcal{MB}}(Y_2)$};
  \end{scope}
\end{tikzpicture}
    \end{adjustbox}
    \caption{\textbf{The Overview of OSCAR}: \uline{O}ne-\uline{S}hot multi-label \uline{C}ausal \uline{A}utoRegressive discovery. Blue: the two pretrained Transformers \(\text{Tf}_x\) (CarFormer) and \(\text{Tf}_y\) (EPredictor); green: operations vectorized across the \(N\) sampled trajectories \(x^{(l)}_{<t}\). Steps (1)--(5) are described in Section~\ref{sec:oneshotmarkov}. An edge \(X_t \rightarrow Y_j\) is kept if \(\hat{I}_N > \tau_j\) and is annotated with the causal indicator \(\mathcal{C}\) (Eq.~\eqref{eq:causal_indic}) and its standard deviation \(\sigma\); \(\widehat{\mathcal{MB}}(Y_1), \widehat{\mathcal{MB}}(Y_2)\) are the resulting estimated Markov Boundaries of \(Y_1, Y_2\).}
    \label{fig:oscar}
\end{figure}

\subsection{Autoregressive Event Sequence Models.}
We begin by repurposing the two Transformer architectures introduced in Part~\ref{part1} to perform next event (\textit{CarFormer} as \(\text{Tf}_{x}\)) and next labels (\textit{EPredictor} as \(\text{Tf}_y\)) prediction.
These two autoregressive Transformers model the conditional probability distribution of the next events and labels conditioned on the past sequence of observed events \(X_{<t} = (X_0, \cdots, X_{t-1})\), where, following the indexing convention of Chapter~\ref{c3:foundation_causal_discovery}, \(X_t\) denotes the event at position \(t\) of the sequence. The predictive distributions are defined as:
\begin{align}
P_{\theta_x}(X_t| X_{<t}) &\triangleq \textit{Softmax}(\boldsymbol{h}^{x}_{t-1}) = \text{Tf}_x(x_{<t}) \label{eq:prob_x} \\
P_{\theta_y}(Y|X_t, X_{<t}) &\triangleq \textit{Sigmoid}(\boldsymbol{h}^{y}_{t}) = \text{Tf}_y(x_{\leq t})\label{eq:prob_y}
\end{align}
Here, \(\boldsymbol{h}^{x}_{t-1}, \boldsymbol{h}^{y}_{t} \in \mathbb{R}^{d}\) are the logits produced by the two Transformer heads of \( \text{Tf}_x\) and \(\text{Tf}_y\) parametrized by \(\theta_x, \theta_y\). The majority of \(\text{Tf}_x\) (except the heads) serves as a backbone for \(\text{Tf}_y\), see Fig.~\ref{fig:oscar}.

\subsection{Assumptions}

Working with causal structure learning from observed data requires several assumptions. As stated in the foundation, we assume: (1) the causal Markov assumptions \cite{pearl_1998_bn}, which state that a variable is conditionally independent of its non-descendants given its parents, (2) no hidden confounders (A\ref{assumption:causal_sufficiency}), (3) temporal precedence (A\ref{assumption:temporal_precedence}) and  (4) a faithfulness via the defined BN~(Def.~\ref{def:bn_faithfulness}). Formally, we assume the following throughout this chapter:

\begin{assumption}[Stationary Markovian Dynamic System]\cite{koller2009probabilistic}
\label{assumption:stationarity}
Let \(X_{\leq t} = (X_0, \ldots, X_t)\) denote the trajectory observed up to step \(t\). We assume that the transition probability \(P(Y^{(t+1)}_{j}|X_{\leq t})\) does not depend on the specific step \(t\) at which it is evaluated. We therefore write it as a time-invariant transition probability \(P(Y_{j}|X_{\leq t})\), such that for any step \(t \geq 0\) and any observed trajectory \(x_{\leq t}\):
\begin{equation}
    P(Y^{(t+1)}_j = y|X_{\leq t} = x_{\leq t}) = P(Y_{j} = y|X_{\leq t} = x_{\leq t})
\end{equation}
\end{assumption}
\noindent Stationarity allows us to compute the correct posteriors \(P(Y^{(t)}_j|X_{<t}, X_t)\) even at different steps \(t\).

\begin{assumption}[Oracle Models]
\label{assumption:oracle}
We assume that two autoregressive Transformer models, \(\text{Tf}_x\) and \(\text{Tf}_y\), are trained via maximum likelihood on a dataset of multi-labeled event sequences \(\mathcal{D} = \{S^{(1)}_l, \cdots, S^{(n)}_l\}\), and can perfectly approximate the true conditional distributions of events and labels:
\begin{align}
    P(X_t|\textit{Pa}_{\mathcal{G}}(X_t)) &= P_{\theta_x}(X_t|\textit{Pa}_{\mathcal{G}}(X_t)) = \text{Tf}_x(x_{<t}) \\
    P(Y_j|\textit{Pa}_{\mathcal{G}}(Y_j)) &= P_{\theta_y}(Y_j|\textit{Pa}_{\mathcal{G}}(Y_j))=\text{Tf}_y(x_{\leq t})\label{eq:oracle}
\end{align}
\end{assumption}
This is a mild to strong assumption. Although we do not take any structural assumption about the data generating process, we assume that the two autoregressive models perfectly captures the true distribution. We challenge this assumption in the experiments as well as in Chapter~\ref{c7:event_to_event}.

\begin{assumption}[Bounded Lagged Effects]\label{assumption:lagged_effects}
Once events up to step \(t\) have been observed as \(X_{<t} = \{X_0, \cdots, X_{t-1}\}\), any lagged copy \(X^{(t + \Delta)}_t\) of the event \(X_t\), i.e., its influence carried \(\Delta\) steps into the future, is independent of \(Y_j\) conditioned on \(X_{<t}\):
\[
Y_j \perp X^{(t + \Delta)}_t | X_{<t}, \quad \Delta \geq 1,
\]
where the bound of one step (\(\Delta \geq 1\), i.e., the gap to the next event \(X_{t+1}\)) is the finite allowed delay for causal influence.
\end{assumption}
\noindent In other words, we allow the causal influence of event \(X_t\) on \(Y_j\) until the next event \(X_{t+1}\) is observed. We acknowledge that this may not hold for data with strong lagged effects (e.g., financial transactions), but it is relevant for log-based and error code-based data. An extended discussion on the impact of assumptions is provided later in Section~\ref{sec:6_discussion}

\subsection{Lemmas}
We can derive the Markov Boundary of each label in the Bayesian Network constructed from a multi-labeled sequence (Fig.~\ref{fig:markov_boundary_identification}). Note that the figure uses two different indexing conventions. On the left (observed sequence), the subscript of \(X_t\) denotes the sequential \emph{position} \(t \in \{0, \ldots, L-1\}\) of an event (Chapter~\ref{c3:foundation_causal_discovery}). On the right (causal graph), the subscript of \(Y_j\) denotes the \emph{identity} of a label \(j \in \{1, \ldots, |\mathcal{Y}|\}\), since all labels co-occur at the terminal position \(L\) of the sequence and are therefore indexed by label identity rather than by position.

\begin{figure}[!b]
    \centering
    \begin{tikzpicture}

        \draw[thick] (-2,0) -- (3.5,0) node[right] {Time};
        \draw[thick] (-2,0) -- (-2,2) node[right] {Event Type};

        \node[state,fill=blue] (x1) at (-1.5,0.5) {};
        \node[state,fill=blue] (x2) at (-0.5,0.5) {};
        \node[state,fill=blue] (x3) at (1,0.5) {};
        \node[state,fill=blue] (x4) at (2 ,0.5) {};

        \node[state,fill=red] (y) at (3,0.5) {};
        \node[state,fill=red] (y) at (3, 1) {};

        \draw[dashed] (-1.5,0) -- (-1.5,0.5);
        \draw[dashed] (-0.5,0) -- (-0.5,0.5);
        \draw[dashed] (1,0) -- (1,0.5);
        \draw[dashed] (2,0) -- (2,0.5);
        \draw[dashed] (3,0) -- (3,0.5);

        \node[below] at (-1.5,0) {\(X_0\)};
        \node[below] at (-0.5,0) {\(\cdots\)};

        \node[below] at (1,0) {\(X_{L-2}\)};
        \node[below] at (2,0) {\(X_{L-1}\)};

        \node[below] at (3,0) {\(\boldsymbol{Y}_{L}\)};

        \draw[thick,->] (4,-1) -- (5,-1) node[midway,above] {Identification};

        \node[state] (cx1) at (5.3,0) {\(X_{0}\)};
        \node[state] (cxd) at (7,0) {\(\cdots\)};
        \node[state] (cx2) at (9,0) {\(X_{L-2}\)};
        \node[state] (cxi) at (11,0) {\(X_{L-1}\)};
        \node[state] (cy) [above =of cxd] {\(Y_1\)};
        \node[state] (cy2) [above =of cxi] {\(Y_2\)};

        \path (cx1) edge (cxd);
        \path (cxd) edge  (cy);
        \path (cx1) edge[bend right=30] (cx2);
        \path (cx1) edge[bend right=30] (cxi);
        \path (cx2) edge (cxi);
        \path (cx2) edge[bend left=60] (cy2);
        \path (cxd) edge (cx2);
        \path (cxi) edge (cy2);
        \path (cxd) edge[bend left=60] (cxi);

        \node[draw=blue,dotted,fit=(cxd) (cxd), inner sep=0.2cm] (mb1) {};
        \node[anchor=south east, blue] at (mb1.north west) {\(\mathcal{MB}(Y_1)\)};

        \node[draw=cyan,dotted,fit=(cx2) (cxi), inner sep=0.2cm] (mb2) {};
        \node[anchor=west, cyan] at (mb2.east) {\(\mathcal{MB}(Y_2)\)};
    \end{tikzpicture}
    \caption{\textbf{Example of a Causal Graph} extracted from a multi-label \textcolor{blue}{event} sequence where \textcolor{blue}{\(\mathcal{MB}(Y_1)\)} represents the Markov Boundary of \textcolor{red}{\(Y_1\)} and \textcolor{cyan}{\(\mathcal{MB}(Y_2)\)} the Markov Boundary of \textcolor{red}{\(Y_2\)}. Left: the bold vector \(\boldsymbol{Y}_L\) denotes all labels co-occurring at the final position \(L\); right: after identification, each individual label \(Y_j\) is shown with its own Markov Boundary \(\mathcal{MB}(Y_j)\).}
    \label{fig:markov_boundary_identification}
\end{figure}

\begin{lemma}[Markov Boundary Equivalence]\label{lemma:mb_par}
In a multi-label event sequence \(S_l\) and under the temporal precedence assumption A\ref{assumption:temporal_precedence}, the Markov Boundary of each label \(Y_j\) is only its parents such that \(\forall X \in \{\mathbf{U} - \textit{Pa}_{\mathcal{G}}(Y_j)\},  X \perp Y_j|\textit{Pa}_{\mathcal{G}}(Y_j) \Leftrightarrow \mathcal{MB}(Y_j) = \textit{Pa}_{\mathcal{G}}(Y_j) \).
\end{lemma}

\begin{proof}\label{proof:lemma_mb_equivalence}
    Let \(<\mathbf{U}, \mathcal{G}, P>\) be the BN composed of the events from the multi-labeled sequence. Following the
temporal precedence assumption A\ref{assumption:temporal_precedence}, the labels \(\boldsymbol{y_L}\) can only be caused by past events \((x_0, \cdots, x_L)\); we assume that labels do not cause any other labels. Thus, \(Y_j\) has no descendants, so no children and spouses. Therefore, together with the Markov Assumption we know that \(\forall X \in \{\mathbf{U} - \textit{Pa}_{\mathcal{G}}(Y_j)\}: Y_j \perp X|\textit{Pa}_{\mathcal{G}}(Y_j)\). Which is the definition of the MB (Def. \ref{def:markov_boundary}). Consequently \(\mathcal{MB}(Y_j) = \textit{Pa}_{\mathcal{G}}(Y_j)\).
\end{proof}

\noindent To recover the causal graph \(\mathcal{G}\), we must make sure it is identifiable from the observational data.

\begin{lemma}[Identifiability of \(\mathcal{G}\)]\label{lemma:oracle_identifiability}
    Assuming the faithfulness condition holds for the true causal graph \(\mathcal{G}\). Let \(\text{Tf}_x\) and \(\text{Tf}_y\) be Oracle models perfectly approximating the actual conditional distributions of events and labels, respectively. The joint distribution \(P_{\theta_x, \theta_y}\) can then be constructed, and any conditional independence detected from the distributions estimated by \(\text{Tf}_x\) and \(\text{Tf}_y\) corresponds to a conditional independence in \(\mathcal{G}\):
    \[
    X_t \perp_{\theta_x, \theta_y} Y_j \mid X_{<t} \quad \implies \quad X_t \perp_{\mathcal{G}} Y_j \mid X_{<t}.
    \]
    Where \( \perp_{\theta_x, \theta_y}\) denotes the independence entailed by the joint probability \(P_{\theta_x, \theta_y}\).
\end{lemma}

\begin{proof}\label{proof:lemma_identifiyability_g_mb}
We assume that the data is generated by the associated causal graph \(\mathcal{G}\) following the BN from a multi-labeled sequence \(S_l\). Given that the Oracle models \(\text{Tf}_x\) and \(\text{Tf}_y\) are trained to perfectly approximate the true conditional distributions, for any variable \(U\) in the graph, we have:
\[
    P(U | \text{Pa}_{\mathcal{G}}(U)) =
    \begin{cases}
        P(Y_j | \text{Pa}_{\mathcal{G}}(Y_j)) = P_{\theta_y}(Y_j | \text{Pa}_{\mathcal{G}}(Y_j)), & \text{if } U = Y_j \in \boldsymbol{Y} \\
        P(X_t | \text{Pa}_{\mathcal{G}}(X_t)) = P_{\theta_x}(X_t | \text{Pa}_{\mathcal{G}}(X_t)), & \text{if } U = X_t.
    \end{cases}
\]
The joint distribution \(P_{\theta_x, \theta_y}\) can then be constructed using the chain rule
\[
P_{\theta_x, \theta_y}(X_0, \cdots, X_t, Y_1, \cdots, Y_{|\mathcal{Y}|}) = \prod^t_{k=0}P(X_k|Pa_{\mathcal{G}}(X_k)) \prod^{|\mathcal{Y}|}_{l=1}P(Y_l|\text{Pa}_{\mathcal{G}}(Y_l))
\]
\noindent By the faithfulness assumption \cite{pearl_1998_bn}, if the conditional independencies hold in the data, they must also hold in the causal graph \(\mathcal{G}\):
\[X_t \perp Y_j | X_{<t} \implies X_t \perp_{\mathcal{G}} Y_j | X_{<t}\]
Since we can approximate the true conditional distributions, it follows that:
\[X_t \perp_{\theta_x, \theta_y} Y_j |X_{<t} \implies X_t \perp Y_j | X_{<t} \implies X_t \perp_{\mathcal{G}} Y_j | X_{<t}\]
Where \( \perp_{\theta_x, \theta_y}\) denotes the independence entailed by the joint probability \(P_{\theta_x, \theta_y}\).
As a result, the graph \(\mathcal{G}\) can be identified from the observational data.
\end{proof}

\subsection{Conditional Mutual Information Estimation via Autoregressive Models}

OSCAR works like a constraint-based causal discovery algorithm, where the conditioning set of nodes \(X_{<t}\) increases over time as we unfold the Bayesian Network in Fig.~\ref{fig:markov_boundary_identification}. Specifically, we would like to assess how much additional information an event \(X_t\) occurring at step \(t\) provides about label \(Y_{j}\) when we already know the past sequence of events \(X_{<t} = \{X_0, \cdots, X_{t-1}\}\). We essentially try to answer if:
\[P(Y_{j}|X_t, X_{<t}) = P(Y_{j}|X_{<t}) \Leftrightarrow
D_{KL}(P(Y_{j}|X_t, X_{<t})\|P(Y_{j}|X_{<t})) = 0\]
where \(D_{KL}\) denotes the \textit{Kullback-Leibler divergence} (Def.~\ref{def:dkl}). The distributional difference between the conditionals \(P(Y_j|X_t, X_{<t}), P(Y_j| X_{<t})\) is akin to Information Gain \(I_G\) conditioned on past events \(x_{<t} = \{x_0, \cdots, x_{t-1}\}\) (Def.~\ref{def:information_gain}):

\begin{equation}
    I_G(Y_j, x_t|x_{<t}) \triangleq D_{KL}(P(Y_j|X_t=x_t, X_{<t} = x_{<t}) || P(Y_j|X_{<t} = x_{<t}))
\end{equation}

\noindent Throughout this thesis, the event follows its conditional distribution given the trajectory, \(x_t \sim p(x_t \mid x_{<t})\), and the trajectory follows its marginal distribution, \(x_{<t} \sim p(x_{<t})\). The expectation of the information gain over the event is the difference between the conditional entropies, denoted as \(H\):
\begin{equation}
    \mathbb{E}_{x_t \sim p(x_t|x_{<t})}[I_G(Y_j, x_t|x_{<t})] = H(Y_j|x_{<t}) - H(Y_j|X_t, x_{<t}) \label{eq:info_gain}
\end{equation}

\noindent and its further expectation over the trajectory is the CMI, which assesses conditional independence (Def.~\ref{def:conditional_independence}):
\begin{equation}
    I(Y_{j}, X_t|X_{<t}) \triangleq H(Y_j|X_{<t}) - H(Y_j|X_t, X_{<t})\label{eq:cmi_theorique}
    = \mathbb{E}_{x_{<t} \sim p(x_{<t})}\Big[\mathbb{E}_{x_t \sim p(x_t|x_{<t})}\big[I_{G}(Y_j, x_t|x_{<t})\big]\Big]
\end{equation}

\noindent OSCAR works on a single observed sequence, in which the event at position \(t\) is known. The question is therefore whether this observed event \(x_t\) carries information about \(Y_j\) beyond its trajectory. We keep \(x_t\) fixed and average the information gain over the trajectories only, which defines the \emph{sample-level CMI} of the observed event:
\begin{equation}\label{eq:cmi_sample}
    I(Y_j, x_t \mid X_{<t}) \triangleq \mathbb{E}_{x_{<t} \sim p(x_{<t})}\big[I_G(Y_j, x_t \mid x_{<t})\big]
\end{equation}
It is non-negative and equals zero if and only if \(P(Y_j \mid x_t, x_{<t}) = P(Y_j \mid x_{<t})\) for almost every trajectory; replacing the fixed \(x_t\) by \(x_t \sim p(x_t \mid x_{<t})\) inside the expectation gives back the CMI of Eq.~\ref{eq:cmi_theorique}.
To approximate Eq.~\ref{eq:cmi_sample}, a naive Monte Carlo estimation~\cite{Doucet2001} is performed where we draw $N$ random trajectories
\(x^{(l)}_{<t} = \{x^{(l)}_0, \ldots, x^{(l)}_{t-1}\}\), denoting the $l$-th sampled \emph{trajectory} (Monte Carlo particle), from the model \(P_{\theta_x}\)\footnote{Averaging the information gain over multiple sampled past trajectories enables tractable computation as we do not need to enumerate over all possible trajectories \(\{x_0, x_1, x_2\}, \{x_2, x_1, x_0\}, \{x_1, x_2, x_0\}, \cdots\)}:
\begin{align}\label{eq:cmi_approx}
\hat{I}_N(Y_{j}, x_t \mid X_{<t})
&= \frac{1}{N} \sum_{l=1}^N I_G(Y_j, x_t \mid x^{(l)}_{<t})
\end{align}
In practice (Algorithm~\ref{alg:oscar}), only the first \(c\) positions of a trajectory are resampled, each from \(\text{Tf}_x\) given the observed events before it, while the observed events from position \(c\) on are kept; this resampling approximates \(p(x_{<t})\). Under Assumption~\ref{assumption:oracle} (Oracle model), \(P_{\theta_x}\) equals the true distribution, and \(\hat{I}_N\) is an unbiased estimator of Eq.~\ref{eq:cmi_sample} when the trajectories follow \(p(x_{<t})\).
Additionally, the label probabilities are clipped to \([\varepsilon, 1-\varepsilon]\) (Algorithm~\ref{alg:oscar}), and the KL divergence between two Bernoulli distributions satisfies \(D_{KL}(\mathrm{Bern}(p) \,\|\, \mathrm{Bern}(q)) \leq -\log \min(q, 1-q)\). Each term of Eq.~\ref{eq:cmi_approx} is therefore bounded:
\[0 \leq I_G(Y_{j}, x_t \mid x^{(l)}_{<t}) \leq \log(1/\varepsilon)\]

\noindent Thus the posterior variance of \(f_t(x_{<t}) = I_G(Y_{j}, x_t \mid x_{<t})\) satisfies \(\sigma^2_{f_t} \triangleq \mathbb{E}_{x_{<t} \sim p(x_{<t})}[f^2_t(x_{<t})] - I(Y_j, x_t \mid X_{<t})^2 < +\infty\) \cite{Doucet2001}, the variance of \(\hat{I}_N\) is equal to \(\textit{var}(\hat{I}_N) = \frac{\sigma^2_{f_t}}{N}\), and from the strong law of large numbers the estimate \(\hat{I}_N\) converges almost surely to the sample-level CMI:
\begin{align}
\hat{I}_N(Y_j, x_t \mid X_{<t}) &\xrightarrow[N \to +\infty]{\text{a.s.}}
\mathbb{E}_{x_{<t} \sim p(x_{<t})}\!\left[ I_G(Y_{j}, x_t \mid x_{<t}) \right]
= I(Y_j, x_t \mid X_{<t}).
\end{align}

\subsection{Sequential Markov Boundary Recovery}

Under conditional independence, the sample-level CMI is exactly zero, but its estimate \(\hat{I}_N\) is not, because of sampling and model errors. In practice, a label-specific threshold \(\tau_j > 0\) is therefore applied to Eq.~\ref{eq:cmi_approx} to decide whether the observed event belongs to the Markov Boundary of \(Y_j\):
\begin{equation}\label{eq:cmi_epsilon}
    X_t \in \widehat{\mathcal{MB}}(Y_j) \quad \Leftrightarrow \quad \hat{I}_N(Y_j,  x_t \mid X_{<t}) > \tau_j
\end{equation}
\(\tau_j\) is dynamically computed for each label based on the mean and standard deviation of the estimates \(\hat{I}_N\) across the sequence, such that: \(\tau_j = \mu_{Y_j} + k \cdot \sigma_{Y_j}\),
where \(k\) controls the confidence interval. We analyze the effect of \(k\) in Appendix.~\ref{abl:threshold_selection}.

\begin{theorem}[Markov Boundary Identification in Event Sequences]
\label{th:mb-recovery}
If  \(S^{(k)}_l\) a multi-labeled sequence drawn from a dataset  \(\mathcal{D} = \{S^{(1)}_l, \cdots, S^{(n)}_l\}\) where two Oracle Models \(\text{Tf}_x\) and \(\text{Tf}_y\) were trained on, then under causal sufficiency (A\ref{assumption:causal_sufficiency}), bounded lagged effects (A\ref{assumption:lagged_effects}) and temporal precedence (A\ref{assumption:temporal_precedence}), the Markov Boundary of each label \(Y_j\) in the causal graph \(\mathcal{G}\) can be identified using conditional mutual information for CI-testing.
\end{theorem}

\begin{proof}\label{proof:th1}
    By induction over the sequence length \(L\) of the multi-label sequence \(S^{(k)}_l\), we want to show that under temporal precedence A\ref{assumption:temporal_precedence}, bounded lagged effects A\ref{assumption:lagged_effects}, causal sufficiency A\ref{assumption:causal_sufficiency}, Oracle Models A\ref{assumption:oracle} the Markov Boundary of label \(Y_j\) can be identified in the causal graph \(\mathcal{G}\). %
Let \(\widehat{\mathcal{MB}}^{(L)}(Y_j)\) denote the estimated Markov Boundary of \(Y_j\) after observing \(L\) events \(X_0, \ldots, X_{L-1}\).

\paragraph{Base Case} \(\mathbf{L=1}\):
Consider the BN for step \(L=1\) following the Markov assumption \cite{pearl_1998_bn} with two nodes \(X_0, Y_j\). Using \(\text{Tf}_x, \text{Tf}_y\) as Oracle Models A\ref{assumption:oracle}, we can express the conditional probabilities for any node \(U\):

\begin{equation}
    P(U | \text{Pa}_{\mathcal{G}}(U)) =
    \begin{cases}
            P(X_0) = P_{\theta_x}(X_0 | [CLS]) \; \text{if}\; U \in \boldsymbol{X} \\
            P(Y_j|X_0) = P_{\theta_y}(Y_j | X_0)\; \text{otherwise}
    \end{cases}
\end{equation}
where \(P(.|[CLS])\) is considered the marginal distribution \(P(.)\).
\noindent Assuming that P is faithful (Def.~\ref{def:bn_faithfulness}) to \(\mathcal{G}\), no hidden confounders bias the estimate (A\ref{assumption:causal_sufficiency}) and temporal precedence (A\ref{assumption:temporal_precedence}), we can estimate the CMI \ref{eq:cmi_theorique} such that iff \(I(X_0, Y_j|\emptyset) > 0 \Leftrightarrow Y_j \not\perp_{\theta_x, \theta_y} X_0 \implies Y_j \not\perp_{\mathcal{G}} X_0\) (Lemma \ref{lemma:oracle_identifiability}).

Since we assume temporal precedence A\ref{assumption:temporal_precedence}, we can orient the edge such that \(X_0\) must be a parent of \(Y_j\) in \(\mathcal{G}\). Using Lemma \ref{lemma:mb_par}, we know that \(\text{Pa}_{\mathcal{G}}(Y_j) = \mathcal{MB}(Y_j) \implies X_{0} \in \mathcal{MB}(Y_j)\), thus we must include \(X_0\) in \(\widehat{\mathcal{MB}}^{(1)}(Y_j)\), otherwise not.

\paragraph{Heredity}
For \(L = t\), we obtained \(\widehat{\mathcal{MB}}^{(t)}(Y_j)\) with the BN over the events \(X_0, \ldots, X_{t-1}\).
Now for \(L = t+1\), the BN has \(t+2\) nodes denoted as \(\boldsymbol{U'} = (X_0, \cdots, X_{t-1}, X_{t}, Y_j)\). Using the Oracle Models A\ref{assumption:oracle} and following the Markov assumption \cite{pearl_1998_bn}, we can estimate the following conditional probabilities for any nodes \(U \in \boldsymbol{U'}\):

\begin{equation}
    P(U | \text{Pa}_{\mathcal{G}}(U)) =
    \begin{cases}
        P(Y_j | \text{Pa}_{\mathcal{G}}(Y_j)) = P_{\theta_y}(Y_j | \text{Pa}_{\mathcal{G}}(Y_j)), & \text{if } U \in \boldsymbol{Y} \\
        P(X| \text{Pa}_{\mathcal{G}}(X)) = P_{\theta_x}(X|\text{Pa}_{\mathcal{G}}(X)), & \text{otherwise}.
    \end{cases}
\end{equation}
By bounded lagged effects (A\ref{assumption:lagged_effects}) we know that the causal influence of past \(X_{<t}\) on \(Y_j\) has expired. In addition, no hidden confounders (A\ref{assumption:causal_sufficiency}) bias the independence testing. Finally,
using Eq.~\ref{eq:cmi_theorique} with \(X_{<t} = \{X_0, \ldots, X_{t-1}\}\), we can estimate the CMI such that iff
\(I( Y_j, X_{t}| X_{<t}) > 0 \Leftrightarrow Y_j \not\perp_{\theta_x, \theta_y}  X_{t} | X_{<t} \implies Y_j \not\perp_{\mathcal{G}} X_{t} | X_{<t}\) (Lemma \ref{lemma:oracle_identifiability}).

Since we assume temporal precedence A\ref{assumption:temporal_precedence}, we can orient the edge so that \(X_{t}\) must be a parent of \(Y_j\) in \(\mathcal{G}\). Using Lemma \ref{lemma:mb_par}, we know that \(\text{Pa}_{\mathcal{G}}(Y_j) = \mathcal{MB}(Y_j) \implies X_{t} \in \mathcal{MB}(Y_j)\).
Thus, \(X_{t} \in \widehat{\mathcal{MB}}^{(t+1)}(Y_j)\), which represents \(\mathcal{MB}(Y_j)\) after \(t+1\) observed events.

Finally, \(\widehat{\mathcal{MB}}^{(t+1)}(Y_j)\) still recovers the Markov Boundary of \(Y_j\) such that \[\forall U \in \{\boldsymbol{U'} - \widehat{\mathcal{MB}}^{(t+1)}(Y_j)\}, Y_j \perp U|\widehat{\mathcal{MB}}^{(t+1)}(Y_j)\]
\end{proof}

\noindent Intuitively, Theorem~\ref{th:mb-recovery} states that if our Transformers perfectly approximate the true joint distribution, then testing conditional mutual information at each step is sufficient to recover the Markov Boundary of each label sequentially. By induction, we prove that with bounded lagged effects of the previous events, we can restrict their causal influence and recover the correct \(\mathcal{MB}\) of each label in the associated Bayesian Network (Fig.~\ref{fig:markov_boundary_identification}).

\subsection{Computation}
\paragraph{Context}
\noindent Following the EPredictor architecture seen in Chapter~\ref{c2:ep_prediction_based_on_live_data}, to ensure stable conditional entropy estimates and reliable predictions from \(\text{Tf}_y\), the CMI is computed after observing \(c\) events (\textit{context}). This design choice also enables out-of-the-box parallelization. By sampling \(N\) trajectories \(x^{(l)}_{\leq c}\) for the context, the CMI is independently computed across positions \(t \in [c, L]\).
One caveat is the phenomenon of entropy saturation \cite{shannon1951prediction}, whereby the conditional entropy \(H(Y_j \mid X_{\leq t})\) diminishes as the trajectory \(X_{\leq t}\) grows longer. Recall the Lemma~\ref{lemma:entropy_conditionning}:
\[
H(Y_j \mid X_{t+1}, X_{\leq t}) \leq H(Y_j \mid X_t, X_{<t}).
\]
In other words, once a sufficiently informative context is observed, future uncertainty becomes minimal. Therefore, context \(c\) and sequence length \(L\) must be carefully selected to balance informativeness and computational efficiency. In our experiments, we set \(c=15 \) and \(L=128\). An ablation on \(c\) and the quality of the NADEs can be found in the Appendix \ref{appendix:nades}.
\paragraph{Parallelization}
A key advantage of our approach is its scalability. Fig.~\ref{fig:oscar} shows all parallelized steps on GPUs. Unlike traditional methods whose complexity depends on the event and label cardinality \(|\mathcal{X}|\) and \(|\mathcal{Y}|\) \cite{feature_selection_review}, \uline{our method is agnostic to both}. CMI estimations are independently performed for all positions \(t \in [c, L]\), with the sampling pushed into the batch dimension and results averaged across the \(N\) trajectories using Eq.~\ref{eq:cmi_approx}. This reduces the per-sample computational overhead via batch parallelism from \(\mathcal{O}(\text{BS} \times N \times L)\) sequential forward passes to a single batched forward pass as \(\mathcal{O}(1)\). The full procedure, including the top-\(k\)/nucleus sampling step, is given in Algorithm~\ref{alg:oscar} (Appendix~\ref{sec:oscarimplementation}). Concretely, \(\text{Tf}_x\) is called once on the batch of observed sequences; the \(N\) trajectories of all sequences are stacked into a single batch, so that \(\text{Tf}_y\) evaluates all of them in one forward pass; the CMI estimates of all positions \(t\) and labels \(Y_j\) are then obtained by element-wise operations and an average over the trajectory dimension; finally, the per-label threshold is applied along the position dimension.

\section{Causal Indicator}
While deterministic DAGs reveal structural dependencies, they often obscure the \textit{magnitude} and \textit{direction} of influence between variables. In many settings, a small subset of causal events may exert disproportionate influence on the probability of a label. Also, causal relationships can be either \textit{excitatory} or \textit{inhibitory}. That is, the presence of a cause may either increase or decrease the likelihood of its effect.

For instance, if
\(P(Y_j \mid X_t, X_{<t}) < P(Y_j \mid X_{<t})\)
then \(X_t\) negatively influences \(Y_j\), yet still constitutes a valid causal relationship \cite{pearl_2009}. Without quantifying the effect direction and strength, such cases may mislead an operator.
Given that we can estimate both conditionals \(P(Y_j \mid X_t, X_{<t})\) and \(P(Y_j \mid X_{<t})\), we define the \textit{causal indicator} \(\mathcal{C} \in [-1, 1]\) between an event \(X_t\) and a label \(Y_j\) under causal sufficiency and the trajectory \(X_{<t}\) that we assume fixed for every measurement \cite{measure_of_causal_strength_oxford}:

\begin{equation}\label{eq:causal_indic}
\mathcal{C}(Y_j, X_t; X_{<t}) := \mathbb{E}_{p(x_{<t})}[P(Y_j =1\mid X_t=x_t, X_{<t} = x_{<t}) - P(Y_j =1\mid X_{<t} = x_{<t})]
\end{equation}
where the expectation is taken over trajectories \(x^{(l)}_{<t}\) sampled from the learned model distribution \(P_\theta\) using the same Monte Carlo estimation as Eq.~\ref{eq:cmi_approx}. We compute the mean and standard deviations over trajectories \(x^{(l)}_{<t}\) to provide uncertainty estimates.
It measures how much the observed event \(x_t\) shifts the probability of the label relative to the prediction from the trajectory alone. If the conditioning \(X_{<t}\) fully blocks all unwanted backdoor paths, i.e., non-causal paths between \(X_t\) and \(Y_j\) that enter \(X_t\) through a common cause (A~\ref{assumption:causal_sufficiency}), the first term equals the interventional probability \(P(Y_j = 1 \mid do(X_t = x_t), x_{<t})\), and Eq.~\ref{eq:causal_indic} is an average causal effect (ACE)~\cite{pearl_2009} whose reference is the trajectory-only prediction. Here, \(\mathcal{C} > 0\) indicates a positive influence and \(\mathcal{C} < 0\) reflects an inhibitory effect.
While several metrics for causal strength exist—including Causal Power \cite{Cheng1997} and Good’s measure \cite{measure_of_causal_strength_oxford}, we adopt this measure for its simplicity of interpretation. An operator can easily read it and get a sense of the rise in likelihood of the label \(Y_j\). We employ the term causal \textit{indicator} to distinguish it from causal strength measures, which, if using this formulation, can be problematic as pointed out by \cite{quantifyingcausalinfluence}. Ours serves more as an indication than a strength, which is here the conditional mutual information.

\section{Empirical Evaluation}\label{c7_one_shot:section:empirical_eval}
\subsection{Settings}\label{sec:settings}
We used a \(g4dn.12xlarge\) instance from AWS Sagemaker to run comparisons. It contains 48 vCPUs and 4 NVIDIA T4 GPUs. During inference, we used 16-bit (fp16) floating-point precision for \(\text{Tf}_y\) and 32-bit (fp32) for \(\text{Tf}_x\). We used a combination of F1-Score, Precision, and Recall with different averaging \cite{reviewmultilabellearning} to perform the comparisons. The code for OSCAR, \(\text{Tf}_x, \text{Tf}_y\) and the evaluation are provided  \footnote{\url{https://github.com/Mathugo/OSCAR-One-Shot-Causal-AutoRegressive-discovery}} as well as the anonymized version of the dataset for reproducibility purposes.

\subsection{Vehicle Event Sequences Dataset}
We evaluated our method on an anonymized real-world vehicular test set of \(n=300,000\) sequences. It contains \(|\mathcal{Y}|=474\) different error patterns and about \(|\mathcal{X}| = 29,100\) different DTCs forming sequences of \( \approx 150 \pm 90\) events. We used 105m parameters backbones as \(\text{Tf}_x, \text{Tf}_y\). The two NADEs did not see the test set during training. %
We set the elements of the EP rules (as defined in \ref{sec:ep_def}, Eq.~\ref{eq:ep_def}) as the correct Markov Boundary for each label \(y_j\) in the tested sequences. Importantly, the rules are subject to change over time by automotive engineers, contain some noise, and are biased, which makes it more difficult to extract the true \(\mathcal{MB}\). Further, there is about \(12\%\) missing ground truth \(\mathcal{MB}\) rules for certain \(Y_j\).

\subsection{Comparisons}
Although no existing method directly targets sample-level multi-label causal discovery \cite{cd_temporaldata_review}, we benchmark OSCAR against local structure learning (LSL) algorithms that estimate global Markov Boundaries. This includes established approaches such as CMB \cite{cmb}, MB-by-MB \cite{WANG2014252}, PCD-by-PCD \cite{pcdpcd}, IAMB \cite{iamb} from the \textit{PyCausalFS} package \cite{causality_based_feature_selection_2019}, as well as the more recent, state-of-the-art MI-MCF \cite{mimcf}. Nine random folds of the test data were created and converted into a multi-one-hot data-frame where one row represents one sequence, and each column represents an event type or label \((\mathcal{X, Y})\). We set the target nodes as the labels with \textit{PyCausalFS}.

\subsection{Results}
We first drew \(n=50,000\) random sequences from our dataset and performed comparisons (Table~\ref{tab:performance_comparison}). We observed that even under this reduced setup, LSL algorithms failed to compute the Markov Boundaries within a 3-day timeout, far exceeding practical limits for deployment. OSCAR, on the other hand, shows robust classification over a large amount of events \((29,100)\), especially \(55\%\) precision, in a matter of minutes.
This behavior highlights the current infeasibility of multi-label causal discovery in massive discrete event sequences. This positions OSCAR as a more feasible approach for large-scale sample-level causal discovery in production environments.

\noindent To enable at least partial comparison, we further sub-sampled to \(n=500\) sequences (Table~\ref{tab:performance_comparison_n_500}) to enable a faster computation. However, the number of labels in the test set is approximately the same for \(n=500\) samples. This resulted in a poor number of CI tests for the baselines. As a result, LSL algorithms output empty \(\mathcal{MB}\) sets after multiple hours. Especially MI-MCF with even \(500\) samples suffers from its expensive CMI testing. Thus, traditional algorithms suffer from having either too many samples and taking days to compute or too little data to even function.

\begin{table}[h]
\centering

\begin{tabular}{lcccc}
\hline
\textbf{Algorithm} & \textbf{Precision}↑ & \textbf{Recall}↑ & \textbf{F1}↑ & \textbf{Running Time (min)}↓ \\ \hline
IAMB & - & - & - & \(>4320\) \\
CMB & - & - & - & \(>4320\) \\
MB-by-MB & - & - & - & \(>4320\) \\
PCDbyPCD & - & - & - & \(>4320\) \\
MI-MCF & - & - & - & \(>4320\) \\
OSCAR & \(\mathbf{55.26 \pm 1.42}\) & \(\mathbf{31.37 \pm 0.82}\) & \(\mathbf{40.02 \pm 1.03}\) & \(\boldsymbol{11.7}\)\\
\hline
\end{tabular}\caption{Comparisons of \(\mathcal{MB}\) retrieval with \(n=50,000\) samples, \(|\mathcal{X}| = 29,100, |\mathcal{Y}|=474\) averaged over \(6-\)folds. Classification metrics averaging is 'weighted' and shown as sample-level for OSCAR. The symbol ’-’ indicates that the algorithm did not output the \(\mathcal{MB}\) within 3 days. Metrics are given in \(\%\).}
\label{tab:performance_comparison}
\end{table}
\begin{table}[h]
\centering
\begin{tabular}{cccccc}
\hline
\textbf{Algorithm} & \textbf{Precision}↑ & \textbf{Recall}↑ & \textbf{F1 }↑ &  \textbf{Running Time (min)}↓ \\ \hline
IAMB &\(0.0 \pm 0.0\) & \(0.0 \pm 0.0 \) & \(0.0 \pm 0.0\) & \(129.4\) \\
CMB & \(0.0 \pm 0.0\) & \(0.0 \pm 0.0 \) & \(0.0 \pm 0.0\) & \( 128.7\)\\
PCDbyPCD & \(0.0 \pm 0.0\) & \(0.0 \pm 0.0 \) & \(0.0 \pm 0.0\) & \(129.1\) \\
MB-by-MB & \(0.0 \pm 0.0\) & \(0.0 \pm 0.0 \) & \(0.0 \pm 0.0\) & \( 140.3\) \\
MI-MCF & \(0.0 \pm 0.0\) & \(0.0 \pm 0.0\) & \(0.0 \pm 0.0\) & \(> 1440\) \\
OSCAR & \( \mathbf{54.78 \pm 2.91} \) & \( \mathbf{30.39 \pm
 2.39}\) & \( \mathbf{39.92\pm 2.25}\) & \( \mathbf{0.14}\) \\
 \hline
\end{tabular}
\caption{Comparisons of \(\mathcal{MB}\) retrieval with \(n=500\) samples over \(9\) folds.}
\label{tab:performance_comparison_n_500}
\end{table}

\noindent We illustrate the explainability provided by our method for the task of explaining error patterns observed in a vehicle (Fig.~\ref{fig:graph_steering_wheel_degradation}). A concrete use case for OSCAR in this context is to refine or develop new error-pattern rules based on OSCAR output predictions, such as non-common causal variables \cite{learningcommoncausalvarlabel} between labels, thereby improving the automation of quality processes.

\section{Limitations}\label{sec:6_discussion}

Our approach relies on several assumptions that enable sample-level causal discovery under practical and computational constraints.

\paragraph{Temporal Precedence}
Temporal precedence (A\ref{assumption:temporal_precedence}) simplifies causal directionality. However, this relies heavily on precise event time-stamping. Even though we only test \(X_t \rightarrow Y_j\), this could falsify the conditioning test \(X_{<t}\) and introduce spurious relationships.%

\paragraph{Bounded Lagged Effects}
The bounded lagged effects (A\ref{assumption:lagged_effects}) assumption enables us to restrict causal influence and recover the \(\mathcal{MB}\) of each label using Theorem~\ref{th:mb-recovery}. In most real-world sequences where the causally relevant part of the trajectory is short, this holds empirically. Nonetheless, in highly delayed causal chains, like financial transactions, some influences may be missed.

\paragraph{Causal Sufficiency}
As with many causal discovery approaches, we assume all relevant variables are observed (A\ref{assumption:causal_sufficiency}). It constitutes a mild assumption, especially in sample-level causal discovery where fewer data points are observed.

\paragraph{Inter-label Effects}
By definition, the labels are explained solely by events. While simplifying causal discovery, this assumption could be relaxed in future work by using the \textit{do} operator \cite{pearl_2009} to perform interventions on common causal variables of multiple labels. For example, our current framework estimates the Markov Boundaries for each label independently. However, inter-label dependencies can exist, particularly when labels share overlapping Markov Boundaries (e.g., \(\mathcal{MB}(Y_1) = \{X_1, X_3\}, \mathcal{MB}(Y_2) = \{X_1, X_2\}\)). We aim to investigate a 'Phase 2' for OSCAR, focusing on inter-label dependencies through simulated interventions. For instance, if we consider a sequence of two labels \(Y_1, Y_2\) with the Markov Boundaries above, we could perform counterfactual interventions by applying \(do(X_1=0), do(X_3=0)\). We then compute the average change in the likelihood of \(Y_1\), which, if non-zero, indicates a dependence between \(Y_1\) and \(Y_2\). \cite{learningcommoncausalvarlabel} points out that the assumptions of these inter-label dependencies are already anchored in the Markov Boundaries; we do the same in this chapter.

\paragraph{NADEs}
Using NADEs allows for relaxation of structural assumption about the underlying data-generating processes (e.g., Poisson Processes or SCMs). As demonstrated by the Ablations~\ref{appendix:nades}, the effectiveness of OSCAR hinges on the capacity of \(\text{Tf}_x\) and \(\text{Tf}_y\) to approximate true conditional probabilities (A\ref{assumption:oracle}) and provide Oracle CI-test. While assuming Oracle tests is common in the literature~\cite{mbb-by-mbb, feature_selection_review} and necessary to recover faithful causal structures, this remains a strong assumption. It is valid only to the extent that the models are perfectly trained. In multi-label classification settings, performance may degrade in underrepresented regions of the data distribution. For example, we analyze on a reduced dataset, the performance of OSCAR as a function of the \(\mathcal{MB}\) length:

\begin{figure}[ht]
    \centering
    \begin{adjustbox}{width=1.3\textwidth, center}
    \includegraphics[width=\textwidth]{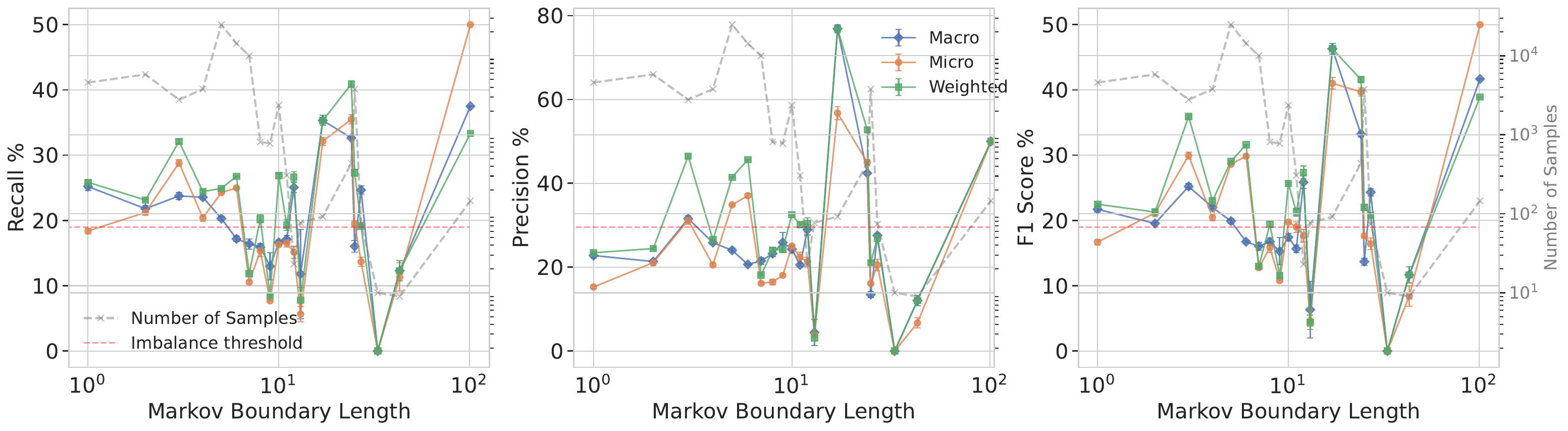}
    \end{adjustbox}
    \caption{\textbf{Evolution of the Classification Performance as a Function of the Markov Boundary Length} $|\mathcal{MB}(Y_j)|$ ($n=45969$ samples). We can identify that the Markov Boundary complexity is not the bottleneck but rather the number of samples per class (\textcolor{red}{imbalance threshold}), which reduces classification performance.}
    \label{fig:markov_len}
\end{figure}

\noindent Fig.~\ref{fig:markov_len} reveals the classification performance depending on the number of events in the ground truth \(\mathcal{MB}\). On the same plot is drawn in \textcolor{gray}{grey} the number of samples that each \(\mathcal{MB}\) length contains (to account for imbalance). We observe that generally, a bigger \(|\mathcal{MB}(Y_j)|\) does not imply a reduction in performance, highlighting the capability of OSCAR to retrieve complex Markov Boundaries in high-cardinality data. However, we observe that past a \emph{certain number of samples (imbalance threshold} in \textcolor{red}{red} \(\approx 7 \times 10^2\) samples), the classification metrics are directly correlated with the number of samples per \(|\mathcal{MB}(Y_j)|\). This indicates that \(\text{Tf}_x, \text{Tf}_y\) struggle to output proper conditional probabilities for rare classes, this results in a non faithful CI-test. Therefore, when using OSCAR and more generally assumption A\ref{assumption:oracle}, one should carefully assess class imbalance in the pretraining phase.

\section{Summary}
OSCAR extends autoregressive models toward causal discovery by repurposing them as NADEs.
The proposed methodology combines theoretical guarantees with parallelized computation on GPUs.
Under assumptions, such as temporal precedence, causal sufficiency, and bounded lagged effects, OSCAR can reliably recover the local causal structure underlying each observed sequence, as we have shown.
Thanks to its parallelized implementation on GPUs, the approach scales seamlessly to tens of thousands of event types and hundreds of labels, achieving causal discovery for \(50, 000\) of sequences in minutes where traditional constraint-based methods fail to compute within days.
By introducing a \emph{causal indicator}, OSCAR further quantifies the direction and magnitude of influence between events and outcomes, making causal discovery not only feasible but also interpretable for practitioners.
Applied to a large-scale vehicular dataset, OSCAR demonstrated the ability to uncover the correct EP rules. However, we highlighted the limitations of the assumptions, including the impact of under-training of EPredictor on rare classes, which can, in practice, break the Oracle assumption. Nevertheless, this constitutes an important conceptual change in our overall framework—from predicting \emph{when} and \emph{what} faults will occur, to understanding \emph{why} they arise. Such causal interpretability is essential in many industrial problems, in fault diagnostics but also in production lines where defects arrive sequentially.

\section{Outlook}
Chapter~7 addresses a distinct challenge: while OSCAR recovers causal structure within a single observed sequence, practitioners require a consensus causal graph that is stable across the full vehicle fleet. To this end, Chapter~\ref{c7:multi_label_causal_discovery} introduces CARGO, a graph aggregation method that uses the extracted per-sample causal graph of OSCAR to provide a global population-level structure.

\chapter{Population-Level Multi-Label Causal Discovery}\label{c7:multi_label_causal_discovery}
\glsresetall

It is essential in many real-world systems to recover the global set of causes for each outcome (e.g., diseases, manufacturing defects) across the entire available data. We aim to recover all the causes of each error pattern to potentially suggest new rules or automate the rules of unknown error patterns. This chapter is based on the following publication:
\begin{description}[style=nextline,leftmargin=0cm,labelsep=0em]
\item[\textbf{Towards Practical Multi-label Causal Discovery in High-Dimensional Event Sequences via One-Shot Graph Aggregation.}] \cite{math2025towards}
Hugo Math, Rainer Lienhart.
\textit{Workshop on Structured Probabilistic Inference \& Generative Modeling at NeurIPS 2025}, San Diego, USA, December 2025.
\end{description}

\section{Introduction}

Although the local view of causal relationships enables interpretable causal reasoning within individual sequences, real-world systems such as fleets of vehicles, patient populations, or distributed networks often share underlying causal mechanisms that recur across many realizations. Understanding these global causal dependencies is essential to building coherent diagnostic or decision-making frameworks that generalize beyond a single observation. To date, it remains largely unknown how to perform such population-level causal discovery across industrial-scale event sequences \cite{MANOCCHIO2024122564, MedBERT, bihealth, flight_service_cd, pdm_dtc_feature_extraction} as standard methods are computationally intractable. %

To address this, we reinterpret multi-label causal discovery for event sequences as a form of Bayesian model averaging \cite{HOET1999, pearl_1998_bn}, i.e., combining several candidate models instead of selecting a single one, where each sequence is treated as a sample from a local causal model. Specifically, each sequence induces a sample-level causal graph (i.e., a directed acyclic graph (DAG)) \cite{math2025oneshot} inferred from \gls{oscar}. Together, they are fused to form a global structure \cite{divide_conquer_mb_discovery}. This process, known as structural fusion \cite{consensus_bn_jose}, aggregates local graphs into a consensus causal graph over all observed samples.
For this purpose, we introduce \gls{cargo}: (\uline{C}ausal \uline{A}ggregation via \uline{R}obust \uline{G}raph \uline{O}perations), a scalable two-phase causal discovery framework. %

\section{Related Work}

\paragraph{Bayesian Model Averaging}
Fusing BNs has two primary applications: averaging models from different experts to learn a global representation~\cite{HOET1999}, or performing causal discovery in distributed settings with federated learning algorithms~\cite{fedpc_fedfci, Guo_Yu_Liu_Li_2024}.

\begin{figure}[!h]
    \centering
    \begin{tikzpicture}[->,>=stealth',shorten >=1pt,auto,node distance=1.5cm,
                        semithick, every state/.style={circle,draw,minimum size=1.2em}]

        \node[state] (a1) at (0,2) {\(X_1\)};
        \node[state] (b1) at (1.5,2) {\(X_2\)};
        \node[state] (c1) at (0.75,0.5) {\(Y_1\)};
        \path (a1) edge (c1);
        \path (b1) edge (c1);

        \node[state] (a2) at (0,-1) {\(X_2\)};
        \node[state] (b2) at (1.5,-1) {\(X_3\)};
        \node[state] (c2) at (0.75,-2.5) {\(Y_2\)};
        \path (a2) edge (c2);
        \path (b2) edge (c2);

        \node[state] (a3) at (-2,0.5) {\(X_1\)};
        \node[state] (b3) at (-0.5,0.5) {\(X_3\)};
        \node[state] (c3) at (-1.25,-1) {\(Y_2\)};
        \path (a3) edge (c3);
        \path (b3) edge (c3);

        \draw[thick,->] (2.5,-0.75) -- (4.5,-0.75) node[midway,above] {Structural Fusion};

        \node[state] (x1) at (6,1.5) {\(X_1\)};
        \node[state] (x2) at (7.5,1.5) {\(X_2\)};
        \node[state] (x3) at (9,1.5) {\(X_3\)};
        \node[state] (y1) at (6.75,0) {\(Y_1\)};
        \node[state] (y2) at (8.25,0) {\(Y_2\)};

        \path (x1) edge (y1);
        \path (x2) edge (y1);
        \path (x2) edge (y2);
        \path (x3) edge (y2);
        \path (x1) edge[bend right=20] (y2);
    \end{tikzpicture}
     \caption{\textbf{Structural Fusion Example}: individual set of causal graphs \(\{\mathcal{G}^{(k)}\}\) (left) aggregated into a fused DAG for multi-label event sequences (right) using a simple union.}
    \label{fig:structural_fusion_example}
\end{figure}
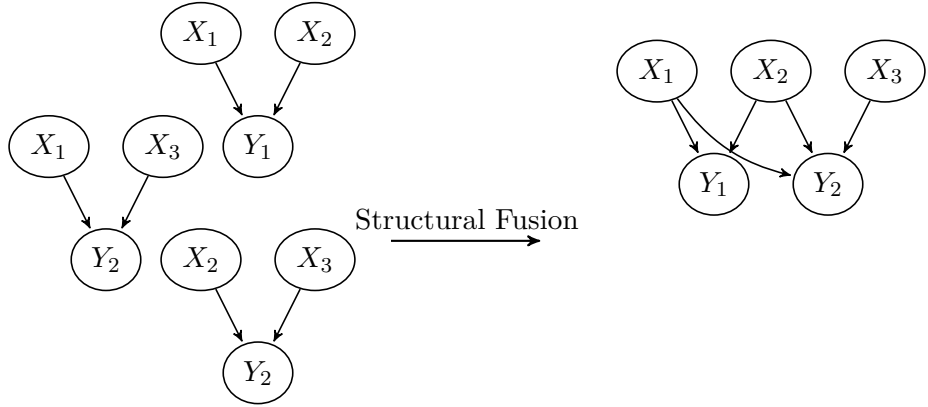

\noindent Formally, given a set of Bayesian Networks \(\{B_k\}^m_{k=1}\) with associated DAGs \(\{\mathcal{G}^{(k)} = (\mathbf{V}_k, E_k)\}^m_{k=1}, \mathbf{V}_k \in \mathbf{U}\) sharing the same finite set of node \(\mathbf{U}\), structural fusion aims to construct the DAG \(\mathcal{G}^* = (V, E), V \in \mathbf{U}\). Multiple  fusion methods exist and leverage either the probability distribution \(P\) by doing Bayesian Model Averaging \cite{HOET1999} or focus on the structural learning (Fig.~\ref{fig:structural_fusion_example}) of \(\mathcal{G}^*\) \cite{delSagrado2001, consensus_bn_jose, go2022robust, approx_fusionPUERTA2021155, Guo_Yu_Liu_Li_2024}, i.e., only merging the edges \(\{E_k\}^m_{k=1}\)to the correct nodes \(\{\mathbf{V}_k\}^m_{k=1}\).
Bayesian Model Averaging is considered to be an NP-hard \cite{consensus_bn_jose} problem. We focus on the structural learning and seek the merged edges \(E = \bigcup^m_{i=1} E^\sigma_i\). The consistent node ordering \(\sigma\) ensures acyclicity. The fused DAG \(\mathcal{G}^*\) is the minimal I-map (Def.~\ref{def:imaps}) of the intersection of the conditional independencies across all DAGs \(\mathcal{G}^{(k)} = (\mathbf{V}_k, E_k)^m_{k=1}\).

\paragraph{Greedy Equivalence Search}
Greedy equivalence search (GES) \cite{ges} is one of the most theoretically sound methods to recover a Markov equivalence class (MEC: Def.~\ref{def:mec}) of a DAG. Particularly in the context of infinite samples, GES provides theoretical guarantees of reaching the true graph.
Formally, GES searches for the MEC of the graph \(\mathcal{G}^*\) from the observational dataset \(\mathcal{D} \) with distribution \(P\) using a scoring function \(f_s\). This defines the optimization problem as:

\begin{equation}\label{eq:ges_opti}
    \mathcal{G}^* = \arg_{\mathcal{G}} \text{max} \; f_s(\mathcal{G}, \boldsymbol{D})
\end{equation}
\cite{ges} proved that, under parametric assumption, a large number of samples and using the Bayesian Information Criterion (BIC) as criterion \(f_s\), GES is guaranteed to recover an MEC of \(\mathcal{G}^*\) since it is score equivalent (Def.~\ref{def:score_equivalent}) and locally consistent (Def.~\ref{def:local_consistency}).
Some contemporary works point toward decomposing classical causal discovery for high-cardinality datasets into a subproblems (\emph{divide-and-conquer} approaches). \cite{laborda_ring_based_distirbuted} introduce a ring-based distributed algorithm for learning BN, \cite{divide_conquer_mb_discovery} explores a distributed approach for large-scale causal structure learning and \cite{recursive_mb_approach_pmlr_large_scale} for Markov Boundaries.

\paragraph{Frequency-Based}
Multiple heuristics have been developed to merge multiple BNs.
One uses an edge frequency cutoff \cite{Steele2008Consensus},
another relies on estimating the proportion of false positive edges and integer linear programming (ILP) \cite{consensus_bn_freq_cutoff_no} or a mix of both: \cite{torrijos2025informedgreedyalgorithmscalable}. As pointed out by \cite{consensus_bn_freq_cutoff_no}, choosing a frequency cutoff \(\tau\) is particularly challenging. Moreover, under class imbalance and the long-tail problem in labeled datasets \cite{longtailissue}, where a few labels have many samples and most labels have very few, the theoretical property of frequency approaches, such as the law of large numbers, might not hold.

\section{Methodology}
We now explain the two phases of CARGO, which are (1) the sample-level causal discovery (2) graph aggregation using adaptive thresholding.

\begin{figure}[!b]
    \centering
    \begin{adjustbox}{width=1.15\textwidth, center}
    \begin{tikzpicture}[x=1cm, y=1cm, -, font=\footnotesize]
  \tikzset{
    mdl/.style={draw=blue!55!black, fill=blue!10, rounded corners=4pt, align=center, thick},
    cmp/.style={draw=green!45!black, fill=green!10, rounded corners=4pt, align=center, thick},
    agg/.style={draw=orange!70!black, fill=yellow!15, rounded corners=4pt, align=center, thick},
    arr/.style={-{Stealth[length=1.8mm]}, thick},
    big/.style={-{Stealth[length=2.6mm, width=3.2mm]}, line width=2.2pt, gray!60},
    sn/.style={circle, draw, thick, minimum size=5.2mm, inner sep=0pt, fill=white, font=\scriptsize},
    phase/.style={rotate=90, font=\small\bfseries, text=gray!60!black},
  }
  \node[phase] at (-0.6,4.3) {Phase 1};
  \foreach \k/\y/\lab in {1/5.0/S^{(1)}, 2/4.4/S^{(2)}, 3/3.2/S^{(m)}}{
    \node[anchor=east] at (0.55,\y) {$\lab$};
    \draw[thick] (0.65,\y) -- (2.9,\y);
    \foreach \x in {0.9,1.3,1.7,2.1,2.5} \fill (\x,\y) circle (0.05);
    \fill[red!70!black] (2.85,\y) circle (0.07);
  }
  \node at (1.75,3.85) {$\vdots$};
  \node[font=\scriptsize, align=center] at (1.75,5.55) {$m$ labeled sequences};
  \draw[big] (3.15,4.1) -- (3.85,4.1);
  \node[cmp, minimum width=2.6cm, minimum height=2.0cm] (osc) at (5.3,4.1) {OSCAR\\(Fig.~\ref{fig:oscar})\\on every sequence,\\in parallel};
  \draw[big] (6.75,4.1) -- (7.45,4.1);
  \begin{scope}[shift={(7.7,4.6)}]
    \node[font=\scriptsize] at (0.65,1.05) {$\mathcal{G}^{(1)}$};
    \node[sn] (a1) at (0,0.5) {$X_1$}; \node[sn] (a2) at (1.3,0.5) {$X_2$}; \node[sn] (ay) at (0.65,-0.35) {$Y_1$};
    \draw[arr] (a1) -- (ay); \draw[arr] (a2) -- (ay);
  \end{scope}
  \begin{scope}[shift={(10.0,4.6)}]
    \node[font=\scriptsize] at (0.65,1.05) {$\mathcal{G}^{(2)}$};
    \node[sn] (b1) at (0,0.5) {$X_2$}; \node[sn] (b2) at (1.3,0.5) {$X_3$}; \node[sn] (by) at (0.65,-0.35) {$Y_2$};
    \draw[arr] (b1) -- (by); \draw[arr] (b2) -- (by);
  \end{scope}
  \node at (12.0,4.25) {$\cdots$};
  \begin{scope}[shift={(12.7,4.6)}]
    \node[font=\scriptsize] at (0.65,1.05) {$\mathcal{G}^{(m)}$};
    \node[sn] (c1) at (0,0.5) {$X_1$}; \node[sn] (c2) at (1.3,0.5) {$X_3$}; \node[sn] (cy) at (0.65,-0.35) {$Y_1$};
    \draw[arr] (c1) -- (cy); \draw[arr] (c2) -- (cy);
  \end{scope}
  \node[font=\scriptsize, align=center] at (11.0,3.35) {$m$ sample-level graphs (over event types)};
  \draw[dashed, gray] (-0.95,2.8) -- (14.4,2.8);

  \node[phase] at (-0.6,1.1) {Phase 2};
  \draw[big] (13.35,3.05) -- (13.35,2.3);
  \node[agg, minimum width=3.6cm, minimum height=2.3cm] (frq) at (12.55,1.0) {};
  \node[font=\scriptsize, align=center] at (12.55,1.55) {edge frequency\\$\hat{\pi}_{i,j} = \frac{1}{m}\sum_k B^{(k)}_{i,j}$};
  \foreach \i/\h in {0/0.75, 1/0.22, 2/0.62, 3/0.55, 4/0.15}
    \fill[blue!35] (11.1+0.6*\i,0.2) rectangle ++(0.35,\h);
  \draw[dashed, red!70!black, thick] (10.95,0.6) -- (14.15,0.6) node[above left, font=\tiny, text=red!70!black] {$\tau_j$};
  \node[font=\tiny] at (12.55,0.02) {candidate edges $X_i\to Y_j$};
  \draw[big] (10.55,1.0) -- (9.85,1.0);
  \node[agg, minimum width=3.6cm, minimum height=2.3cm] (thr) at (7.85,1.0) {};
  \node[font=\scriptsize] at (7.85,1.72) {adaptive threshold $\tau_j(m_j)$};
  \draw[-{Stealth[length=1.4mm]}] (6.35,0.3) -- (9.45,0.3) node[below, font=\tiny, anchor=north east] {$m_j$ (log)};
  \draw[-{Stealth[length=1.4mm]}] (6.35,0.3) -- (6.35,1.65);
  \draw[red!70!black, thick, domain=0:2.9, samples=50] plot ({6.4+\x}, {0.5+1.0/(1+exp(4*(\x-1.4)))});
  \draw[dotted] (7.8,0.3) -- (7.8,1.0) node[pos=0, below, font=\tiny] {$m_0$};
  \draw[big] (5.85,1.0) -- (3.8,1.0) node[midway, below=4pt, text=black, font=\scriptsize, align=center] {keep if\\$\hat{\pi}_{i,j} \geq \tau_j(m_j)$};
  \begin{scope}[shift={(0.6,0.3)}]
    \node[font=\scriptsize] at (1.3,1.85) {global graph $\mathcal{G}^{*}$};
    \node[sn] (g1) at (0,1.15) {$X_1$}; \node[sn] (g2) at (1.3,1.15) {$X_2$}; \node[sn] (g3) at (2.6,1.15) {$X_3$};
    \node[sn] (gy1) at (0.65,0) {$Y_1$}; \node[sn] (gy2) at (1.95,0) {$Y_2$};
    \draw[arr] (g1) -- (gy1); \draw[arr] (g2) -- (gy2); \draw[arr] (g3) -- (gy2);
  \end{scope}
\end{tikzpicture}
    \end{adjustbox}
    \caption{\textbf{Overview of CARGO}. Phase 1 (sample level, top): OSCAR (Fig.~\ref{fig:oscar}) is applied in parallel to each of the \(m\) labeled sequences and yields one sample-level graph \(\mathcal{G}^{(k)}\) per sequence, recorded over event types \(X_i\). Phase 2 (aggregation, bottom): for every candidate edge \(X_i \rightarrow Y_j\), the edge frequency \(\hat{\pi}_{i,j}\) across the \(m\) graphs is compared with the label-specific adaptive threshold \(\tau_j(m_j)\), a logistic decay in the label support \(m_j\) with midpoint \(m_0\) (Eq.~\eqref{eq:adaptive_threshold}); the kept edges form the global graph \(\mathcal{G}^*\).}
    \label{fig:cargo}
\end{figure}

\subsection{Structural Fusion of Markov Boundaries}

Let \(\{\mathcal{G}^{(k)} = (\mathbf{V}_k, E_k)\}^m_{k=1}\) be the set of generated DAGs by the Phase 1 from the dataset \(\mathcal{D}\) containing \(m\) i.i.d multi-labeled sequences \(\{S^{(k)}_l\}^m_{k=1}\) drawn from a joint distribution \(P(X,Y)\). Each graph \(\mathcal{G}^{(k)}\) represents local Markov Boundaries identified within a multi-labeled sequence \(S^{(k)}_l\). Our objective is to fuse these local graphs into a single, global consensus graph \(\mathcal{G}^* =  (\boldsymbol{U}, E)\) (see Fig.~\ref{fig:structural_fusion_example}) with the events always as parents of labels \(Y_j\) such as:
\[\text{Pa}_{\mathcal{G}^*}(Y_j) \subseteq \{X_1, \cdots, X_{|\mathcal{X}|}\}\]
\noindent\textbf{Index convention.} Unlike Chapter~\ref{c7:multi_label_one_shot_causal_discovery}, where \(X_t\) denotes the event at position \(t\) of a single sequence, the fused graph \(\mathcal{G}^*\) is defined over event \emph{types}, since positions are not comparable across sequences. In this chapter, \(X_i\) therefore denotes the event type \(i \in \{1, \ldots, |\mathcal{X}|\}\): a local edge \(X_t \rightarrow Y_j\) found by OSCAR at position \(t\) of sequence \(S^{(k)}_l\) is recorded in \(\mathcal{G}^{(k)}\) as the type-level edge \(X_i \rightarrow Y_j\), where \(i\) is the type of \(x_t\). This is the same position-to-type projection that defines the summary graph in Chapter~\ref{c7:event_to_event} (Def.~\ref{def:iscg}).

\noindent A naive fusion approach, such as taking the simple union of all edges \(E = \bigcup^m_{k=1} E^\sigma_k\) works if and only if the Oracle models (A\ref{assumption:oracle}) yield faithful CI-tests in Phase 1 such that the local graphs \(\mathcal{G}^{(k)}\) are faithful to \(P(X,Y)\)  (\cite{consensus_bn_jose}, Theorem 4).
Here, the ordering \(\sigma\) does not matter since we are dealing with Markov Boundaries. Thus, \(\mathcal{G}^*\) is naturally a DAG because we considered previously that outcome labels are solely explained by events, which simplifies acyclicity. We define the Bernoulli variable \(B^{k}_{i,j}\) for each potential edge \((X_i \rightarrow Y_j)\) within each sequence \(S^{(k)}_l\):
\[B^k_{i,j} = \begin{cases}
    1 \; \text{if the edge} \; X_i \rightarrow Y_j \; \text{is present in } \; \mathcal{G}^{(k)} \\
    0 \; \text{otherwise}
\end{cases}\]
\noindent Under the Oracle Models assumption (A\ref{assumption:oracle}), Phase 1 acts as a perfect conditional independence tester. Consequently, the detection of an edge in a local graph \(\mathcal{G}^{(k)}\) corresponds precisely to a true causal dependency in the global graph \(\mathcal{G}^*\). The probability of this event \(P(B^k_{i,j} = 1)\) is therefore the true marginal probability of the edge's existence, which we denote as \(\pi_{i,j}\). The empirical frequency, \(\hat{\pi}_{i,j}(m)\), of the edge \((X_i \rightarrow Y_j)\) after observing \(m\) sequences is the sample mean of these i.i.d Bernoulli variables \(
\hat{\pi}_{i,j}(m) = \frac{1}{m}\sum_{k=1}^{m} B_{ij}^{k}\).
By the Law of Large Numbers (LLN), as the number of i.i.d sequences \(m\) tends to infinity, the empirical frequency converges in probability to the true expected value of the random variable:
\[
\hat{\pi}_{i,j}(m) \xrightarrow{p} \mathbb{E}[B_{i,j}^{(k)}] = \pi_{i,j}
\]
 Given a sufficiently large number of sequences, the empirical frequency \(\hat{\pi}_{i,j}(m)\) serves as a consistent estimator for the true probability of the edge's existence in the global DAG \(\mathcal{G}^*\).

\subsection{Aggregation under Imperfect CI-tests}\label{sec:agg_imperfect_ci_tests}
The assumption of an Oracle CI-tester, while necessary for initial theoretical guarantees and common in the literature~\cite{recursive_mb_approach_pmlr_large_scale}, is invariably violated in practice due to factors like model capacity, limited data or class imbalance as seen in the experiments of Chapter~\ref{c7:multi_label_one_shot_causal_discovery}. The extracted sample-level graphs will most likely violate the independencies in \(\mathcal{G}^{(k)}\) and thus in \(\mathcal{G}^*\). Let us model the performance of our sample-level CI-test for any potential edge \(X_i \rightarrow Y_j\) with the following error rates: (1) False Positive Rate (Type I Error): \(\alpha = P(\text{detect \(|\) edge is spurious)}\) (2) True Positive Rate (Sensitivity): \(1-\beta = P(\text{detect \(|\) edge is causal)}\).
We operate under the reasonable assumption that our classifier is significantly better than random, implying that \(1-\beta \gg \alpha\).
The expected value of our Bernoulli variable \(B_{i,j}^k\) is now:
\begin{align*}
\mathbb{E}[B_{ij}^{k}] &= P(B_{ij}^{k}=1) \\
&= P(\text{detect}\;|\;\text{causal})P(\text{causal}) + P(\text{detect}\;|\;\text{spurious})P(\text{spurious}) \\
&= (1-\beta)\pi_{ij} + \alpha(1-\pi_{ij})
\end{align*}
The empirical frequency now converges to this new expectation. For a true edge \((\pi_{i,j} = 1)\), the empirical frequency converges to a high value:
\(
\hat{\pi}_{ij}(m) \xrightarrow{p} 1 - \beta
\) and for spurious edge \((\pi_{i,j} = 0)\) it converges to a low value: \(\hat{\pi}_{ij}(m) \xrightarrow{p} \alpha\). This reveals the critical role of frequency aggregation as a mechanism for separating signal from noise.

\subsection{Adaptive Fusion for Structural Discovery in Long-Tail Distributions}
A primary challenge in real-world data is the long-tail distribution of outcome labels, where a few "head" labels possess abundant data while the vast majority of "tail" labels are data-sparse  \cite{longtailissue}. %

\begin{figure}[!h]
    \centering
    \includegraphics[width=0.9\linewidth]{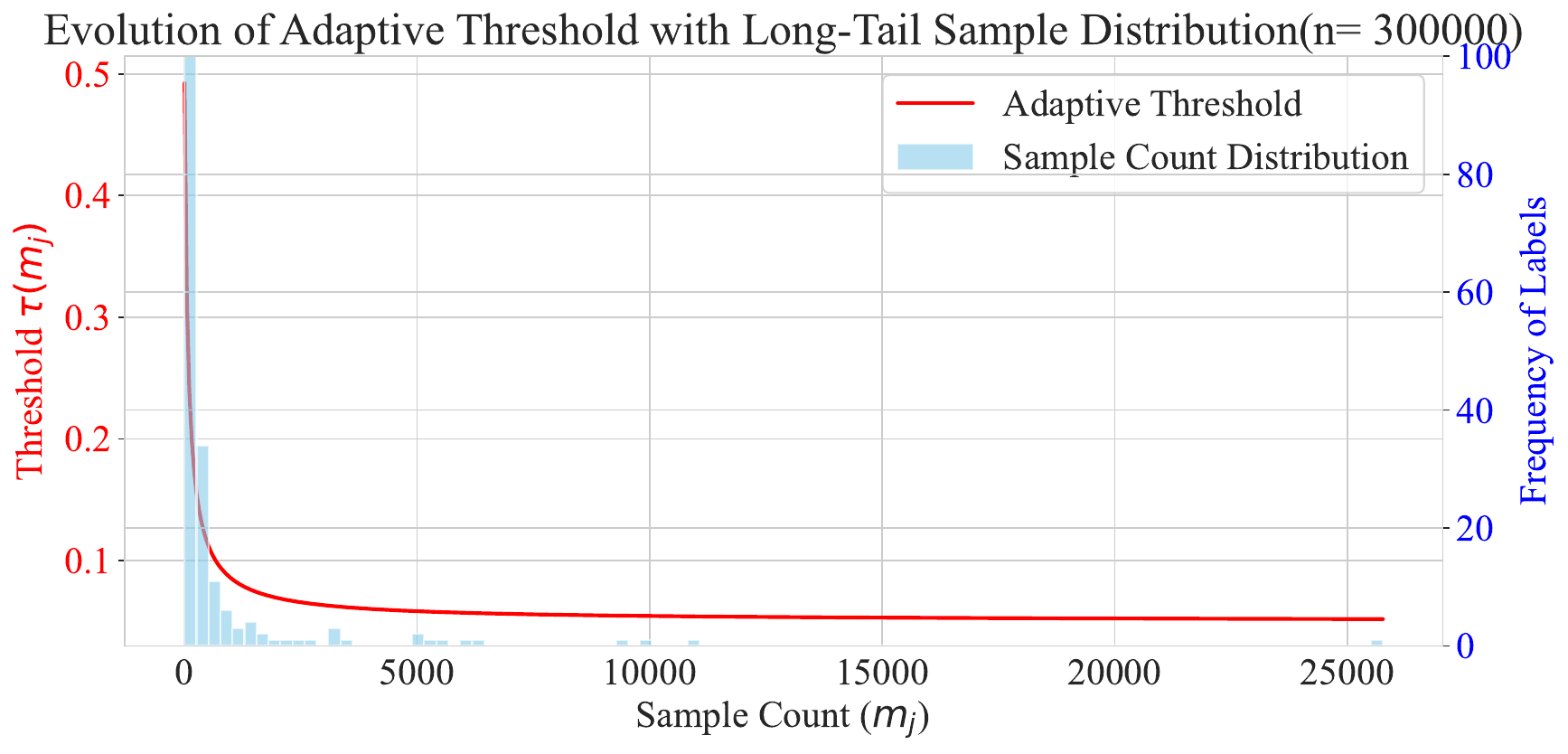}
    \caption{\textbf{Adaptive Thresholding Function \(\tau_j(m_j)\)} across varying support \(m_j\), illustrating the logistic decay from \(\tau_{\max}\) to \(\tau_{\min}\).}
    \label{fig:adaptive_threshold}
\end{figure}
\noindent To resolve this, we introduce an \emph{adaptive thresholding} strategy (Fig.~\ref{fig:adaptive_threshold}) that tailors the edge inclusion criterion to the statistical power available for each label. We define a label-specific threshold \(\tau_j,\) as a logistic decay function of its sample support \(m_j\):

\begin{equation}
\tau_j(m_j)= (\tau_{\max} - \tau_{\min}) \cdot \frac{1}{1 + e^{k(\log m_j - \log m_0)}} + \tau_{\min}
\label{eq:adaptive_threshold}
\end{equation}
This function smoothly interpolates between a user-defined maximum threshold, \(\tau_{max}\) (prioritizing precision for the tail), and a minimum, \(\tau_{min}\) (prioritizing recall for the head). The function's behavior is calibrated by the data distribution. Such that decay midpoint \(m_0\) is set to the median of all label supports, providing a robust anchor point against skew.
The decay rate \(k\) is made inversely proportional to the log-inter-quartile range of supports such that:
\begin{equation}\label{eq:k_log_inter_quartile_range}
k= \frac{2 \log 3}{\log q_{75} - \log q_{25}}
\end{equation}
Consequently, for rare labels with small \(m_j\), the high variance of the frequency estimate necessitates a high threshold \(\tau_j(m_j)\) that acts as a strong regularizer. For common labels with large \(m_j\), the LLN guarantees the convergence of \(\hat{\pi}_{i,j} \) to the true edge probability, justifying a lower threshold to capture a more complete causal structure.

\section{Empirical Evaluation}

The empirical evaluation of CARGO follows the same experimental setting (baselines and dataset configuration) as introduced in Chapter~\ref{c7:multi_label_one_shot_causal_discovery}. The aggregation phase is summarized in Algorithm~\ref{alg:cargo} (Appendix~\ref{sec:phase2implementation}). This task is challenging, as it involves handling cumulative noise across sequences from the imperfect CI-tests, managing severe label imbalance, and ensuring that aggregated structures remain both interpretable and statistically valid.

\subsection{Benchmark Against Multi-Label Causal Discovery Methods}

\begin{table}[!b]
\centering
\begin{adjustbox}{width=1.25\textwidth, center} %
\begin{tabular}{lcccc}
\hline
\textbf{Algorithm} & \textbf{Precision}↑ & \textbf{Recall}↑ & \textbf{F1}↑ & \textbf{Running Time (min)}↓ \\ \hline
IAMB & - & - & - & \(>4320\) \\
CMB & - & - & - & \(>4320\) \\
MB-by-MB & - & - & - & \(>4320\) \\
PCDbyPCD & - & - & - & \(>4320\) \\
MI-MCF & - & - & - & \(>4320\) \\
OSCAR (sample-level) & \(55.3 \pm 1.4\) & \(31.3 \pm 0.8\) & \(40.0 \pm 1.0\) & \(\boldsymbol{11.7}\) \\
CARGO (population-level) & \(\mathbf{ 60.6 \pm 1.5 }\) & \(\mathbf{ 45.8\pm 1.7}\) & \(\mathbf{ 45.8\pm 1.2}\) & \(11.8\) \\
\hline
\end{tabular}
\end{adjustbox}
\caption{Comparisons of \(\mathcal{MB}\) retrieval with \(m=50,000\) samples averaged over \(6-\)folds with \(|\mathcal{Y}| = 474, |\mathcal{X}|=29,100\) nodes. Averaging is 'weighted'. The symbol ’-’ indicates that the algorithm did not output the Markov Boundaries within 3 days. Metrics are given in \(\%\). OSCAR reports the sample-level performance (Phase 1).}
\label{tab:performance_comparison_plus_agg}
\end{table}

\paragraph{Results}
Table~\ref{tab:performance_comparison_plus_agg} shows the performance of the multi-label causal discovery algorithms, the sample-level performance of OSCAR and the global Markov boundary retrieval of CARGO. Since the evaluation setup is the same, all baseline LSL algorithms failed to compute complete Markov Boundaries within a three-day timeout, as demonstrated in Chapter~\ref{c7:multi_label_one_shot_causal_discovery}.
We observe that CARGO still achieves a significant performance improvement, even though the imperfect conditional independence tests introduce cumulative errors. The classification task is also considerably more challenging: in the global setting, the model aims to recover the underlying global causal rules rather than the more straightforward local rules identifiable within a single sequence. It surpasses OSCAR by a clear margin across all metrics, e.g., the Precision is (\(60.6 \pm 1.5\) vs. \(55.26 \pm 1.42\)) for the sample-level performance, confirming the benefit of tailored graph aggregation for large-scale causal discovery.
\begin{figure}[!b]
    \centering
    \begin{adjustbox}{width=1.3\textwidth, center} %
     \includegraphics[width=1\linewidth]{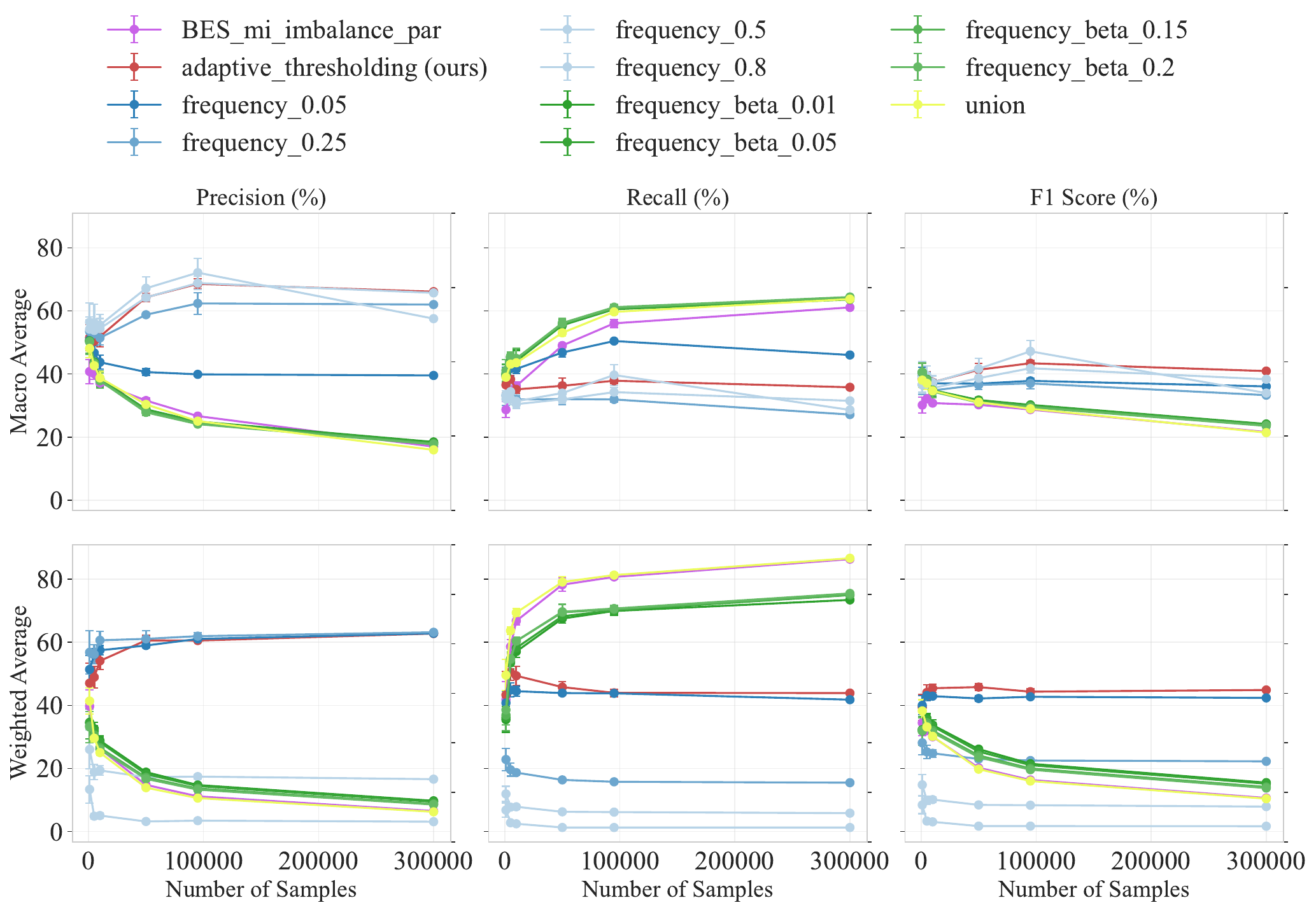}
     \end{adjustbox}
     \caption{\textbf{Criterion Ablation for Structural Fusion}. Comparison of different criteria for the structural fusion (Phase 2) as a function of the number of samples \(m\). With \(|\mathcal{Y}| = 474, |\mathcal{X}|=29,100\) nodes.}\label{fig:ablation_comparaison_criterions}
\end{figure}
\subsection{Ablation on Aggregation Criteria for Phase 2}
\paragraph{Criteria}
We provide an Ablation of the different criteria used in the structural fusion of Markov Boundaries.
\emph{Union} stands for a simple union over all edges without any removal.
\emph{Frequency} or edge voting, counts how often is \(X_i \in \mathcal{MB}(Y_j) \). Then apply a static frequency threshold \(\tau = [0.05, 0.25, 0.5, 0.8]\). \emph{BES mi} uses the mutual information between events and labels in the CAIG score optimized by a Backward Equivalence Search \cite{ges} (Appendix~\ref{appendix:criterions}). \emph{Expected FPR} (false positive ratio) \cite{consensus_bn_freq_cutoff_no} describes two beta distributions which are fitted using the distribution of the mutual information \(I(X_i, Y_j)\) extracted from Phase 1. The lower tail is used for outlier detection, with target false-positive rates of \(0.01, 0.05, 0.15\), and \(0.2\) (\emph{frequency beta} in Fig.~\ref{fig:ablation_comparaison_criterions}).
Detailed definitions can be found in Appendix \ref{appendix:criterions}.

\paragraph{Results}
Figure~\ref{fig:ablation_comparaison_criterions} illustrates the impact of aggregation choices during Phase 2. A naive \emph{Union} maximizes recall (\(84 \%  \; \text{for weighted}\)) but suffers from poor precision. Optimizing a local scoring criterion (\emph{BES mi}) did not enhance the performance over a basic \emph{Union}.
In contrast, fitting Beta distributions to detect outliers using their mutual information appears to perform better; hence, \emph{frequency beta} outperforms alternatives, particularly in terms of lower FPR. Experimental results for frequency approaches with static thresholds corroborate the analysis in Section \ref{sec:agg_imperfect_ci_tests}. Specifically, we observe a trade-off dictated by class support: strict cut-offs ($\tau \in [0.5, 0.8]$) yield the lowest weighted F1-scores, as they penalize classes with large support ($m_j$). Conversely, lowering the threshold ($\tau \in [0.05, 0.25]$) leads to a substantial gain in weighted precision ($+40\%$), but at the expense of macro-averaged metrics ($-20\%$ precision), as rare classes become overwhelmed by noise.

Our adaptive thresholding criterion outperformed all baselines by applying conservative thresholds to small supports and lenient thresholds to large supports. Consequently, it achieves state-of-the-art performance across all metrics, ranking first with F1-scores of $44.88\%$ (weighted) and $40.9\%$ (macro), and precision scores of $62.8\%$ and $66.1\%$, respectively.

\section{Summary}
CARGO completes the population-level events-to-outcome causal discovery cell (Table~\ref{tab:causal_framework}).
Moving from sample-level to population-level discovery significantly increases the task's complexity, as it requires the model to recover generalizable causal mechanisms rather than sequence-specific associations. Yet CARGO achieves strong performance (\(60.6\%\) precision), especially across a remarkable order of magnitude of \(10^4\) event types and \(10^2\) labels, demonstrating both robustness and scalability in real-world automotive diagnostics. It demonstrates that complex causal dependencies can be efficiently inferred from noisy and imbalanced data.
Under the temporal assumptions, large samples, faithfulness, and a perfect CI-test, CARGO recovers the set of Markov Boundaries. However, we highlighted the practical limitations, in particular long-tail distributions, which will most likely introduce spurious edges due to the imperfect CI tests.

\section{Outlook}
OSCAR and CARGO together form a complete pipeline for multi-label causal discovery, identifying, for each outcome label, the set of events that causally drive it. Nevertheless, both methods share an important scope restriction: they model events-to-outcome dependencies, treating the causal structure between events themselves as opaque. In many diagnostic scenarios, however, understanding how one DTC causally triggers another is equally essential for root-cause analysis and cascading failure diagnosis, e.g., why does DTC A tend to precede DTC B? This limitation motivates a change in problem formulation: rather than asking which events cause a given label, we now ask how events cause each other. Chapter~\ref{c7:event_to_event} addresses this more general challenge with TRACE, a framework for scalable sample-level event-to-event causal discovery in discrete sequences.

\chapter{Sample-Level Event-to-Event Causal Discovery}\label{c7:event_to_event}
\glsresetall
A complete diagnostic picture requires understanding how events causally influence each other, the event-to-event setting tackled in this chapter. In this setting, it becomes possible to identify root-causes, mediating effects and recover the global causal graphs. We therefore study causal discovery from a single observed sequence of discrete events generated by
a stochastic process, as encountered in vehicle
logs, manufacturing systems, or patient trajectories. This regime is particularly challenging due to the absence of repeated samples, the high cardinality of the event vocabulary, and long-range temporal dependencies, all encountered within a single sequence at inference time. This chapter is based on the following contribution:

\begin{description}[style=nextline,leftmargin=0cm,labelsep=0em]
\item[\textbf{Your Autoregressive Model Already Reveals the Causal Graph.}]~\cite{math2026tracescalableamortizedcausal} Hugo Math, Rainer Lienhart, Workshop on \emph{Structured Probabilistic Inference \& Generative Modeling} at ICML 2026, Seoul, South Korea, July 2026.
\end{description}

\section{Introduction}
Suppose we observe a single realization of a discrete stochastic process, for instance, the symptoms, tests, and disease evolution of a patient \cite{Li2021CausalHM, bihealth}, a manufacturing line's tests, or some diagnostic codes generated by a vehicle.
Which of these discrete events influences the occurrence of the others?%

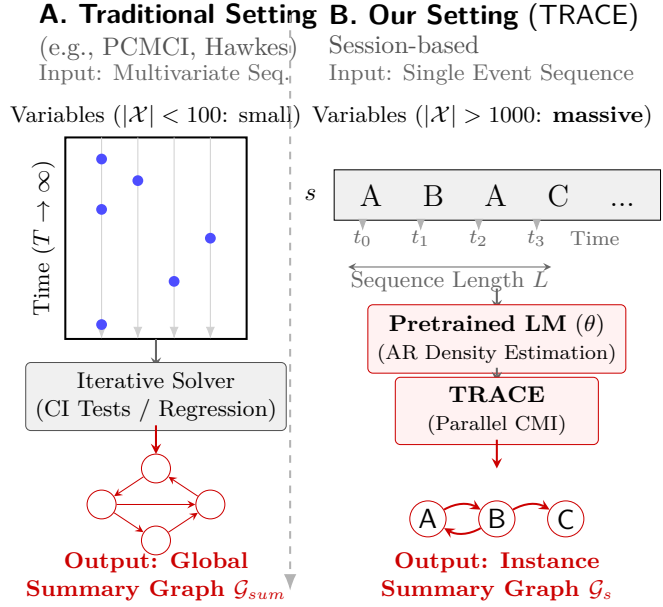
\begin{figure}[!t]
    \centering
    \resizebox{0.6\columnwidth}{!}{%
    \begin{tikzpicture}[font=\sffamily]

    \node[anchor=north west] at (0,0.8) {\textbf{A. Traditional Setting}};
    \node[anchor=north west, text=gray!80!black, font=\small] at (0,0.35) {(e.g., PCMCI, Hawkes)};
    \node[anchor=north west, text=gray, font=\footnotesize] at (0,-0.05) {Input: Multivariate Seq.};

    \node[anchor=south, font=\footnotesize] at (1.75,-1.2) {Variables ($|\mathcal{X}| < 100$: small)};

    \draw[thick] (0.5,-1.2) rectangle (3.0,-4);
    \foreach \x in {1, 1.5, 2, 2.5} { \draw[gray!30] (\x,-1.2) -- (\x,-4); }
    \node[rotate=90, font=\footnotesize] at (0.2,-2.6) {Time ($T \to \infty$)};
    \foreach \x/\y in {1/-1.5, 1.5/-1.8, 1/-2.2, 2.5/-2.6, 2/-3.2, 1/-3.8} {
        \node[circle, fill=blue!70, inner sep=1.5pt] at (\x,\y) {};
    }

    \draw[->, thick, gray] (1.75,-4.0) -- (1.75,-4.4);

    \node[draw=gray!80!black, fill=gray!10, rounded corners=2pt, minimum width=2.5cm, minimum height=0.8cm, align=center, font=\footnotesize] (solver) at (1.75, -4.8) {Iterative Solver\\(CI Tests / Regression)};

    \draw[->, thick, red!80!black] (1.75,-5.2) -- (1.75,-5.6);

    \node[circle, draw=red!80!black, inner sep=1pt, minimum size=0.4cm] (n1) at (1.0, -6.3) {};
    \node[circle, draw=red!80!black, inner sep=1pt, minimum size=0.4cm] (n2) at (2.5, -6.3) {};
    \node[circle, draw=red!80!black, inner sep=1pt, minimum size=0.4cm] (n3) at (1.75, -5.8) {};
    \node[circle, draw=red!80!black, inner sep=1pt, minimum size=0.4cm] (n4) at (1.75, -6.8) {};

    \draw[->, red!80!black] (n1) -- (n2); \draw[->, red!80!black] (n3) -- (n1);
    \draw[->, red!80!black] (n2) -- (n3); \draw[->, red!80!black] (n1) -- (n4);
    \draw[->, red!80!black] (n4) -- (n2);

    \node[text=red!80!black, font=\bfseries\footnotesize, align=center] at (1.75,-7.3) {Output: Global\\Summary Graph $\mathcal{G}_{sum}$};

    \draw[dashed, thick, gray!50] (3.6,0.5) -- (3.6,-7.5);

    \node[anchor=north west] at (4.0,0.8) {\textbf{B. Our Setting} (TRACE)};
    \node[anchor=north west, text=gray!80!black, font=\small] at (4.0,0.35) {Session-based};
    \node[anchor=north west, text=gray, font=\footnotesize] at (4.0,-0.05) {Input: Single Event Sequence};

    \node[anchor=south, font=\footnotesize] at (6.25,-1.2) {Variables ($|\mathcal{X}| > 1000$: \textbf{massive})};

    \node[draw, fill=gray!10, minimum width=4.5cm, minimum height=0.7cm, anchor=west, font=\large] (seq) at (4.2,-2.0) {A \quad B \quad A \quad C \quad ...};
    \node[anchor=east, font=\bfseries\small] at (4.1,-2.0) {$s$};

    \foreach \x/\t in {4.6/0, 5.4/1, 6.2/2, 7.0/3} {
         \node[font=\scriptsize, text=gray] at (\x, -2.6) {$t_{\t}$};
         \draw[gray!50] (\x, -2.35) -- (\x, -2.5);
    }
    \node[font=\scriptsize, text=gray] at (7.8, -2.6) {Time};
    \draw[<->, gray] (4.4,-3.0) -- (7.2,-3.0);
    \node[font=\footnotesize, text=gray] at (5.8,-3.2) {Sequence Length $L$};

    \draw[->, thick, gray] (6.45,-3.3) -- (6.45,-3.6);

    \node[draw=red!80!black, fill=red!5, rounded corners=2pt, minimum width=2.8cm, minimum height=0.6cm, font=\footnotesize, align=center] (lm) at (6.45, -4.0) {\textbf{Pretrained LM} ($\theta$)\\ \scriptsize (Autoregressive Density Estimation)};

    \draw[->, thick, gray] (6.45,-4.4) -- (6.45,-4.6);

    \node[draw=red!80!black, fill=red!5, rounded corners=2pt, minimum width=2.8cm, minimum height=0.6cm, font=\footnotesize, align=center] (vec) at (6.45, -5.0) {\textbf{TRACE}\\ \scriptsize (Parallel CMI)};

    \draw[->, thick, red!80!black] (6.45,-5.4) -- (6.45,-5.8);

    \node[circle, draw=red!80!black, inner sep=1pt, minimum size=0.5cm] (nA) at (5.5, -6.5) {A};
    \node[circle, draw=red!80!black, inner sep=1pt, minimum size=0.5cm] (nB) at (6.45, -6.5) {B};
    \node[circle, draw=red!80!black, inner sep=1pt, minimum size=0.5cm] (nC) at (7.4, -6.5) {C};

    \draw[->, thick, red!80!black] (nA) to[bend left] (nB);
    \draw[->, thick, red!80!black] (nB) to[bend left] (nC);
    \draw[->, thick, red!80!black] (nB) to[bend left] (nA);

    \node[text=red!80!black, font=\bfseries\footnotesize, align=center] at (6.45,-7.3) {Output: Sample\\Summary Graph $\mathcal{G}_{s}$};

    \end{tikzpicture}
    }
    \caption{\textbf{Methodological Shift.} \textbf{(A) Traditional Causal Discovery in Sequences} (e.g., PCMCI, Hawkes, Granger) relies on iterative solvers (CI-tests) over long multivariate time series ($T \to \infty$). \textbf{(B) Our TRACE Approach} processes a single sequence (e.g., event logs, user interactions, patient trajectories) through a pretrained autoregressive (AR) model as density estimator to compute the Conditional Mutual Information (CMI) in parallel, enabling scalable causal discovery over massive vocabularies ($|\mathcal{X}| > 1000$).}
    \label{fig:data_paradigm}
\end{figure}
\noindent
Contemporary causal discovery algorithms for sequences infer relations across multiple parallel streams (e.g., distinct users or sensors). These methods are ill-suited for recovering causal structure within a single event stream, nor do they scale to the long, noisy, and heterogeneous sequences encountered in modern industrial systems. Moreover, sample-level causal discovery (see Chapter~\ref{c7:multi_label_one_shot_causal_discovery}) poses a challenging problem, as it requires identifying conditional independencies among event types from a single realization of the process.

\noindent In the era of large-scale pretraining, Autoregressive Language Models (AR LMs) \cite{gpt, touvron2023llamaopenefficientfoundation} have emerged as powerful density estimators, encoding rich conditional distributions over complex contexts to predict the next token \cite{draxler2025transformers}. Consequently, amortized causal discovery gained a surge of interest in the recent literature~\cite{amortized_cd_time, balazadeh2025causalpfn}. In this setting, the heavy computational cost is shifted to pretraining a single model, so that inferring the causal structure of a new dataset only requires a cheap forward pass. We propose to take this a step further by repurposing existing predictive priors rather than training specialized models. This effectively transforms a production forecaster into a causal discovery engine without the need for task-specific retraining.

\section{Related Work}
\subsection{Causal Discovery in Event Sequences}

Distinguishing between the multi-stream and single-stream paradigm is key, as they fundamentally address different causal problems (Fig.~\ref{fig:data_paradigm}). We point out the current methods' limitations in Table~\ref{tab:related_work}.

\begin{table*}[!t]
\centering
\small
\setlength{\tabcolsep}{4pt}
\begin{adjustbox}{width=1.25\textwidth, center} %
\begin{tabular}{lcccccc}
\toprule
\textbf{Method Class}
& \parbox{2.1cm}{\centering \textbf{Discrete} \textbf{Events}}
& \parbox{1.8cm}{\centering \textbf{High} \textbf{Card.}}
& \parbox{1.8cm}{\centering \textbf{Non-}\textbf{Param.}}
& \parbox{0.7cm}{\centering \textbf{Lags}}
& \parbox{2.1cm}{\centering \textbf{Sample-} \textbf{Level}}
& \parbox{2.5cm}{\centering \textbf{Linear} \textbf{Complexity}} \\
\midrule
Constraint-based (PCMCI, FCI)
& $\times$ & $\times$ & \cmark & \cmark & $\times$ & $\times$ \\

Score-based (DYNOTEARS)
& $\times$ & $\times$ &  $\times$ & \cmark & $\times$ & $\times$ \\

Granger (TCDF, CAUSE)
& $\times$ & $\times$ & $\times$ & \cmark & $\times$ & $\times$ \\

Noise-based (VarLiNGAM)
& $\times$ & $\times$ & $\times$ & \cmark & $\times$ & $\times$ \\

Hawkes / TPP Models (THP, SHP)
& \cmark & $\times$ & $\times$ & \cmark & $\times$ & $\times$ \\

Info-Theoretic (NPHC, CASCADE)
& \cmark & $\times$ & \cmark & \cmark & $\times$ & $\times$ \\

\textbf{TRACE (Ours)}
& \cmark & \cmark & \cmark & \cmark & \cmark & \cmark \\
\bottomrule
\end{tabular}
\end{adjustbox}
\caption{Comparison of causal discovery methods. \textbf{Discrete Events}: Operates on discrete event sequences (e.g., text, logs) rather than multivariate time series. \textbf{High Card.}: Computationally tractable for large vocabularies ($|\mathcal{X}| > 10^3$). \textbf{Non-Param.}: Agnostic to the functional form (e.g., linearity) of causal relationships. \textbf{Lags}: Models delayed causal effects. \textbf{Sample-Level}: Infers a local causal graph specific to a single sequence, rather than a global graph. \textbf{Linear Complexity}: Complexity scales linearly with the vocabulary size \(|\mathcal{X}|\).}
\label{tab:related_work}
\end{table*}

\paragraph{Multiple Streams (Standard).}

The most common paradigm considers a long multivariate time series, each corresponding to an individual entity (e.g., sensor, user, or machine). The goal is to uncover how the occurrence of events in one sequence
influences the occurrences in others. To recall the main methods, it is traditionally done via Granger-based \cite{granger_causality, Shojaie2010DiscoveringGG, tcdf, cause}~(TCDF, CAUSE), constraint-based~\cite{pcmci}~(PCMCI), functional~\cite{varlingam}~(VARLiNGAM) or optimization-based methods ~\cite{dynotears}~(DYNOTEARS). Modern information-theoretic variants (NPHC \cite{nphc}, CASCADE~\cite{cueppers2024causal}) and neural point processes (THP~\cite{transformerhawkeprocess}, SHTP~\cite{shtp}). As established in Chapters~\ref{c3:foundation_causal_discovery} and ~\ref{c7:multi_label_one_shot_causal_discovery}, these methods are limited to small number of nodes in the resulting graph.%

\paragraph{Single Stream (The Sample-Level Regime).}
Our setting differs fundamentally (Fig.~\ref{fig:data_paradigm}) as we operate in the 'session-based' regime common to NLP and system logs: we observe many short, independent sequences over a massive vocabulary (often with \(|\mathcal{X}| > 1000\)).
During inference, only one realization of the process is often available. Here, the goal is to understand if event type $A$ causes type $B$. This regime is significantly harder due to the sparsity of specific event pairs in high-cardinality vocabularies and the lack of independent trials during inference.
While recent works attempt to interpret attention weights in Transformers as causal graphs~\cite{tf_causalinterpretation_neurips_2023}, this is heavily criticized for being a poor proxy of causality~\cite{filippova-2020-elephant}.

\section{Methodology}
\begin{figure*}[!t]
    \centering
    \resizebox{\textwidth}{!}{%
    \begin{tikzpicture}[font=\sffamily, >=stealth]

    \node[anchor=north west] at (-0.5, 2.7) {\textbf{Phase 1: Self-Supervised Training}};
    \draw[dashed, gray!40, rounded corners] (-0.5, 2.2) rectangle (3.8, -1.0);

    \foreach \x/\y/\n in {0/0/s_N, 0.1/0.1/\dots, 0.2/0.2/s_2, 0.3/0.3/s_1} {
        \draw[fill=white, draw=gray] (\x, \y) rectangle (\x+1.0, \y+1.2);
        \node[font=\tiny] at (\x+0.5, \y+0.6) {$\n$};
    }
    \node[align=center, font=\scriptsize] at (0.65, -0.5) {Training\\Sequences};

    \draw[->, thick, gray] (1.3, 0.8) -- (2.0, 0.8) node[midway, above, font=\tiny] {Train};

    \node[draw=red!80!black, fill=red!5, rounded corners, minimum width=1.4cm, minimum height=1cm, align=center, font=\small] (lm_train) at (2.8, 0.8) {\textbf{AR Model}\\($\theta$)};
    \node[align=center, font=\scriptsize, text=gray] at (2.6, 0) {(language model, rnn)};

    \node[anchor=north west] at (6.5, 2.7) {\textbf{Phase 2: Sample-Level Causal Discovery (Inference)}};
    \draw[dashed, gray!40, rounded corners] (4.2, 2.2) rectangle (16.5, -1.0);

    \node[anchor=west] at (4.7, 1.4) {\footnotesize Input Sequence $s$};
    \node[draw, fill=gray!10, rounded corners=2pt, minimum width=2.5cm] (seq) at (5.6, 0.8) { A \quad B \quad A \quad C};

    \node[draw=red!80!black, fill=red!5, rounded corners, minimum width=1.4cm, minimum height=1cm, align=center, font=\small] (lm_train) at (2.8, 0.8) {\textbf{AR Model}\\($\theta$)};
    \node[align=center, font=\scriptsize, text=gray] at (6.1, 0) {A single sequence is observed};

    \node[draw=red!80!black, fill=red!5, rounded corners, minimum width=1.4cm, minimum height=1cm, align=center, font=\small] (lm_train) at (2.8, 0.8) {\textbf{AR Model}\\($\theta$)};
    \node[align=center, font=\scriptsize, text=gray] at (10.5, 0) {Use the learned probabilistic dynamics};

    \draw[->, thick, gray] (seq) -- (7.8, 0.8);

    \node[draw=red!80!black, fill=red!5, rounded corners, minimum width=1.4cm, minimum height=1cm, align=center, font=\small] (lm_inf) at (8.5, 0.8) {\textbf{AR Model}\\($\theta$)};

    \draw[->, thick, gray] (lm_inf) -- (9.8, 0.8) node[midway, above, font=\tiny, align=center] {$P_\theta$};

    \node[draw=blue!80!black, fill=blue!5, rounded corners, minimum width=2.4cm, minimum height=0.8cm, align=center, font=\footnotesize] (cmi) at (11.3, 0.8) {\textbf{Parallelized CMI}\\(CI-Tests)};

    \draw[->, thick, gray] (cmi) -- (13.2, 0.8);

    \node[circle, draw, inner sep=1pt, minimum size=0.5cm] (t0) at (13.5, 1.9) {A}; \node[right=1pt, font=\tiny, gray] at (t0.east) {$t_0$};
    \node[circle, draw, inner sep=1pt, minimum size=0.5cm] (t1) at (13.5, 1.1) {B}; \node[right=1pt, font=\tiny, gray] at (t1.east) {$t_1$};
    \node[circle, draw, inner sep=1pt, minimum size=0.5cm] (t2) at (13.5, 0.3) {A}; \node[right=1pt, font=\tiny, gray] at (t2.east) {$t_2$};
    \node[circle, draw, inner sep=1pt, minimum size=0.5cm] (t3) at (13.5, -0.5) {C}; \node[right=1pt, font=\tiny, gray] at (t3.east) {$t_3$};

    \draw[->, gray!20] (t0) to[bend right=20] (t2); %
    \draw[->, thick, blue] (t0) -- (t1); %
    \draw[->, thick, blue] (t1) -- (t2); %
    \draw[->, thick, blue] (t2) -- (t3); %
    \draw[->, thick, blue] (t1) to[bend right=45] (t3); %

    \node[align=center, font=\small, text=blue!80!black] at (11.4, -0.5) {Sample Time Graph\\$\mathcal{G}_{t,s}$};

    \draw[->, ultra thick, blue!80!black] (14, 0.3) -- (14.8, 0.3) node[midway, above, font=\scriptsize] {Projection};

    \node[circle, draw, minimum size=0.7cm] (nA) at (15.2, 1.4) {A};
    \node[circle, draw, minimum size=0.7cm] (nB) at (16.2, 1.4) {B};
    \node[circle, draw, minimum size=0.7cm] (nC) at (15.7, 0.2) {C};

    \draw[->, thick, blue] (nA) to[bend left=15] (nB);
    \draw[->, thick, blue] (nB) to[bend left=15] (nA); %
    \draw[->, thick, blue] (nB) -- (nC);
    \draw[->, thick, blue] (nA) -- (nC);

    \node[align=center, font=\small, text=blue!80!black] at (15.7, -0.5) {Summary Graph\\$\mathcal{G}_{s}$};

    \end{tikzpicture}}
    \caption{\textbf{TRACE Methodology.} \textbf{Phase 1 (Training):} An autoregressive (AR) model (e.g., LM, RNN) is pretrained on a corpus of event sequences via next-token prediction to learn the process dynamics (\(P_\theta\)). \textbf{Phase 2 (Inference):} A single sequence $s$ is passed through the frozen model. We then estimate conditional mutual information (\textbf{Parallelized CMI} module) to prune non-causal edges and form the \textit{Sample Time Causal Graph} $\mathcal{G}_{t,s}$. Finally, this graph is projected onto the event types to recover the \textit{Summary Causal Graph} $\mathcal{G}_s$.}
    \label{fig:methodology_pipeline}
\end{figure*}
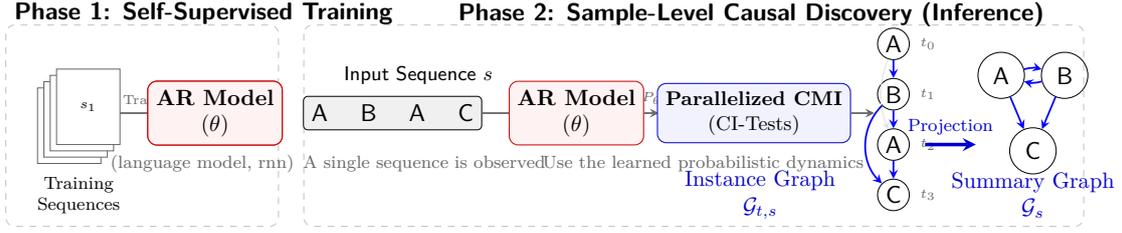

An overview of TRACE can be found in Fig.~\ref{fig:methodology_pipeline}. Concretely, the pipeline consists of the same two phases as OSCAR (Chapter~\ref{c7:multi_label_one_shot_causal_discovery}), reused here at the event level instead of the label level. In Phase 1 (left), an autoregressive model \(\text{AR}_\theta\) is pretrained once, offline, via next-token prediction on a corpus of event sequences, exactly as in Section~\ref{c2:ep_prediction_based_on_live_data}. In Phase 2 (right), at inference time, a single previously unseen sequence \(s\) is passed through the frozen \(\text{AR}_\theta\) to obtain its learned transition probabilities \(P_\theta\). These are then fed to the \textbf{Parallelized CMI} module, which evaluates the conditional mutual information estimator \(\hat{I}_N\) for every candidate event pair simultaneously (detailed in Fig.~\ref{fig:trace_diagram} below) and prunes non-causal edges to obtain the \textit{Sample Time Causal Graph} \(\mathcal{G}_{t,s}\). Finally, this per-sequence graph is projected onto the underlying event types to recover the aggregated, human-readable \textit{Summary Causal Graph} \(\mathcal{G}_s\) (Def.~\ref{def:iscg}). A key advantage of TRACE is its architectural agnosticism. The framework decouples the density estimation (Phase 1) from causal discovery (Phase 2). Consequently, TRACE can leverage any state-of-the-art autoregressive backbone (e.g., Transformers, Mamba, RNNs).

\subsection{Data-Generating Process}
We model the \textit{data-generating process} (DGP) as a non-stationary stochastic process\footnote{As fixed in Chapter~\ref{c3:foundation_causal_discovery} and already used for OSCAR in Chapter~\ref{c7:multi_label_one_shot_causal_discovery}, \(X_t\) denotes the event at position (time step) \(t\) of a sequence, in line with the dynamical systems and stochastic processes literature~\cite{koller2009probabilistic}.} \(\{X_t, t \in \mathbb{N}\}\) taking values in a finite, discrete alphabet $\mathcal{X}$. We assume that the process forms a high-order Markov chain~\cite{koller2009probabilistic} with transition distribution \(P(X_{t}|X_{<t})\), where \(X_{<t} = X_{0:t-1} \triangleq \{X_0, \cdots, X_{t-1}\}\). This implies that the past \(X_{0:t-1}\) is not independent of the future \(X_{t+1}\) conditioned on the present \(X_{t}\).

\subsection{Sample-Level Causality}
Our analysis begins at the level of the specific observed sequence. To bridge the gap between the multinomial realization \(s = (x_0, \dots, x_L)\) and causal structure, we define the Binary Event Process \(\{E_t\}_{t=0}^L\) as
\[E_t \triangleq \mathds{1}_{X_t = x_t}\]

\noindent Here, \(E_t\) is a binary random variable representing the realization of the specific event type observed at time \(t\). This transformation allows us to represent the causal dependencies specific to this sequence as a DAG, which we term the \emph{Sample Time Causal Graph}.

\begin{definition}[Sample Time Causal Graph]\label{def:itcg}
    Let \(s\) be a realization of the process \(\{X_t\}\). The Sample Time Causal Graph \(\mathcal{G}_{t,s} = (\mathbf{E}_s, \mathcal{E}^{\text{time}}_{s})\) is a DAG whose nodes \(\mathbf{E}_s = \{E_0, \dots, E_L\}\) are the binary event variables of the sequence, one per time step. A directed edge \(E_t \to E_{t'}\), with \(t < t'\) and time lag \(k = t' - t \geq 1\) between cause and effect, exists in \(\mathcal{E}^{\text{time}}_{s}\) if and only if the realization of the event at \(t\) is a cause of the event at \(t'\).
\end{definition}
\noindent This graph \(\mathcal{G}_{t,s}\) (illustrated in Fig.~\ref{fig:methodology_pipeline}, right) represents the causal structure of the sequence \(s\), unrolled in time.

\subsection{Summary-Level Causality}
Operators are often confronted with observing a single sequence during inference to understand generalizable rules (e.g., "Fire causes Smoke") rather than specific timestamps \((\mathcal{G}_{t,s})\). We adapt the terminology from~\cite{assaad_survey_ijcai_cd_time_series} regarding summary causal graph (SCG), which treats the unique event types in \(\mathcal{X}\) as the variables of interest. Therefore, for a single sequence \(s\), we aim to recover its \emph{Sample Summary Causal Graph}, ``sample'' referring, as throughout this thesis, to one observed sequence.
\begin{definition}[Sample Summary Causal Graph]\label{def:iscg}
    Let \(\mathcal{X}_s \subseteq \mathcal{X}\) be the set of unique event types in \(s\). The Sample SCG \(\mathcal{G}_{s} = (\mathcal{X}_s, \mathcal{E}_s)\) is the surjective projection of the Sample Time Causal Graph \(\mathcal{G}_{t,s}\) onto \(\mathcal{X}_s\). Specifically, a type-level edge \(u \to v\) exists in \(\mathcal{G}_s\) if and only if it appears at least once in the sample time graph:
    \begin{align}
        u \to v \in \mathcal{E}_s \iff \exists\, t < t' \text{ s.t. } (E_t \to E_{t'}) \in \notag\\
        \mathcal{E}^{\text{time}}_{s} \land x_{t} = u \land x_{t'} = v \notag
    \end{align}
\end{definition}
\noindent Hence, our causal discovery objective is to identify the parents \(\text{Pa}_{\mathcal{G}_{t,s}}(E_{t'})\) of each node \(E_{t'}\) in the sequence (e.g., detecting Fire@\(t_1\) \(\rightarrow\) Smoke@\(t_2\)) and aggregate them to reconstruct the sample summary causal graph \(\mathcal{G}_{s}\) (e.g., Fire \(\rightarrow\) Smoke), illustrated in Fig.~\ref{fig:methodology_pipeline} (right).
\begin{remark}[Cyclicity in Summary Graphs]\label{rem:cyclic_summary}
    Consistent with the standard literature~\cite{assaad_survey_ijcai_cd_time_series}, the summary graph \(\mathcal{G}_{s}\) is an abstraction of the time-unrolled graph and is therefore permitted to contain cycles.
\end{remark}

\subsection{Autoregressive Density Estimation}
TRACE leverages the same density estimation capability as OSCAR~(Chapter~\ref{c7:multi_label_one_shot_causal_discovery}): a pretrained AR model denoted \(\text{Tf}_\theta\) (CarFormer) estimates the next token probabilities \(P_\theta\). We map it to the binary event process as:
\begin{equation}\label{eq:softmax_to_tf}
    P_\theta(E_t = 1 \mid X_{<t} = x_{<t}) = [\text{Softmax}(\text{Tf}_\theta(x_{<t}))]_{x_t}
\end{equation}
The model is trained on a single event stream  rather than multi-labeled sequences (Chapter~\ref{c2:ep_prediction_based_on_live_data}). This formulation allows us to utilize the joint distributions learned by the AR model for the structure learning of \(\mathcal{G}_{t,s}\). Therefore, in contrast to recent works probing causal reasoning abilities of AR models (e.g., language models) via prompting \cite{long2022can, kiciman2024causal}, we focus on extracting causal structure from the learned probabilistic dynamics.
\subsection{Assumptions}
In particular, \(\mathcal{G}_{t,s}\) implies (1) the \emph{Markov assumption}~\cite{pearl_1998_bn}, such that a variable is conditionally independent of its non-descendants given its parents, and (2) \emph{consistency through time}~\cite{assaad_survey_ijcai_cd_time_series}. Following Chapter~\ref{c7:multi_label_one_shot_causal_discovery}, we assume temporal precedence~(A\ref{assumption:temporal_precedence}) and causal sufficiency~(A\ref{assumption:causal_sufficiency}).
Specifically to this chapter, we assume:

\begin{assumption}[\(\epsilon-\)Oracle Model] \label{assumption:oracle_trace}
We assume that the AR model \(\text{Tf}_\theta\), trained via maximum likelihood estimation on a dataset generated by the true distribution \(P\), has converged such that the Kullback-Leibler divergence is bounded by $\epsilon$:
\begin{equation}\label{eq:oracle_approx}
D_{KL}(P(X_t \mid X_{<t}) ~||~ P_\theta(X_t \mid X_{<t})) \leq \epsilon
\end{equation}
And that the total variation distance \(\delta(P, P_\theta) \leq \frac{1}{2}\). We thus define \(\text{Tf}_\theta\) as an \emph{$\epsilon$-Oracle} model.
\end{assumption}
By Pinsker's inequality~\cite{cover1999elements}, Eq.~\ref{eq:oracle_approx} implies the total variation distance $\delta(P,P_\theta)$ to be bounded by $\sqrt{\epsilon/2}$:
\begin{align}
& D_{KL}(P(X_t \mid X_{<t}) ~||~ P_\theta(X_t \mid X_{<t})) \leq \epsilon \notag \\
\implies & \sqrt{\frac{1}{2}D_{KL}(P(X_t \mid X_{<t}) ~||~ P_\theta(X_t \mid X_{<t}))} \leq \sqrt{\frac{\epsilon}{2}} \notag \\
\implies& \delta(P, P_\theta) \leq \sqrt{\frac{\epsilon}{2}} \; \text{By Pinsker's inequality}
\end{align}
Where \(\delta(P, P_\theta) = \sup_A |P(A) - P_\theta(A)|\), the supremum taken over all measurable events \(A\) in the sample space.

Essentially, we aim at being more flexible than Chapter~\ref{c7:multi_label_one_shot_causal_discovery} via a possible non-stationary stochastic process and an imperfect density estimator (A\ref{assumption:oracle_trace}).

\section{Single Stream Causal Discovery}
In this section, we describe how we can derive a CI-test to construct the Sample Time Causal Graph \(\mathcal{G}_{t,s}\) within single streams. %

\subsection{Conditional Mutual Information}
In a sequence \(s = (x_0, \cdots, x_L)\), we would like to assess how much additional information the realization \(x_t\) \((E_t = 1)\) provides about the next event occurrence \(X_{t+1} = x_{t+1}\) when we already know the past observation \(X_{<t}\). We essentially try to answer whether:
\begin{align*}
    P(E_{t+1}|E_t, X_{<t}) = P(E_{t+1}|X_{<t})
\end{align*}
As seen in Chapter~\ref{c3:foundation_causal_discovery}, Information Gain \(I_G\)~(Def.~\ref{def:information_gain}) measures how much the realization \(e_t\) changes the distribution of \(E_{t+1}\) given \(x_{<t}\):

\begin{align}
    I_G(E_{t+1}, e_t|x_{<t}) &= D_{KL}(P(E_{t+1}|e_t, x_{<t})\|P(E_{t+1}|x_{<t}))\label{eq:info_gain_trace}
\end{align}
\noindent whose expectation over \(e_t \sim P(E_t \mid x_{<t})\) equals \(H(E_{t+1}|x_{<t}) - H(E_{t+1}|E_t, x_{<t})\).
\noindent The conditional independence (Def.~\ref{def:conditional_independence}) between event \(E_t\) and event \(E_{t+1}\) is assessed with the CMI such as:
\begin{align}
    I(E_{t+1}, E_t|X_{<t}) &\triangleq H(E_{t+1}|X_{<t}) - H(E_{t+1}|E_t, X_{<t}) \notag \\
    &= \mathbb{E}_{x_{<t} \sim p(x_{<t})}\Big[\mathbb{E}_{e_t \sim P(E_t \mid x_{<t})}\big[I_{G}(E_{t+1}, e_t|x_{<t})\big]\Big]\label{eq:cmi_theorique_trace}
\end{align}
We can deduce the following CI-test:
\begin{equation}\label{eq:ci_test_adjacent}
     E_{t+1} \not\perp E_t | X_{<t} \Leftrightarrow I(E_{t+1}, E_t|X_{<t}) > 0
\end{equation}
\begin{remark}
   We condition on the full trajectory \(X_{<t}\) rather than the coarsened binary trajectory \(E_{<t}\) to test for all observed events in the trajectory, thereby blocking potential back-door paths (confounders) that would otherwise remain hidden in the binary projection.
\end{remark}

\subsection{Estimation and Approximation Error}
The previous Eq.~\ref{eq:cmi_theorique_trace} involves an expectation over the distribution of trajectories \(x_{<t}\). Since the true distribution $P$ is unknown, we utilize the \(\epsilon\)-Oracle model $\text{Tf}_\theta$ as a proxy to simulate $N$ i.i.d.\ trajectories (particles) $\{x^{(l)}_{<t}\}_{l=1}^N \sim P_\theta(X_{<t})$. We define the empirical estimator $\hat{I}_N$ of the CMI as the Monte Carlo estimation:

\begin{equation}\label{eq:cmi_estimation_naive_monte_carlo}\hat{I}_N(E_{t+1}, E_t \mid X_{<t})= \frac{1}{N} \sum_{l=1}^N \underbrace{\mathbb{E}_{e_t \sim P_\theta(E_t \mid x^{(l)}_{<t})} [I_G(E_{t+1}, e_t \mid x^{(l)}_{<t})]}_{f_\theta(x^{(l)}_{<t})}
\end{equation}
where \(P_\theta(E_t = 1 \mid x^{(l)}_{<t}) = P_\theta(X_t = x_t \mid x^{(l)}_{<t})\) is the probability that the model assigns to the observed event type at position \(t\) given the sampled trajectory (Eq.~\ref{eq:softmax_to_tf}), so that \(e_t \in \{0, 1\}\) follows a Bernoulli distribution with this parameter.

\begin{proposition}[Convergence to the $\epsilon$-Proxy]\label{prop:consistency_cmi_trace}
The estimator $\hat{I}_N$ is a consistent estimator of the \(\epsilon\)-Oracle induced CMI denoted as $I_\theta$. By the Strong Law of Large Numbers, as $N \to \infty$:
$$\hat{I}_N \xrightarrow[N \to +\infty]{\text{a.s.}} I_\theta(E_{t+1}, E_t \mid X_{<t})$$
\end{proposition}

\textit{(Proof Sketch)}
It follows directly from the Strong Law of Large Numbers (SLLN), as the samples are drawn i.i.d. from $P_\theta$ and the information gain term \(f_\theta(x^{(l)}_{<t})\) is bounded by $\log 2$, ensuring finite variance.

\begin{proof}\label{proof:consistency_estimator_cmi_trace}
The trajectories $x^{(l)}_{<t}$ are sampled directly from the model $\text{Tf}_\theta$.
Let $f_\theta(x^{(l)}_{<t})$ represents the estimation CMI for a fixed trajectory \(x_{<t}\) and \(I_\theta\) the CMI with the approximated distribution \(P_\theta\). Expressing this as a difference of conditional entropies:
\begin{equation}\label{eq:info_gain_bounded}
\begin{aligned}
0
&\leq \mathbb{E}_{e_t \sim P_\theta(E_t \mid x^{(l)}_{<t})} I_G\!\left(E_{t+1}, e_t| x^{(l)}_{<t}\right) \\
&= H_\theta\!\left(E_{t+1} | x^{(l)}_{<t}\right)
   - H_\theta\!\left(E_{t+1}| E_t, x^{(l)}_{<t}\right) \\
&\leq H_\theta\!\left(E_{t+1} | x^{(l)}_{<t}\right)
\leq \log{2}.
\end{aligned}
\end{equation}
Thus the posterior variance of \(f_\theta(x^{(l)}_{<t})\) satisfies \(\sigma^2_{f} \triangleq \mathbb{E}_{x_{<t}}[f_\theta^2(x_{<t})] - I_\theta^2(f_\theta) < +\infty\) \cite{Doucet2001} then the variance of \(\hat{I}_N(f)\) is equal to \(\textit{var}(\hat{I}_N(f)) = \frac{\sigma^2_{f}}{N}\) and from the strong law of large numbers:
\begin{align}
\hat{I}_N &\xrightarrow[N \to +\infty]{\text{a.s.}}
\mathbb{E}_{x_{<t} \sim P_\theta(X_{<t})}\!\left[ \mathbb{E}_{e_t \sim P_\theta(E_t \mid x_{<t})}\big[I_G(E_{t+1}, e_t \mid x_{<t})\big] \right]
\triangleq I_\theta(f_\theta)
\end{align}
\end{proof}

\noindent Since we know that the estimator \(\hat{I}_N\) converges to the approximated CMI as \(I_\theta\), we derive an approximation error bound to characterize the remaining noise induced by the imperfect \(\epsilon\)-Oracle model.

\begin{theorem}[Total Error Bound in the $\epsilon$-Regime]\label{thm:total_error_bound}
Let $\hat{I}_N$ be the Monte Carlo estimator of the approximated CMI \(I_\theta\). Assuming the AR model $P_\theta$ approximates the true DGP $P$ as an \(\epsilon\)-Oracle model (A\ref{assumption:oracle_trace}), the asymptotic error of the true CMI \(I\) is bounded by:
\begin{small}
\begin{equation*}
\limsup_{N \to \infty} | I - \hat{I}_N  | \leq 2 \sqrt{\epsilon/2} \ln(2) + 2(1+\sqrt{\epsilon/2} )h_b\left(\frac{\sqrt{\epsilon/2} }{1+\sqrt{\epsilon/2} }\right)
\end{equation*}
\end{small}
\noindent where $h_b(\cdot)$ is the binary entropy function $h_b(p) = -p \ln p - (1-p) \ln (1-p)$.
\end{theorem}
\textit{(Proof Sketch)} We use the Alicki-Fannes-Winter \cite{Winter_2016} inequality for the difference between conditional entropies of two distributions with a small total variation distance \(\delta(P, P_\theta) \) corresponding to an \(\epsilon\)-Oracle model.

\begin{proof}
By definition, the Conditional Mutual Information is the difference of two conditional entropies (Eq. \ref{eq:cmi_theorique_trace}); by its symmetry (Lemma~\ref{lemma:mi_symmetry}), we write it with the cause \(E_t\) as target:

    $$I(E_{t'}, E_t | X_{<t}) = H(E_t | X_{<t}) - H(E_t | E_{t'}, X_{<t})$$

    Our framework operates in the teacher forcing regime of sequence-to-sequence models: the model is always conditioned on the observed events, never on its own predictions. We do not sample the trajectory $X_{<t}$ from the model's joint distribution. Instead, we estimate the CMI conditioned on the observed trajectory $x_{<t}^{(l)}$. Consequently, the relevant error metric is the per-step conditional divergence at time $t$, given the fixed trajectory. %

    Let $\Delta = |I - I_\theta|$ be the CMI estimation error. By the triangle inequality:

    $$\Delta \leq \underbrace{|H_P(E_t \mid X_{<t}) - H_\theta(E_t \mid X_{<t})|}_{\text{Term A}} + \underbrace{|H_P(E_t \mid E_{t'}, X_{<t}) - H_\theta(E_t \mid E_{t'}, X_{<t})|}_{\text{Term B}}$$

    We apply the sharp continuity bound for conditional entropy in classical systems (\cite{Winter_2016}, Lemma 2). Let $\rho$ and $\sigma$ be the true and model distributions respectively. Let $\delta = \frac{1}{2}\|P(E_t|\cdot) - P_\theta(E_t|\cdot)\|_1$ be the total variation distance. Since the target variable $E_t$ is binary (\(d_A = 2\)), the bound is: $$|H_\rho - H_\sigma| \leq \delta \ln(d_A) + (1+\delta)h_b\left(\frac{\delta}{1+\delta}\right)$$

    From Assumption~\ref{assumption:oracle_trace}, $\delta \leq \frac{1}{2}$. Substituting $d_A=2$ (so $\ln 2 $ in nats since we are using cross-entropy loss in PyTorch~\cite{pytorch}):$$|H_P(E_t \mid X_{<t}) - H_\theta(E_t \mid X_{<t})| \leq \delta \ln(2) + (1+\delta)h_b\left(\frac{\delta}{1+\delta}\right)$$

    Term B represents the same entropy difference conditioned on an augmented set $\{E_{t'}, X_{<t}\}$. Since the target dimension $d_A$ remains 2, the same bound applies, with $\delta$ now the total variation distance between the true and model joint distributions of $(E_t, E_{t'})$ given the trajectory (see the remark below). Summing the terms:
    $$ |I - I_\theta| \leq 2\delta \ln(2) + 2(1+\delta)h_b\left(\frac{\delta}{1+\delta}\right)$$
    Substituting the Pinsker upper bound $\delta = \sqrt{\epsilon/2}$ and noting that the right-hand side ($h_b(x)$) is monotonically increasing in \(\delta\), we obtain the final bound in terms of the oracle score $\epsilon$:

\begin{equation}
| I - I_\theta  | \leq 2 \sqrt{\epsilon/2} \ln(2) + 2(1+\sqrt{\epsilon/2} )h_b\left(\frac{\sqrt{\epsilon/2} }{1+\sqrt{\epsilon/2} }\right)
\end{equation}

Finally, we decompose the total error into estimation variance and approximation bias using the triangle inequality:
\begin{equation}
    | \hat{I}_N - I | = | (\hat{I}_N - I_\theta) + (I_\theta - I) | \leq \underbrace{| \hat{I}_N - I_\theta |}_{\text{Estimation Error}} + \underbrace{| I_\theta - I |}_{\text{Approximation Bias}}
\end{equation}
Given that $\hat{I}_N$ is a consistent estimator of the model's internal CMI, $I_\theta$ (Prop.~\ref{prop:consistency_cmi_trace}). By the Strong Law of Large Numbers, $\hat{I}_N \xrightarrow{a.s.} I_\theta$ as $N \to \infty$ the stochastic estimation error vanishes, leaving only the irreducible approximation bias:
$$ \limsup_{N \to \infty} | \hat{I}_N - I | \leq 0 + |I - I_\theta|$$

We thus obtain the final bound in terms of the oracle score as:

$$ \limsup_{N \to \infty} | \hat{I}_N - I | \leq 2 \sqrt{\epsilon/2} \ln(2) + 2(1+\sqrt{\epsilon/2} )h_b\left(\frac{\sqrt{\epsilon/2} }{1+\sqrt{\epsilon/2} }\right)$$
\end{proof}
\begin{remark}
Assumption~\ref{assumption:oracle_trace} bounds the divergence of a single next-event distribution. For Term~B, the relevant joint distribution of \((E_t, E_{t'})\) given \(x_{<t}\) involves the steps \(t, \ldots, t'\). By the chain rule of the KL divergence and the data-processing inequality~\cite{cover1999elements}, its divergence is at most \((t'-t+1)\,\epsilon\). The bound of Theorem~\ref{thm:total_error_bound} is therefore exact for Term~A, and holds for Term~B with \(\epsilon\) replaced by \((t'-t+1)\,\epsilon\); the error still vanishes as \(\epsilon \to 0\).
\end{remark}
This confirms that minimizing the cross-entropy loss ($\epsilon$) directly minimizes the upper bound on structural causal error using the CMI as causal strength. Crucially, this bound implies that as the AR model approaches the true distribution ($\epsilon \to 0$), the causal identification error vanishes.

\subsection{Identifiability}
Standard Faithfulness~\cite{constrainct_based_cd} (Def.~\ref{def:bn_faithfulness}) relies on Oracle CI-tests returning exact zeros (for the conditional mutual information), an unrealistic premise under finite samples and imperfect density estimation. We instead adopt a variant of Strong Faithfulness~\cite{uhler2013geometry}, which requires valid causal associations to exceed a detection threshold $\tau$ (i.e., $I > \tau$). Crucially, in our setting, this lower bound $\tau$ is by the estimator's bias (Theorem~\ref{thm:total_error_bound}) (if we assume \(N \rightarrow +\infty\)).
\begin{definition}[$\epsilon$-Strong Faithfulness]\label{def:strong_faithfulness}
Let $\tau_\epsilon$ be the asymptotic approximation error bound of the model (Theorem~\ref{thm:total_error_bound}). A distribution $P$ is \textbf{$\epsilon$-Strongly Faithful} to a causal graph $\mathcal{G}$ with respect to the estimator $P_\theta$ if, for every active edge $E_{t} \to E_{t'} \; \text{with} \; t < t'$, the \textbf{true} CMI satisfies:
\begin{equation}
I(E_{t'}, E_t \mid X_{<t}) > 2\tau_\epsilon
\end{equation}
\end{definition}
\noindent Identifiability is guaranteed provided the true causal signal dominates the model's approximation error (verified experimentally in Appendix~\ref{appendix:validation_of_epsilon_faithfulness}).

\begin{lemma}[Identifiability of the Sample Time Causal Graph]\label{lemma:identifiability_trace}Let $\hat{I}_N$ be the consistent Monte Carlo estimator of the CMI derived from the $\epsilon$-Oracle model $P_\theta$. Under the assumption of $\epsilon$-Strong Faithfulness (Def.~\ref{def:strong_faithfulness}), the sample time causal graph \(\mathcal{G}_{t,s}\) is identifiable with probability 1 as $N \to \infty$.
\end{lemma}
\begin{proof}
Let $\Delta = | \hat{I}_N - I |$ be the total estimation error. From Theorem~\ref{thm:total_error_bound}, we have a finite bound corresponding to a noise floor \(\tau_\epsilon\) such as $\limsup_{N \to \infty} \Delta \leq \tau_\epsilon$. Hence, for any \(\eta > 0\), \(\Delta \leq \tau_\epsilon + \eta\) almost surely for all sufficiently large \(N\). Since \(\eta\) can be chosen arbitrarily small (in particular \(\eta < \gamma/2\) below), we write \(\tau_\epsilon\) for \(\tau_\epsilon + \eta\) in the following, which does not change the conclusions.

We analyze the two cases for binary classification of the edge $E_t \to E_{t'}$:

\textbf{Case 1: No Edge ($H_0$).}
If the edge is absent, $I = 0$. The estimator is bounded by the noise floor:
\begin{align}
  &  0 \le |\hat{I}_N -I| \le \tau_\epsilon\notag  \\
  &\iff 0 \leq \hat{I}_N \leq \tau_\epsilon\notag
\end{align}
The CI-test is rejected (Correct Rejection).

\textbf{Case 2: Active Edge ($H_1$).}
If the edge exists, by Definition~\ref{def:strong_faithfulness},
$I = 2\tau_\epsilon + \gamma$ for some $\gamma > 0$.
We have:
\begin{align}
  &  |\hat{I}_N - I| = \Delta \notag\\
  &\implies - \Delta \leq \hat{I}_N - I \leq \Delta \notag\\
  &\implies I - \Delta \leq \hat{I}_N \leq \Delta + I \notag\\
  &\implies 2\tau_\epsilon + \gamma - \tau_\epsilon \leq \hat{I}_N \leq \Delta + I \notag\\
  &\implies \tau_\epsilon + \gamma \leq \hat{I}_N \leq \Delta + I \\
  &\implies \tau_\epsilon < \hat{I}_N
\end{align}
Since \(\gamma > 0\) we have \(\hat{I}_N > \tau_\epsilon\). The estimator detects an edge  (Correct Detection). Thus the graph is identifiable.
\end{proof}

\subsection{Lagged Effects via Simulated Interventions}
To evaluate the lagged effects of an event \(E_{t}\) on \(E_{t'}\) with \(t< t'\), we control for the intermediate events, so-called \textit{mediators} \(\boldsymbol{M} = X_{t+1:t'-1}\). To this end, we simulate a Controlled Direct Effect (CDE) \cite{pearl_2009} of \(E_t\) on \(E_{t'}\), i.e., the effect of \(E_t\) on \(E_{t'}\) when the mediators are set by intervention rather than observed.

\begin{definition}[Randomized Interventional Do-Operator]\label{def:do_in_sequences_expected}
Let \(\boldsymbol{M} = X_{t+1:t'-1}\) be the set of intermediate events between cause \(E_t\) and effect \(E_{t'}\). We define the intervention \(do(\boldsymbol{M}\sim Q)\) as the expectation over counterfactual realizations sampled from a proposal \(Q\) (e.g., Uniform over \(|\mathcal{X}|\)) and average this effect for \(N\) samples \(m^{(l)}\) as with the Monte Carlo estimation (Eq.~\ref{eq:cmi_estimation_naive_monte_carlo}) such as:
\begin{align}
  P(E_{t'}|do(\boldsymbol{M} \sim Q), X_{<t}) &\triangleq \mathbb{E}_{\boldsymbol{M} \sim Q} \left[P(E_{t'}|\boldsymbol{M}, X_{<t}) \right] \\
    &\approx \frac{1}{N} \sum_{l=1}^{N} P(E_{t'} \mid \mathbf{m}^{(l)},  X_{<t})\notag
\end{align}
\end{definition}
\begin{remark}
    This effectively marginalizes out the intermediate causal mechanisms only if we assume that there are no hidden confounders (Assumption~\ref{assumption:causal_sufficiency}).
\end{remark}
Using the previous definition, we modify Eq.~\ref{eq:info_gain_trace} to detect lagged information gain from \(E_t\) to \(E_{t'}\), namely \(I^\mathcal{L}_G\):

\begin{definition}[Lagged Information Gain]\label{def:lagged_ig}
    Let \(E_t\) be the cause, \(E_{t'}\) be the effect (\(t < t'\)) and \(\boldsymbol{M} = X_{t+1:t'-1}\) the set of intermediate events. The Lagged Information Gain \(I^{\mathcal{L}}_G\) is defined:
\begin{align}\label{eq:lagged_info_gain}
I^{\mathcal{L}}_{G}(E_{t'}, e_t \mid x_{<t}) \triangleq D_{\mathrm{KL}}\Big( P\big(E_{t'} \mid do(\boldsymbol{M} \sim Q), e_t, x_{< t}\big)
\big\| P(E_{t'} \mid  do(\boldsymbol{M} \sim Q), x_{< t}) \  \Big)
\end{align}
\end{definition}
Under the do-operator, conditioning on the realized \(e_t\) is still admissible since the intervention affects \(M\) only (the intermediates).

\section{Algorithm: Parallel Causal Discovery}\label{sec:sequential_cd}

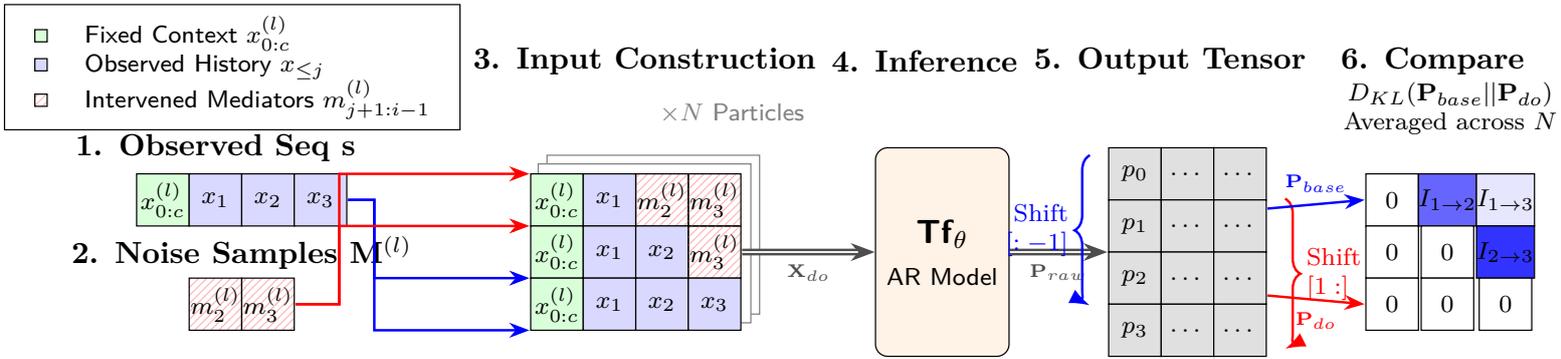
\begin{figure*}[!b]
    \centering
    \begin{adjustbox}{width=1.4\textwidth, center} %
\begin{tikzpicture}[
    font=\sffamily,
    >=Stealth,
    tensor/.style={matrix of nodes, nodes={draw, minimum size=6mm, anchor=center, font=\scriptsize}, column sep=-\pgflinewidth, row sep=-\pgflinewidth, inner sep=0pt},
    obs/.style={fill=blue!15},
    inter/.style={fill=red!15, pattern=north east lines, pattern color=red!30},
    context/.style={fill=green!15},
    gray_out/.style={fill=gray!20},
    label_text/.style={font=\bfseries\small, align=center}
]

\node[label_text] (label_seq) at (-0.3, 4.4) {1. Observed Seq $\mathbf{s}$};
\matrix[tensor] (seq) at (0, 3.8) {
    |[context]| $x^{(l)}_{0:c}$ & |[obs]| $x_1$ & |[obs]| $x_2$ & |[obs]| $x_3$ \\
};

\node[label_text] (label_noise) at (0, 3.2) {2. Noise Samples $\mathbf{M}^{(l)}$};
\matrix[tensor] (noise) at (0, 2.6) {
 |[inter]| $m^{(l)}_2$ & |[inter]| $m^{(l)}_3$  \\
};

\node[label_text] (label_broadcast) at (4.6, 5.4) {3. Input Construction};

\matrix[tensor] (staircase) at (4.5, 3.2) {
    |[context]| $x^{(l)}_{0:c}$ & |[obs]| $x_1$ & |[inter]| $m^{(l)}_2$ & |[inter]| $m^{(l)}_3$ \\
    |[context]| $x^{(l)}_{0:c}$ & |[obs]| $x_1$ & |[obs]| $x_2$ & |[inter]| $m^{(l)}_3$ \\
    |[context]| $x^{(l)}_{0:c}$ & |[obs]| $x_1$ & |[obs]| $x_2$ & |[obs]| $x_3$ \\
};

\begin{scope}[on background layer]
    \draw[fill=white, draw=black!50] ($(staircase.north west)+(0.2,0.2)$) rectangle ($(staircase.south east)+(0.2,0.2)$);
    \draw[fill=white, draw=black!50] ($(staircase.north west)+(0.1,0.1)$) rectangle ($(staircase.south east)+(0.1,0.1)$);
    \node at (5.6, 4.8) [text=black!50] {\scriptsize $\times N$ Trajectories};
\end{scope}

\draw[->, thick, blue] (seq.east) -- ++(0.3,0) |- ($(staircase.west)+(0, -0.3)$); %
\draw[->, thick, blue] (seq.east) -- ++(0.3,0) |- ($(staircase.west)+(0, -0.9)$); %

\draw[->, thick, red] (noise.east) -- ++(0.5,0) |- ($(staircase.west)+(0, 0.9)$); %
\draw[->, thick, red] (noise.east) -- ++(0.5,0) |- ($(staircase.west)+(0, 0.3)$);

\node[label_text] (label_inference) at (7.8, 5.4) {4. Inference};
\node[draw, fill=orange!10, minimum width=1.5cm, minimum height=2.4cm, rounded corners, align=center] (model) at (8, 3.2) {\textbf{Tf}$_\theta$ \\ \scriptsize AR Model};

\draw[->, thick, double, black!70] (staircase.east) -- (model.west)
    node[midway, below, font=\tiny\bfseries] {$\mathbf{X}_{do}$};

\node[label_text] at (10.6, 5.4) {5. Output Tensor};

\matrix[tensor] (out_tensor) at (10.8, 3.2) {
    |[gray_out]| $p_0$ & |[gray_out]| $\dots$ & |[gray_out]| $\dots$ \\
    |[gray_out]| $p_1$ & |[gray_out]| $\dots$ & |[gray_out]| $\dots$ \\
    |[gray_out]| $p_2$ & |[gray_out]| $\dots$ & |[gray_out]| $\dots$ \\
    |[gray_out]| $p_3$ & |[gray_out]| $\dots$ & |[gray_out]| $\dots$ \\
};

\node[label_text] (label_cmi) at (13.6, 5.4) {6. Compare};

\matrix[tensor] (cmi_mat) at (13.8, 3.2) {
    |[fill=white]| 0 & |[fill=blue!60]| $I_{1 \to 2}$ & |[fill=blue!10]| $I_{1 \to 3}$ \\
    |[fill=white]| 0 & |[fill=white]| 0 & |[fill=blue!80]| $I_{2 \to 3}$ \\
    |[fill=white]| 0 & |[fill=white]| 0 & |[fill=white]| 0 \\
};

\draw[->, thick, double, black!70] (model.east) -- (out_tensor.west)
    node[midway, below, font=\tiny\bfseries] {$\mathbf{P}_{raw}$};

\draw[decoration={brace, mirror, amplitude=5pt}, decorate, thick, blue]
    ($(out_tensor.north west)+(-0.2,-0.1)$) -- ($(out_tensor.south west)+(-0.2, 0.6)$)
    node[midway, left=3pt, font=\scriptsize, align=right] {Shift \\ $[:-1]$};

\draw[decoration={brace, amplitude=5pt}, decorate, thick, red]
    ($(out_tensor.north east)+(0.2,-0.6)$) -- ($(out_tensor.south east)+(0.2, 0.1)$)
    node[midway, right=3pt, font=\scriptsize, align=left] {Shift \\ $[1:]$};

\draw[->, thick, blue] ($(out_tensor.east)+(0, 0.5)$) -- (cmi_mat.west|-0, 3.8)
    node[midway, above, font=\tiny\bfseries] {$\mathbf{P}_{base}$};

\draw[->, thick, red] ($(out_tensor.east)+(0, -0.5)$) -- (cmi_mat.west|-0, 2.6)
    node[midway, below, font=\tiny\bfseries] {$\mathbf{P}_{do}$};

\node[above=0.3cm of cmi_mat, align=center, font=\scriptsize] (eq) {
    $D_{KL}(\mathbf{P}_{do} || \mathbf{P}_{base})$ \\
    Averaged across $N$
};

\node[draw, anchor=south east, fill=white] at (2.5, 4.6) {
    \scriptsize
    \begin{tabular}{cl}
         \tikz\node[draw, fill=green!15, inner sep=2pt] {}; & Fixed Context $x^{(l)}_{0:c}$ \\
         \tikz\node[draw, fill=blue!15, inner sep=2pt] {}; & Observed Trajectory $x_{\le t}$ \\
         \tikz\node[draw, fill=red!15, pattern=north east lines, pattern color=red!30, inner sep=2pt] {}; & Intervened Mediators $m^{(l)}_{t+1:t'-1}$ \\
    \end{tabular}
};

\end{tikzpicture}
\end{adjustbox}
\caption{\textbf{Overview of TRACE Parallel CI-tests.} We construct a single broadcasted tensor $\mathbf{X}_{do}$ where each row $t$ incrementally fixes the trajectory $x_{\le t}$ while randomizing the future (staircase pattern). The model processes this tensor in parallel to produce raw probabilities $\mathbf{P}_{raw}$ (grey). We then compute the Causal Mutual Information by comparing adjacent rows: the distribution at row $t-1$ serves as the baseline ($\mathbf{P}_{base}$, blue) for the intervention at row $t$ ($\mathbf{P}_{do}$, red); entry $I_{t \to t'}$ is the estimated CMI of the edge from cause $x_t$ to effect $x_{t'}$.}
\label{fig:trace_diagram}
\end{figure*}

TRACE uses a series of parallelizable tensor operations on GPUs. Instead of iterating sequentially, \(I^{\mathcal{L}}_G\) is evaluated for all candidate edges simultaneously. We start from an unknown \(\mathcal{G}_{un}\) representing the single stream and iteratively remove edges based on the CMI estimation \(\hat{I}_N\) to obtain \(\mathcal{G}_{t,s}\). An overview of the process can be found in Fig.~\ref{fig:trace_diagram}. Concretely, for a fixed observed trajectory $x_{\le t}$ (blue region), we build a broadcasted tensor $\mathbf{X}_{do}$ that stacks, row by row, one intervened continuation with mediator events $m^{(l)}_{t+1:t'-1}$ (red, hatched region). The earlier fixed context $x^{(l)}_{0:c}$ (green region) is shared across all rows. A single forward pass of $\text{AR}_\theta$ over this tensor yields, for every row simultaneously, the model's predicted distribution over the next event. We compare the distribution of the effect \(E_{t'}\) obtained at row $t$ (the intervened, "do", distribution $\mathbf{P}_{do}$) against the distribution at the preceding row $t-1$ (the baseline distribution $\mathbf{P}_{base}$) via $D_{KL}(\mathbf{P}_{do}||\mathbf{P}_{base})$, the same order as in the information gain (Eq.~\ref{eq:info_gain_trace}). Averaged across the $N$ sampled trajectories, this comparison yields exactly the CMI estimator $\hat{I}_N$ used to prune the edge $(x_t \rightarrow x_{t'})$ from $\mathcal{G}_{un}$. This staircase construction is what allows all $L$ candidate edges incident to a given effect event to be tested in a single batched forward pass, rather than $L$ sequential ones.

We introduce Theorem~\ref{thm:trace_soundness}, which guarantees the soundness of our algorithm when returning the strong sample time causal graph.
\begin{theorem}[Soundness of TRACE for the Sample Time Causal Graph]
\label{thm:trace_soundness}
Let $\mathcal{G}_{t,s}$ be the Sample Time Causal Graph of a sequence $s$ generated by a stochastic process.
Assume the underlying distribution $P$ is $\epsilon$-Strongly Faithful to $\mathcal{G}_{t,s}$ (Def.~\ref{def:strong_faithfulness}) and that TRACE uses a consistent CMI estimator $\hat{I}_N$ with threshold $\tau_\epsilon$ (Lemma~\ref{lemma:identifiability_trace}) and the corresponding \(\epsilon\)-Oracle Model \(P_\theta\). Then,
 under Causal Sufficiency (A\ref{assumption:causal_sufficiency}) and Temporal Precedence (A\ref{assumption:temporal_precedence}), it recovers the correct Sample Time Causal Graph $\mathcal{G}_{t,s}$ asymptotically as $N \to \infty$.
\end{theorem}
\textit{(Proof Sketch)}
By induction, we show that for each effect event \(E_{t'}\), we can recover its causes among the earlier events \(E_t\), \(t < t'\), using the consistent CMI estimator (Prop.~\ref{prop:consistency_cmi_trace}) which generates a \(\epsilon\)-Strong Faithful CI-test (Lemma~\ref{lemma:identifiability_trace}). We then control for lagged effects using Def.~\ref{def:lagged_ig}. By temporal precedence (A\ref{assumption:temporal_precedence}) and causal sufficiency (A\ref{assumption:causal_sufficiency}), we can conclude that no other events will affect the effect event \(E_{t'}\) and thus verify the heredity.

\begin{proof}
    (Throughout this proof, we identify the graph node at time \(t\) with both the realized event \(x_t\) and its binary indicator \(E_t \triangleq \mathds{1}_{X_t=x_t}\) from Section~\ref{c7:event_to_event}, using whichever notation is more convenient at each step.)

    We proceed by induction on the index of the effect event $t' \in \{0, \dots, L\}$ knowing temporal precedence (Assumption~\ref{assumption:temporal_precedence}).

    \textbf{Goal:} We show that for every $t'$, the estimated parent set $\widehat{Pa}(x_{t'})$ is exactly the true parent set $Pa(x_{t'})$ in the sample time causal graph $\mathcal{G}_{t,s}$.

    \textbf{Base Case ($t'=0$):}
    Consider the first event $x_0$. By temporal precedence and causal sufficiency (Assumption \ref{assumption:causal_sufficiency}), $x_0$ has no ancestors in the observed sequence. Thus, the true parent set is $Pa(x_0) = \emptyset$.
    TRACE evaluates candidate causes $E_t$ with $t < 0$. Since no such events exist in the sequence, the candidate set is empty and TRACE returns $\widehat{Pa}(E_0) = \emptyset$.
Thus, $\widehat{Pa}(E_0) = Pa(E_0)$.

\textbf{Heredity:}
Assume that for all effect indices smaller than $t'$, the algorithm has correctly identified the parent sets (note that the decision for a pair $(t, t')$ depends only on the trajectory $x_{<t}$ and the intervened mediators, not on previous graph decisions).
We consider the effect event $E_{t'}$. With the full variant, the algorithm iterates through all earlier events $E_t$, $t < t'$, as candidate parents. \emph{For each candidate}, we apply the decision rule based on the estimator $\hat{I}_N$ (Def.~\ref{def:strong_faithfulness}) assuming that no hidden confounders alters the CI-tests (Assumption~\ref{assumption:causal_sufficiency}):

\begin{itemize}\item \textbf{Case 1: $E_t$ is a True Parent ($E_t \in Pa(E_{t'})$).} By the $\epsilon$-Strong Faithfulness assumption (Def.~\ref{def:strong_faithfulness}), the true conditional mutual information satisfies $I > 2\tau_\epsilon$. By Lemma~\ref{lemma:identifiability_trace} (Identifiability), this ensures that the estimator satisfies $\hat{I}_N > \tau_\epsilon$ asymptotically. Consequently, TRACE \textbf{accepts} the edge.\item \textbf{Case 2: $E_t$ is Not a Parent ($E_t \notin Pa(E_{t'})$).}
Every path from $E_t$ to $E_{t'}$ other than a direct edge either goes back through the trajectory $X_{<t}$, which is conditioned on, or through the mediators $X_{t+1:t'-1}$, which are set by intervention (Def.~\ref{def:do_in_sequences_expected}); paths through later events cannot reach $E_{t'}$ (temporal precedence), and there are no hidden confounders (causal sufficiency). By the Causal Markov Condition, $E_{t'}$ is therefore independent of $E_t$ given $X_{<t}$ under the intervention on the mediators, and the lagged information gain (Def.~\ref{def:lagged_ig}) is zero.
By Lemma~\ref{lemma:identifiability_trace}, the estimator is bounded by the noise floor: $\hat{I}_N \le \tau_\epsilon$.
Consequently, TRACE \textbf{rejects} the edge.
\end{itemize}Since the algorithm makes the correct decision for every candidate $E_t$ individually, the resulting set $\widehat{Pa}(E_{t'})$ is identical to $Pa(E_{t'})$.

\textbf{Conclusion:} By induction, $\widehat{Pa}(E_{t'}) = Pa(E_{t'})$ for all $t'=0, \dots, L$. Since the graph $\mathcal{G}_{t,s}$ is defined by the union of these parent sets, TRACE recovers $\mathcal{G}_{t,s}$ exactly. Consequently, by Def.~\ref{def:iscg}, TRACE recovers the sample summary causal graph \(\mathcal{G}_s\).
\end{proof}

\subsection{Scalability of TRACE}
To parallelize the Monte-Carlo estimation (Eq.~\ref{eq:cmi_theorique_trace}) and counterfactual sampling (Eq.~\ref{eq:lagged_info_gain}) we avoid computing the CMI for the full trajectory \(x_{0:t-1}\) but a truncated version, called \emph{context} similarly to Chapter~\ref{c7:multi_label_one_shot_causal_discovery}, such as:
\[x_{<t} \approx  x_{0:c} \; \text{for} \; c < t \; \text{and} \; 0 < c \ll L\]
We argue that truncating the trajectory enables parallelized CI-tests on GPUs. In all experiments, we set \(c = \text{max}(0.1L, 20)\). Although it might break Markovianity for long sequence, empirical results show robustness to this truncation. We provide ablation to unseen sequence lengths during training and show our method to be robust to high delayed effects in Fig~\ref{fig:main_results}.
\paragraph{Sparse Approximation}
In contrary to the sample-level multi-label causal discovery case, for each effect \(E_{t'}\) we must perform \(t'\) CI-tests (one for every candidate cause \(E_t\), \(t < t'\)). For a sequence of length \(L\), the total number of CI-tests is given by \(\sum^L_{t'=1} t' = \frac{L(L+1)}{2}\) which grows quadratically with the sequence length. As a result, even on multiple GPUs, inference becomes computationally intractable.

To solve this, we propose a \emph{sparse variant} for which we bound the lagged effects of previous events on future events up to a memory \(m\) (Assumption~\ref{assumption:bounded_lagged_effects}). Thus the DGP \(\{X_t\}\) becomes an \(m\)-order Markov chain.
With \(m \ll L\), TRACE scales linearly with the sequence length \(L\). The memory complexity transitions from:
\begin{equation*}\label{eq:linearization_complexity}
    \mathcal{O}( N \cdot (L-c) \cdot L \cdot |\mathcal{X}|) \xrightarrow{\text{Bounded Memory}} \mathcal{O}(N \cdot m \cdot L \cdot |\mathcal{X}|)
\end{equation*}

\section{Experiments}
We evaluate TRACE on synthetic linear Structural Causal Models (SCMs) and real-world vehicle logs. All baselines utilize the same frozen backbone to evaluate the contribution of the inference mechanism.
TRACE is implemented in Python.
A more complete protocol description can be found in Appendix~\ref{appendix:evaluation} as well as additional ablations.
\subsection{Settings}
\paragraph{Synthetic Linear-SCM}
We validate TRACE on sequences generated by linear SCMs with controllable memory $m$, sequence length \(L\), and vocabulary size $|\mathcal{X}|$. Our evaluation proceeds in two phases: (1) We train a standard AR LM (LLaMA architecture \cite{touvron2023llamaopenefficientfoundation}) on the SCM and validate the training by monitoring the oracle scores \(\hat{\epsilon}\) (Eq.~\ref{eq:oracle_score}, normalized \(\epsilon\)) (2) We then apply TRACE to recover the summary causal graph of each single observation. To evaluate performance, we perform atomic interventions, i.e., interventions on a single event, by uniformly randomizing
\(E_{t}\) and measuring the average KL divergence over 10 counterfactual between post-intervention and observational distributions of \(E_{t'}\). If the divergence is above a margin \(\delta = 0.05\) (distinct from the estimator's CMI-detection threshold \(\tau\) discussed throughout this chapter), an edge \(E_t \rightarrow E_{t'}\) exists in \(\mathcal{G}_{t,s}\).
We then report the Precision, Recall, and Structural Hamming Distance (SHD) against this ground truth.

We benchmark TRACE against four distinct baseline types:
\begin{itemize}[leftmargin=*, noitemsep, topsep=1pt]
    \item \textbf{Neural Granger}: A Granger-causal discovery method that uses the same AR Model as TRACE but computes the difference in probability rather than the CMI to detect causality.
    \item \textbf{Attention}: We train a BERT~\cite{bert} model on the same SCM with the same model capacity and training steps as \(\text{Tf}_\theta\) and extract the attention scores at the deepest layer. A threshold \(\tau = 0.02\) is applied to get the adjacency matrix.
   \item \textbf{Saliency} (Input $\times$ Gradient): A local sensitivity baseline that estimates feature importance by computing the gradient of the target token's log-probability with respect to the input embeddings~\cite{inputxgradient}. %
    \item \textbf{Shapley Value Sampling}: An axiomatic attribution method rooted in cooperative game theory~\cite{shap}. Shapley values estimate the \textit{average marginal contribution} of a token $e_{t}$ to the prediction of $e_{t'}$ by sampling permutations of the input trajectory.
    \item \textbf{Naive baselines}: A random guesser that predicts edges $(E_t \to E_{t'})$ with a fixed probability $\rho=0.01$ and a frequency baseline that assumes the top-$k$ most frequent event types are universal causes for all other events. These tests reveal whether the task is non-trivial.
\end{itemize}
\subsection{Comparative Analysis}
As detailed in Table \ref{tab:baselines}, TRACE establishes state-of-the-art performance for causal discovery on discrete sequences generated from single streams and outperforms the strongest baseline by over 0.20 F1 Score. While attention scores alone fail to distinguish correlation from causation (F1 \(0.50\)), the Neural Granger baseline achieves a respectable F1 of \(0.69\) but suffers from high variance (SHD \(100.2 \pm 14.6\)). Saliency methods, traditional in NLP, also fail to distinguish causal links between events, especially with poor precision \(0.51\). This disparity highlights a critical insight:
\textbf{measuring the CMI is a far more robust signal of causality} than monitoring the probability fluctuation of a single target token (Granger) or relying on metrics that are not anchored in causal discovery (Saliency, attention scores).

\begin{table}[!h]
  \begin{center}
    \begin{small}
        \resizebox{\columnwidth}{!}{
        \begin{tabular}{lccc}
          \toprule
          Method & SHD ($\downarrow$) & F1 ($\uparrow$) & Precision ($\uparrow$) \\
          \midrule
          Random                       & 218.8{\scriptsize $\pm$3.2} & 0.01{\scriptsize $\pm$0.00} & 0.04{\scriptsize $\pm$0.01} \\
          Frequency                    & 723.0{\scriptsize $\pm$6.7} & 0.09{\scriptsize $\pm$0.00} & 0.06{\scriptsize $\pm$0.00} \\
          Attention (BERT)             & 321.0{\scriptsize $\pm$15}  & 0.50{\scriptsize $\pm$0.01} & 0.35{\scriptsize $\pm$0.01} \\
          Saliency (Input x Gradient LLaMA)       & 160.2{\scriptsize $\pm$6.55}  & 0.67{\scriptsize $\pm$0.01} & 0.51{\scriptsize $\pm$0.01} \\
          Shapley Value Sampling (LLaMA)       & 148.0{\scriptsize $\pm$5.09}  & 0.60{\scriptsize $\pm$0.01} & 0.55{\scriptsize $\pm$0.01} \\
          Neural Granger (LLaMA)       & 100.2{\scriptsize $\pm$14.6}  & 0.69{\scriptsize $\pm$0.04} & 0.71{\scriptsize $\pm$0.04} \\
          \midrule
          \textbf{TRACE} (LLaMA)       & \textbf{28.6}{\scriptsize $\pm$2.8} & \textbf{0.91}{\scriptsize $\pm$0.01} & \textbf{0.89}{\scriptsize $\pm$0.01} \\
          \bottomrule
        \end{tabular}
        }
    \end{small}
  \end{center}
   \caption{\textbf{Identifiability Comparison.} Comparison of causal discovery performance on synthetic SCMs with \(|\mathcal{X}|=1000, L=64, \epsilon=0.05, \tau=3.10^{-5}, N=128\). \textbf{TRACE} significantly outperforms local (Saliency) and global (Granger/Shapley) baselines, achieving over 20 points higher F1 while maintaining high precision. Results across 10 runs are reported.}
  \label{tab:baselines}
  \vskip -0.1in
\end{table}

\begin{figure}[!h]
    \centering
   \includegraphics[width=1\linewidth]{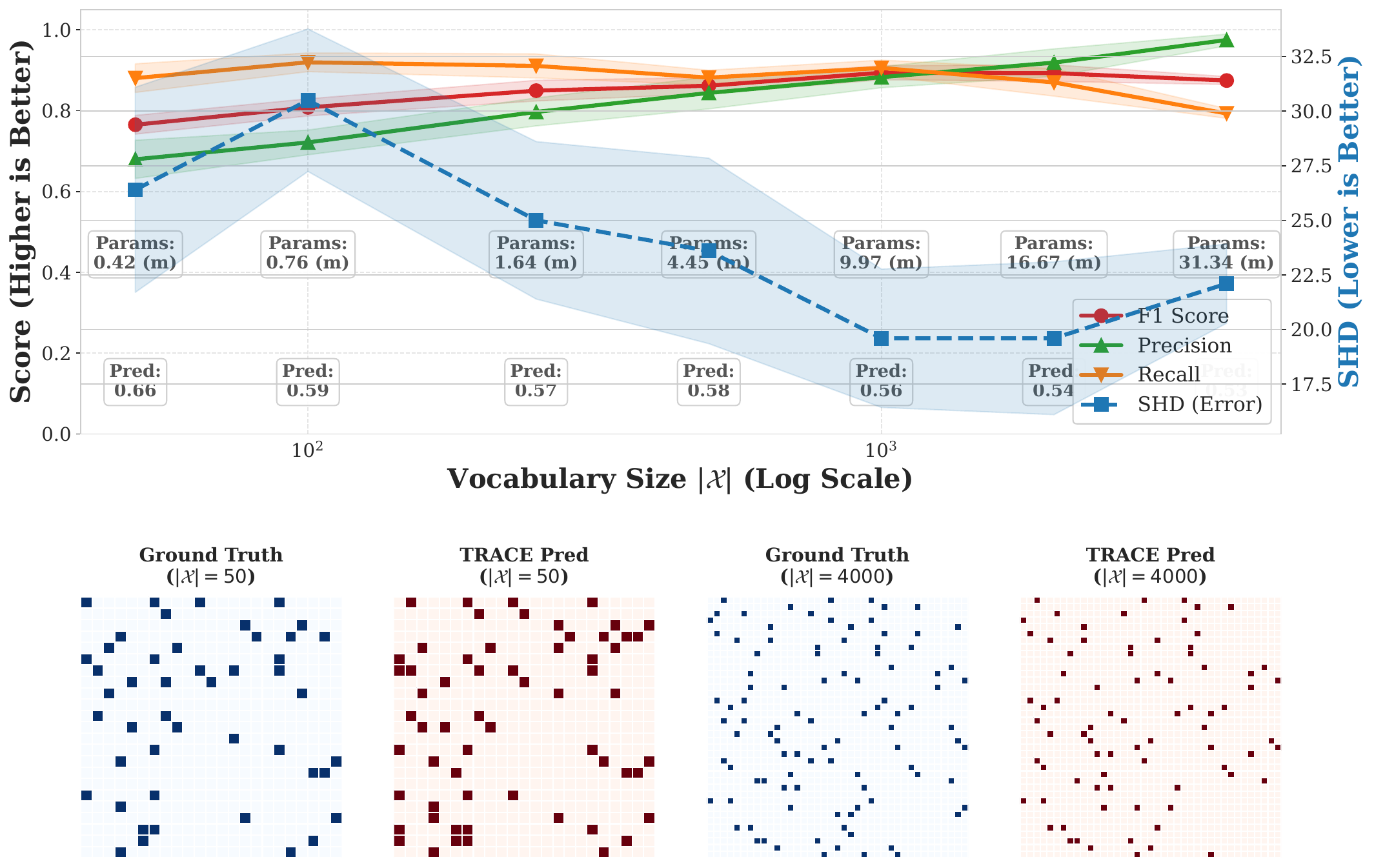}
       \caption{\textbf{Scalability to High-Cardinality Event Spaces.} Evaluation of structural identifiability across exponentially growing vocabulary sizes. \textbf{Top:} Evolution of discovery metrics. TRACE exhibits \textbf{performance invariance}, maintaining F1 $\approx 0.81$ even as the combinatorial search space explodes. \textbf{Bottom:} Visual examples of recovered summary graphs $\mathcal{G}_s$ at scale. The Shannon redundancy \textit{Pred} = $1-H(P)/H_{max}$ (0 for uniform noise, 1 for a deterministic process) confirms that TRACE succeeds even in high-entropy regimes. ($\hat{\epsilon}=0.01, L=64, N=64, \tau=10^{-4}$).}
    \label{fig:Comparison_identifiability_n_scalability}
\end{figure}

\subsection{Scalability and Robustness Analysis}

\paragraph{Scaling to Large Event Vocabularies}
Standard causal discovery algorithms suffer from combinatorial explosions when the variable count or state space increases. In Fig.~\ref{fig:Comparison_identifiability_n_scalability}, we challenge TRACE with massive vocabularies to validate its scalability to massive event types. The top panel reports the discovery metrics for vocabularies from \(|\mathcal{X}| = 50\) to \(4000\) (log scale); the boxes give the size of the AR model trained for each vocabulary and the predictability \(\textit{Pred}\) of the SCM. The bottom panel compares the ground-truth and the recovered summary graphs, drawn as adjacency matrices, for \(|\mathcal{X}| = 50\) and \(4000\). As the vocabulary grows, precision increases (from about \(0.68\) to \(0.97\)) while recall decreases slightly (from about \(0.9\) to \(0.8\)), and the SHD stays between about \(20\) and \(30\) edges, with its lowest values for the largest vocabularies. Remarkably, we observe that \textbf{discovery performance is invariant to the vocabulary size} $|\mathcal{X}|$. The F1 score remains stable, between about $0.77$ and $0.89$, even as the search space grows exponentially and the underlying SCM entropy increases. This confirms a key advantage of our approach: by leveraging a pre-trained AR model, TRACE enables causal discovery over massive event vocabularies at scales that were infeasible before.

\begin{figure*}[!h]
    \centering
    \begin{adjustbox}{width=1.35\textwidth, center} %
    \includegraphics[width=1\linewidth]{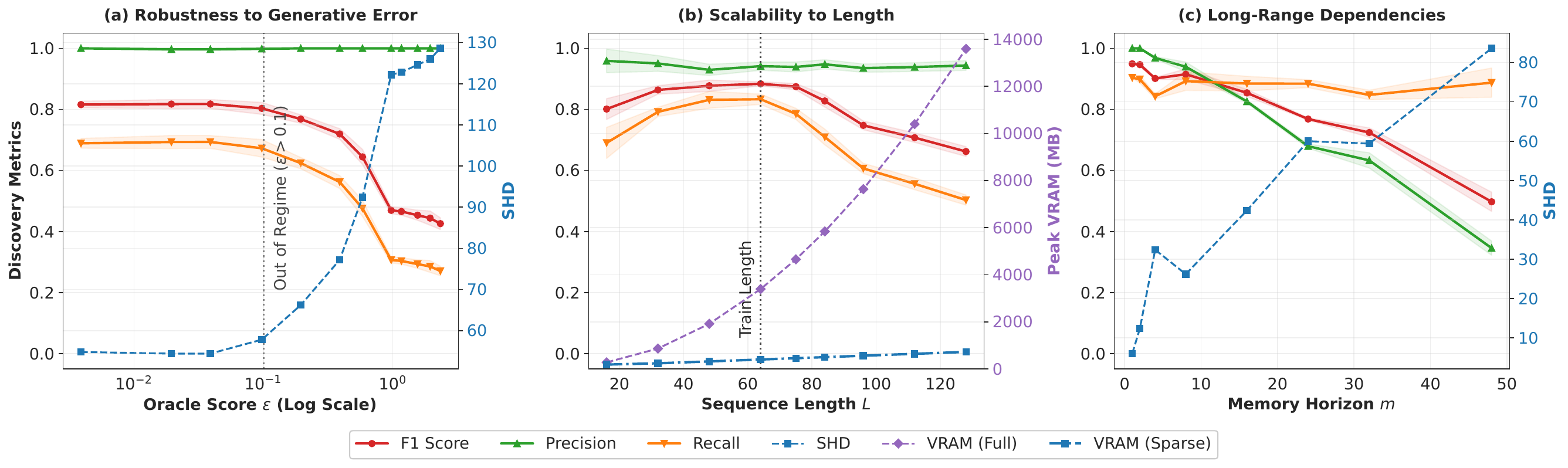}
    \end{adjustbox}
    \caption{\textbf{Robustness and Scalability Analysis} ($|\mathcal{X}|=1000, N=128, \tau=10^{-4}, L=64$). Evolution of causal discovery performance (F1, Precision, Recall, SHD). \textbf{(a) Robustness to Generative Error:} Performance as a function of the model's oracle score $\epsilon$. TRACE exhibits a phase transition, recovering structure even for imperfect models ($\epsilon < 0.1$) and maintaining high Precision even as fidelity degrades. \textbf{(b) Scalability to Length:} Performance and GPU memory usage vs. sequence length $L$. The \textbf{Sparse} variant demonstrates linear memory scaling ($O(mL)$), enabling inference on sequences far exceeding the training length ($L=64$), whereas the Full variant scales quadratically. \textbf{(c) Long-Range Dependencies:} Robustness to increasing delayed-effects $m$. TRACE maintains F1 $>0.8$ even as dependencies span one third of the sequence ($m=20$), confirming the method's ability to capture distant causal mechanisms.}
    \label{fig:main_results}
\end{figure*}

\paragraph{Unseen Sequence Lengths}
We evaluate TRACE's ability to scale to unseen sequence length during training of the AR Model, as well as the maximum GPU memory consumption for the \textit{Full} and the \textit{Sparse} variant, which computes up to \(m\) lagged effects per event. As shown in Fig.~\ref{fig:main_results} (b), the sparse variant of TRACE achieves linear memory scaling, whereas the full variant exhibits quadratic memory growth. For the classification metrics, we observe a quick degradation in Recall as the sequence length exceeds the training window. Therefore, the model's ability to identify all causal links diminishes as the temporal context becomes increasingly out-of-distribution. Precision remains remarkably high and stable \((\approx 0.95\)) across all tested lengths. This suggests that the model becomes more conservative with longer sequences. This \textbf{conservative failure mode} is highly desirable and further supported by Fig.~\ref{fig:main_results} (a).
\paragraph{Identifiability Precedes Convergence.}
A central question is whether the AR model must be perfect ($\epsilon \to 0$) to recover causal structure. Fig.~\ref{fig:main_results} (a) reveals a critical finding: \textbf{exact convergence is not a necessary condition for identifiability}. We observe a distinct phase transition where the coarse-grained causal graph is recovered early in the training dynamics ($\epsilon \approx 0.1$), before the model masters fine-grained transition probabilities. TRACE exhibits a \textbf{conservative failure mode}. As approximation error $\epsilon$ increases, the model defaults to "blindness" (lower Recall) rather than "hallucination" (lower Precision), maintaining near-perfect precision ($>0.95$) even in high-entropy regimes. This suggests that model uncertainty manifests as a failure to detect weak signals rather than the generation of false positives—a desirable property for safety-critical applications.
\paragraph{Deep Temporal Dependencies.}
In Fig.~\ref{fig:main_results} (c), we stress-test the method by extending the memory horizon of the underlying SCM up to $m=48$. TRACE maintains high stability (F1 $> 0.80$) up to \(m=20\), confirming that our parallelized intervention mechanism effectively \textbf{captures long-range dependencies}. While precision naturally softens as the number of events behind detected increases, hence necessitating a bigger threshold \(\tau\).
\begin{figure}[!h]
    \centering
    \includegraphics[width=0.85\linewidth, trim=0 0 0 28pt, clip]{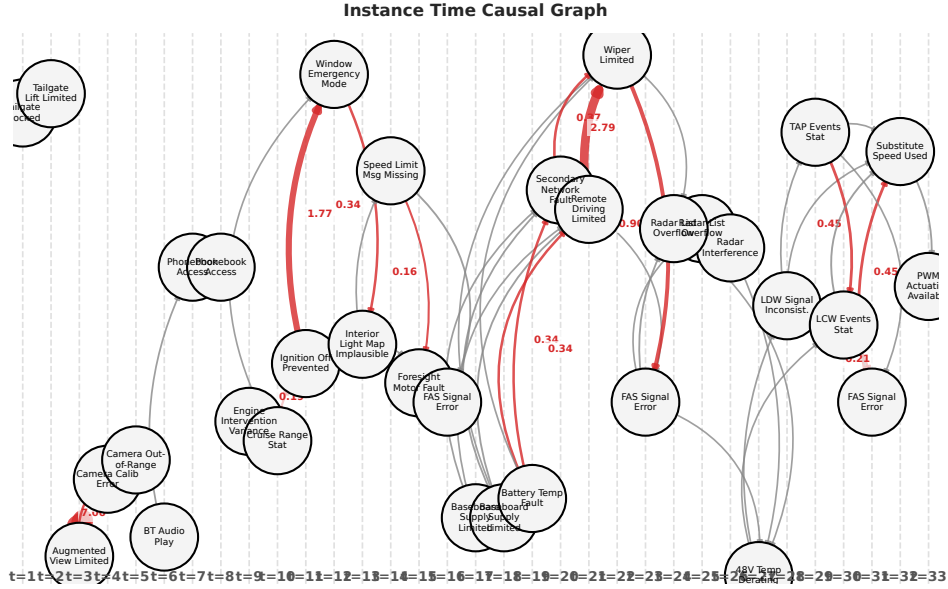}
    \caption{\textbf{Sample Time Causal Graph.} Temporal evolution of a diagnostic defect cascade in a vehicle ($|\mathcal{X}| \approx 29,100$). TRACE effectively captures causal relationships, revealing distinct \textbf{error clusters} at different time steps (e.g., initial sensor failures at $t=3$ triggering mechanical faults at $t=12$, battery at issue \(t=17\)). This enables actionable root-cause analysis by isolating the specific onset of a failure mechanism and their strength using the CMI \(\hat{I}_N\)}%
    \label{fig:time_instance_graph_dtc}
\end{figure}
\subsection{Application to Vehicles Diagnostics}\label{sec:application}
To evaluate TRACE, we use the same dataset as in Chapters~\ref{c7:multi_label_one_shot_causal_discovery} and \ref{c7:multi_label_causal_discovery} ($|\mathcal{X}| = 29{,}100$ DTCs). The training loss reported of \(\text{Tf}_x\) is $\mathcal{L}_{AR} = 1.91$ nats, corresponding to an oracle score \(\hat{\epsilon}\) (Eq.~\eqref{eq:oracle_score}) that places the model well within the identifiable regime characterized in Section~\ref{sec:additional_ablations}. We apply TRACE to analyze complex electrical cascades, specifically focusing on battery and sensor degradation scenarios, as shown in Fig.~\ref{fig:time_instance_graph_dtc} with the sample time causal graph \(\mathcal{G}_{t,s}\) and in Fig.~\ref{fig:carformer_summary_appendix} (Appendix~\ref{appendix:part2}) with the corresponding sample summary causal graph \(\mathcal{G}_s\) of the same sequence.

The sample time graph in Fig.~\ref{fig:time_instance_graph_dtc} exhibits three qualitatively distinct clusters that align with the physical layout of the diagnosed vehicle. An early cluster of sensor- and camera-related events appears around \(t=3\), a subsequent cluster of mechanical/motor faults around \(t=12\), and a late-stage cluster of battery- and network-related faults around \(t=17\). This temporal separation is consistent with a cascading failure mechanism, in which an initial sensor malfunction propagates through the vehicle's control network before eventually degrading the power system, rather than all faults being independently and simultaneously triggered. In standard approaches, these events are often collapsed into a single static correlation graph, obscuring both the order of operations and the directionality of their causal relationship. TRACE instead preserves the temporal ordering while still recovering the compact type-level summary graph \(\mathcal{G}_s\) needed for root-cause analysis. We regard this qualitative case study as a proof of concept illustrating TRACE's applicability to industrial diagnostic logs; a systematic, expert-validated evaluation of many such cascades remains an important direction for future work (Section~\ref{sec:limitation}). %

\section{Limitations}\label{sec:limitation}
We now include a discussion regarding the main assumptions taken in this chapter.

\subsection{Causal Sufficiency}\label{sec:limitation_causal_sufficency}
A fundamental assumption in causal discovery is \textit{Causal Sufficiency} (Assumption~\ref{assumption:causal_sufficiency})—the premise that no unobserved confounders influence the system. Since TRACE relies on pre-trained backbones which may have learned from noisy or incomplete data, we empirically evaluate the robustness of TRACE under controlled violations of causal sufficiency, focusing on two realistic forms of hidden confounding.

\begin{figure}[!h]
    \centering
    \includegraphics[width=0.9\linewidth]{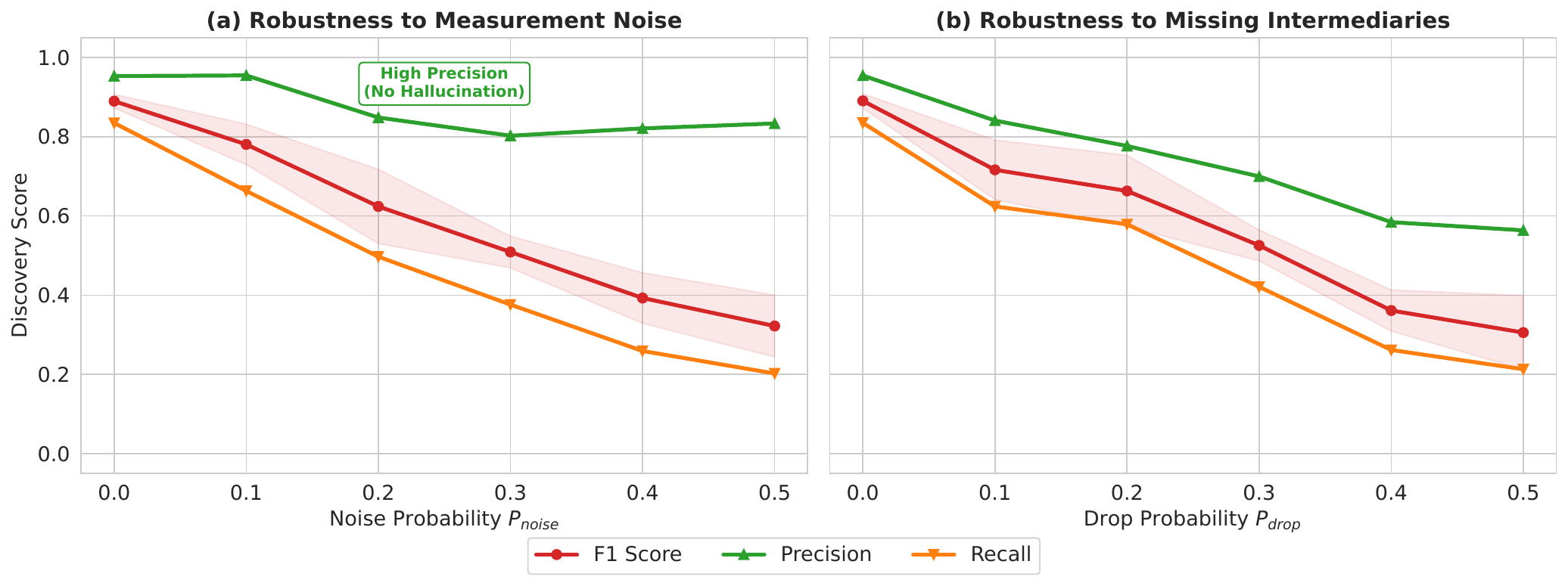}
\caption{\textbf{Robustness to Hidden Confounding.} Evaluation of TRACE under violations of causal sufficiency. \textbf{(a)} Measurement Error: Random noise is injected into the trajectory. Precision stays high, indicating resistance to hallucination. \textbf{(b)} Temporal Drops: Time steps are randomly deleted, thus conditioning sets are broken. TRACE still recovers structure despite missing intermediaries but the discovery scores quickly decrease.}
\label{fig:limitations}
\end{figure}

\paragraph{Measurement Error (Noise Injection).} In Fig. \ref{fig:limitations}(a), we simulate measurement error by randomly replacing valid tokens in the trajectory with noise (at a fixed corruption probability \(P_{noise}\)). While Recall naturally degrades as the true causal parents are obscured, \textbf{Precision remains high} ($>0.8$) even when 40\% of the trajectory is corrupted. This confirms that TRACE does not hallucinate false edges from noisy inputs; if the causal signal $X \to Y$ is destroyed by measurement error, the model assigns $CMI \approx 0$ rather than hallucinating a spurious link.

\paragraph{Missing Intermediaries (Temporal Drops).} In the same Fig. \ref{fig:limitations}(b), we simulate missing data by randomly dropping time steps, effectively hiding intermediate nodes in the causal chain ($X \to Z_{hidden} \to Y$). This is a more critical scenario where the Markov conditioning sets are incomplete as intermediate causal nodes are unobserved. %

\subsection{Temporal Precedence \& Instantaneous Effects}
TRACE assumes that causal influence respects temporal precedence, which means observations are perfectly recorded over time and therefore does not model instantaneous causal effects between events occurring at the same time index. This assumption is standard in sequential causal discovery and ensures that the recovered causal graph is acyclic and identifiable in the single observed sequence setting.

From a theoretical standpoint, instantaneous effects are not identifiable from a single observed trajectory without additional parametric assumptions, repeated samples, or access to interventions. In practice, apparent simultaneity often \emph{arises from time discretization}, logging resolution, or batching effects. TRACE interprets such cases \emph{through the earliest observable temporal ordering}, yielding a conservative but identifiable causal structure. As a result, the recovered summary graph captures directed causal influence with positive temporal delay, rather than true simultaneity.

\subsection{Consistency Through Time}
TRACE operates in a sample-level regime, where causal structure must be inferred from a single observed trajectory rather than from repeated i.i.d.\ samples \emph{during inference}. To make this statistically meaningful, we assume \emph{consistency through time}: the causal mechanisms governing the generation of events are invariant across time indices. In other words, causal directionality does not reverse over time. For instance, if \(A \rightarrow B\), it is subsequently assumed that \(B \not\rightarrow A\).

This assumption is strictly weaker than stationarity~\cite{assaad_survey_ijcai_cd_time_series}. While the marginal distribution of $\{X_t\}$ may vary over time, the underlying causal dependencies—encoded by the directed edges of the sample time causal graph remain stable. This form of causal invariance is standard in sequential settings and underlies the validity of summary causal graphs that collapse time-indexed relations into event-to-event dependencies (Def.~\ref{def:iscg}).

\subsection{Amortization}
TRACE leverages amortized causal discovery, shifting the computational burden to the self-supervised pretraining of an autoregressive model. However, because the base entropy of the true data-generating process is rarely known in practice, the generative error $\epsilon$ cannot be directly deduced from the cross-entropy loss. This makes identifying the optimal model capacity and training duration challenging; as a practical heuristic, we recommend slight over-parameterization and extended training. Furthermore, while TRACE achieves linear scaling during inference with respect to sequence length and vocabulary size ($|\mathcal{X}|$), this efficiency does not account for the prerequisite pretraining phase, which scales non-linearly with $|\mathcal{X}|$.

\subsection{Threshold Calibration and the Mediator-Randomization Trade-off}\label{sec:limitation_threshold_mediator}
An independent replication of TRACE~\cite{chadyuk2026replicatingtrace}, reimplemented from the public \texttt{seq2cause} release (Appendix~\ref{sec:appendixc_reproducibility}) without access to the BMW data used in this thesis, reproduces the headline result of Section~\ref{sec:application} (F1 $\approx 0.90$--$0.91$ against exact interventional ground truth, on the synthetic benchmark of Appendix~\ref{appendix:scm}) across vocabulary sizes $|\mathcal{X}| \in \{100, \ldots, 2000\}$. It also refines two practical aspects of the method, which we summarize here.

First, on threshold selection, they show the CMI-detection threshold $\tau$ that maximizes F1 is not a universal constant. Instead, it tracks the causal strength of the underlying data-generating process, concretely its lag decay: across regimes, the selected $\tau^*$ consistently straddles the ground-truth margin $\delta$ (Section~\ref{sec:application}) rather than any fixed value, drifting with vocabulary size, redundancy, and the sharpness of the lag decay. This corroborates selecting $\tau$ by validation on held-out data, as already practiced throughout this thesis's own experiments (e.g., Appendix~\ref{sec:additional_ablations}), and indicates that a $\tau$ value reported for one dataset should not be transplanted to another without re-selection.

Second, on the mediator-randomization do-operator (Def.~\ref{def:do_in_sequences_expected}), they identify a stability limitation at longer lags. Uniformly randomizing every mediating position, as the staircase construction (Fig.~\ref{fig:trace_diagram}) does by design, drives the predicted effect token into a context far outside the pretraining distribution. Its probability then collapses toward its marginal frequency, and the resulting CMI reads orders of magnitude below its true value at lag $\geq 2$, independently of the number of sampled trajectories $N$. Consistent with this diagnosis, in practice we favor an \emph{atomic} intervention when probing such longer-range dependencies: only the candidate cause position is randomized, while mediating positions are held at their observed, factual values, keeping the effect token's predicted probability at its natural, in-distribution scale. This is a deliberate trade-off rather than a strict improvement: conditioning on the observed mediators no longer corresponds to an interventional average over trajectories and can, in principle, be confounded by the specific factual configuration, whereas Theorem~\ref{thm:trace_soundness}'s identifiability guarantee is stated for the fully randomized construction. We view formally characterizing this trade-off, rather than resolving it by fiat, as an open problem.

\section{Summary}

We presented \textsc{TRACE}, a framework for single-sequence causal discovery in
high-cardinality discrete event streams. By repurposing a pretrained autoregressive
model as a conditional density estimator, \textsc{TRACE} recovers the sample time
and summary causal graphs without any task-specific retraining, scaling
\emph{linearly} with the event vocabulary size.

Our comparative analysis demonstrates that measuring causal influence via conditional
mutual information is fundamentally more discriminative than attribution-based
alternatives: \textsc{TRACE} achieves an F1 score of $0.91 \pm 0.01$, outperforming
the strongest attribution baseline (Shapley value sampling, $0.60$) by more than 30
points and Neural Granger causality ($0.69$) by over 20 points. Critically, this gap
is not merely a matter of model capacity, all methods share the same pretrained
backbone, but reflects a fundamental distinction between \emph{correlation}
(captured by saliency and Granger-type scores) and \emph{causal dependence}
(captured by the CMI under controlled interventions). This result provides direct
empirical evidence that causal structure learning is necessary for reliable discovery
in event sequences, and cannot be replaced by post-hoc attribution.

On the theoretical side, Theorem~\ref{thm:total_error_bound} establishes a finite bound
on the gap between the estimated CMI $\hat{I}_N$ and the true causal strength $I$,
showing that this gap is governed exclusively by the model's cross-entropy
approximation error $\epsilon$. Lemma~\ref{lemma:identifiability_trace} then guarantees that
the sample time causal graph is identifiable with probability 1 as $N \to \infty$,
provided the true causal signal exceeds the noise floor $\tau_\epsilon$ induced by
$\epsilon$, a condition we term $\epsilon$-Strong Faithfulness. Taken together, these
results confirm that structural identifiability is achievable under imperfect density
estimation, and that minimizing the pretraining cross-entropy loss directly tightens
the causal identification bound. \textsc{TRACE} thus represents a principled,
GPU-scalable step toward causal discovery in real-world, high-cardinality event
streams, where both the scale and the noise level of data preclude oracle-level
approximations.

\section{Outlook}

The discovery frameworks presented in this part \textsc{OSCAR}, \textsc{CARGO},
and \textsc{TRACE} collectively establish a scalable pipeline for uncovering causal
structures in high-cardinality single-event streams. Several frontiers nonetheless
remain open.

\paragraph{Toward Causal Scoring Grounded in the Minimum Description Length (MDL).}

A conceptually significant direction concerns the theoretical foundations of the
causal scoring function underlying \textsc{TRACE}. At its core, \textsc{TRACE}
decides whether event $E_t$ causes $E_{t'}$ by measuring how much knowing $E_t$
reduces uncertainty about $E_{t'}$, given the observed trajectory. This quantity is the conditional mutual
information. This choice is empirically
effective, yet it can be given a deeper justification through the lens of the
\emph{Minimum Description Length} (MDL) principle~\cite{rissanen1978modeling,
grunwald2004tutorialintroductionminimumdescription}.

MDL is a general principle of model selection grounded in information theory:
given a dataset, the best explanation is the one that leads to the shortest
description of the data~\cite{grunwald2004tutorialintroductionminimumdescription}. Concretely, one compares two
descriptions of the same observations, a model $M$ and the data encoded
\emph{given} that model and selecting the model minimizing their combined
length $L(M) + L(\mathcal{D} \mid M)$. It is a mathematical formulation of the Occam's razor principle, stating that when faced with competing explanations for the same phenomenon, the simplest one, i.e., the one requiring the fewest assumptions, is usually the best. Applied to causal discovery, this
principle instantiates \emph{causal minimality}: the true causal graph is the
simplest one that fully accounts for the observed dependencies, and any
additional edge would only increase the total description length without
improving the fit. MDL-based methods have been
successfully applied to causal inference in discrete
data~\cite{cueppers2024causal} and, most closely related to the present
setting, to learning Granger-causal networks in \gls{tpp}~\cite{mdlh}.

\textsc{TRACE}'s approximation error bound (Theorem~\ref{thm:total_error_bound})
reveals a direct structural parallel with these ideas. Cross-entropy
\emph{is} a compression objective: a model achieving a lower cross-entropy provides a shorter expected code for the data under Shannon's source-coding theorem~\cite{cover1999elements}. Minimizing $\epsilon$ during pretraining therefore simultaneously tightens the noise floor $\tau_\epsilon$, sharpens the identifiability threshold, and, in the language of MDL, corresponds to finding a more compact two-part description of the event sequence.
Formalizing this connection constitutes a promising theoretical direction. %

\paragraph{Unified Event-to-Event and Events-to-Outcome Discovery.}
A framework supporting simultaneous recovery of both event-to-event and
events-to-outcome causal dependencies within a single sequence would provide a
more complete picture of system dynamics. As indicated by the open cell in
Table~\ref{tab:causal_framework}, the population-level counterpart of \textsc{TRACE}
remains an open problem, requiring new aggregation strategies and, likely, explicit
acyclicity constraints to reconcile sample time graphs across sequences.

\paragraph{From Causal Graphs to Actionable Rules.}
Despite providing a robust causal backbone, the numerical and graphical
representations produced by \textsc{TRACE} remain a silent language for the domain expert. Causal graphs identify statistical dependencies but do not
inherently capture the symbolic logic—such as Boolean combinations of
events—or the contextual nuance required for formal diagnostic standards.
There remains a substantial gap between numerical causal evidence and the actionable rules that practitioners can validate and trust. Part~\ref{pa:p3} closes this methodological arc. %
By transitioning from causal discovery to automated
reasoning, it embeds the causal structures recovered by \textsc{OSCAR}, \textsc{CARGO} within a multi-agent framework powered by \glspl{llm}. As a result, the system synthesizes interpretable Boolean rules alongside natural language explanations, effectively transforming raw causal evidence into human-interpretable diagnostic expertise.

\part{Automated Causal Reasoning and Logic Synthesis}
\label{pa:p3}

\chapter{Multi-Agent Causal Reasoning for Automated Error Pattern Rule Synthesis}\label{c8:carep}
\chaptermark{Multi-Agent Causal Reasoning for Automated Error Pattern Rule Synthesis}
\glsresetall

While Part~\ref{part1} focused on learning event representations and Part~\ref{pa:p2} on uncovering causal dependencies, Part~\ref{pa:p3} explores how these causal insights can be operationalized. It is based on the following contribution:

\begin{description}[style=nextline,leftmargin=0cm,labelsep=0em]

\item[\textbf{Neuro-Symbolic Rule Discovery: Empowering LLMs with Causality for Vehicle Diagnostics.}]
\cite{math2026multiagentcausalreasoningerror}
Hugo Math, Julian Lorenz and Rainer Lienhart. Workshop on \emph{Logical Reasoning of Large Language Models} at ICLR 2026, April 2026.
\end{description}
LLMs and agentic systems have demonstrated the capacity to integrate heterogeneous information sources for structured reasoning tasks. From this perspective, we introduce \gls{carep}: (\uline{C}ausal \uline{A}utomated \uline{R}easoning for \uline{E}rror \uline{P}atterns), a multi-agent reasoning system that leverages the causal structures discovered by OSCAR and CARGO. It automatically generates interpretable EP rules\footnote{Unlike Chapters~\ref{c2:ep_prediction_based_on_live_data}--\ref{c7:multi_label_causal_discovery}, where an EP denotes a labeled outcome, this chapter's goal is to recover the EP's underlying Boolean rule itself; see Section~\ref{sec:ep_def}.} together with their natural language explanation.
CAREP integrates three coordinated agents: (1) a \textit{causal discovery agent} that refines DTC–EP relationships using the underlying causal graphs, (2) a \textit{contextual information agent} that enriches reasoning with semantic and metadata knowledge, and (3) an \textit{orchestrator agent} that synthesizes candidate Boolean rules and produces transparent reasoning traces.

We evaluated both the capacity of LLMs and CAREP to generate accurate Boolean rules (i.e., correct DTCs) and the accuracy of those rules based on their truth tables relative to the ground truth. CAREP demonstrates the ability to automatically infer accurate and human-readable EP rules, outperforming purely language-model-based baselines while providing interpretability.
This framework thus completes the methodological trajectory of the thesis, transforming predictive modeling into causal discovery and finally into automated causal reasoning.

\section{Introduction}

Beyond causality, automation requires interpretability. Presenting unexplained rules is risky in safety-critical domains such as the automotive industry or in medical data. \Glspl{llm}~\cite{OpenAI_GPT4_2023, deepseekai2025deepseekr1incentivizingreasoningcapability, touvron2023llamaopenefficientfoundation} have recently demonstrated emergent capabilities in planning, explanation, abduction, reasoning, and in-context learning, i.e., solving a new task from examples given in the prompt without updating the model weights~\cite{icl, zhang2024largelanguagemodelsinterpolated}. Especially when using Chain-of-Thought \cite{chainofthoughts} (CoT), a prompting technique that asks the model to write intermediate reasoning steps before its answer, to guide them through step-by-step reasoning processes. Since CoT is not grounded in the external world, it is limited in its ability to reason reactively or update its knowledge, thereby propagating factual hallucinations and errors throughout the reasoning process. ReAct (Reason and Act) \cite{Yao2022ReActSR} goes further
by creating a \emph{reason to act} and \emph{act to reason} framework where the LLM has access to external environments (e.g., Wikipedia) to incorporate additional information into the reasoning.

\begin{figure}[!h]
    \centering
    \includegraphics[width=0.8\textwidth]{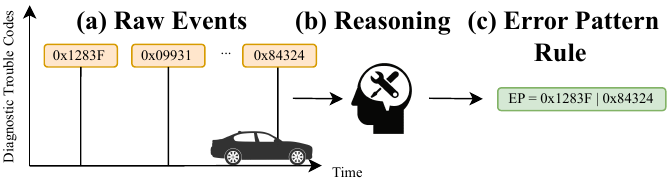} %
    \caption{\textbf{Illustration of Error Pattern Automation in Modern Vehicles}. (a) A vehicle with an unknown defect generates a sequence of Diagnostic Trouble Codes (DTCs) over time. (b) A domain expert analyzes the DTCs and the vehicle's metadata (e.g., model, descriptions). (c) Provide a Boolean rule to identify this error pattern.}
    \label{fig:illustration}
\end{figure}

Through \gls{rag},
external memory access, and context injection~\cite{rag_or},
LLMs can use raw text as references to improve question answering. Importantly, although causal graphs provide a theoretically grounded, valuable source of information, they alone rarely yield a complete, interpretable rule. A DTC may
be discovered as a strong causal parent of an EP, but paired with the textual
metadata can reveal whether the DTC reflects a root cause, a downstream
effect, or an irrelevant side-symptom. LLMs offer a practical
computational mechanism for integrating these heterogeneous evidence
sources, resolving contradictions, and translating causal signals into
symbolic form through the Boolean expression of an EP.

Nevertheless, a single model is often insufficient to handle the diverse subtasks
involved in automation.
The growing paradigm of \textit{agentic systems}~\cite{agents, zhao2025agenticrarediseasediagnosis, biomedicalagent} highlights how
specialized agents, each with distinct capabilities, can collaborate
through structured dialogues or planning loops. This opens the door to human-interpretable, automated causal reasoning in complex industrial pipelines \cite{2024arXiv240715073D}. For example, often referred to as \gls{cml}, it offers a more nuanced and interpretable approach to black box models and has found interesting applications \cite{GUPTA2025100116}.
That is, for LLMs, as noted by \cite{mirzadeh2025gsmsymbolic}, they are sensitive and fragile without semantic or neuro-symbolic knowledge. For tasks requiring the selection of correct tokens, accuracy decreases exponentially with the number of tokens, making them unsuitable for high-cardinality event sequences alone.

\section{Related Work}

\subsection{Multi-Agent Fault Diagnosis}
The application of multi-agent systems for industrial diagnostics and failure analysis has become a prominent research area. \cite{8283595} established an early framework using an agent-based inference engine to provide efficient and reliable assistance for vehicle failure diagnosis. In the domain of autonomous safety, \cite{autonomous_collision_problems} proposed an explainable system designed to interpret complex vehicle accidents, emphasizing the need for transparent decision-making. More recently, the scope of agentic systems has expanded to the manufacturing sector. \cite{agentic_production_line} introduced an agentic AI framework for "self-healing" production lines capable of autonomous root cause analysis and fault correction.
While these works share the goal of automating diagnostics through agents, CAREP distinguishes itself by addressing four critical challenges:
\begin{itemize}
    \item Extreme Cardinality: CAREP is specifically designed for large-scale event streams containing over 29,100 unique DTC types, a scale where traditional inference engines, causal inference and discovery typically struggle.
    \item Causal Integration: It integrates a dedicated causal discovery agent (utilizing OSCAR and CARGO) to filter irrelevant events before reasoning begins.
    \item Metadata Integration: Through embedding models, the metadata is integrated as references and used by a contextual information agent.
    \item Symbolic Synthesis: Unlike systems that provide static alerts, CAREP automates the synthesis of symbolic Boolean rules (Error Patterns) alongside natural language reasoning traces, effectively translating raw statistical evidence into actionable human-readable expertise.
\end{itemize}

\noindent We deliberately adopt a multi-agent design rather than a single monolithic pipeline for three reasons. First, \emph{separation of concerns}: the causal discovery agent and the contextual information agent draw on structurally different evidence, numerical causal graphs versus unstructured text, and forcing a single LLM call to parse both simultaneously empirically increases hallucination. Second, \emph{modularity}: each agent can be upgraded independently, for instance replacing OSCAR/CARGO with the unified causal graph discussed in the outlook of this chapter, or the retrieval corpus with a domain-adapted knowledge base, without retraining the orchestrator. Third, \emph{traceability}: the orchestrator agent must explicitly cite which piece of evidence (a causal indicator, a co-occurrence statistic, or a retrieved document) supports each element of the synthesized rule. Routing all evidence through this agent therefore yields an auditable reasoning trace that a single end-to-end model would not naturally expose.

\subsection{Error Pattern Rule Automation}
\noindent
The creation of EPs has emerged as production and vehicle manufacturing have become increasingly complex. Fig.~\ref{fig:illustration} illustrates the manual process: (a) a vehicle produces a sequence of DTCs over time (e.g., \texttt{0x1283F}, \texttt{0x09931}, \texttt{0x84324}); (b) an expert analyzes them together with the vehicle's metadata; and (c) expresses the defect as a Boolean rule, here \(\text{EP} = \texttt{0x1283F} \,|\, \texttt{0x84324}\), i.e., the EP is detected if either of the two DTCs occurs, while \texttt{0x09931} is irrelevant.
As demonstrated in the foundations, to identify new defects, experts analyze the diagnostic trouble code (DTC) sequence for each vehicle. Intuitively, this reasoning task is not trivial. Experts must draw on their prior observations and knowledge of specific DTCs that may cause each EP.
That is where the Boolean operators are useful. They allow us to construct different rules to characterize precise EPs. It sometimes requires the absence of certain DTCs (\(!\) NOT) to separate overlapping EPs. Or the presence of multiple DTCs at the same time (\(\&\) AND) or a particular culprit (\(|\) OR). Therefore, the different operators \((\&, |, !)\) involve different statistical perspectives. Intuitively, the \(!\) operator measures inhibitory strength \cite{measure_of_causal_strength_oxford}, where the likelihood of observing a label \(y_1\) is lowered if an event \(x_{10}\) is observed. On the other hand, \(\&\) and \(|\) are more associated with excitatory events that will raise the likelihood of an EP. Knowing only the causes of an EP is not sufficient to elaborate a new rule. It is necessary to have indicators that quantify how DTCs are related to one another and to EPs.
Finally, rules are subject to change and are dynamically updated by a domain expert based on the new incoming data. As a result, automating a rule is a complex, dynamic problem.

\section{Methodology}
\noindent Let \(\mathcal{D} = \{S_l^1, \cdots, S_l^m\}\) be a dataset of multi-labeled sequences. Each label \(y_j\) is defined as a Boolean rule between some events \(\{x_i\}_i, x \in \mathcal{X}\). Such that if in a sequence \(S_l^{(k)}\), the Boolean rule is true for some label \(y_j\), then it is present in \(S_l^{(k)}\).
\(\mathcal{D}\) contains multiple labels with unknown Boolean rules denoted as \(y^{un} \in \mathcal{Y}\).
We aim to find the unknown rule of each label \(y^{un}\) using the observed sequences in \(D\).

\noindent The general architecture of CAREP is shown in Fig.~\ref{fig:carep}. The agentic system outputs a reasoning explanation for each estimated error pattern rule. To increase prediction's plurality, CAREP outputs \(5\) rules per EP with different levels of confidence (high, medium, low) that are reflected in the provided explanation.
Each agent receives a system prompt that specifies their task and uses CoT \cite{chainofthoughts}. The orchestrator is responsible for structuring comprehensive, traceable reasoning for each rule based on the other agents' outputs. We now examine each component separately.

\begin{figure*}[!t]
    \centering
    \resizebox{\textwidth}{!}{\providecommand{\carepFn}{\fontsize{8}{9.5}\selectfont}
\providecommand{\carepFs}{\fontsize{7}{8.3}\selectfont}
\providecommand{\carepFl}{\fontsize{9}{11}\selectfont}
\begin{tikzpicture}[x=1cm, y=1cm, -, font=\carepFn]
  \tikzset{
    panel/.style={draw=gray!60, rounded corners=6pt, thick},
    inbox/.style={rounded corners=4pt, thick, align=left, text width=#1, inner sep=4pt},
    mdl/.style={draw=blue!55!black, fill=blue!10, rounded corners=4pt, align=center, thick},
    cmp/.style={draw=green!45!black, fill=green!10, rounded corners=4pt, align=center, thick},
    agg/.style={draw=orange!70!black, fill=yellow!15, rounded corners=4pt, align=center, thick},
    agent/.style={rounded corners=5pt, thick, align=center, minimum height=1.35cm, font=\carepFl\bfseries},
    arr/.style={-{Stealth[length=1.8mm]}, thick},
    big/.style={-{Stealth[length=2.6mm, width=3.2mm]}, line width=2.2pt, gray!60},
    link/.style={{Stealth[length=1.8mm]}-{Stealth[length=1.8mm]}, thick, double distance=1.2pt},
    ttl/.style={font=\carepFl\bfseries, anchor=west},
    sn/.style={circle, draw, thick, minimum size=5.2mm, inner sep=0pt, fill=white, font=\carepFs},
  }

  \draw[panel, fill=gray!5] (0,5.7) rectangle (5.0,10.6);
  \node[ttl] at (0.15,10.3) {Inputs};
  \node[inbox=4.5cm, draw=orange!70!black, fill=white, anchor=north west] (main) at (0.25,9.95) {%
    \textbf{Sequences and models}\\[1pt]
    $\bullet$ $m$ DTC sequences of the\\\hphantom{$\bullet$ }unknown EP $y^{un}$\\
    $\bullet$ $\text{Tf}_x$ (CarFormer) and\\\hphantom{$\bullet$ }$\text{Tf}_y$ (EPredictor)};
  \node[inbox=4.5cm, draw=green!45!black, fill=white, anchor=north west] (ctx) at (0.25,7.85) {%
    \textbf{Context documents}\\[1pt]
    $\bullet$ DTC descriptions\\
    $\bullet$ Rules of the known EPs\\
    $\bullet$ Check control messages (CCMs)\\
    $\bullet$ Metadata (e.g., model range)};

  \draw[panel, fill=orange!3] (5.6,5.7) rectangle (16.4,10.6);
  \node[ttl] at (5.75,10.3) {(a) Causal discovery (Chapters 6 and 7)};
  \draw[big] (5.05,9.2) -- (5.85,9.2);
  \foreach \y/\lab in {9.55/x_0, 9.0/x_1, 8.05/x_{27}, 7.5/x_{51}}{
    \node[anchor=east, font=\carepFs] at (6.55,\y) {$\lab$};
    \draw[arr] (6.6,\y) -- (7.15,\y);
  }
  \node[font=\carepFs] at (6.3,8.6) {$\vdots$};
  \node[font=\carepFs] at (6.3,7.05) {$\vdots$};
  \node[font=\carepFs, align=center] at (6.5,10.0) {};
  \node[cmp, minimum width=2.3cm, minimum height=2.4cm] (osc) at (8.45,8.5) {\textbf{OSCAR}\\sample-level\\causal discovery\\on each sequence};
  \draw[big] (9.65,8.5) -- (10.25,8.5);
  \begin{scope}[shift={(10.45,8.85)}]
    \node[font=\carepFs] at (0.55,0.85) {$\mathcal{G}^{(1)}, \ldots, \mathcal{G}^{(m)}$};
    \node[sn] (g1) at (0,0.2) {$x_1$}; \node[sn] (g2) at (1.1,0.2) {$x_{27}$}; \node[sn, fill=red!12] (gy) at (0.55,-0.75) {$y^{un}$};
    \draw[arr] (g1) -- (gy); \draw[arr] (g2) -- (gy);
  \end{scope}
  \draw[big] (11.85,8.5) -- (12.45,8.5);
  \node[agg, minimum width=1.8cm, minimum height=2.4cm] (cargo) at (13.1,8.5) {\textbf{CARGO}\\population-\\level fusion\\into $\mathcal{G}^*_{un}$};
  \draw[big] (14.05,8.5) -- (14.4,8.5);
  \node[inbox=1.6cm, draw=orange!70!black, fill=white, font=\carepFs, anchor=west] (ind) at (14.45,8.5) {%
    \textbf{Causal}\\\textbf{indicators}\\[1pt]
    frequency $\hat{\pi}$\\
    ACE\\$\hat{\mathcal{C}}_\mu \pm \hat{\mathcal{C}}_\sigma$\\
    PMI};

  \node[agent, draw=green!45!black, fill=green!10, minimum width=3.6cm] (ctxag) at (4.9,4.45) {(b) Contextual\\information agent};
  \node[agent, draw=blue!55!black, fill=blue!10, minimum width=3.2cm] (orch) at (9.0,4.45) {(c) Orchestrator\\agent};
  \node[agent, draw=orange!70!black, fill=yellow!15, minimum width=3.6cm] (cdag) at (13.3,4.45) {(a) Causal\\discovery agent};
  \node[cmp, font=\carepFs, minimum width=2.1cm, minimum height=0.9cm] (emb) at (1.35,4.45) {Retrieval with\\Titan V2 embeddings};
  \draw[arr] (1.35,5.7) -- (emb.north);
  \draw[arr] (emb) -- (ctxag);
  \draw[link] (ctxag) -- (orch);
  \draw[link] (orch) -- (cdag);
  \draw[arr] (ind.south) |- (cdag.east);
  \draw[arr] (3.6,5.7) |- ([yshift=0.3cm]orch.north) -- (orch.north)
     node[pos=0.25, above, font=\carepFs, fill=white, inner sep=1pt] {10 sample sequences and metadata, in the prompt};

  \node[ttl] at (0.15,3.05) {Outputs};
  \draw[big] (orch.south) -- (9.0,2.95);
  \node[inbox=3.75cm, draw=blue!55!black, fill=blue!5, anchor=north west, font=\carepFs] (refs) at (0.25,2.3) {%
    \textbf{\carepFn References}\\[3pt]
    {[1]} \texttt{dtc\_descriptions.csv}\\
    {[2]} \texttt{ccms\_description.csv}\\
    {[3]} \texttt{ep\_known\_descriptions.csv}};
  \node[inbox=11.55cm, draw=blue!55!black, fill=blue!5, anchor=north west, font=\carepFs, line width=1.2pt] (reas) at (4.5,2.9) {%
    \textbf{\carepFn Candidate rules with traceable reasoning}\\[3pt]
    Looking at the data, I notice several key DTCs causing this error pattern. \textbf{I propose the following rules:}\\[2pt]
    $\boldsymbol{x_1 \,\&\, x_{27}}$: the most frequent DTCs in the samples. They co-occur (PMI $>0$) in most samples triggering CCM~3~\textcolor{blue!60!black}{[2]}. Their high causal indicators (ACE $0.45$ and $0.30$) support this rule.\\[2pt]
    $\boldsymbol{x_1 \,|\, x_{27} \,|\, x_{57}}$: given the frequencies and causal indicators, either $x_1$, $x_{27}$, or $x_{57}$ suffices to trigger the EP. This aligns with the known EP \emph{Fuel level sensors do not reach the threshold}~\textcolor{blue!60!black}{[3]}.\\[2pt]
    $\boldsymbol{x_1 \,|\, x_{80}}$: the function of $x_{80}$ combined with $x_1$ creates a pattern similar to known EPs involving fuel problems. The PMI of $x_{80}$ and $x_1$ is strong ($1.86$), but this rule has medium confidence due to the lower frequency of these DTCs.\\[3pt]
    From the sample sequences, I observe that CCM~3~\textcolor{blue!60!black}{[2]} is consistently triggered across all samples. In the retrieved documents, several of these DTCs are related to the fuel system: $x_{80}$ and $x_1$ are likely related to fuel tank issues~\textcolor{blue!60!black}{[1]}. I also found a known EP that seems relevant: \emph{Fuel level sensor in the left tank does not reach the empty threshold}, with logic $x_1 \,\&\, (x_{27} \,|\, x_{80})$, which is related to fuel level sensors~\textcolor{blue!60!black}{[3]}. The negative PMI values between many of the DTCs suggest that they are alternative paths to the same error condition rather than co-occurring conditions.};
  \draw[thick, double distance=1.2pt] (refs.east) -- (refs.east -| reas.west);
  \draw[dashed, thick, gray!70] (0,3.35) rectangle ([yshift=-0.25cm]reas.south -| 16.4,0);
\end{tikzpicture}}
    \caption{\textbf{CAREP: A Causal Reasoning Agentic System for Error Pattern Automation}. (a) Causal discovery: OSCAR (Chapter~\ref{c7:multi_label_one_shot_causal_discovery}) extracts one sample-level graph from each sequence of the unknown EP, CARGO (Chapter~\ref{c7:multi_label_causal_discovery}) fuses them into the population-level graph \(\mathcal{G}^*_{un}\), and the causal indicators of its candidate causes (frequency, ACE, PMI) are passed to the causal discovery agent. (b) The context documents are retrieved with Titan V2 embeddings by the contextual information agent. (c) The orchestrator agent combines both, together with ten sample sequences and the metadata given in the prompt, and outputs candidate Boolean rules with a traceable reasoning that cites the retrieved documents [1]--[3].}
    \label{fig:carep}
\end{figure*}
\subsection{Causal Discovery}\label{c8:sec:cd}

\paragraph{Population-Level Causal Graphs}
We explicitly model the unknown error patterns \(y^{un}\) as the target in the autoregressive classifier \(\text{Tf}_y\)'s (EPredictor) output logits\footnote{The training labels are used but the Boolean rule is generated at inference time.}. This forms a supervised unknown class learning strategy and enables us to extract the posteriors \(P_{\theta_y}(Y = y^{un}|.)\).
To provide new causes for each unknown EPs, we employ OSCAR (\(k=2.75, N=64\)) and CARGO to extract causal relationships from the dataset \(\mathcal{D}\). As a result, we obtain one population-level graph \(\mathcal{G}^*_j\) for each unknown error pattern \(y_j \in \mathcal{Y}^{un}\), where \(\mathcal{Y}^{un} \subset \mathcal{Y}\) denotes the set of EPs whose rules are unknown; each graph contains the aggregated causal relationships of its EP, alongside causal indicators.

\paragraph{Average Causal Effect}
\noindent We reuse the causal indicator of the sample-level phase (OSCAR, Eq.~\eqref{eq:causal_indic}) as \emph{ACE} (Average Causal Effect). This indicates, on average, how much the DTC increased or decreased the likelihood of an error pattern. We report its expected value as the mean and its standard deviation over the sampled trajectories \(x^{(l)}_{<t}\) used in the Monte Carlo estimation of Eq.~\eqref{eq:cmi_approx}:
\begin{align}
    \Delta^{(l)}_{t,j} &= P(Y_j = 1 \mid X_t = x_t, X_{<t} = x^{(l)}_{<t}) - P(Y_j = 1 \mid X_{<t} = x^{(l)}_{<t}) \notag \\
    \hat{\mathcal{C}}_\mu (Y_j, X_t) &\triangleq \frac{1}{N}\sum^N_{l=1} \Delta^{(l)}_{t,j} \label{eq:ace_mean} \\
     \hat{\mathcal{C}}_\sigma(Y_j, X_t) &\triangleq \sqrt{\frac{1}{N}\sum^N_{l=1}\big(\Delta^{(l)}_{t,j} - \hat{\mathcal{C}}_\mu(Y_j, X_t)\big)^2} \label{eq:ace_std}
\end{align}

\paragraph{Co-occurrence}\label{sec:co_occurence}
\noindent The different Boolean operators in the defined EP rules in Eq.~\eqref{eq:ep_def} imply different statistical perspectives. In the previous section, we captured the elements in the rule, i.e., the set of DTC causes for a given label. However, we should also provide co-occurrence estimates of event pairs to properly address the OR, AND, and NOT operators.
Traditionally, Point-wise Mutual Information (PMI) \cite{cover1999elements} is used to measure co-occurrence strength. It is defined as:
\begin{equation}
    pmi(x_i, x_j) = \log{\frac{p(x_i,x_j)}{p(x_i)p(x_j)}}
\end{equation}
where \(p(x_i), p(x_j)\) are probability mass functions.

\noindent Along with the previous empirical frequency of event \(x_i\) in the set of causes of label \(y_j\), denoted as \(\hat{\pi}_{i,j}\), we also compute empirical joint frequencies of pairs of events within the same label-specific graph \(\mathcal{G}^*_j\).
Formally, given a collection of DAGs \(\{(\mathbf{V}_k, E_k)\}_{k=1}^m\), the marginal and joint empirical probabilities are estimated as
\[
\hat{p}(x_i \mid y_j) = \frac{1}{m_j} \sum_{k=1}^m \mathds{1}\{X_i \in \mathbf{V}_k, \, Y_j \in \mathbf{V}_k\},
\]
\[
\hat{p}(x_i, x_\ell \mid y_j) = \frac{1}{m_j} \sum_{k=1}^m \mathds{1}\{X_i \in \mathbf{V}_k, \, X_\ell \in \mathbf{V}_k, \, Y_j \in \mathbf{V}_k\},
\]
where \(m_j\) denotes the number of samples in \(\mathcal{D}\) where label \(Y_j\) is present, and \(\mathds{1}\{\cdot\}\) is the indicator function.

\noindent The PMI between two events \(x_i\) and \(x_\ell\) conditioned on a label \(y_j\) is then given by:
\begin{equation}\label{eq:pmi}
    \widehat{pmi}(x_i, x_\ell \mid y_j)
= \log \frac{\hat{p}(x_i, x_\ell \mid y_j)}{\hat{p}(x_i \mid y_j)\,\hat{p}(x_\ell \mid y_j)}.
\end{equation}
\noindent Intuitively, \(\widehat{pmi}(x_i, x_\ell \mid y_j) > 0\) indicates that events \(x_i\) and \(x_\ell\) co-occur more often than expected under independence, while a negative value suggests mutual exclusivity.

\subsection{Contextual Information}
\noindent We add to each explanation of \(y^{un}\) contextual information about the description of DTCs, EPs, and the known error pattern rules. Concretely, the retrieval corpus consists of the textual descriptions of all DTCs and of the Boolean rules of every EP that is \emph{not} masked in the current evaluation fold (Section~\ref{c8:sec:settings}). These are the same expert-authored rules introduced in Section~\ref{sec:ep_def}, whose provenance is further documented in Appendix~\ref{appendix:bmw_confidentiality}. This design mirrors how a human diagnostic engineer would consult the existing rule catalog when characterizing a new fault. These descriptions are processed into embedding vectors using a \textit{Titan V2 Embedding}\footnote{https://aws.amazon.com/de/blogs/aws/amazon-titan-text-v2-now-available-in-amazon-bedrock-optimized-for-improving-rag/} from AWS Bedrock. They are then used as a Retrieval-Augmented Generation (RAG) \cite{rag_or} system, in which the contextual information agent queries are matched against the description embeddings generated by the embedding model. It is incorporated into the context of the contextual information and orchestrator agents.

Now, directly in the prompt (\textit{in-context learning} \cite{icl}), we randomly inject 10 DTC sequences that exhibit the same unknown EP we aim to identify. As well as metadata, such as the vehicle model range and the triggered message printed on the board (check control message: CCM). The overall data inputs are shown in Fig.~\ref{fig:carep}, and the JSON input to the orchestrator is shown in Appendix.~\ref{verbatim:output_json}.

\section{Experiments}

\subsection{Evaluation Metrics}
\noindent This task requires multiple levels of evaluation. Suppose that we have an automated method to extract a Boolean rule \(R^*\) for each unknown EP \(y^{un}\) in a batch of sequences. We then want to evaluate how good this generated rule is compared to the ground truth.

Hence, we distinguish between two evaluation types: (1) \textit{structural}, where we compare the Boolean rules directly as a classification set, and (2) \textit{semantic}, where we compare the corresponding truth-tables of the estimated Boolean expressions to the truth-table of the ground truth using the Sympy package \cite{sympy}.

\subsubsection{Structural Evaluation}
\noindent Specifically, for (1) we evaluate: \textit{Are the DTCs in the estimated rules correct?}
For this, we divide the estimated sets and ground truth using the Boolean operators as separators and perform a standard multi-label classification:
    \[\text{dtc1 \& dtc2 \& !dtc5} \;|\; \text{dtc3} \Longrightarrow [dtc1, dtc2, dtc3, dtc5]\]
This evaluation is the same as the one described in Part~\ref{pa:p2} for the causal discovery algorithms recovering Markov Boundaries, i.e., sets of causes of an EP.

\subsubsection{Semantic Evaluation}

\noindent For (2) we enumerate all possible assignments of the present Boolean variables and compute the truth table of the estimated rules and the ground truth to express: \textit{Is the rule logically correct?} We calculate the accuracy, precision, recall, F1 of the predicted value of the truth tables (e.g., Tab~\ref{tab:truth_table_example}).

\begin{table}[!h]
\centering
\begin{tabular}{|c|c|c|c|}
\hline
\textbf{$x_1$} & \textbf{$x_2$} & \textbf{Ground Truth ($x_1 \; \& \; x_2$)} & \textbf{Predicted Rule ($x_1 \;|\; x_2$)} \\
\hline
0 & 0 & 0 & \textcolor{green}{0} \\
0 & 1 & 0 & \textcolor{red}{1} \\
1 & 0 & 0 & \textcolor{red}{1} \\
1 & 1 & 1 & \textcolor{green}{1} \\
\hline
\end{tabular}
\caption{Illustrative truth table for semantic evaluation. The accuracy using the predicted rule is \(50\%\).}
\label{tab:truth_table_example}
\end{table}

\subsection{Settings}\label{c8:sec:settings}
\begin{figure}[!ht]
    \centering
    \includegraphics[width=0.95\linewidth]{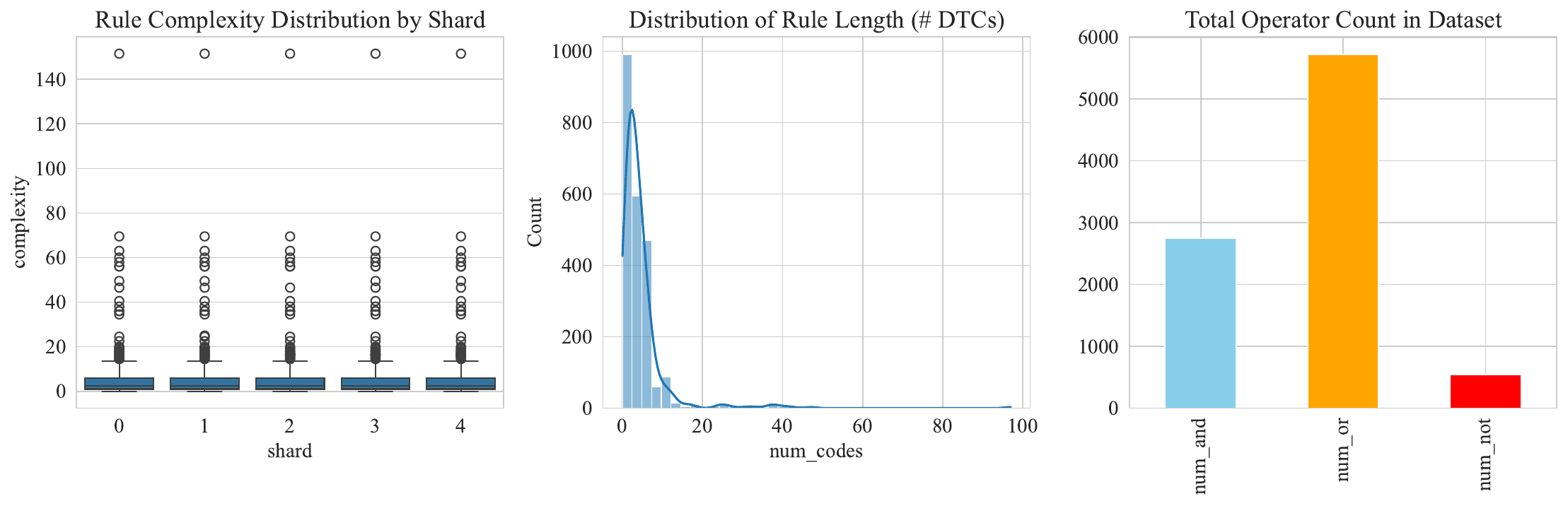}
    \caption{\textbf{Analysis of Ground Truth Error Patterns.}
(Left) \textbf{Complexity Distribution by Fold:} The consistency of rule complexity scores across fold confirms a balanced data split strategy.
(Middle) \textbf{Rule Length Histogram:} The distribution of Diagnostic Trouble Codes (DTCs) per rule reveals a long-tail nature, with some complex rules involving over 20 unique codes.
(Right) \textbf{Operator Frequency:} The significant presence of logical NOT ($\neg$) and OR ($\vee$) operators highlights the non-trivial, inhibitory causal structures.}
\label{fig:complexity_analysis}
\end{figure}

\noindent We used the same experimental setup and dataset as in Chapters~\ref{c7:multi_label_one_shot_causal_discovery} and \ref{c7:multi_label_causal_discovery}. Fig.~\ref{fig:complexity_analysis} characterizes the ground-truth rules: their complexity is distributed in the same way in all five folds (left), most rules involve fewer than ten DTCs but a few involve several dozen (middle), and OR is the most frequent operator, about twice as frequent as AND, while NOT is the rarest (right). All ground-truth EP rules used in this evaluation were authored and are maintained by BMW's diagnostic and quality engineering teams as part of their standard error-pattern definition process (Appendix~\ref{appendix:bmw_confidentiality}); we treat them as-is and do not modify or re-validate their content. We created 5 folds of \(50,000\) sequences and randomly masked \(20\) different EPs with at least \(n \geq 100\) sequences. We set their masked rule as the ground truth for each label \(y^{un}_j\) and average the results across the five folds. We used a combination of F1-Score, Precision, and Recall with different averaging methods \cite{reviewmultilabellearning} to compare with LLM-only baselines.

\subsection{Comparison}
We compared CAREP against multiple LLMs, such as Claude Sonnet 3.5 and 3.7\footnote{https://www.anthropic.com/news/claude-3-7-sonnet}, GPT4.1 and GPT4.1 mini\footnote{https://openai.com/index/gpt-4-1/}. We further investigated whether smaller LLMs \cite{belcak2025smalllanguagemodelsfuture} with fewer parameters impacted the evaluation.
The LLMs are compared using only the observed DTC sequences and their descriptions, hence in Fig.~\ref{fig:carep} the orange region is removed.
We also added CAREP without the causal indicators. We then performed the structural and semantic evaluation of the generated Boolean rules presented in the foundation chapter. The top-\(3\) and top-\(5\) metrics are computed from the five estimates, ordered by decreasing confidence (high, medium, low), to provide a clearer picture. We average the metrics across labels (macro average).

\subsection{Results}
\begin{figure*}[!t]
    \centering
    \includegraphics[width=1\textwidth]{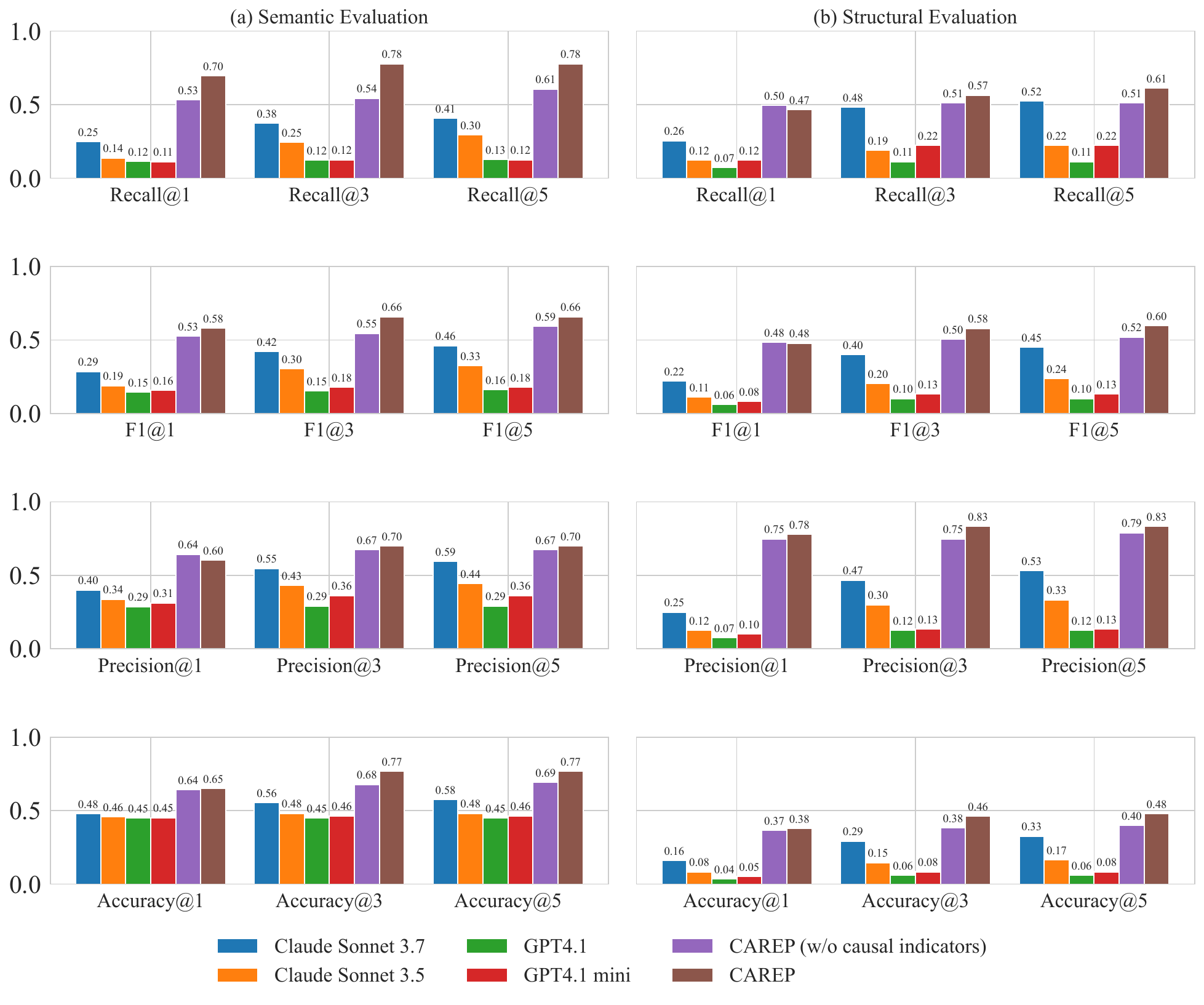} %
    \caption{\textbf{Performance Comparison of CAREP Against Standalone LLMs}. (a) Semantic evaluation represents how well the estimated error pattern rules fit the ground truth in terms of Boolean expression. (b) Structural evaluation or multi-label classification reveals if the estimated rules contain the correct DTCs.}
    \label{fig:carep_eval_grid}
\end{figure*}
\noindent
Figure~\ref{fig:carep_eval_grid} reports the performance of our method CAREP against multiple LLM baselines (Claude Sonnet 3.5/3.7, GPT4.1, and GPT4.1 mini). CAREP consistently outperforms all standalone LLMs across both evaluation protocols. The ground-truth EP rules used for evaluation were manually defined by diagnostic experts and may contain noise or inconsistencies. Results should be interpreted accordingly.

\paragraph{Semantic evaluation}
In terms of truth-table agreement, CAREP substantially improves over LLMs in capturing the logical structure of error patterns. CAREP achieves a \emph{Recall@1} of $0.70$, compared to $0.25$ for the best-performing baseline (Claude Sonnet 3.7). Precision and F1 scores follow the same trend, showing that CAREP’s causal discovery step yields more faithful Boolean rules rather than over-generalized expressions. Interestingly, LLMs still reach $\approx 0.50$ semantic \emph{Accuracy@5}, indicating that DTCs' observations and descriptions alone allow them to approximate the ground truth partially. However, they lack the consistency and reliability required for deployment that CAREP provides.

\begin{table}[!h]
\centering
\resizebox{\columnwidth}{!}{%
\begin{tabular}{lcccc}
\hline
\textbf{Method} & \textbf{Precision} $\uparrow$ & \textbf{Recall} $\uparrow$ & \textbf{F1} $\uparrow$ & \textbf{Time (min)} $\downarrow$ \\ \hline
\multicolumn{5}{l}{\textit{Phase 1: Sample-Level Causal Discovery (OSCAR)}} \\
Sample-Level Graph \(\mathcal{G}^{(k)}\) & $55 \pm 1.4$ & $31 \pm 0.8$ & $40 \pm 1.0$ & $\mathbf{11.7}$ \\ \hline
\multicolumn{5}{l}{\textit{Phase 2: Population-Level Causal Discovery (CARGO)}} \\
Consensus Graph \(\mathcal{G}^*_y\) & $61 \pm 1.5$ & $46 \pm 1.7$ & $46 \pm 1.2$ & $11.8$ \\ \hline
\multicolumn{5}{l}{\textit{Phase 3: CAREP (End-to-End)}} \\
\textbf{CAREP} & $\mathbf{83 \pm 1.1}$ & $\mathbf{60 \pm 1.3}$ & $\mathbf{61 \pm 1.1}$ & $12$ \\ \hline
\end{tabular}
}
\caption{\textbf{Ablation of causal discovery performance across the pipeline stages}: (1) Local Discovery, (2) Global Aggregation, and (3) The complete framework (CAREP). The baselines are evaluated using the structural (does the rule contain the correct variables, ignoring the logic). 'Time' represents cumulative runtime. By adding semantic information (metadata) and LLMs' reasoning, CAREP adds a significant prediction margin in selecting the correct \textit{causes} of an outcome (EP).}
\label{tab:pipeline_performance}
\end{table}

\paragraph{Structural evaluation}
When evaluating whether the predicted rules include the correct DTCs, CAREP's advantage becomes even more pronounced. For \emph{Precision@1}, CAREP achieves $0.78$ versus only $0.25$ for Claude Sonnet 3.7 and $0.05$ for GPT4.1 mini. Similar margins are observed in F1 and accuracy. GPT4.1 and mini, in particular, perform poorly (\emph{F1@1} $< 0.1$), demonstrating that without causal discovery, LLMs fail to correctly identify the true DTCs present in the error pattern rules. In addition, unlike OSCAR and CARGO, which were previously evaluated using only the structural evaluation, CAREP benefits from the contextual information and from multiple output Boolean rules. It obtains \(75\%\)  \emph{Precision@1} and \(83\%\) \emph{Precision@5}, compared with \(60\%\) \emph{Precision} (Table~\ref{tab:performance_comparison_plus_agg}) for CARGO and \(55\%\) for OSCAR.

\section{Summary}
\noindent
We introduced CAREP, a multi-agent causal-reasoning framework that automates the discovery of error pattern rules from large-scale event sequences of error codes. By combining causal discovery, RAG systems, and agents, CAREP consistently outperforms state-of-the-art LLM baselines while producing interpretable reasoning traces essential for safety-related deployment. The ablation across the three pipeline stages (Table~\ref{tab:pipeline_performance}) confirms that neither causal structure alone nor
language reasoning alone is sufficient: the performance gap between CARGO and CAREP
demonstrates that contextual grounding is indispensable for correct Boolean rule
synthesis. We emphasize, however, that CAREP is designed as an assistive tool rather than a fully autonomous decision-maker. We further note a methodological limitation of our evaluation: precision, recall, and F1 measure only the \emph{structural} and \emph{logical} agreement between a generated rule and its expert-authored ground truth. They do not assess whether a domain expert would find the generated rule, together with its natural-language explanation, genuinely understandable or actionable in practice. Establishing this form of human-centered interpretability would require a dedicated user study with certified diagnostic engineers, which we leave to future work. Given the high-stakes nature of automotive diagnostics, unsupervised rule deployment carries inherent risks; therefore, human experts must remain in the loop to validate and supervise the system's outputs. The methods and findings of this chapter therefore conclude the full pipeline of this thesis, from raw event-sequence modeling through causal structure recovery to symbolic rule generation. The question of how to scale, adapt, and generalize this pipeline remains the central open problem motivating the directions below.
\section{Outlook}

\noindent
Several directions offer high potential for extending CAREP toward a more general and
robust automated reasoning system. A first and most immediate direction is to replace
the events-to-outcome causal backbone (OSCAR and CARGO) with a \emph{unified
causal graph} that jointly encodes event-to-event and events-to-outcome dependencies, the
open cell of Table~\ref{tab:causal_framework}. Such a graph would allow CAREP to
reason not only about which DTCs cause a given error pattern, but also about the
cascading sequence of fault propagation leading to it. Incorporating event-to-event
causal edges from TRACE into CAREP's causal discovery agent is a natural first step. The richer structural evidence would enable the orchestrator agent to synthesize
temporally ordered Boolean rules, capturing not just the presence but the sequential
ordering of causes, a property that current EP rules do not yet cover. Scaling this to the
full population-level setting, where a unified DAG over both event types and outcome
labels is aggregated across a fleet, constitutes the primary research goal of the
\textit{seq2cause} project~\cite{math2026seq2cause}.

A second direction concerns the language backbone of the reasoning agents. The current CAREP implementation relies on general-purpose LLMs that lack the specialized
knowledge of automotive subsystems, DTC taxonomy, descriptions and diverse technical jargon. Domain-adapted language models, pretrained or
continuously fine-tuned on automotive documentation, service manuals, and
fault-knowledge bases would reduce hallucination rates and improve the semantic
fidelity of generated rules, particularly for rare error patterns where contextual
retrieval alone is insufficient.

\part*{}
\addtocontents{toc}{\protect\bigskip}
\chapter{Conclusion and Outlook}\label{c7:conclusion_n_outlook}
\glsresetall
This thesis demonstrated that high-cardinality discrete event sequences, long treated as intractable by statistical models and causal discovery methods, can be systematically modeled, causally analyzed, and transformed into interpretable diagnostic knowledge using a unified pipeline of autoregressive Transformers, information-theoretic causal tests, and multi-agent reasoning.
\section{Conclusion}
This thesis aimed to address one of the core challenges in modern industrial systems: transforming massive, high-cardinality event streams into interpretable and actionable knowledge for practitioners and researchers. By focusing on the automotive domain, a setting that epitomizes complexity, scale, and safety-critical constraints, we progressively built a pipeline that evolves from sequence modeling to causal discovery and, finally, to automated reasoning for error pattern automation.

In Part~\ref{part1}, we demonstrated that event sequences, such as diagnostic trouble codes (DTCs), can be modeled as a language. By introducing CarFormer and EPredictor, we showed that Transformer-based architectures can accurately learn spatio-temporal dependencies in irregular, unbalanced event streams and forecast both when and what failures are likely to occur. We also showed that incorporating additional modalities, such as sensor readings, improves predictive performance with BiCarFormer, a bidirectional Transformer model. This laid the foundation for predictive maintenance systems capable of real-time, self-diagnostic behavior.

In Part II, we moved beyond prediction to understanding. We introduced three complementary frameworks for causal discovery in sequences: OSCAR and TRACE, which perform sample-level causal discovery using autoregressive Transformers as density estimators, and CARGO, which aggregates these sample-level graphs into coherent global causal graphs. A key technical insight underpinning this is the functional duality between prediction and causal discovery: an autoregressive model trained to predict the next event implicitly learns the joint distribution of the sequence. It can therefore be directly reused, without any fine-tuning, as a neural density estimator for conditional mutual information estimation. Critically, the teacher-forcing mechanism of sequence-to-sequence training (i.e., conditioning on observed rather than model-generated trajectory) allows all \(N\) conditional log-probabilities required for a CMI estimate to be computed in a single forward pass. This collapses what would otherwise be \(N\) sequential inference calls into a single vectorized operation and makes the approach tractable on the scale of tens of thousands of distinct events.
Together, they enable, for the first time, causal discovery in sequences with a considerable number of distinct events and labels, exceeding tens of thousands of DTCs and hundreds of error patterns in our experiments. These data settings have not been explored in the literature and remain a major obstacle to the adoption of causal discovery in large-scale datasets. These three methods are anchored in modern parallelization of algorithms on GPUs, where Transformers and more generally autoregressive models are the fundamental starting point. The methods are implemented within common frameworks such as Hugging Face and PyTorch. This results in three practical causal discovery methods that practitioners can deploy alongside their event sequence model, and scale its infrastructure using only GPUs. In addition, they are theoretically grounded in causal discovery and probability theory. We proved both experimentally and theoretically that a perfect oracle model is not required to recover the causal graph and provided the soundness of our approaches under several assumptions. We analyzed the limitations of such assumptions and proposed robust methods, such as adaptive thresholding for aggregating graphs with imbalanced label distributions.

Finally, in Part~\ref{pa:p3}, we transitioned toward automating error patterns in vehicles, a complex reasoning task. We highlighted how we could build an entire framework powered by efficient causal discovery, contextual information retrieval, and LLMs to reason over dense, multimodal inputs. We show that a multi-agent-based system (\gls{carep}) can generate plausible rules that are highly similar to those created by domain experts. Beyond their direct automotive applications, the methods proposed in this dissertation have broader implications. The combination of autoregressive modeling, scalable causal inference, and multi-agent reasoning opens new pathways for domains such as healthcare, cybersecurity, and industrial process monitoring, wherever complex event-driven systems demand reliable and explainable decision support.

It is important to acknowledge the assumptions and limitations that bound these contributions. The causal discovery frameworks developed in Part II rely on causal sufficiency, the absence of hidden confounders, which may not hold in all industrial deployments where unmeasured variables influence observed events. The evaluation of CAREP in Part III is conducted against EP rules that were themselves constructed by domain experts; while this provides a practical benchmark, it introduces a degree of circularity that future work should address through independent validation. Finally, all experiments are conducted on proprietary BMW data, which, while ensuring industrial relevance, limits direct reproducibility; the open-source seq2cause library and accompanying synthetic datasets are provided to partially mitigate this constraint.

In summary, this dissertation establishes a clear progression from statistical learning to causal understanding and automated reasoning. By integrating large-scale modeling with causal discovery and agent-based frameworks, we provided a unified foundation for the next generation of autonomous systems—ones capable not just of prediction, but of interpretable, causally-aware decision-making across science and industry.

\section{Outlook}
The work presented in this thesis brings together sequence modeling, causal discovery, and multi-agent reasoning into a unified framework for automated fault understanding in industrial-scale event sequences. While the proposed methods demonstrate strong performance in automotive diagnostics and point toward a new paradigm for large-scale event understanding, many open problems remain, both theoretical and practical. These challenges define a rich landscape for future research that extends far beyond the automotive domain.

\subsection{Toward a Unified Causal Graph over Events and Outcomes}
Table~\ref{tab:causal_framework} organizes the causal discovery landscape of this thesis along two axes: causal scope (sample-level vs. population-level) and dependency type (events-to-outcome vs. event-to-event). Three of the four cells are addressed by the contributions of this thesis (OSCAR, TRACE, and CARGO), yet the fourth cell, population-level event-to-event discovery, remains perhaps the most scientifically important open problem. %
Filling the open cell requires aggregating sample time causal graphs produced by TRACE across thousands of sequences into a single, general consistent summary graph over event types. This is the natural event-to-event counterpart of CARGO, and its realization faces three non-trivial obstacles. First, sample time graphs encode when a causal edge fires within a specific sequence, making naive frequency-based aggregation sensitive to temporal inconsistencies across sequences of varying length and context. Second, enforcing acyclicity constraints on a graph with tens of thousands of nodes after aggregation is computationally demanding and requires careful post-processing beyond what CARGO's Phase 2 currently provides. Third, robust uncertainty quantification over aggregated edges, such as distinguishing persistent structural causes from spurious co-occurrences driven by rare events demands principled statistical thresholding at a scale that has not yet been studied.
However, filling this cell individually still leaves the deeper problem unresolved: the two dependency types are not independent. In practice, a DTC may causally trigger another DTC (event-to-event), which in turn causes an error pattern (events-to-outcome), forming a causal chain that spans both columns of Table~\ref{tab:causal_framework} simultaneously. At the sample-level, OSCAR and TRACE can in principle be run jointly on the same sequence, but their outputs, a Markov Boundary per label and an sample time causal graph over events are not currently unified into a single coherent object. At the population-level, the problem compounds: one would require a mixed graph that simultaneously encodes directed edges between event types and directed edges from event types to outcome labels, with consistent semantics across both. We refer to this as the joint population-level causal graph, and it constitutes the most ambitious open problem in this line of work. Such a graph would allow a practitioner to trace a complete causal path from a root-cause DTC, through intermediate triggering events, to the final error pattern entirely from observational data, without any manual rule definition. Addressing this problem will require extending the aggregation framework of CARGO to handle heterogeneous node types (events and labels), developing acyclicity constraints that respect the asymmetry between the two node classes, and designing evaluation protocols that go beyond current set-based and Hamming-distance metrics. The seq2cause library is architected with this unification in mind, and we consider the joint population-level causal graph the single most impactful direction for future research emerging from this thesis.

\subsection{Multimodal and Multi-Resolution Causal Discovery}
Throughout this thesis, causal discovery primarily operates on discrete event streams possibly with outcomes. Yet modern vehicles, like many stochastic systems, generate multimodal data, including environmental conditions, textual descriptions, or camera-based observations. Extending causal discovery to this multimodal setting, by combining discrete events with continuous signals, images, or text, would enable richer causal graphs that can explain failures through finer-grained interactions. Foundations from multimodal Transformers, contrastive pretraining, and cross-modal attention could support such expansion. A promising direction is causal fusion learning, where causal edges are allowed to form across modalities and at multiple temporal resolutions (millisecond sensor bursts vs. day-level DTCs), similar to summary graphs in time series \cite{pmlr-assadsummary}. This would help address phenomena such as cascading failures arising from the joint interaction of software and hardware subsystems, and enrich the EP rules by incorporating environmental conditions.

\subsection{Unseen Domain and Continuous Learning}
A central challenge for deploying intelligent diagnostic systems in real-world environments is their ability to generalize beyond the distribution from which they were trained. Vehicles evolve across software versions, model years, component suppliers, and even regional driving conditions. %
Models trained once and deployed indefinitely, whether sequence predictors, causal discovery frameworks, or multi-agent reasoning systems, inevitably face degradation as the data distribution drifts. Such a system needs to continuously learn from new data while retaining previously acquired knowledge.

This opens a rich avenue for future research at the intersection of continual learning and multi-agent adaptation. Classical continual learning methods based on regularization, replay, or architectural expansion \cite{Goodfellow2013AnEI} provide the first building blocks, but remain insufficient for highly structured, relational tasks such as diagnostic rule synthesis. More recent theoretical developments, such as the “Nested Learning” perspective \cite{behrouz2025nested}, challenge the assumption that deep architectures inherently build hierarchical abstractions, suggesting instead that models must be explicitly encouraged to discover stable mechanisms that persist across contexts. They use different updating frequencies of neurons across layers, mimicking brain oscillations.

Future systems may therefore incorporate continual updates during Transformers' training and enable smooth, continuous synthesis of EP rules, with agents detecting when new vehicles introduce unseen failure modes, unseen DTC combinations, or entirely novel clusters of behaviors. Multi-agent architectures are particularly promising in this setting: one agent may monitor distribution shift, another may propose updated causal graphs for new domains. Beyond adaptation, such systems could also identify \emph{when} previously learned EP rules should be revised, pruned, or merged, enabling long-term autonomous maintenance of diagnostic knowledge bases.

\subsection{Industrial Foundation Models}
Recent progress in large language models raises the question of whether analogous foundation models for event sequences can be developed. Such models would unify sequence modeling, forecasting, causal inference, anomaly detection, and reasoning capabilities across domains such as healthcare, cybersecurity, aviation, and industrial operations. The introduced Transformer-based architectures, coupled with causal discovery tools, could serve as stepping stones toward such foundation models. Pretraining on large-scale, heterogeneous event-log corpora, comparable to natural-language corpora, would enable these models to generalize across systems and support in-context learning \cite{icl}, zero-shot, or few-shot diagnostics on unseen devices, procedures, or scenarios. It is essential to strike a balance between the world of words and events through adapted pretraining and fine-tuning tasks. Recent approaches adapt diffusion models for discrete data \cite{sahoo2024simple}. This could help build more robust bidirectional models rather than traditional left-to-right models \cite{nie2025large}, particularly for sequence classification and machine translation tasks.

\bookmarksetup{startatroot}
\listoffigures
\listoftables

\printbibliography

\KOMAoptions{open=any} %
\appendix
\chapter{Appendix of Part~\ref{part1}}\label{appendix:part1}
\section{Error Pattern Prediction}
 This section contains the appendix material for Chapter~\ref{c2:ep_prediction_based_on_live_data}.

\subsection{Multivariate Hawkes Process Log-likelihood Derivation}\label{appendix:derivation_ll_hawke_mutual}
The likelihood for any point process parametrized by \(\theta\) can be computed as the sum of the log-likelihood for each process~\cite{hawke_process_and_app_2025}. Let \(N^k(T)\) denote the number of events observed in stream \(k\) by time \(T\), and \(t^k_1, \cdots, t^k_{N^k(T)}\) their occurrence times (as introduced for the mutually-exciting Hawkes process in Chapter~\ref{c2:foundation_event_sequence_modeling}):

\begin{equation*}
    \ln \mathcal{L} (\theta|T) = \sum^{|\mathcal{X}|}_{k=1} \ln \mathcal{L}_k (\theta)
\end{equation*}
where each term is defined by:
\begin{align}
    \ln{\mathcal{L}_k}(\theta) &= \ln \left(\left[\prod^{N^k(T)}_{i=1} \lambda^{*}_k(t^k_i)\right]\exp{(-\int^T_0 \lambda^{*}_k(t) dt)}\right) \notag\\
\end{align}
This expression reduces to:
\begin{align}
    &= \sum^{N^k(T)}_{i=1}\ln{\lambda^{*}_k(t^k_i)} + \ln{\left(\exp{(-\int^T_0 \lambda^{*}_k(t) dt)}\right)} \notag\\
    &= \sum^{N^k(T)}_{i=1}\ln{\lambda^{*}_k(t^k_i)} -\int^T_0 \lambda^{*}_k(t) dt\label{eq:mll_mutual_hawkes_per_event}
\end{align}
Finally, we have for the total log-likelihood:
\begin{equation}\label{eq:mll_mutual_hawkes_appendix}
     \ln \mathcal{L} (\theta|T) = \sum^{|\mathcal{X} |}_{k=1}\sum^{N^k(T)}_{i=1}\ln{\lambda^{*}_k(t^k_i)} - \sum^{|\mathcal{X}|}_{k=1}\int^T_0 \lambda^{*}_k(t)dt
\end{equation}

\subsection{Models Definition}
A GPT \cite{gpt} model is added to improve comparison over baseline models.
We kept the original implementation of the feed-forward layers \cite{tf} but initialize the intermediate layers with: $ W_l \sim \mathcal{N}(0, \frac{2}{L \sqrt{d_l}}) $, where \(l\) indexes the layer, \(d_l\) is that layer's input dimension, and \(L\) is the total number of layers (unrelated to the sequence length \(L\) used elsewhere in this thesis).
We used root-mean-square normalization (RMS Norm) from \cite{rmsnorm} instead of traditional layer normalization. We initialized all other linear layers using the SMALLINIT schema \cite{transformernotears} $ W_l \sim \mathcal{N}\left(0, \sqrt{\frac{2}{d_l + 4d_l}}\right)$.
The names printed in the Tab \ref{tab:pretraining_exp} are model variations based on the embeddings given to CarFormer:
\begin{description}
  \item[time]: Along with the event type $\mathbf{E}$ an absolute time embedding is added such as input $\mathbf{U} = \mathbf{E} + \mathbf{T}$
  \item [mileage]: Same as time except we add a mileage embedding $\mathbf{M}$ such as $\mathbf{U} = \mathbf{E} + \mathbf{M}$
  \item [rot]: A RoPE \cite{Roformer} is applied to $\mathbf{Q,K}$ such as $\mathbf{Q} = \mathbf{R}^d_{\Theta}\mathbf{W}^Q\mathbf{E}$,  $\mathbf{K} =\mathbf{R}^d_{\Theta}\mathbf{W}^K\mathbf{E}$
  \item [ce]: We add a context embedding $\mathbf{CE} = \mathbf{T} + \mathbf{M}$ to $\mathbf{Q,K}$ such as $\mathbf{Q = U W}^Q \mathbf{ + CE}$ to every attention layers.
  \item [m2c, c2m]: Inspired by the Disentangled Attention mechanism from DeBerta \cite{Deberta}. Attention score $\mathbf{A}_{i,j}$ between tokens $i$ and $j$ is computed from hidden states vector $\{\mathbf{H}_i\}$ at event step $i$ and a mileage vector  $\{\mathbf{M}_i\}$ at event step $i$ such as: $\mathbf{A}_{i,j} = \{\mathbf{H}_i, \mathbf{M}_{i} \} \times \{\mathbf{H}_j, \mathbf{M}_{j} \}^T = \mathbf{H}_i\mathbf{H}^T_j + \mathbf{H}_i\mathbf{M}^T_{j} + \mathbf{M}_{i}\mathbf{H}^T_j + \mathbf{M}_{i}\mathbf{M}^T_{j}$ = \textit{"content-to-content"} + \textit{"content-to-mileage"} + \textit{"mileage-to-content"} + \textit{"mileage-to-mileage"} = \textit{"c2c"} + \textit{"c2m"} + \textit{"m2c"} + \textit{"m2m"}.
\end{description}
\subsection{Random Event Injection}
\begin{figure}[t]
    \centering
\includegraphics[width=0.8\columnwidth]{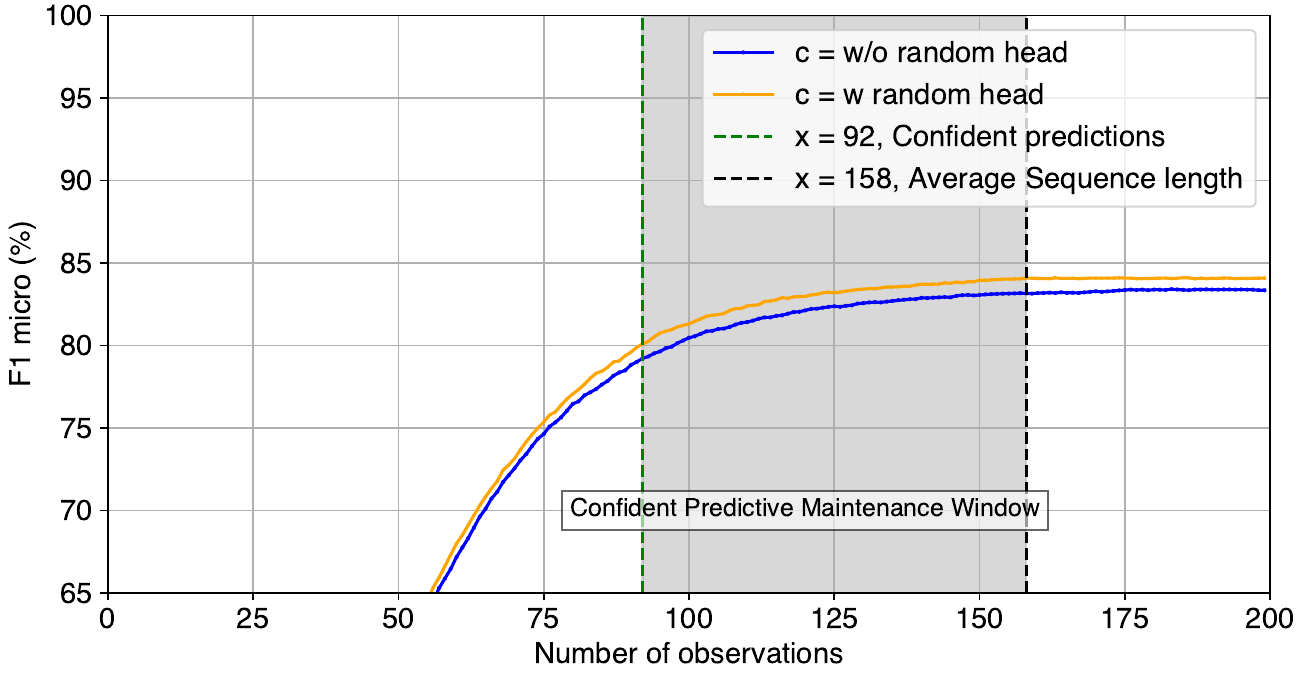}
    \caption{\textbf{Confident Predictive Maintenance Window Evolution} with and w/o the random head.}
    \label{fig:randomhead}
\end{figure}

 Let \( A_i \) be the random variable representing the number of random events injected at step \( i \). We can say that:
\begin{align*}
P(A_i = r) &= (1-p)p^r  \quad \text{for} \quad r = 0, 1, 2, \ldots
\end{align*}
Where \( A_i \) is the number of random events injected at step \( i \), \( r \) is the number of trials (injected events) until the first failure, \( p \) is the probability of injecting a random event. The expected total number of fake events injected over a sequence \(S\) of length \(L\) is therefore:
$$
\mathbb{E}\left[\sum_{i=1}^{L} A_i\right] = \frac{Lp}{1-p}
$$ since each \(A_i\) is geometrically distributed with mean \(p/(1-p)\). In our case, this translates to $150 \times 0.05/0.95 \approx 7.9$ random events per sequence. Fig.~\ref{fig:randomhead} shows the practical benefit of this injection procedure: it shifts the onset of the Confident Predictive Maintenance Window (Def.~\ref{def:cpmw}) to fewer observed events compared to a model trained without the Random Event Head. In other words, EPredictor reaches a reliable prediction earlier in the sequence.

\section{Multimodal Error Pattern Offline Prediction}
\begin{figure}[!htb]
      \centering
      \includegraphics[width=0.6\columnwidth]{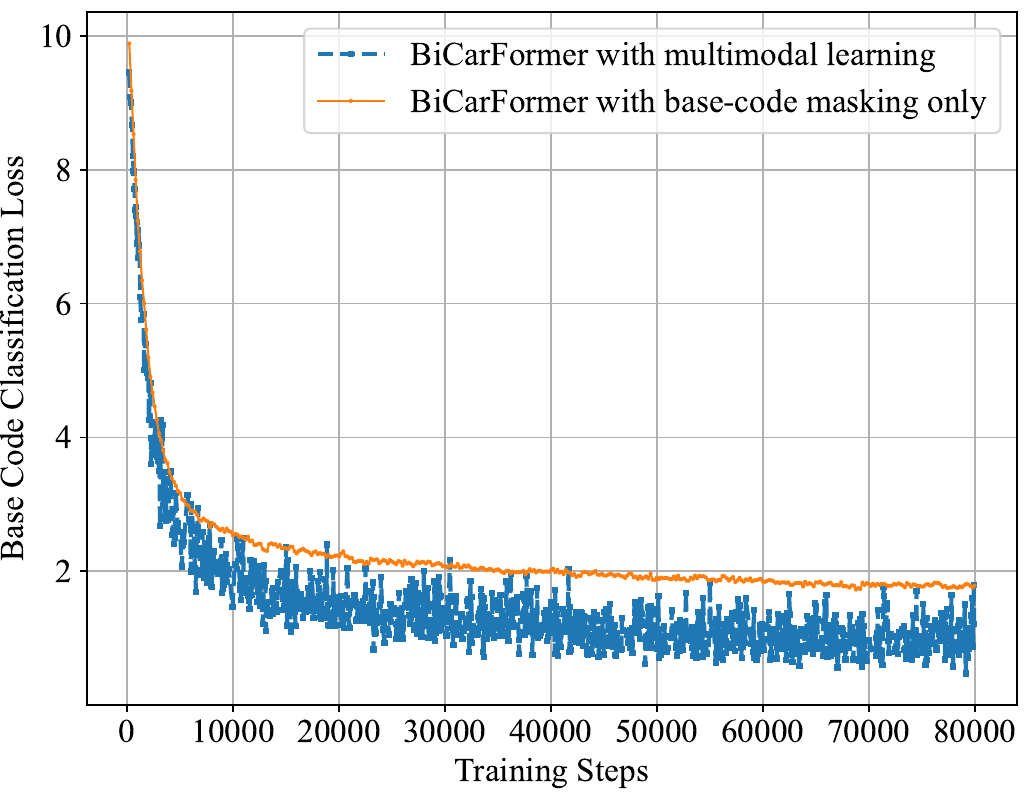}
      \caption{\textbf{Pretraining Loss Comparison}. Base-DTC classification loss comparison with and without multimodal learning.}
      \label{fig:pretraining_comp}
\end{figure}

This section contains the appendix material for Chapter~\ref{c4:multimodal}. Fig.~\ref{fig:pretraining_comp} below compares the Base-DTC classification loss during BiCarFormer's pretraining with and without the multimodal (environmental-condition) objective described in Chapter~\ref{c4:multimodal}. The lower loss obtained with multimodal learning indicates that jointly predicting the masked DTC and the masked environmental condition provides a useful auxiliary training signal rather than merely adding noise to the optimization.

\chapter{Appendix of Part~\ref{pa:p2}}\label{appendix:part2}
\raggedbottom

\section{Sample-Level Multi-Label Causal Discovery}
This section contains the appendix material for the sample-level multi-label causal discovery Chapter~\ref{c7:multi_label_one_shot_causal_discovery}.

\subsection{OSCAR Algorithm}\label{sec:oscarimplementation}
Algorithm~\ref{alg:oscar} summarizes OSCAR for one observed sequence; it complements the derivation of Section~\ref{sec:oneshotmarkov} and the overview of Fig.~\ref{fig:oscar}. Lines~\ref{alg:oscar:px}--\ref{alg:oscar:sample} correspond to steps (1)--(2) of Fig.~\ref{fig:oscar}, line~\ref{alg:oscar:py} to steps (3)--(4), and lines~\ref{alg:oscar:cmi}--\ref{alg:oscar:mb} to step (5). The sampling of line~\ref{alg:oscar:sample} is detailed in Appendix~\ref{abl:sampling}, and \(\mu_{Y_j}\), \(\sigma_{Y_j}\) denote the mean and standard deviation of \(\hat{I}_N(Y_j, x_t \mid X_{<t})\) over the positions \(t\). In practice, every step is applied to a batch of sequences at once, and the \(N\) trajectories are stacked along the batch dimension, so that each of the two models is called in a single forward pass. A reference implementation is available in the \texttt{seq2cause} library~\cite{math2026seq2cause}.

\begin{algorithm}[!t]
\DontPrintSemicolon
\caption{OSCAR: sample-level Markov Boundary recovery for one sequence}\label{alg:oscar}
\KwIn{observed sequence \(x_{0:L-1}\); pretrained models \(\text{Tf}_x\) and \(\text{Tf}_y\); context length \(c\); number of trajectories \(N\); top-\(k\)/nucleus sampling parameters; threshold multiplier \(k\)}
\KwOut{estimated Markov Boundaries \(\widehat{\mathcal{MB}}(Y_j)\) and causal indicators \(\hat{\mathcal{C}}(Y_j, X_t)\) for every label \(Y_j\)}
\(P_{\theta_x}(X_s \mid x_{<s}) \gets \text{Tf}_x(x_{<s})\) for all \(s < c\) \tcp*{Eq.~\eqref{eq:prob_x}}\label{alg:oscar:px}
\For{\(l = 1, \ldots, N\) \textbf{in parallel}}{
  draw \(x^{(l)}_s \sim P_{\theta_x}(X_s \mid x_{<s})\) for all \(s < c\) \tcp*{top-\(k\)/nucleus}\label{alg:oscar:sample}
  \(x^{(l)} \gets (x^{(l)}_{0:c-1}, x_{c:L-1})\) \tcp*{observed suffix}
}
\(p^{(l)}_{t,j} \gets P_{\theta_y}(Y_j = 1 \mid x^{(l)}_{\leq t})\) for all \(l\), \(t \geq c\), \(j\), clipped to \([\varepsilon, 1-\varepsilon]\) \tcp*{Eq.~\eqref{eq:prob_y}}\label{alg:oscar:py}
\ForEach{position \(t > c\) and label \(Y_j\)}{
  \(\hat{I}_N(Y_j, x_t \mid X_{<t}) \gets \frac{1}{N}\sum_{l=1}^{N} D_{KL}\big(\mathrm{Bern}(p^{(l)}_{t,j}) \,\|\, \mathrm{Bern}(p^{(l)}_{t-1,j})\big)\) \tcp*{Eq.~\eqref{eq:cmi_approx}}\label{alg:oscar:cmi}
  \(\hat{\mathcal{C}}(Y_j, X_t) \gets \frac{1}{N}\sum_{l=1}^{N} \big(p^{(l)}_{t,j} - p^{(l)}_{t-1,j}\big)\) \tcp*{Eq.~\eqref{eq:causal_indic}}
}
\ForEach{label \(Y_j\)}{
  \(\tau_j \gets \mu_{Y_j} + k \cdot \sigma_{Y_j}\) \tcp*{Eq.~\eqref{eq:cmi_epsilon}}
  \(\widehat{\mathcal{MB}}(Y_j) \gets \{X_t : \hat{I}_N(Y_j, x_t \mid X_{<t}) > \tau_j\}\)\;\label{alg:oscar:mb}
}
\Return{\(\{\widehat{\mathcal{MB}}(Y_j)\}_j\), \(\hat{\mathcal{C}}\)}
\end{algorithm}

\noindent Since \(\text{Tf}_y\) reuses \(\text{Tf}_x\) as its backbone, the distribution of line~\ref{alg:oscar:px} can also be read from the same forward pass of \(\text{Tf}_y\); the two models are kept separate in Algorithm~\ref{alg:oscar} for clarity.

\subsection{PyCausalFS}
To ensure exact reproducibility of the local-structure-learning baselines reported in Chapter~\ref{c7:multi_label_one_shot_causal_discovery}, we specify their common hyperparameter here. Local structure learning algorithms were all used with \(\alpha=0.1\) (the significance level for the underlying conditional-independence tests) in the associated code: \url{https://github.com/wt-hu/pyCausalFS/tree/master/pyCausalFS/LSL}.
\subsection{MI-MCF}
We similarly report the configuration of the strongest baseline, MI-MCF, for reproducibility. MI-MCF \cite{mimcf} was used for comparison following the official implementation at \url{https://github.com/malinjlu/MI-MCF} we used \(\alpha = 0.05, L = 268, k_1 = 0.7, k_2=0.1\), the significance level, sample size, and two internal weighting coefficients as named in that implementation's own configuration (unrelated to this thesis's sequence-length \(L\) or the CARGO decay-rate \(k\)).

\subsection{NADEs Quality.}\label{appendix:nades}
We did several ablations on the quality of the NADEs and their impact on the sample-level multi-label causal discovery phase. In particular, Table \ref{tab:ablation_nades_comparison} presents multiple \(\text{Tf}_x, \text{Tf}_y\) with respectively 90 and 15 million parameters or 34 and 4 million parameters. We also varied the context window (conditioning set \(X_{<t}\)), trained on different amounts of data (Tokens), and reported the classification results on the test set of \(\text{Tf}_y\) alone. We did not output the running time since it was always the same for all NADEs: \(1.27\) minutes for 50,000 samples and \(0.14\) for 5000.
A longer context \(c\) improves the label model alone (the F1 of \(\text{Tf}_y\) rises from \(88.6\%\) at \(c=4\) to \(92.6\%\) at \(c=30\)), but OSCAR performs best at \(c=15\) and drops sharply at \(c=30\). Two effects explain this: events in the first \(c\) positions are only resampled and never tested as causes, and a long context already explains most of the label, so that each further event adds little information (entropy saturation, Chapter~\ref{c7:multi_label_one_shot_causal_discovery}). The smaller model trained on fewer tokens (47m parameters, 300m tokens) loses about 7.5 points of F1 compared with the larger model at the same context \(c=20\), which shows that the quality of the density estimators directly limits causal discovery. The ranking is the same for \(n=500\) samples, with larger standard deviations.

\begin{table}[ht]
\centering
\footnotesize %
\setlength{\tabcolsep}{4pt} %
\begin{tabular}{lcccccc}
\toprule
\textbf{Tokens} & \textbf{Param.} & \textbf{Context} & \textbf{Prec. (↑)} & \textbf{Rec. (↑)} & \textbf{F1 (↑)} & \textbf{Tfy F1 (↑)} \\
\midrule
\multicolumn{7}{c}{\textit{For \(n = 50{,}000\) samples}} \\
\midrule
1.5B & 105m & \(c = 4\)  & \(47.95 \pm 1.05\) & \(30.65 \pm 0.51\) & \(37.39 \pm 0.67\) & 88.6 \\
1.5B & 105m & \(c = 12\) & \(54.62 \pm 1.03\) & \(29.88 \pm 0.73\) & \(38.63 \pm 0.85\) & 90.43 \\
1.5B & 105m & \(c = 15\) & \(\mathbf{55.26 \pm 1.42}\) & \(\mathbf{31.37 \pm 0.82}\) & \(\mathbf{40.02 \pm 1.03}\) & 90.57 \\
1.5B & 105m & \(c = 20\) & \(49.52 \pm 1.59\) & \(\mathbf{31.76 \pm 0.85}\) & \(36.54 \pm 1.10\) & 91.19 \\
1.5B & 105m & \(c = 30\) & \(36.65 \pm 1.18\) & \(22.75 \pm 0.78\) & \(26.57 \pm 0.91\) & \textbf{92.64} \\
300m & 47m & \(c = 20\) & \(39.49 \pm 1.77\) & \(26.30 \pm 0.89\) & \(29.01 \pm 1.10\) & 83.6 \\
\midrule
\multicolumn{7}{c}{\textit{For \(n = 500\) samples}} \\
\midrule
1.5B & 105m & \(c = 12\) & \(54.84 \pm 4.55\) & \(\mathbf{31.45 \pm 2.23}\) & \(\mathbf{39.95 \pm 2.83}\) & 90.43 \\
1.5B & 105m & \(c = 15\) & \(55.04 \pm 3.36\) & \(29.90 \pm 1.78\) & \(38.74 \pm 2.24\) & 90.57 \\
1.5B & 105m & \(c = 20\) & \(48.84 \pm 4.01\) & \(\mathbf{31.65 \pm 2.37}\) & \(36.19 \pm 2.65\) & \textbf{91.19} \\
300m & 47m & \(c = 20\) & \(38.23 \pm 2.91\) & \(25.31 \pm 2.39\) & \(27.92 \pm 2.25\) & 83.6 \\
\bottomrule
\end{tabular}
\caption{Ablations of the performance of Phase 1 (Sample-Level Markov Boundary retrieval) as a function of different NADEs with \(n=50{,}000\) and \(n=500\) samples, averaged over 5-folds. Classification metrics use weighted averaging. Metrics are given in \(\%\).}
\label{tab:ablation_nades_comparison}
\end{table}

\subsection{Sampling Procedure for CMI}\label{abl:sampling}
We performed an ablation (Tab \ref{tab:sampling_comparison_transposed}) on the effect of sampling methods to estimate the expected value over all possible trajectories \(X_{<t}\). We used one A10 GPU on a sample of the test dataset (4000 random samples) composed of 205 labels with a batch size of 4 during inference.
We tested top-\(k\) sampling with \(k \in \{20, 25, 35\}\) \cite{fan-etal-2018-hierarchical}, with and without a temperature \(T \in \{0.8, 0.9, 1.1, 1.2\}\) applied to the log-probabilities \(\boldsymbol{\hat{x}}\) such that \[\boldsymbol{\hat{x}}' = \text{softmax}(\log{\boldsymbol{\hat{x}}}/T)\]

\noindent the combination of top-\(k\) and nucleus sampling \cite{Holtzman2020The} with probability mass \(p \in \{0.8, 0.9\}\), and finally a random permutation of the observed events within the context \(c\) instead of sampling. The exact combinations are listed in Table~\ref{tab:sampling_comparison_transposed}.
We fixed the threshold multiplier of the dynamic threshold (Eq.~\ref{eq:cmi_epsilon}) to \(3\) and performed 10 runs. We then report the mean and standard deviation of each classification metric and of the elapsed time (sec).

\noindent Sampling trajectories from \(\text{Tf}_x\) roughly doubles the F1 scores of OSCAR (micro F1 from \(14.1\%\) without sampling to up to \(28.8\%\)), at the cost of an about eleven times longer run time. Averaging over trajectories smooths the CMI along the sequence, which makes the threshold easier to set when several labels are present. Permuting the observed events instead of sampling new ones helps much less (micro F1 \(18.2\%\)). The sampling variants differ by less than \(2.5\) points; we choose \textbf{top-\(k\)=35 with \(p=0.8\)}, which gives the best micro F1 and is within \(0.6\) points of the best macro and weighted F1.
\begin{table}[ht]
    \centering
    \footnotesize %
    \setlength{\tabcolsep}{4pt} %
    \begin{tabular}{@{}lcccc@{}}
        \toprule
        \textbf{Proposal} & \textbf{F1 Micro (\%)} & \textbf{F1 Macro (\%)} & \textbf{F1 Wtd. (\%)} & \textbf{Time (s)} \\
        \midrule
        w/o Sampling         & \(14.07\)                & \(12.29\)                & \(16.67\)                &  \(\mathbf{49.30 \pm 0.30}\)       \\
        Permutation          & \(18.22 \pm 0.36\)       & \(13.75 \pm 0.09\)       & \(19.21 \pm 0.03\)       & \(557.82 \pm 0.13\)       \\
        Top-k=20             & \(26.77 \pm 0.71\)       & \(23.83 \pm 0.19\)       & \(29.25 \pm 0.07\)       & \(557.40 \pm 0.13\)       \\
        Top-k=35             & \(26.57 \pm 0.96\)       & \(24.08 \pm 0.23\)       & \(29.30 \pm 0.07\)       & \(557.35 \pm 0.10\)       \\
        Top-k=35+T=0.8       & \(27.36 \pm 0.65\)       & \(23.77 \pm 0.21\)       & \(28.98 \pm 0.07\)       & \(557.45 \pm 0.11\)       \\
        Top-k=35+T=1.2       & \(26.59 \pm 1.49\)       & \(\mathbf{24.62 \pm 0.29}\) & \(\mathbf{29.52 \pm 0.06}\) & \(557.45 \pm 0.12\)       \\
        Top-k=25+p=0.8       & \(27.98 \pm 0.67\)       & \(23.82 \pm 0.28\)       & \(29.18 \pm 0.07\)       & \(558.07 \pm 0.07\)       \\
        \textbf{Top-k=35+p=0.8} & \(\mathbf{28.82 \pm 0.75}\) & \(24.06 \pm 0.25\)       & \(29.17 \pm 0.07\)       & \(558.16 \pm 0.14\)       \\
        Top-k=35+p=0.9       & \(26.39 \pm 0.99\)       & \(24.12 \pm 0.31\)       & \(29.26 \pm 0.11\)       & \(558.11 \pm 0.12\)       \\
        Top-k=35+p=0.9+T=0.9 & \(27.63 \pm 0.75\)       & \(23.90 \pm 0.24\)       & \(29.04 \pm 0.09\)       & \(558.07 \pm 0.12\)       \\
        Top-k=35+p=0.9+T=1.1 & \(26.75 \pm 1.30\)       & \(24.47 \pm 0.24\) & \(29.45 \pm 0.09\)       & \(558.06 \pm 0.11\)       \\
        \bottomrule
    \end{tabular}
    \caption{Sample-level classification performance and elapsed time (sec) of OSCAR across different sampling methods (\(k\): top-\(k\) size, \(p\): nucleus probability mass, \(T\): temperature), mean and standard deviation over 10 runs (\emph{w/o Sampling} is deterministic). Best results are in \textbf{bold}.}
    \label{tab:sampling_comparison_transposed}
\end{table}

\subsection{Sampling Number}\label{sec:samplingnumber}
\begin{figure}[!h]
    \centering
    \includegraphics[width=0.62\linewidth]{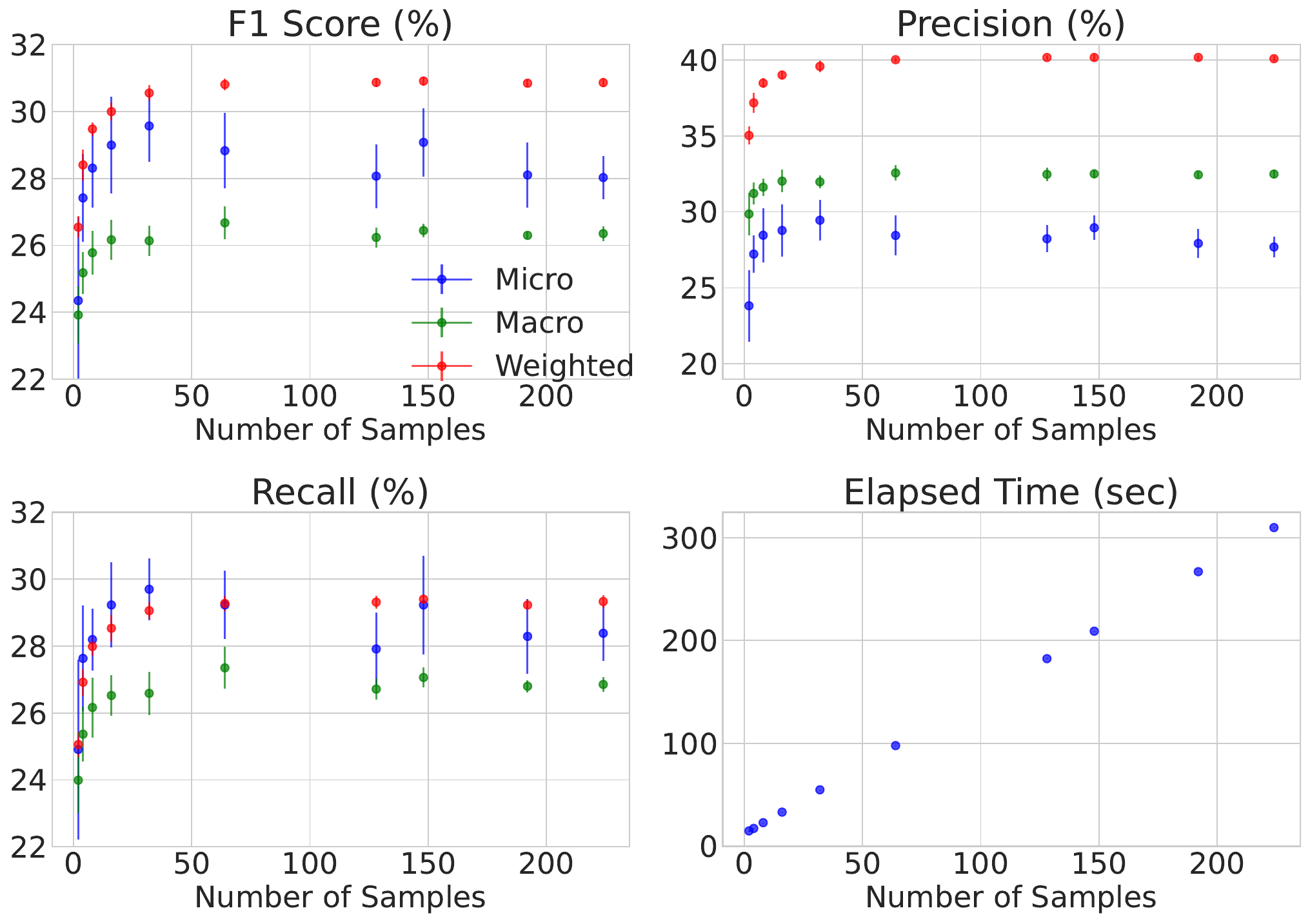}
    \caption{\textbf{Ablation of the Number of Sampled Trajectories \(N\)}. Results are reported using 1-sigma error bar.}
    \label{abl:number_sampling}
\end{figure}

We experimented with different numbers of \(N\) for the sampling method across different averaging methods (micro, macro, weighted), Fig~.\ref{abl:number_sampling}. We performed eight different runs and reported the average, standard deviation, and elapsed time. In general, sampling with a larger \(N\) tends to decrease the standard deviation and yield more reliable Markov Boundary estimation. As we process more samples, the model gradually improves with logarithmic growth until it converges to a final score. We also verify that our time complexity is linear with the number of samples \(N\).
In Fig.~\ref{abl:number_sampling}, all F1 scores rise quickly up to about \(N=32\) and then stay flat (around \(31\%\) weighted, \(28\%\) micro, and \(26\%\) macro), and precision stops improving at about \(N=64\), whereas the run time grows linearly, from about \(15\) seconds for small \(N\) to about \(300\) seconds for \(N=224\).
Based on these results, we generally choose \(N=68\).

\subsection{Dynamic Thresholding}
\begin{figure}[!ht]
    \centering
    \includegraphics[width=0.62\linewidth]{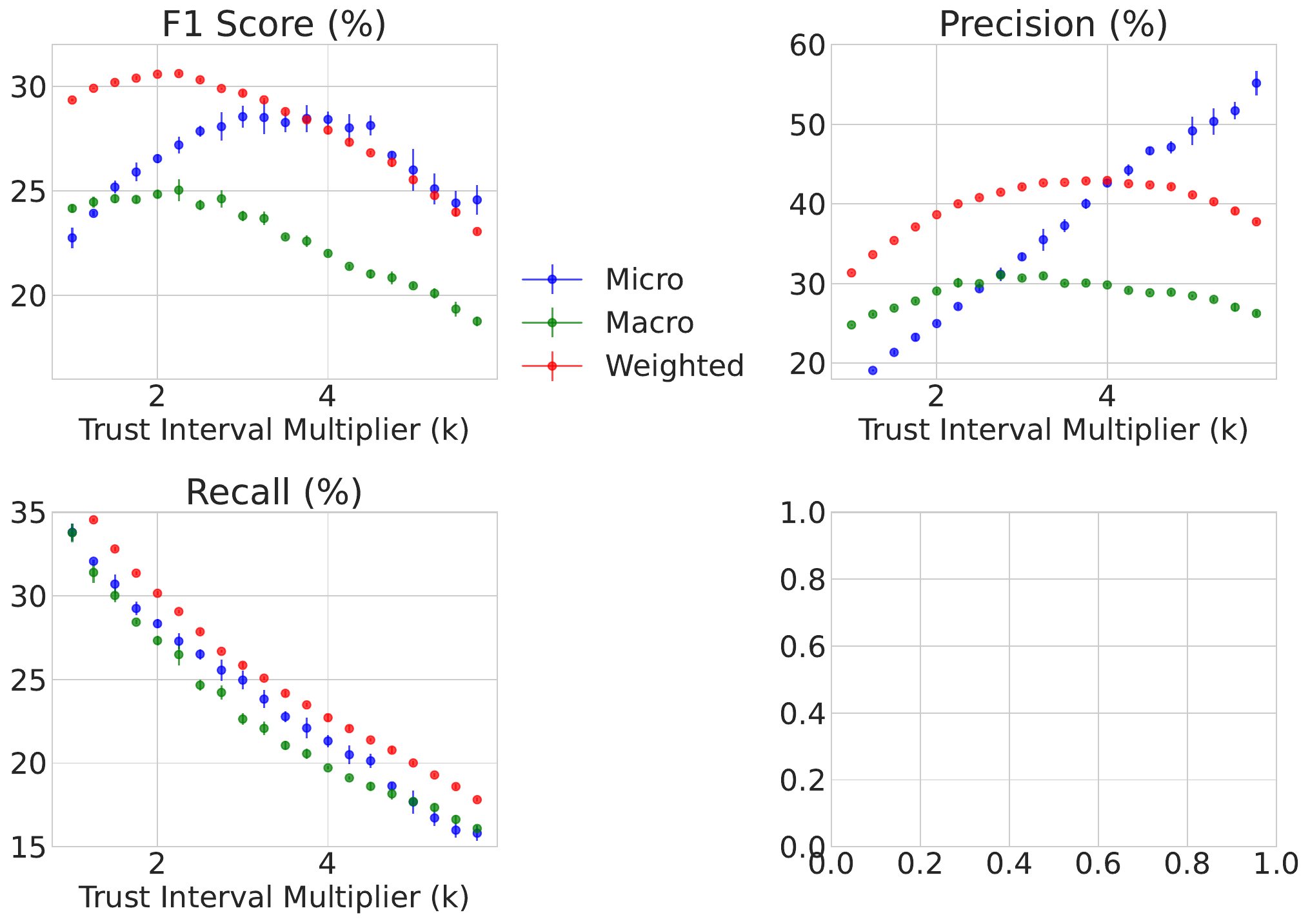}
    \caption{\textbf{Sensitivity Analysis}. Evolution of the sample-level F1 Score, Precision, and Recall as a function of coefficient \(k\). Results are reported using 1-sigma error bar.}
    \label{abl:threshold_selection}
\end{figure}

We performed ablations on the effect of \(k\) during the dynamic thresholding of the CMI in Eq.~\ref{eq:cmi_epsilon} to assess conditional independence in Fig.~\ref{abl:threshold_selection}. A larger \(k\) raises the threshold, so fewer events are declared causes: recall decreases steadily for all averages, while micro precision increases. The F1 scores therefore peak at different values: around \(k=2\) for the weighted and macro averages and around \(k=3\) to \(4\) for the micro average. To balance the classification metrics across the different averages, we set \(k=2.75\).

\section{Population-Level Multi-Label Causal Discovery}
This section contains the appendix material for Chapter~\ref{c7:multi_label_causal_discovery}.

\subsection{Definition of Scoring Criteria}
Before introducing the criteria compared in Appendix~\ref{appendix:criterions}, we first recall the formal properties, decomposability, score-equivalence, and local consistency, that a scoring criterion for structural fusion should ideally satisfy.

\begin{definition}\label{def:decomp}
    (Decomposable Criterion). We say that a scoring criterion \(S(\mathcal{G}, \mathcal{D})\) is decomposable if it can be written as a sum of measures, each of which is a function only of one node and its parents. In other words, a decomposable scoring criterion \(f_s\) applied to a DAG \(\mathcal{G}\) can always be expressed as:
\begin{equation}\label{eq:decomp}
    S(\mathcal{G}, \mathcal{D}) = \sum^n_i f_s(X_i, \text{Pa}_{\mathcal{G}}(X_i))
\end{equation}
\end{definition}

\begin{definition}[Score equivalent]\label{def:score_equivalent}\cite{ges}.
A score \(f_s\) is score equivalent if it assigns the same score to all the graphs in the same MEC.
\end{definition}
\begin{definition}[Local Consistency]\label{def:local_consistency}\cite{ges}
Let \(\mathcal{D}\) contain \(m\) i.i.d samples from some distribution \(P(.)\), and let \(\perp_p\) denote conditional independence in \(P\). Let \(\mathcal{G}\) be any possible DAG and \(\mathcal{G}'\) a different DAG obtained by adding the edge \(i \rightarrow j\) to \(\mathcal{G}\). A score \(f_s\) is locally consistent if both hold:
\begin{itemize}
    \item If \(X_i \not\perp_p X_j|Pa_{\mathcal{G}}(X_j), \) then \(f_s(\mathcal{G'}, \mathcal{D}) > f_s(\mathcal{G}, \mathcal{D})\)
    \item If \(X_i \perp_p X_j|Pa_{\mathcal{G}}(X_j), \) then \(f_s(\mathcal{G'}, \mathcal{D}) < f_s(\mathcal{G}, \mathcal{D})\)
\end{itemize}
\end{definition}

\subsection{Aggregation Criteria}\label{appendix:criterions}
This section defines the aggregation criteria compared in the ablation of Chapter~\ref{c7:multi_label_causal_discovery} (Fig.~\ref{fig:ablation_comparaison_criterions}). Each criterion decides, for every label \(Y_j\) and event type \(X_i\), whether an edge \(X_i \rightarrow Y_j\) found in the sample-level graphs \(\mathcal{G}^{(1)}, \ldots, \mathcal{G}^{(m)}\) of Phase 1 is kept in the global graph \(\mathcal{G}^*\). \emph{Union} keeps every edge, and \emph{adaptive thresholding} is CARGO's own criterion (Algorithm~\ref{alg:cargo}); the remaining baselines are described below. Their names in Fig.~\ref{fig:ablation_comparaison_criterions} are given in parentheses.

\subsubsection{Frequency}
Static frequency voting is the simplest baseline (\texttt{frequency\_}\(\tau\)). For each label \(Y_j\), the frequency \(\hat{\pi}_{i,j}\) of the edge \(X_i \rightarrow Y_j\) across the \(m\) sample-level graphs is computed as in Algorithm~\ref{alg:cargo}, and the edge is kept if \(\hat{\pi}_{i,j} \geq \tau\) for a fixed threshold \(\tau \in \{0.05, 0.25, 0.5, 0.8\}\), the same for all labels.

\subsubsection{Expected False-Positive Rate}
This baseline (\texttt{frequency\_beta}) is adapted from the consensus-network literature~\cite{consensus_bn_freq_cutoff_no}. There, two Beta distributions are fitted to the edge frequencies, one for spurious and one for true edges, and the cutoff is chosen such that the expected rate of false-positive edges stays below a target value. We follow the same principle but fit the two Beta distributions to the mutual information \(I(Y_j, X_i)\) of the candidate edges (estimated as in Eq.~\ref{eq:mi_heuristic}) instead of their frequencies, and remove the edges in the lower tail. The target false-positive rates are \(0.01, 0.05, 0.15\), and \(0.2\).

\subsubsection{Mutual Information with Backward Equivalence Search}
This baseline (\texttt{BES\_mi\_imbalance\_par}) searches for the graph that maximizes a score based on the mutual information between events and labels.

\paragraph{Backward Equivalence Search.}
The Backward Equivalence Search (BES) is the second phase of GES~\cite{ges}: starting from the union of all sample-level edges, edges are removed one at a time as long as the score of the graph increases. Heuristic fusion approaches~\cite{approx_fusionPUERTA2021155, delSagrado2001} instead select the edges independently, by maximizing a sum of per-edge scores:
\begin{equation}\label{eq:criterion_edges}
E = \underset {E_i \in \varepsilon}\arg {\text{max}} \sum_{e \in E_i} S(e)
\end{equation}
where \(\varepsilon\) denotes the search space (all sets of possible edges over \(\boldsymbol{U}\)) and \(S(e)\) a criterion for the relevance of a single edge (e.g., its frequency). Such a per-edge criterion ignores the overall network structure, its complexity, and missing data~\cite{consensus_bn_jose}; a graph-level score optimized by BES is therefore used here, in the spirit of consensus fusion~\cite{torrijos2025informedgreedyalgorithmscalable}.

\paragraph{Estimating the Mutual Information.}
We use the mutual information \(I(Y_j, X_i)\) as the strength of an edge \(X_i \rightarrow Y_j\), because it can be computed from the quantities already estimated in Phase 1 (Fig.~\ref{fig:cargo}). It satisfies two of the properties required from a causal strength measure \(C_{X_i \rightarrow Y_j}\)~\cite{quantifyingcausalinfluence}: if \(C_{X_i \rightarrow Y_j} = 0\), the joint distribution satisfies the Markov condition with respect to the DAG obtained by removing the edge \(X_i \rightarrow Y_j\); and if the DAG consists of the single edge \(X_i \rightarrow Y_j\), then \(C_{X_i \rightarrow Y_j} = I(Y_j, X_i)\).
The mutual information is, however, difficult to estimate directly~\cite{estimating_mi}. Using the chain rule~\cite{cover1999elements}, we decompose it with the interaction information \(I(Y_j, X_i, X_{<t}) \triangleq I(Y_j, X_{<t}) - I(Y_j, X_{<t} \mid X_i)\), which indicates whether knowing \(X_{<t}\) weakens (negative) or strengthens (positive) the dependence between \(X_i\) and \(Y_j\):
\begin{align}
    I(Y_j, X_i) &= I(Y_j, X_i \mid X_{<t}) + I(Y_j, X_i, X_{<t}) \notag\\
    &= I(Y_j, X_i \mid X_{<t}) + I(Y_j, X_{<t}) - I(Y_j, X_{<t} \mid X_i) \label{eq:mi_recomp}
\end{align}
The first term is the CMI estimated in Phase 1 (Eq.~\ref{eq:cmi_approx}). The second term requires the marginal \(P(Y_j)\), which we estimate by the empirical label frequency \(\hat{P}(Y_j)\) in the large dataset \(\mathcal{D}\) (for a small dataset, \(\hat{P}(Y_j)\) may differ from \(P(Y_j)\)):
\begin{equation}\label{eq:mi_y_z}
    I(Y_j, X_{<t}) = \mathbb{E}_{p(x_{<t})} D_{KL}(P(Y_j|x_{<t})||\hat{P}(Y_j)) = \mathbb{E}_{p(x_{<t})} I_G(Y_j, x_{<t})
\end{equation}
The third term would require an additional estimation conditioned on \(X_i\). As a heuristic, we drop it and use
\begin{equation}\label{eq:mi_heuristic}
    \hat{I}(Y_j, X_i) = I(Y_j, X_i \mid X_{<t}) + I(Y_j, X_{<t}).
\end{equation}
Since \(I(Y_j, X_{<t} \mid X_i) \geq 0\), this heuristic can only overestimate \(I(Y_j, X_i)\); it is a score that reuses the Phase 1 estimates, not an approximation with a guaranteed error, and its usefulness is assessed empirically in Fig.~\ref{fig:ablation_comparaison_criterions}.

\paragraph{Class-Aware Information Gain (CAIG).}
The score optimized by BES balances three factors: the mutual information of the selected edges, the class imbalance, and the network complexity. For each label \(Y_j\) with parent set \(\text{Pa}^{\mathcal{G}}_j \subseteq \{X_1, \ldots, X_{|\mathcal{X}|}\}\) in a candidate graph \(\mathcal{G}\), the local score is
\begin{equation}\label{eq:caig_local}
    s_I(Y_j, \text{Pa}^{\mathcal{G}}_j) = \sum_{X_i \in \text{Pa}^{\mathcal{G}}_j} \hat{I}(Y_j, X_i) \; - \; \alpha \cdot |\text{Pa}^{\mathcal{G}}_j| \cdot \log{\left(\frac{m}{m_j} + 1\right)},
\end{equation}
and the CAIG score of the graph is the sum of the local scores over all labels:
\begin{equation}\label{eq:caig}
    f_s(\mathcal{G}, \mathcal{D}) = \sum_{j=1}^{|\mathcal{Y}|} s_I(Y_j, \text{Pa}^{\mathcal{G}}_j),
\end{equation}
where \(\hat{I}(Y_j, X_i)\) is the heuristic of Eq.~\ref{eq:mi_heuristic}, \(\alpha > 0\) is a regularization hyperparameter, \(m\) is the number of sequences in \(\mathcal{D}\), and \(m_j\) is the number of sequences in which the label \(Y_j\) is positive. The penalty grows with the number of parents and is larger for rare labels (small \(m_j\)), so that a rare label keeps a parent only if its mutual information is high. Since \(f_s\) is a sum of local scores, CAIG is decomposable (Def.~\ref{def:decomp}), like BIC~\cite{bic}.

\subsection{CARGO Phase 2 Algorithm}\label{sec:phase2implementation}
Algorithm~\ref{alg:cargo} summarizes the aggregation phase of CARGO (Phase 2, Fig.~\ref{fig:cargo}). It takes the \(m\) sample-level graphs produced by OSCAR (Algorithm~\ref{alg:oscar}) in Phase 1 and returns the global graph \(\mathcal{G}^*\) reported in Chapter~\ref{c7:multi_label_causal_discovery}. A reference implementation is available in the \texttt{seq2cause} library~\cite{math2026seq2cause}.

\begin{algorithm}[H]
\DontPrintSemicolon
\caption{CARGO Phase 2: adaptive frequency fusion of sample-level graphs}\label{alg:cargo}
\KwIn{sample-level graphs \(\mathcal{G}^{(1)}, \ldots, \mathcal{G}^{(m)}\) from Phase 1; label supports \(m_j\); bounds \(\tau_{\max}\), \(\tau_{\min}\)}
\KwOut{global graph \(\mathcal{G}^*\)}
\(m_0 \gets \operatorname{median}_j\, m_j\) \tcp*{decay midpoint}
\(q_{25}, q_{75} \gets\) quartiles of \(\{m_j\}_j\)\;
\(k \gets 2\log 3 \,/\, (\log q_{75} - \log q_{25})\), or \(k \gets 1\) if \(q_{75} = q_{25}\) \tcp*{Eq.~\eqref{eq:k_log_inter_quartile_range}}
\(n_{i,j} \gets 0\) for every event type \(X_i\) and label \(Y_j\)\;
\ForEach{local graph \(\mathcal{G} \in \{\mathcal{G}^{(1)}, \ldots, \mathcal{G}^{(m)}\}\)}{
  \ForEach{edge \(X_i \rightarrow Y_j\) in \(\mathcal{G}\)}{
    \(n_{i,j} \gets n_{i,j} + 1\) \tcp*{edge counts}
  }
}
\(\mathcal{G}^* \gets\) graph with no edges\;
\ForEach{label \(Y_j\)}{
  \(\tau_j \gets (\tau_{\max} - \tau_{\min}) \,/\, \big(1 + e^{k(\log m_j - \log m_0)}\big) + \tau_{\min}\) \tcp*{Eq.~\eqref{eq:adaptive_threshold}}
  \ForEach{event type \(X_i\) with \(n_{i,j} > 0\)}{
    \(\hat{\pi}_{i,j} \gets n_{i,j} / m\) \tcp*{edge frequency}
    \If{\(\hat{\pi}_{i,j} \geq \tau_j\)}{
      add the edge \(X_i \rightarrow Y_j\) to \(\mathcal{G}^*\)\;
    }
  }
}
\Return{\(\mathcal{G}^*\)}
\end{algorithm}

\section{Sample-Level Event-to-Event Causal Discovery}
This section contains the appendix material for Chapter~\ref{c7:event_to_event}.
\subsection{Flexible Assumptions}
We first state, in full, the sparse variant of the bounded-lagged-effects assumption used to make TRACE's conditional-independence testing scale linearly with sequence length (Section~\ref{sec:additional_ablations}).

\begin{assumption}[Bounded Lagged Effects]
\label{assumption:bounded_lagged_effects}
Causal influences have a maximal lag. There exists an integer $m\ge1$, the \emph{lag bound} (the memory of the process, as in the synthetic SCMs of Appendix~\ref{appendix:scm}), such that an event can only influence events at most $m$ steps later. Formally, for any sequence $s$, any effect event $X_{t'}$, and any earlier event $X_t$ with lag $k = t'-t > m$,
\[
X_{t'} \;\perp\; X_t \;\big|\; X_{t'-m:t'-1},
\]
i.e., given its $m$ most recent predecessors, $X_{t'}$ is independent of all earlier events: \(X_{t'} \perp X_{0:t'-m-1} \mid X_{t'-m:t'-1}\).
\end{assumption}

\subsection{Evaluation}\label{appendix:evaluation}
We used an \(ml.g5.4xlarge\) instance from AWS Sagemaker, which contains 8 vCPUs and 1 NVIDIA A10G as GPU with 24GiB for training and inference.

\paragraph{Justification for Evaluating on SCMs.}
As we saw, standard causal discovery algorithms for multivariate time series (e.g., PCMCI, Neural Hawkes Processes, CASCADE) are not applicable in our setting. Moreover, evaluating causal discovery in large scale event sequences is notoriously difficult due to the lack of ground-truth annotations in real-world traces (e.g., server logs, medical records). Furthermore, generic high-order MC are computationally intractable to materialize due to the state space \(|\mathcal{X}|^m\) of \(\{X_t\}\) and unlearnable in high dimensions without structural assumptions.

\paragraph{Synthetic Data}\label{appendix:scm}
We introduce a synthetic benchmark based on a linear Structural Causal Model (SCM). This DGP models the transition distribution \(P(X_t | X_{t-m:t-1})\) via an additive mechanism where past events exert independent \textit{excitatory} or \textit{inhibitory} influences on future outcomes such as:
\begin{small}
\begin{equation*}
    P(X_t \mid X_{t-m:t-1}) = \text{softmax}\left( \mathbf{b} + \sum_{k=1}^m \mathbf{W}^{(k)}_{x_{t-k}} \cdot \lambda(k) \right)
\end{equation*}
\end{small}
where \(\mathbf{W}^{(k)} \in \mathbb{R}^{|\mathcal{X}| \times |\mathcal{X}|}\) is a sparse weight matrix representing the causal influence of an event at lag \(k\) (with the cause–effect notation \(t \to t'\) of Chapter~\ref{c7:event_to_event}, \(k = t' - t\)), \(\mathbf{b}\) is a bias vector shared across lags, and \(\lambda(k)\) is a temporal decay function. Importantly, weights \(\mathbf{W}_{i,j}\) can be \emph{positive (\textbf{excitation}) or negative (\textbf{inhibition})}, allowing for real-world causal mechanisms present in sequences. We tune the sparsity of the weight matrix \(\mathbf{W}\) to obtain a predictability score \(\text{Pred} = 1- \frac{H(P)}{H_{max}} = 1-\frac{H(P)}{\log(|\mathcal{X}|)}\) superior or equal to \(58\%\) across the benchmarked SCMs.

\paragraph{Practical Monitoring: The Oracle Score.}
Since the true entropy $H(P)$ and thus the true KL divergence are unknown, we must approximate $\epsilon$ empirically. We observe that the cross-entropy loss decomposes as $\mathcal{L}_{AR}(\theta) = H(P) + D_{KL}(P||P_\theta)$. We propose the \textit{Oracle Score} $\hat{\epsilon}$ as a normalized estimator of the excess entropy:
\begin{equation} \label{eq:oracle_score}
\hat{\epsilon}(P_\theta) = \frac{\mathcal{L}_{AR}(\theta) - H(P)}{H_{max} - H(P)}
\end{equation}
where \(H_{max} = \log |\mathcal{X}|\) is the maximum entropy (uniform noise) and $H$ is the irreducible entropy of the DGP (approximated by the minimum validation loss observed).

This metric provides a vocabulary-agnostic measure of fit: $\hat{\epsilon} \to 0$ implies convergence to the theoretical limit ($P_\theta \to P$).

\subsection{Additional Ablations}\label{sec:additional_ablations}

\subsubsection{Number of Sampled Trajectories \(N\)}
We observe in Fig.~\ref{fig:ablation_n_particles} that increasing the number of sampled trajectories (particles) \(N\) mainly reduces the number of missed causal relationships: recall rises from about \(0.50\) at \(N=2\) to about \(0.77\), the F1 score from about \(0.66\) to about \(0.86\), and the SHD falls from about \(46\) to about \(23\), while precision stays high (above \(0.94\)) for every \(N\). After \(N > 256\), however, we did not observe significant changes.

\subsubsection{Empirical Validation of $\epsilon$-Strong Faithfulness.}\label{appendix:validation_of_epsilon_faithfulness}
\begin{figure}[H]
\centering
\begin{minipage}[t]{0.48\linewidth}
\centering
\includegraphics[width=\linewidth]{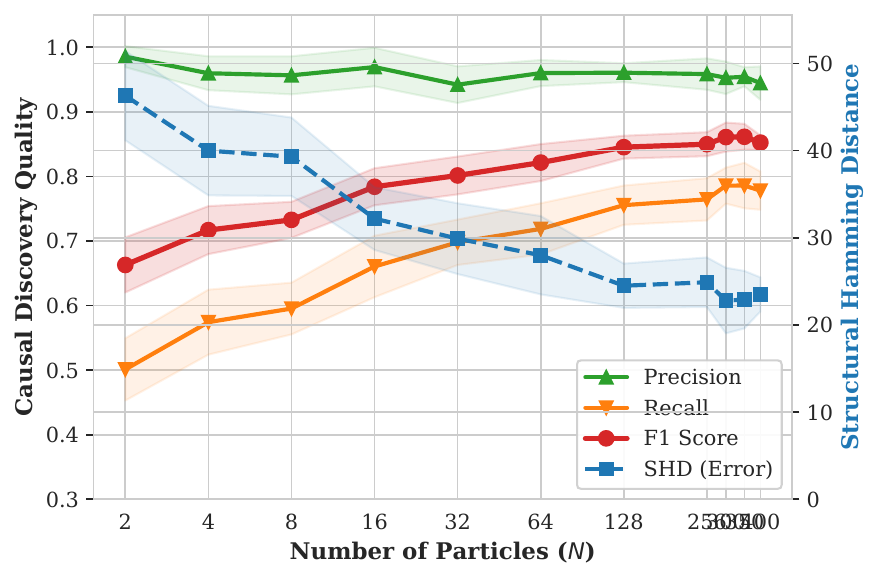}
    \caption{\textbf{Ablation of the Number of Sampled Trajectories \(N\)} (particles). Evolution of causal discovery performance as a function of the number of sampled trajectories \(N\) at \(|\mathcal{X}| = 1000, m=6\).}
\label{fig:ablation_n_particles}
\end{minipage}\hfill
\begin{minipage}[t]{0.48\linewidth}
\centering
\includegraphics[width=\linewidth]{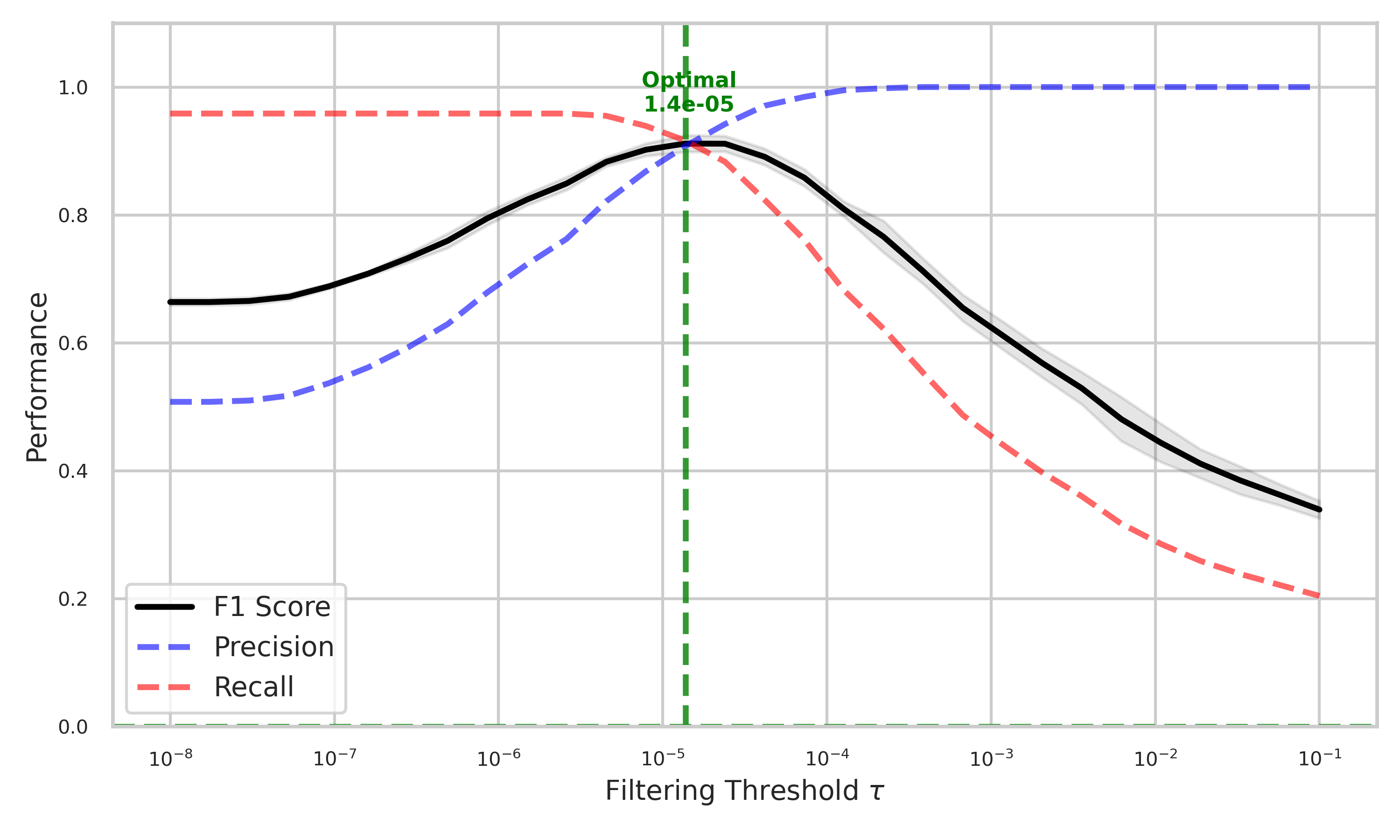}
\caption{\textbf{Sensitivity Analysis.} Classification metrics relative to the filtering threshold $\tau$ with $|\mathcal{X}|=1000, \epsilon = 0.04, c=6, m=6, L=64, N=128$.} %
\label{fig:sensitivity_threshold}
\end{minipage}
\end{figure}

Fig.~\ref{fig:sensitivity_threshold} demonstrates that the standard faithfulness assumption ($\tau = 0$) is untenable in practice, as it fails to distinguish causal signals from finite-sample approximation noise (resulting in low precision). Conversely, the distinct performance peak at $\tau_{opt} \approx 1.4 \times 10^{-5}$ empirically validates the \textbf{$\epsilon$-Strong Faithfulness} hypothesis (Def.~\ref{def:strong_faithfulness}), confirming that causal discovery requires a minimum signal-to-noise ratio. Especially, $\tau_{opt}$ is orders of magnitude lower than the worst-case theoretical bound derived in Theorem~\ref{thm:total_error_bound}, suggesting that the \textit{average-case} estimation error is significantly tighter than the Alicki-Fannes-Winter inequality implies.

\subsection{Additional Summary Causal Graph}
\begin{figure}[H]
    \centering
    \includegraphics[width=0.68\linewidth, trim=0 0 0 30pt, clip]{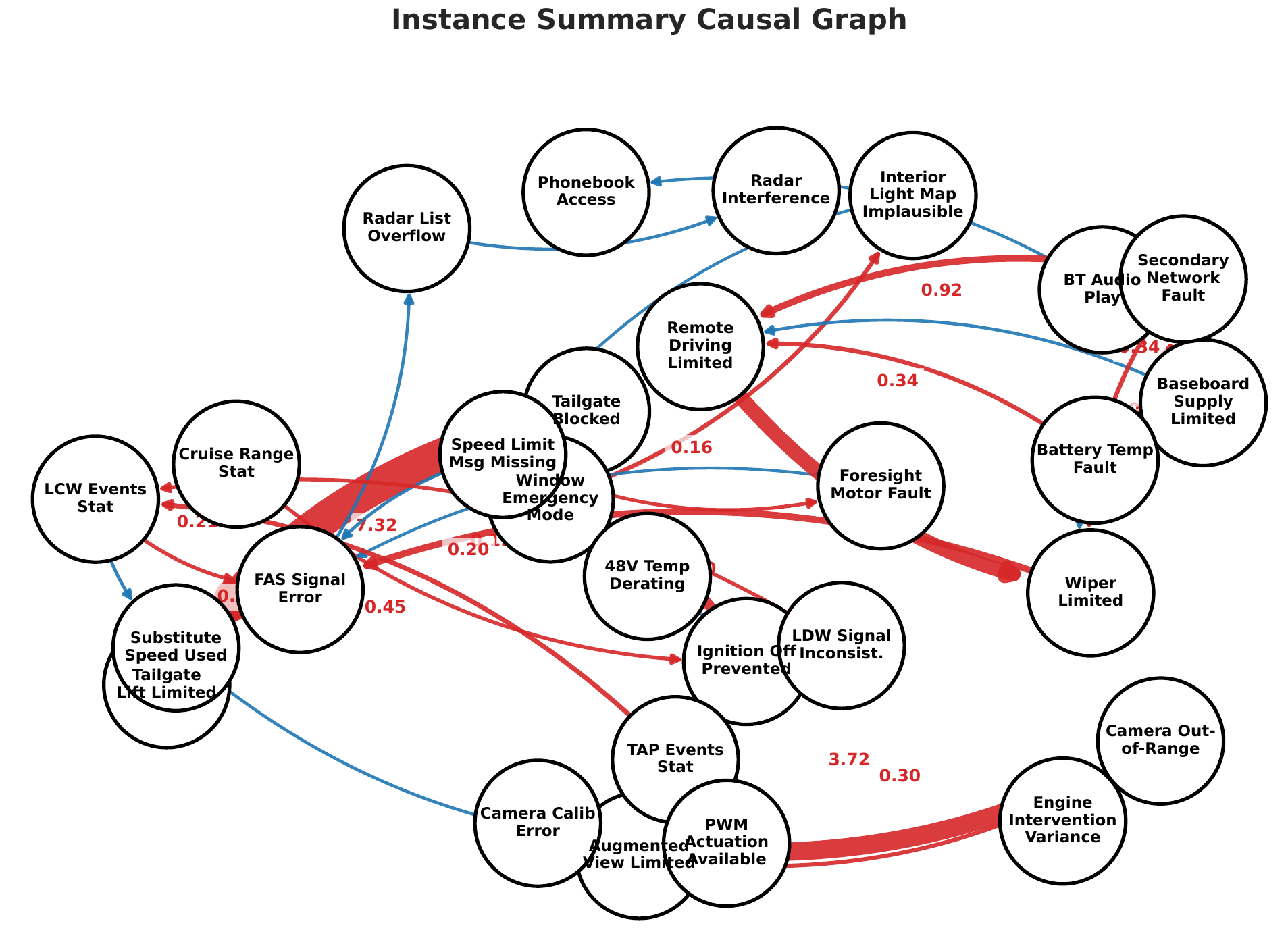}
    \caption{\textbf{Sample Summary Causal Graph \(\mathcal{G}_s\)} of the vehicle sequence shown in Fig.~\ref{fig:time_instance_graph_dtc}, obtained by projecting its sample time edges onto event types (Def.~\ref{def:iscg}). Nodes: DTC types occurring in the sequence. Red edges: strong causal relations, annotated with their estimated CMI \(\hat{I}_N\), line width proportional to \(\hat{I}_N\). Blue edges: weaker relations retained by TRACE (grey in Fig.~\ref{fig:time_instance_graph_dtc}). Event types that recur in the sequence (e.g., \emph{FAS Signal Error}) are merged into a single node and become hubs.}
    \label{fig:carformer_summary_appendix}
\end{figure}

Fig.~\ref{fig:carformer_summary_appendix} complements the sample time causal graph \(\mathcal{G}_{t,s}\) of Fig.~\ref{fig:time_instance_graph_dtc} (Section~\ref{sec:application}). It shows the sample summary causal graph \(\mathcal{G}_s\) (Def.~\ref{def:iscg}) of the \emph{same} vehicle sequence, obtained by projecting every time-indexed edge \(E_t \rightarrow E_{t'}\) of \(\mathcal{G}_{t,s}\) onto the corresponding event types \(x_{t} \rightarrow x_{t'}\).

\paragraph{How to read the figure.} Each node is one diagnostic event type (a DTC, displayed by its description) that occurs in the sequence; each directed edge points from cause to effect. Red edges are the strong causal relations: their estimated conditional mutual information \(\hat{I}_N\) is printed next to them and also encoded by the line width. Blue edges are weaker relations that TRACE still retains (drawn in grey in Fig.~\ref{fig:time_instance_graph_dtc}); their value is omitted for readability. The layout carries no meaning: unlike Fig.~\ref{fig:time_instance_graph_dtc}, there is no time axis.

\paragraph{What the projection changes.} Compared with the sample time graph, three differences are visible. (1) \emph{Time is discarded}: \(\mathcal{G}_s\) states which event types cause which, but no longer when, so the three temporal clusters of Fig.~\ref{fig:time_instance_graph_dtc} (sensor/camera events, mechanical faults, battery/network faults) are no longer separated. (2) \emph{Recurring events are merged into one node}: an event type observed at several steps, such as \emph{FAS Signal Error} (three occurrences in Fig.~\ref{fig:time_instance_graph_dtc}), collects the incoming edges of all its occurrences and therefore becomes a hub of \(\mathcal{G}_s\). (3) \emph{Cycles become possible}: since two event types can be linked in both directions at different steps, \(\mathcal{G}_s\) is not necessarily a DAG (Remark~\ref{rem:cyclic_summary}). The summary graph is thus the compact view an engineer would use for root-cause analysis of this vehicle (which DTC types to inspect first), while the sample time graph remains necessary to establish the order in which the faults unfolded.

\section{CAREP: Multi-Agent System for Automatic Rule Generation}\label{appendix:carep}
This section contains the appendix material for the chapter~\ref{c8:carep}.
\subsection{Input Format}
To illustrate exactly what evidence is handed to the causal reasoner and orchestrator agents (Fig.~\ref{fig:carep}), we reproduce below a representative JSON payload for a single unknown EP. It comprises the candidate causal DTCs, their causal indicators, and a sample of the raw DTC sequences from which they were extracted.

\begin{figure}[H]
\centering
\scriptsize
\begin{lstlisting}[language={}, numbers=none, label={lst:carep}]
"unknown EP": {
  "potential_causes": {
    "DTC3":   {"frequency": 0.78, "ACE_mean": 0.70, "ACE_std": 0.13,
               "pmi": {"DTC345": -1.89}},
    "DTC345": {"frequency": 0.49, "ACE_mean": 0.59, "ACE_std": 0.12,
               "pmi": {"DTC3": -1.89}}
  },
  "dtcs_samples": [
    "0X001 0X0008 0X0120 0X8900 .. META .. CCM 111",
    "0x6052 0X0204 0X0129 0X3410 .. META .. CCM 111",
    ...
  ]
}
\end{lstlisting}
\caption{\textbf{Input Examples Given to the Agentic System}. The \textit{potential causes} are output by the causal discovery algorithm. For instance, \textit{DTC3} is a cause of \textit{unknown EP} and has an Average Causal Effect (\textit{ACE}) of 0.7, i.e., it increases the likelihood of observing the EP on average by 70\% (Eq.~\eqref{eq:ace_mean}). Examples of defective vehicles exhibiting this error pattern are provided in the \textit{DTC samples}.}
\label{verbatim:output_json}
\end{figure}
\chapter{Data Confidentiality and Availability}\label{appendix:bmw_confidentiality}
This appendix documents the confidentiality framework under which the proprietary BMW Group vehicle datasets used throughout Chapters~\ref{c2:foundation_event_sequence_modeling}--\ref{c8:carep} were accessed. It also clarifies the reasons for the limited disclosure of dataset statistics, raw sequences, and expert-authored rule definitions in the main text (cf.\ Section~\ref{sec:dtc_composition} and Section~\ref{sec:ep_def}).

\section{Confidentiality Agreement and Data Protection}
Parts of this dissertation are based on BMW Group internal data and information. Such information may be subject to applicable BMW Group information protection requirements and therefore cannot necessarily be disclosed in full in a publicly available dissertation.
This applies in particular to BMW Group internal datasets, detailed dataset characteristics, vehicle-related information, and expert-authored Error Pattern definitions used in the research.
Consequently, certain BMW Group internal information is presented only at an appropriate level of abstraction or is not included in the publicly available version of this dissertation.

\begin{quote}

Confirmation by the industrial supervisor
I confirm that BMW Group internal data and information used in the context of this research, including applicable internal Error Pattern definitions, are subject to the applicable BMW Group information protection requirements. Their disclosure in a publicly available dissertation is therefore subject to the corresponding information protection restrictions.
This confirmation relates exclusively to the protection and disclosure of BMW Group internal information.

\medskip
\noindent Dr. Stefan Mueller\\
Supervisor, BMW Group\\
Date: \emph{22/10/2026}
\end{quote}

Confidentiality of BMW Group Information

Confirmation by the industrial supervisor
I confirm that BMW Group internal data and information used in the context of this research, including applicable internal Error Pattern definitions, are subject to the applicable BMW Group information protection requirements. Their disclosure in a publicly available dissertation is therefore subject to the corresponding information protection restrictions.
This confirmation relates exclusively to the protection and disclosure of BMW Group internal information.

\section{Reproducibility Beyond BMW Data}\label{sec:appendixc_reproducibility}
To ensure that the methodological contributions of this thesis, CarFormer, EPredictor, BiCarFormer (Part~\ref{part1}), OSCAR, CARGO, TRACE (Part~\ref{pa:p2}), and CAREP (Part~\ref{pa:p3}), remain independently verifiable without access to proprietary BMW data, we release:
\begin{itemize}
    \item the \texttt{seq2cause} library implementing OSCAR, CARGO, and TRACE as a plug-and-play layer on top of any autoregressive backbone (see the Software section of Chapter~\ref{ch:intro});
    \item a synthetic linear Structural Causal Model (SCM) benchmark generator (Appendix~\ref{appendix:scm}), with tunable vocabulary size, memory bound, and oracle score, used to reproduce every ablation reported for TRACE in Chapter~\ref{c7:event_to_event};
    \item anonymized or synthetic example sequences accompanying all public code repositories referenced throughout this thesis.
\end{itemize}
No BMW data, vehicle identifiers, DTC-to-defect mappings, or expert-authored EP rules are included in any of these public artifacts.

These artifacts have already enabled fully independent verification: an unaffiliated group~\cite{chadyuk2026replicatingtrace} reimplemented TRACE from scratch using only the public \texttt{seq2cause} reference code and the synthetic SCM benchmark described above. Entirely without access to BMW data, it reproduced the headline result of Chapter~\ref{c7:event_to_event} (F1 $\approx$ 0.90--0.91 against exact interventional ground truth) across the full range of vocabulary sizes reported in this thesis. Their findings on threshold selection and the mediator-randomization trade-off are discussed in Section~\ref{sec:limitation_threshold_mediator}.

\end{document}